\documentclass[preprint,12pt]{elsarticle}

\usepackage{amssymb}
\usepackage{amsmath}
\usepackage{subfigure}
\usepackage{float}
\usepackage{makecell}
\usepackage{algorithmic}
\usepackage{cleveref}
\Crefname{ALC@unique}{Line}{Lines}
\usepackage{multicol}
\usepackage{multirow}
\usepackage{booktabs}
\usepackage{adjustbox}
\usepackage{array}
\usepackage{tabularx}
\usepackage{cuted}
\usepackage{capt-of}
\usepackage{dblfloatfix}
\usepackage{xcolor}
\usepackage{placeins}
\usepackage{microtype}
\microtypesetup{expansion=false}
\usepackage{algorithm}      
\floatstyle{ruled}
\restylefloat{algorithm}
\usepackage[hyphens]{url}
  \newcommand{\multifieldwidth}{0.86\linewidth}
  \newcommand{\poissonfieldwidth}{0.82\linewidth}
  \newcommand{\kovasfieldwidth}{0.86\linewidth}
  \newcommand{\fullgridadaptwidth}{0.88\linewidth}
  \newcommand{\adaptfieldwidth}{0.82\linewidth}
  \newcommand{\compactdiagnosticwidth}{0.90\linewidth}
  \newcommand{\externalendpointwidth}{0.68\linewidth}
  \newcommand{\widebenchmarkwidth}{0.82\linewidth}
  \newcommand{\wideconditionwidth}{0.82\linewidth}
  \newcommand{\mechanismfigurewidth}{0.78\linewidth}

\graphicspath{{figures/}}

\begin{document}

\begin{frontmatter}
\title{Continual-Learning Physics-Informed Neural Networks for Parameterized Partial Differential Equations}
\author[thu]{Xujia Chen\corref{cor1}}
\ead{chenxj20@mails.tsinghua.edu.cn}
\author[thu]{Xinyue Hu}
\ead{hu-xy25@mails.tsinghua.edu.cn}
\author[thu]{Letian Chen}
\ead{clt21@mails.tsinghua.edu.cn}
\author[thu]{Yi Liu}
\ead{yiliu@tsinghua.edu.cn}
\author[thu]{Wenhui Fan}
\ead{fanwenhui@tsinghua.edu.cn}
\affiliation[thu]{organization={Department of Automation, Tsinghua University},
  city={Beijing},
  postcode={100084},
  country={China}}
\cortext[cor1]{Corresponding author}

\begin{abstract}
The use of artificial intelligence to solve partial differential equations (PDEs) has become an
active research topic in scientific computing. Physics-informed neural networks (PINNs) incorporate
governing equations into neural-network training and can approximate PDE solutions without requiring
large observational datasets. Parameterized PINNs (ParamPINNs) further take physical parameters as
inputs, allowing a single model to represent a family of PDE solutions over a parameter domain.
Existing ParamPINNs, however, still face inefficient training, uneven accuracy across parameters,
and overfitting to a limited set of sampled parameter tasks, which can impair generalization to
unsampled parameters. To address these issues, we propose a continual-learning physics-informed
neural network (CL-PINN), which treats PDE instances at different parameter values as related tasks
and learns them sequentially. CL-PINN combines Bayesian-optimization-based active parameter
selection, task-wise dynamic loss weighting, sparse physics-constrained replay, and an optional
parameter subnetwork to improve task allocation and knowledge retention under bounded active-task
capacity. It requires no observational data and is designed to solve parameterized PDEs over
relatively broad parameter domains under limited computational resources. Multi-seed evaluations on
five benchmarks---one continuous function and four parameterized PDEs---show that Bayesian selection
substantially reduces objective-loss queries relative to grid-greedy search, while sparse replay
mitigates forgetting of earlier tasks. Under the prescribed within-case resource protocols, CL-PINN
generally provides higher and more balanced solution accuracy than fixed-sampling and grid-greedy
baselines. CL-PINN offers a practical route toward learning PDE solutions that generalize across
physical parameters and has the potential to support reusable physics-informed surrogates for
large-scale engineering parameter studies.
\end{abstract}

\end{frontmatter}

\section{Introduction}
\label{sec:intro}

\begin{figure*}[!tp]
\centering
\includegraphics[width=0.98\linewidth]{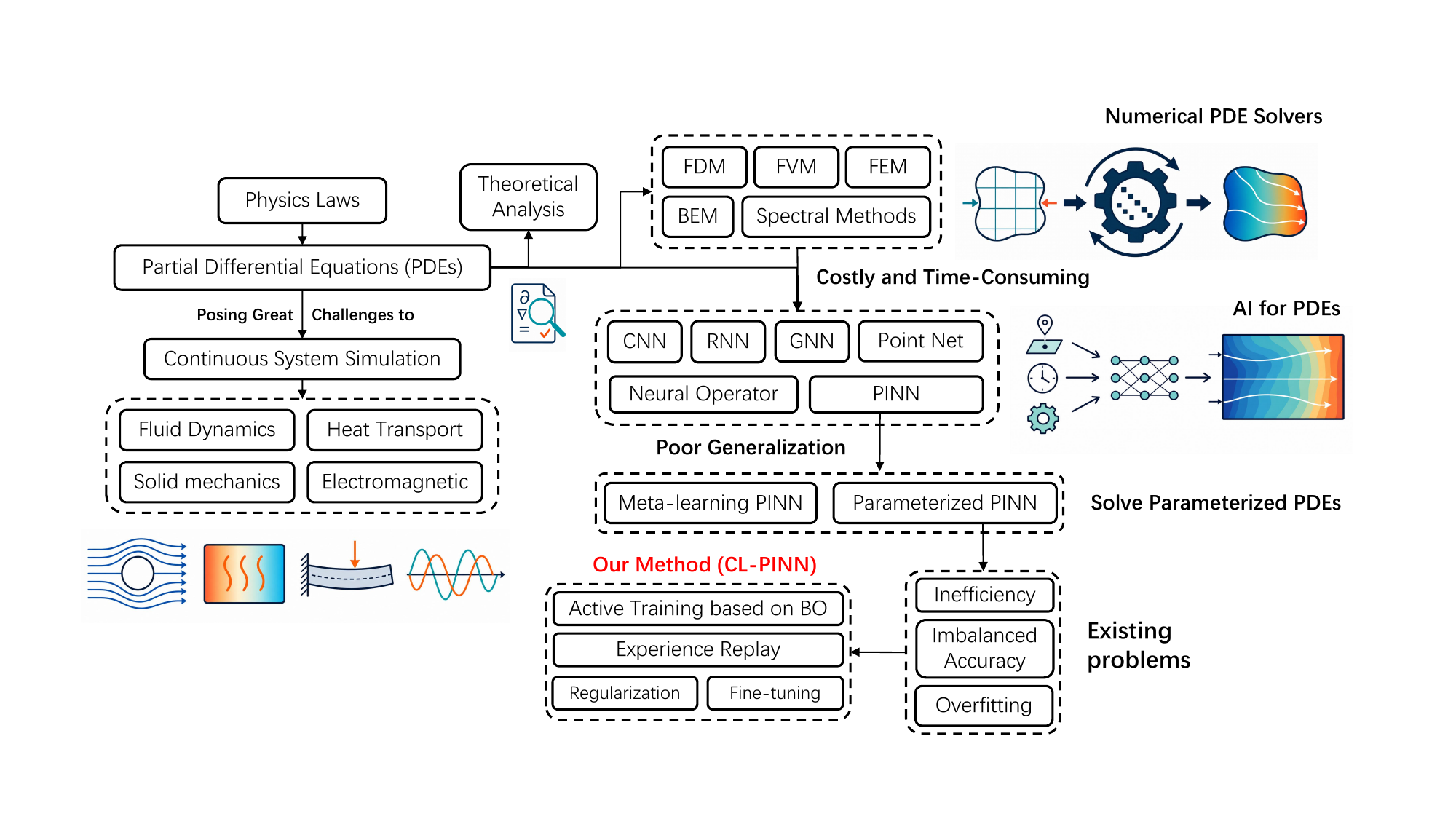}
\caption{\textbf{Background and motivation for CL-PINN: parameterized PINNs face inefficient training, imbalanced cross-parameter accuracy, and poor generalization caused by overfitting to a limited set of sampled tasks.}}
\label{fig:background}
\end{figure*}

Partial differential equations (PDEs) describe continuous systems in fluid mechanics, heat transfer,
solid mechanics, electromagnetism, and many other scientific and engineering fields
\cite{mattheij2005partial}. Nonlinear PDEs often lack closed-form solutions and exhibit complex
multiscale behavior. Conventional finite-difference, finite-volume, finite-element, spectral, and
boundary-element methods discretize the governing equations and rest on well-established numerical
analysis \cite{ames2014numerical}, but high-fidelity simulations can remain computationally
expensive. Figure~\ref{fig:background} summarizes the motivation for this study.

Recent AI-assisted PDE methods use learned models to approximate solution fields. One data-driven
route maps discrete physical states or geometric descriptions to solution fields and trains CNNs
\cite{ribeiro2020deepcfd}, RNNs \cite{ren2022phycrnet}, PointNet \cite{kashefi2021point}, or GNNs
\cite{kumar2021grade} on input--solution pairs generated by numerical solvers or experiments.
Neural operators instead learn mappings between function spaces, for example from an initial
condition, boundary condition, or forcing function to the corresponding solution function; DeepONet
and FNO are also commonly trained from paired input and solution samples
\cite{lu2019deeponet,li2020fourier}.
Other approaches couple machine-learning models to conventional solvers
\cite{meng2020ppinn,aliakbari2022predicting,mitusch2021hybrid}. These methods can combine data with
governing equations \cite{hao2022physics} and, depending on the formulation, reduce some
mesh-preprocessing work.

Physics-informed models have also been applied to inverse problems and equation discovery, for
example to infer unknown diffusion coefficients, source terms, or boundary parameters from limited
state observations, or to identify candidate terms in a governing equation
\cite{both2021deepmod,chen2021physics,stephany2022pde,lu2021physics}. Successful identification
nevertheless depends on observational sufficiency, parameter identifiability, and optimization
conditions; it is not guaranteed by the network alone.

Physics-informed neural networks (PINNs), introduced by Raissi et al. \cite{raissi2019physics},
approximate a PDE solution with a neural network and place governing equations, boundary conditions
(BCs), and initial conditions (ICs) in the training objective. Automatic differentiation evaluates
the required derivatives; Section~\ref{sec:parampinn} formalizes the resulting parameterized task
losses. This formulation converts the PDE solve into a nonlinear optimization problem without
explicitly generating a conventional mesh or solving a large discrete linear system, and it can be
combined naturally with inverse problems, design optimization, and data assimilation. PINN training
can nevertheless be difficult and time-consuming \cite{krishnapriyan2021characterizing}. Proposed
improvements include specialized architectures \cite{haitsiukevich2023improved}, hard boundary
constraints \cite{leake2020deep}, adaptive collocation sampling \cite{lu2021deepxde}, and loss
balancing \cite{xiang2022self}.

A conventional PINN is usually trained for one fixed PDE instance and need not use a large
observation dataset. Repeating that optimization for many parameter values can be expensive.
Meta-learning PINNs \cite{liu2022novel,penwarden2023metalearning} use related PDE tasks to improve
initialization, while parameterized PINNs
\cite{cho2024extension,gasick2023isogeometric,demo2023extended,de2021hyperpinn} represent a family
of PDE instances in one model. Parameterized variants have been studied in heat transfer
\cite{zhang2023parametric}, combustion \cite{liu2023surrogate}, magnetic fields
\cite{beltran2022physics}, and aerodynamics \cite{tangsali2020aerodynamic}.

Latent-space dynamics provide another route to parameterized PDE surrogates. LaSDI learns a
low-dimensional latent representation from full-order solution trajectories and identifies
parameter-dependent dynamics in that space. GP-LaSDI further uses Gaussian processes to connect
physical parameters to latent dynamics and represent interpolation uncertainty
\cite{fries2022lasdi,bonneville2024gplasdi}. These approaches rely on offline full-order
trajectories, whereas the present coordinate-based ParamPINN is trained directly from governing
constraints. They address a related need for repeated parameter queries but differ in data source
and surrogate representation.

A ParamPINN augments the physical coordinates with equation coefficients and control parameters for
the domain, ICs, or BCs. After training, the model can query solutions at different parameter values
without further weight updates. This amortizes repeated inference, stores a solution family in the
model parameters, and supports parameter-sensitivity analysis, although accuracy still depends on
training coverage and optimization.

In existing ParamPINN approaches, the number of physical samples grows with the number of parameter
tasks, increasing memory and computational requirements. Active learning can improve sampling
efficiency by adding selected parameter values incrementally. However, many existing strategies use
costly search procedures or require solver-provided observations. Without observations, multiscale
solution behavior can also produce imbalanced convergence across the parameter space, and
overfitting may occur.

We therefore propose a continual-learning strategy for training ParamPINNs without observational
data under a fixed upper bound on the active sampling set, with the aim of handling wider and
higher-dimensional parameter domains. Its three main components are as follows:

\begin{enumerate}

\item Bayesian optimization iteratively selects new parameter tasks, while dynamic task-loss weighting
mitigates multiscale solution behavior and unequal convergence rates across tasks.

\item Experience replay mitigates catastrophic forgetting after the active set reaches capacity.

\item A separate subnetwork encodes the physical parameters, and parameter-branch regularization and
freezing are evaluated independently for their effects on generalization and optimization stability.
After global training, an optional, separate adaptation stage may improve local accuracy at a
requested parameter.

\end{enumerate}

\section{Problem formulation}
\label{sec:problem}

\subsection{Parameterized physics-informed neural networks (ParamPINNs)}
\label{sec:parampinn}

\begin{figure*}[!tp]
\centering
\includegraphics[width=0.92\linewidth]{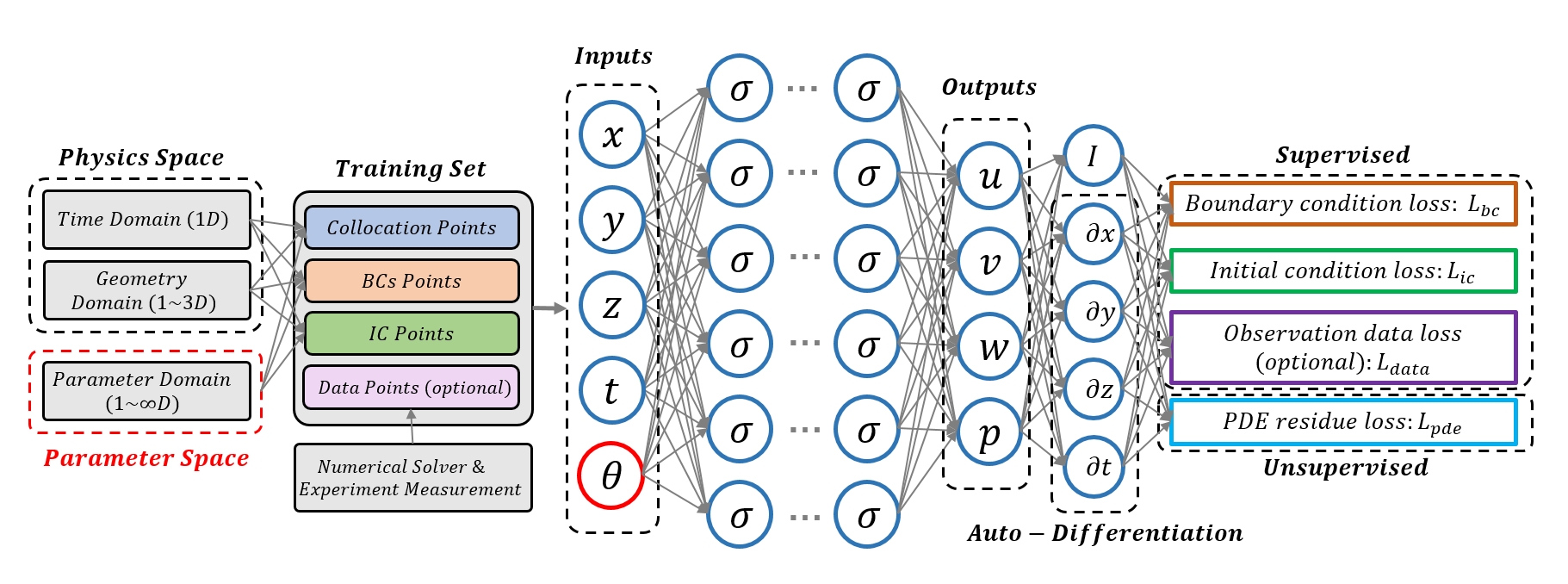}
\caption{\textbf{Structure of a ParamPINN.} A multilayer perceptron maps physical coordinates and PDE parameters to the solution, while automatic differentiation evaluates the governing residuals.}
\label{fig:pinn}
\end{figure*}

A parameterized physics-informed neural network (ParamPINN) extends a conventional PINN by fitting a
family of PDE instances rather than a single instance. After training, it can query solutions
throughout the parameter domain $\Theta$ without updating the network weights. This amortized
inference can reduce repeated-solve and solution-storage costs, expose how the solution varies with
the parameters, and support sensitivity analysis. Its accuracy outside the training parameters is
nevertheless not guaranteed.

Eq.~\ref{eq:parampinn} defines the parameterized PDE, with control parameter $\theta\in\Theta$.
Here, $\mathbf{u}(\mathbf{x},t;\theta)$ is the unknown solution,
$\partial\mathbf{u}(\mathbf{x},t;\theta)/\partial t$ is its time derivative, and
$\mathbf{N}_{\mathbf{x}}$ denotes a possibly nonlinear spatial differential operator. The symbols
$\Omega$ and $\partial\Omega$ denote the spatial domain and its boundary, and the time interval is
$[0,T]$. The functions $f(\mathbf{x};\theta)$ and $g(\mathbf{x},t;\theta)$ specify the initial and
boundary conditions, respectively.

\begin{equation}
\left\{
\begin{array}{@{}l@{\;}l@{}}
\dfrac{\partial \mathbf{u}(\mathbf{x},t;\theta)}{\partial t}
+\mathbf{N}_{\mathbf{x}}(\mathbf{u};\theta)=0,
& \begin{gathered}
\mathbf{x}\in\Omega,\ t\in[0,T],\\
\theta\in\Theta
\end{gathered},\\
\mathbf{u}(\mathbf{x},0;\theta)=f(\mathbf{x};\theta),
& \mathbf{x}\in\Omega,\ \theta\in\Theta,\\
\mathbf{u}(\mathbf{x},t;\theta)=g(\mathbf{x},t;\theta),
& \begin{gathered}
\mathbf{x}\in\partial\Omega,\ t\in[0,T],\\
\theta\in\Theta
\end{gathered}.
\end{array}
\right.
\label{eq:parampinn}
\end{equation}

For a network $\mathbf{u}_\phi$, let $\mathcal{C}_{\mathrm{pde}}^{\theta}$,
$\mathcal{C}_{\mathrm{ic}}^{\theta}$, and $\mathcal{C}_{\mathrm{bc}}^{\theta}$ denote the PDE,
initial-condition, and boundary-condition collocation sets for task $\theta$, with cardinalities
$N_{\mathrm{pde}}^{\theta}$, $N_{\mathrm{ic}}^{\theta}$, and $N_{\mathrm{bc}}^{\theta}$. Define
$\mathbf{r}_{\phi}(\mathbf{x},t;\theta)=\partial_t\mathbf{u}_{\phi}+\mathbf{N}_{\mathbf{x}}(\mathbf{u}_{\phi};\theta)$.
The three physics and condition losses used in this work are defined explicitly below before
they are aggregated across parameter tasks:

\begin{equation}
\begin{aligned}
L_{\mathrm{pde}}(\theta)
&=\frac{1}{N_{\mathrm{pde}}^{\theta}}
\sum_{(\mathbf{x},t)\in\mathcal{C}_{\mathrm{pde}}^{\theta}}
\left\|\mathbf{r}_{\phi}(\mathbf{x},t;\theta)\right\|_2^2,\\
L_{\mathrm{ic}}(\theta)
&=\frac{1}{N_{\mathrm{ic}}^{\theta}}
\sum_{\mathbf{x}\in\mathcal{C}_{\mathrm{ic}}^{\theta}}
\left\|\mathbf{u}_{\phi}(\mathbf{x},0;\theta)-f(\mathbf{x};\theta)\right\|_2^2,\\
L_{\mathrm{bc}}(\theta)
&=\frac{1}{N_{\mathrm{bc}}^{\theta}}
\sum_{(\mathbf{x},t)\in\mathcal{C}_{\mathrm{bc}}^{\theta}}
\left\|\mathbf{u}_{\phi}(\mathbf{x},t;\theta)-g(\mathbf{x},t;\theta)\right\|_2^2.
\end{aligned}
\label{eq:parampinn-components}
\end{equation}

We consider an observation-free forward-solution setting, and hence $L_{\mathrm{data}}=0$. If a
problem contains an auxiliary constraint, $L_{\mathrm{aux}}$ is likewise constructed as a
mean-squared residual over the corresponding sample set.

Let $\Theta_{\mathrm{train}}$ be the finite set of parameter tasks used for training.
Eq.~\ref{eq:parampinn_loss} defines the complete loss $L(\theta)$ for each task and the overall
ParamPINN objective. It includes an optional data term and uses coefficients $\lambda$ to balance
the PDE residual, initial condition, boundary condition, and any auxiliary constraints.

\begin{equation}
\begin{aligned}
\mathcal{L}_{\mathrm{ParamPINN}}
&=\sum_{\theta\in\Theta_{\mathrm{train}}}L(\theta)\\
&=\sum_{\theta\in\Theta_{\mathrm{train}}}\Bigl[
L_{\mathrm{data}}(\theta)
+\lambda_{\mathrm{pde},\theta}L_{\mathrm{pde}}(\theta)\\
&\qquad
+\lambda_{\mathrm{bc},\theta}L_{\mathrm{bc}}(\theta)
+\lambda_{\mathrm{ic},\theta}L_{\mathrm{ic}}(\theta)\\
&\qquad
+\lambda_{\mathrm{aux},\theta}L_{\mathrm{aux}}(\theta)
\Bigr].
\end{aligned}
\label{eq:parampinn_loss}
\end{equation}

Figure~\ref{fig:pinn} illustrates the ParamPINN. Its inputs augment the physical coordinates with
variable PDE coefficients, condition parameters, or geometric parameters. Training follows the
standard PINN procedure, but samples $(x,t,\theta)\in\Omega\times[0,T]\times\Theta$ now lie in the
product of the physical and parameter domains.

\subsection{Existing challenges of ParamPINNs}
\label{sec:issues}

A ParamPINN may require many physical samples to cover its parameter domain, increasing memory and
computational cost. Without observations, multiscale behavior can produce uneven accuracy across
parameters. When only a limited number of parameter values can be sampled, the model may also
overfit the selected tasks and generalize poorly to unsampled parameters.

\textbf{Training-efficiency issues.} ParamPINNs sample the product of the physical and parameter
domains, so large batches place substantial demands on memory and computation. Under a fixed sample
budget, the method must select informative parameter tasks rather than expanding the training set
without bound. Uniform parameter sampling can allocate insufficient effort to rapidly varying
solution regions, while manual task selection depends strongly on experience and becomes difficult
in large or high-dimensional domains. Arthurs and King \cite{arthurs2021active} iteratively add
parameter points through a grid-greedy strategy, but exhaustive grid search becomes costly as
parameter dimension increases. Moreover, a bounded active set requires a mechanism for retaining
earlier tasks after new ones are added.

Mini-batching is a compatible way to reduce the physical-point memory of an individual update,
but it does not determine which parameter tasks should receive a bounded training budget or retain
constraints from tasks displaced from a fixed-capacity active set. It also trades memory against the
number of updates and data exposure, and an unbatched evaluation can retain a large memory peak. We
therefore treat mini-batching as a resource-control option within ParamPINN training rather than a
replacement for task selection and replay; Section~\ref{sec:additional-sensitivity} reports the
measured memory--time trade-off.

\textbf{Accuracy-imbalance issues.} We consider ParamPINN training without observational data and
therefore omit $L_{\mathrm{data}}$, unlike \cite{arthurs2021active}. Optimization is driven mainly
by physics and condition losses, whose scales and convergence rates can differ across parameter
tasks. Similar training losses may therefore coexist with substantially different reference-solution
errors. Parameter selection and task-wise loss weighting are used to reduce this imbalance, but
physics loss remains an imperfect proxy for reference-solution error.

\textbf{Overfitting and local-accuracy issues.} Under limited parameter sampling, increasing network
capacity can improve the fit to complex nonlinear solutions at sampled parameters but may also
intensify overfitting to the finite training tasks and weaken generalization to unsampled
parameters. Even when a global ParamPINN has good average accuracy over the parameter domain, a few
undertrained or strictly unseen parameters may remain in local high-error regions. This creates a
practical tension between global generalization and local accuracy at a requested parameter.

\section{Method}
\label{sec:method}

\begin{figure*}[!tp]
\centering
\includegraphics[width=0.92\linewidth]{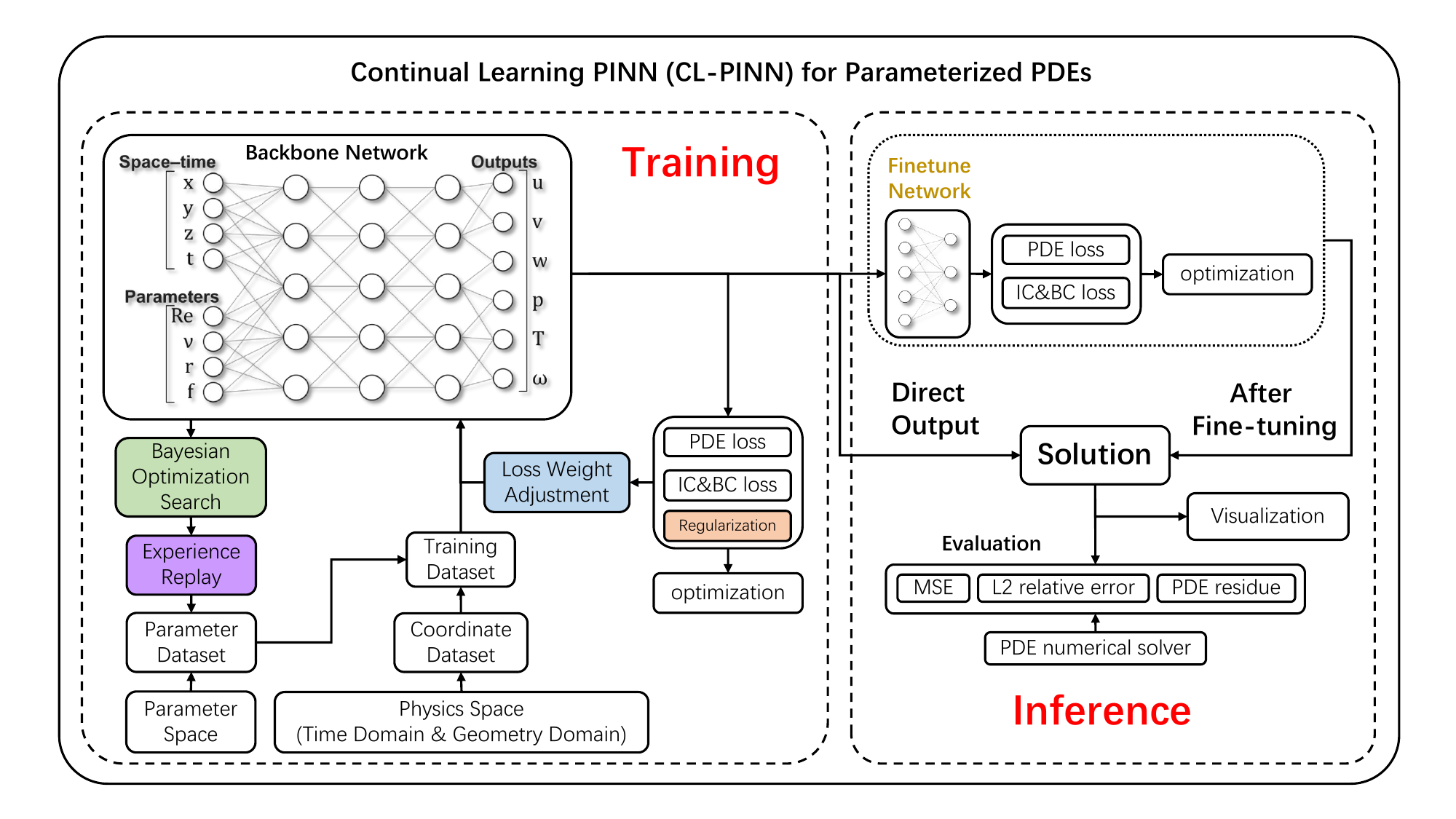}
\caption{\textbf{Overall CL-PINN workflow.} BO selects parameter tasks, dynamic weighting balances admitted tasks, sparse replay preserves constraints from displaced tasks, and the parameter subnetwork represents physical parameters; optional residual adaptation follows global training.}
\label{fig:framework}
\end{figure*}

We introduce CL-PINN, a continual-learning method for ParamPINNs that addresses the three challenges
identified above. Active parameter selection improves training efficiency under a limited query
budget, task-wise dynamic weighting mitigates cross-parameter accuracy imbalance, and a parameter
subnetwork represents variation with the physical parameters; optional single-parameter adaptation
addresses local high-error tasks when needed. Sequentially adding parameter tasks also introduces
the continual-learning-specific risk of forgetting earlier tasks, for which we use sparse
physics-constrained experience replay. The objective is to approximate a family of PDE solutions
with one model, without observational data and under a bounded active-task capacity.
Figure~\ref{fig:framework} summarizes the overall workflow.

Continual learning (CL) \cite{chen2022lifelong} studies how a model can acquire knowledge along a
task sequence while retaining performance on earlier tasks. Common routes replay historical samples
or features, regularize changes to important parameters, or allocate and expand task-specific
structure. Representative replay methods include gradient episodic memory and direct experience
replay \cite{lopezpaz2017gem,rolnick2019experience}. These studies motivate our adaptation to
capacity-limited ParamPINNs: active parameter selection allocates a bounded budget to new tasks,
while sparse replay retains constraints for earlier parameter tasks displaced from the active set.
Together, they form a continual-learning method suited to observation-free ParamPINNs.

The main distinction from supervised continual learning lies in the task signal and replay content.
Each PDE parameter value defines a task, and memory stores only that value and sparse physical
coordinates rather than target solutions. Whenever a task is selected or replayed, the current
network recomputes its PDE, boundary-condition, and initial-condition residuals at those
coordinates. Active selection therefore uses a task-level physics-informed loss to allocate the
new-task budget, dynamic weighting adjusts residual scales and convergence rates among selected
tasks, and sparse replay continues to impose the governing physics of earlier tasks. Together, these
mechanisms define continual learning for coordinate-based, observation-free ParamPINNs.

\subsection{Active training}
\label{sec:active}

Active training prioritizes parameter tasks with large physics losses, but exhaustive search
requires automatic-differentiation residuals at every candidate and becomes rapidly more expensive
as the parameter domain grows in size or dimension. The regularity premise used here is that, over
the domains considered, the parameter-to-solution map is continuous and has a bounded rate of
variation; the current model's task losses can therefore exhibit exploitable correlations between
neighboring parameters. This premise may fail when parameter changes induce abrupt physical-regime
transitions or discontinuities in the solution map.

Bayesian optimization (BO) \cite{frazier2018bayesian} follows a sequential model-based optimization
(SMBO) framework. We use the Gaussian-process regressor (GPR) in Eq.~\ref{eq:gpr} to approximate
task loss over the parameter domain. The surrogate is updated from the parameters and losses already
evaluated and is then used to select the next informative evaluation. This approach is useful when
each residual evaluation is expensive and the candidate space is large or high-dimensional.

BO is introduced here to identify parameters that remain insufficiently resolved by the current
model using fewer complete evaluations of the task objective, hereafter termed objective-loss
queries; it does not solve the nonconvex PINN optimization problem itself. Because the loss
distribution over the parameter domain changes as network training proceeds, the GP is refitted at
every active update and is used to determine subsequent queries and newly admitted training tasks.

$F(x)$ denotes the task-loss function being modeled, $\mathcal{GP}$ denotes a Gaussian process, and
$m(x)$ and $k(x,x')$ are its mean function and covariance kernel, respectively. In the
active-selection problem considered here, the generic input $x$ corresponds to the PDE parameter
$\theta$.

\begin{equation}
\begin{aligned}
F(x)&\sim\mathcal{GP}\!\left(m(x),k(x,x')\right),\\
m(x)&=\mathbb{E}[F(x)],\\
k(x,x')&=\mathbb{E}\!\left[(F(x)-m(x))(F(x')-m(x'))\right].
\end{aligned}
\label{eq:gpr}
\end{equation}

During ParamPINN training, the complete task-level physics-informed loss $L_{\mathrm{BO}}(\theta)$,
before dynamic task weighting, is periodically evaluated as the default BO observation. When a
validated search prior is enabled, the transformed observation is
$S_{\mathrm{BO}}(\theta)=s(\theta)L_{\mathrm{BO}}(\theta)$, where $s(\theta)=1$ is the neutral
choice and $s(\theta)=f_{\mathrm{prior}}(\theta)$ provides problem-specific score shaping. The
case-specific choice is stated in each experimental protocol. Fitting the GP to these observations
yields a posterior mean $\mu(\theta)$ and standard deviation $\sigma(\theta)$ for each candidate
parameter. The upper confidence bound (UCB) in Eq.~\ref{eq:ucb} combines the two: a large $\mu$
favors parameters predicted to have high loss and to be undertrained, while a large $\sigma$
encourages exploration of uncertain regions. The coefficient $\kappa$ controls exploration relative
to exploitation. The BO score selects new tasks, whereas the dynamic weights in
Eq.~\ref{eq:loss_bala} balance tasks already admitted to training; the two serve different purposes.
BO is therefore a query-allocation heuristic and does not guarantee a globally optimal task
sequence, the ranking induced by reference-solution error, or a reduction in wall-clock search time.

Nonconvex network optimization makes the observed task loss depend on initialization and the
current training stage, while multiscale residual components can produce narrow peaks and
heterogeneous variation over the parameter domain. A stationary GP may interpret these effects as
observation noise or incompatible length scales, which can alter acquisition rankings. Refitting the
GP and using UCB exploration mitigate stale or uncertain rankings but do not remove these
limitations.

\begin{equation}
a_{\mathrm{UCB}}(\theta)=\mu(\theta)+\kappa\,\sigma(\theta).
\label{eq:ucb}
\end{equation}

Loss balancing is important in PINN optimization because different residual terms and parameter
tasks can evolve at different scales \cite{krishnapriyan2021characterizing}. To adapt task weights
using both prescribed problem knowledge and recent loss dynamics, we use Eq.~\ref{eq:loss_bala}.

\begin{equation}
\begin{aligned}
L_{q,\mathrm{norm}}(\theta)
&=\frac{L_q(\theta)}{\sum_{\vartheta\in\Theta_{\mathrm{train}}}L_q(\vartheta)},\\
w_{q,\theta}
&=f_{\mathrm{prior}}(\theta)
\exp\!\Bigg[
\lambda_{\mathrm{static}}L_{q,\mathrm{norm}}(\theta)\\
&\qquad {}+\lambda_{\mathrm{dynamic}}
\frac{L_{q-1}(\theta)-L_q(\theta)}{L_q(\theta)+L_{q-1}(\theta)}
\Bigg],\\
\widetilde{w}_{q,\theta}
&=\frac{w_{q,\theta}}
{\sum_{\vartheta\in\Theta_{\mathrm{train}}}w_{q,\vartheta}}
\left|\Theta_{\mathrm{train}}\right|.
\end{aligned}
\label{eq:loss_bala}
\end{equation}

$q$ indexes weight-update events and is not the physical time coordinate. $L_{q,\mathrm{norm}} \in
[0, 1]$ is the normalized task loss, and
$[L_{q-1}(\theta)-L_q(\theta)]/[L_q(\theta)+L_{q-1}(\theta)]\in[-1,1]$ describes the recent loss
decrease. The normalized weight satisfies $\sum_{\theta\in\Theta_{\mathrm{train}}}\widetilde
w_{q,\theta}=|\Theta_{\mathrm{train}}|$. The objective optimized between weight updates is
written explicitly as follows:

\begin{equation}
\mathcal{L}_{q}
=\sum_{\theta\in\Theta_{\mathrm{train}}}
\widetilde{w}_{q,\theta}L_q(\theta).
\label{eq:weighted_total}
\end{equation}

$\Theta_{\mathrm{train}}$ denotes the parameter tasks currently being trained.

$f_{\mathrm{prior}}$ introduces optional problem knowledge into the task weights but is not required
for the framework to operate. It can moderately increase the initial training share of regions
expected to be difficult, small in solution amplitude, or vulnerable to high relative error. The
prior should follow the scale and distribution of the parameter domain, use a simple smooth form
such as a linear or logarithmic transformation, and limit its contrast across parameters so that it
does not overwhelm the current residual signal. In Eq.~\ref{eq:loss_bala}, the static term allocates
weight according to the relative scale of the current loss, whereas the dynamic term adjusts weight
according to recent convergence. When reliable problem knowledge is unavailable, the neutral choice
$f_{\mathrm{prior}}=1$ leaves BO, dynamic adjustment, and experience replay fully operational.

$f_{\mathrm{prior}}$ encodes problem-specific prior weights over parameters.
$\lambda_{\mathrm{static}}$ controls the influence of the normalized current loss, and
$\lambda_{\mathrm{dynamic}}$ controls the influence of the recent relative loss decrease. With the
positive $\lambda_{\mathrm{static}}$ and negative $\lambda_{\mathrm{dynamic}}$ used here, larger
weights are assigned to tasks emphasized by the prior, tasks with larger normalized loss, or tasks
whose loss decreases more slowly.

Algorithm~\ref{algl:ac} summarizes active training. $\Theta$ is the discretized candidate parameter
domain, $\Theta_{\mathrm{train}}$ is the current active task set, $X_{\mathrm{phy}}$ is the physical
collocation set, and $X_{\mathrm{train}}=X_{\mathrm{phy}}\times\Theta_{\mathrm{train}}$.
$N_{\mathrm{resample}}$ is the interval between active updates, $N_{\mathrm{train}}$ is the
requested optimization length, and $N_{\mathrm{bayesian}}$ is the number of Bayesian evaluations per
active update.

\begin{algorithm}[!tbp]
\caption{Active parameter-task selection and dynamic weighting}
\label{algl:ac}
\footnotesize
\begin{algorithmic}[1]
\REQUIRE Candidate parameter domain $\Theta$, physical collocation set $X_{\mathrm{phy}}$,
training length $N_{\mathrm{train}}$, active-update interval $N_{\mathrm{resample}}$, and
$N_{\mathrm{bayesian}}$ BO evaluations per update
\ENSURE Trained network and active task set $\Theta_{\mathrm{train}}$
\STATE Initialize the network and $X_{\mathrm{phy}}$; initialize $\Theta_{\mathrm{train}}$
with the corner points of $\Theta$.
\FOR{$i=1,\ldots,N_{\mathrm{train}}$}
  \STATE Evaluate the physics and condition residuals on
  $X_{\mathrm{phy}}\times\Theta_{\mathrm{train}}$ and update the network parameters.
  \IF{$i\bmod N_{\mathrm{resample}}=0$}
    \STATE Initialize the GP observation set $\mathcal{O}_{\mathrm{gpr}}$ with the tasks in
    $\Theta_{\mathrm{train}}$ and their current scores $S_{\mathrm{BO}}=sL_{\mathrm{BO}}$.
    \STATE Repeat $N_{\mathrm{bayesian}}$ times: fit the GP, maximize UCB, evaluate
    $L_{\mathrm{BO}}$, transform it to $S_{\mathrm{BO}}$, and add the result to
    $\mathcal{O}_{\mathrm{gpr}}$.
    \STATE Among candidate parameters not yet trained, select $\theta_{\mathrm{new}}$ with
    the largest GP-predicted score.
    \STATE Add $\theta_{\mathrm{new}}$ to $\Theta_{\mathrm{train}}$ and update the task
    weights using Eq.~\ref{eq:loss_bala}.
  \ENDIF
\ENDFOR
\end{algorithmic}
\end{algorithm}

\subsection{Experience replay}
\label{sec:replay}

As active training continually adds parameter tasks, the dense collocation set eventually reaches
the capacity allowed by the computational budget. Adding another dense task then requires removing a
previously trained parameter, which can reduce accuracy on the displaced task and cause catastrophic
forgetting. Early loss estimates may also temporarily concentrate active tasks in a restricted part
of the parameter domain, further increasing this risk.

We therefore use sparse replay. Newly added and currently prioritized tasks use dense physical
collocation points, whereas a displaced task retains its parameter value and a small set of physical
coordinates in replay memory. During subsequent training, the current network recomputes the PDE,
boundary-condition, and initial-condition residuals at those coordinates, thereby continuing to
constrain the earlier task without storing any reference-solution labels. Compared with retaining
full collocation sets for all historical parameters, sparse replay keeps a wider range of parameter
tasks under physics constraints at lower memory and computational cost, mitigating degradation on
earlier tasks while the model adapts to new ones. The design follows the experience-replay principle
in continual learning \cite{lopezpaz2017gem,rolnick2019experience}, but replaces labeled examples
with parameter values and physical collocation points.

Algorithm~\ref{algl:replay} summarizes this procedure. $\Theta_{\mathrm{active}}$ and
$\Theta_{\mathrm{replay}}$ are the dense active and sparse replay task sets,
$X_{\mathrm{phy,sparse}}$ is the sparse physical collocation set, and $n_{\mathrm{param,active}}$
and $n_{\mathrm{param,replay}}$ are their respective task capacities.

\begin{algorithm}[!tbp]
\caption{Sparse physics-constrained experience replay}
\label{algl:replay}
\footnotesize
\begin{algorithmic}[1]
\REQUIRE Dense collocation set $X_{\mathrm{phy}}$, sparse set $X_{\mathrm{phy,sparse}}$,
active-task capacity $n_{\mathrm{param,active}}$, and replay-task capacity
$n_{\mathrm{param,replay}}$
\ENSURE Active task set $\Theta_{\mathrm{active}}$ and replay task set
$\Theta_{\mathrm{replay}}$
\STATE Initialize $\Theta_{\mathrm{active}}$ with the parameter-domain corners and set
$\Theta_{\mathrm{replay}}=\varnothing$.
\FOR{$i=1,\ldots,N_{\mathrm{train}}$}
  \STATE Jointly train on $X_{\mathrm{phy}}\times\Theta_{\mathrm{active}}$ and
  $X_{\mathrm{phy,sparse}}\times\Theta_{\mathrm{replay}}$.
  \IF{$i\bmod N_{\mathrm{resample}}=0$}
    \STATE Use Algorithm~\ref{algl:ac} to estimate the losses of candidate, active, and
    replay tasks.
    \STATE Let $\theta_u$ be the highest-loss unseen task and $\theta_a$ the lowest-loss
    active task.
    \IF{the active set is not full}
      \STATE Add $\theta_u$ to $\Theta_{\mathrm{active}}$.
    \ELSE
      \STATE When replay capacity permits, compare $\theta_u$, $\theta_a$, and the
      highest-loss replay task $\theta_r$; retain the task most in need of dense training
      in $\Theta_{\mathrm{active}}$ and store the displaced earlier task sparsely in
      $\Theta_{\mathrm{replay}}$.
    \ENDIF
    \STATE Update the current task weights using Eq.~\ref{eq:loss_bala}.
  \ENDIF
\ENDFOR
\end{algorithmic}
\end{algorithm}

\subsection{Parameter subnetwork and optimizer controls}
\label{sec:param-subnetwork}

A ParamPINN must represent variation in both physical coordinates and parameters, and its accuracy
can decrease at sparsely sampled parameter values. We therefore introduce a separate parameter-input
branch that maps the physical parameters to features before fusing them with the coordinate-backbone
representation. This structure separates parameter encoding from physical-coordinate fitting and
permits optimizer controls to be applied specifically to the parameter branch.

\begin{figure*}[!tp]
\centering
\includegraphics[width=0.80\linewidth]{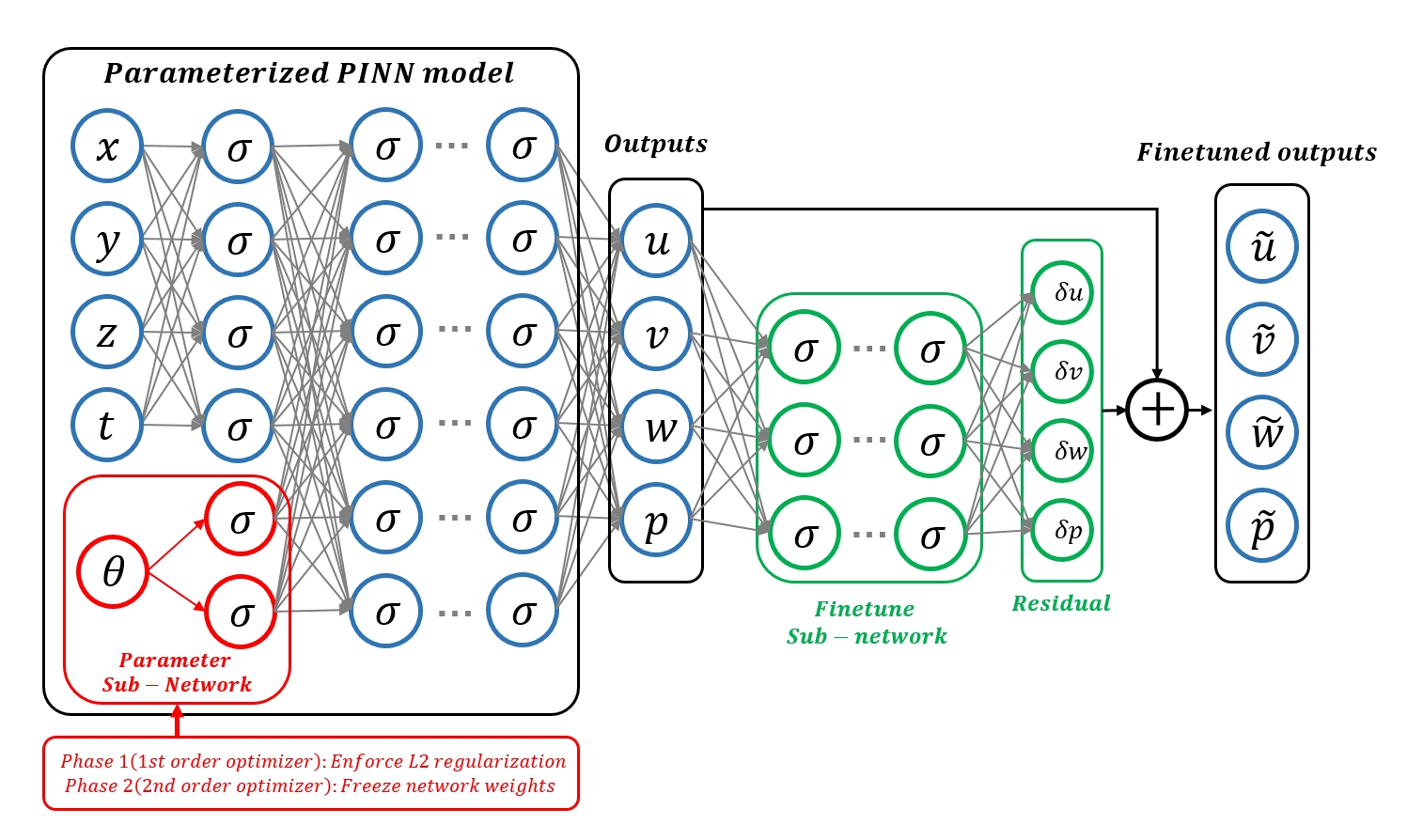}
\caption{\textbf{Parameter subnetwork and optional single-parameter adaptation.} The parameter branch is subject to optional Adam weight decay and L-BFGS-stage freezing; after global training, a compact residual head can be adapted at one specified parameter.}
\label{fig:reg_fine}
\end{figure*}

Under the same regularity premise, the solution varies continuously with the physical parameters and
has a bounded rate of variation over the domains considered here. The parameter inputs therefore
first pass through a separate subnetwork, whose features are then fused with the coordinate
backbone. This design partially separates the representation of how the solution varies with the
parameters from the fitting of its spatial and temporal variation, rather than treating raw
parameters and coordinates identically in one input layer.

When the parameter tasks are few or unevenly distributed, the parameter branch may become overly
sensitive to the sampled values. If targeted decay is enabled, the regularized objective in
Eq.~\ref{eq:weightdecay} is used during Adam, with the penalty applied only to this branch. It
constrains the branch weights and reduces the risk of oversensitivity to the finite sampled tasks.
After switching to L-BFGS, the optional freezing control fixes the learned parameter representation
while the coordinate backbone and output layers continue to be optimized, reducing the risk that
broad second-order updates disrupt the cross-parameter representation. Decay and freezing are
testable optimizer controls motivated by these considerations; their benefits remain equation
dependent.

\begin{equation}
J_{\mathrm{reg}}(\boldsymbol{\phi})
=J(\boldsymbol{\phi})
+\frac{\lambda_{\mathrm{decay}}}{2}
\left\|\boldsymbol{\phi}_{\mathrm{par}}\right\|_2^2.
\label{eq:weightdecay}
\end{equation}

Here, $\boldsymbol{\phi}$ denotes all network parameters, $\boldsymbol{\phi}_{\mathrm{par}}$
contains only the parameter-branch weights, $J$ is the current training objective, and
$\lambda_{\mathrm{decay}}$ is the targeted weight-decay coefficient.

\subsection{ACR2-finetune: optional single-parameter residual fine-tuning}
\label{sec:finetune}

We denote this optional single-parameter downstream adaptation by \textbf{ACR2-finetune}. As shown
on the right of Figure~\ref{fig:reg_fine}, the trained parameterized PINN first produces the base
output. A compact fine-tuning network with one hidden layer branches from the last shared feature
representation, generates a residual correction for each output component, and adds these
corrections componentwise to the base output. Its output layer is initialized to zero, so the
prediction at the start of adaptation is exactly that of the global model. All base-model parameters
remain frozen, and only the residual head is optimized using newly sampled PDE, boundary-condition,
and initial-condition points at the requested parameter. The global model and its solutions at all
other parameter values therefore remain unchanged. ACR2-finetune denotes only this downstream
adaptation procedure and is not part of the global ACR2-arch configuration.

This mode exchanges a small amount of additional online optimization for improved local accuracy at
one requested parameter and is intended for cases in which zero-update inference from the global
model does not meet the local accuracy requirement. Because the base model remains frozen,
single-parameter adaptation does not alter the learned representation over the full parameter
domain.

\section{Experiments}
\label{sec:experiments}

We evaluate CL-PINN on one continuous-function benchmark and four parameterized PDEs. Within each
comparison, common settings are held fixed except for explicitly reported differences in
architecture, replay density, resampling period, and optimization budget. Three parameter-task
baselines are included: uniform sampling (UNI), manual fixed tasks (FIX), and grid-greedy active
selection (AG). UNI and FIX use predetermined training parameters, as in prior studies of
parameterized PINNs
\cite{de2021hyperpinn,cho2024extension,demo2023extended,tangsali2020aerodynamic,liu2023surrogate,baldi2016parameterized,sun2020surrogate}.
AG follows the dynamic grid-greedy selection principle of \cite{arthurs2021active}. The proposed
variants are AC, which combines Bayesian active selection and task-wise weighting without replay;
ACR, which adds fixed-capacity sparse replay; and ACR2-arch, the global ParamFNN configuration
evaluated under the case-specific replay and resampling protocols stated below. Because these
primary configurations may differ in more than one component, single-factor effects are identified
only in the controlled ablations.

For a unified description of the evaluation protocol, let $u_\phi(\mathbf{x},\boldsymbol{\theta})$
and $u(\mathbf{x},\boldsymbol{\theta})$ denote the network prediction and reference solution,
respectively, at parameter $\boldsymbol{\theta}$, and let $\{\mathbf{x}_i\}_{i=1}^{N_\theta}$ be the
corresponding evaluation coordinates. The per-parameter mean-squared error (MSE) and relative $L_2$
error are defined as

\begin{equation}
\begin{aligned}
\mathrm{MSE}(\boldsymbol{\theta})
&=\frac{1}{N_\theta}\sum_{i=1}^{N_\theta}
\left|u_\phi(\mathbf{x}_i,\boldsymbol{\theta})
-u(\mathbf{x}_i,\boldsymbol{\theta})\right|^2,\\
E_{L_2}(\boldsymbol{\theta})
&=\frac{\left\|u_\phi(\cdot,\boldsymbol{\theta})
-u(\cdot,\boldsymbol{\theta})\right\|_2}
{\left\|u(\cdot,\boldsymbol{\theta})\right\|_2}.
\end{aligned}
\label{eq:eval-metrics}
\end{equation}

Throughout the remainder of the paper, $\mathrm{MSE}(\boldsymbol{\theta})$ denotes the per-parameter
mean-squared error and $E_{L_2}(\boldsymbol{\theta})$ the per-parameter relative $L_2$ error. Table
entries labeled ``MSE'' and ``$E_{L_2}$'' are aggregates over the fixed test parameters and random
seeds under the stated protocol.

Let $\Theta_{\mathrm{test}}$ denote the fixed test-parameter set. The macro relative $L_2$ first
evaluates $E_{L_2}$ for each parameter task and then assigns equal weight to all tasks, whereas the
worst relative $L_2$ is the maximum over this set:

\begin{equation}
\begin{aligned}
E_{L_2}^{\mathrm{macro}}
&=\frac{1}{|\Theta_{\mathrm{test}}|}
\sum_{\boldsymbol{\theta}\in\Theta_{\mathrm{test}}}E_{L_2}(\boldsymbol{\theta}),\\
E_{L_2}^{\mathrm{worst}}
&=\max_{\boldsymbol{\theta}\in\Theta_{\mathrm{test}}}E_{L_2}(\boldsymbol{\theta}).
\end{aligned}
\label{eq:eval-macro-worst}
\end{equation}

For a PDE, let $\mathbf{r}_\phi(\mathbf{x};\boldsymbol{\theta})\in\mathbb{R}^{q}$ contain the $q$
governing-equation residual components obtained by substituting the prediction into the PDE, and let
$\{\mathbf{x}^{r}_i\}_{i=1}^{N^r_\theta}$ be the residual-evaluation points. The reported
per-parameter mean PDE residual (Res-PDE) is the mean absolute residual over all points and residual
components; it can then be macro-averaged with equal weight across test parameters:

\begin{equation}
\begin{aligned}
R_{\mathrm{PDE}}(\boldsymbol{\theta})
&=\frac{1}{qN^r_\theta}
\sum_{i=1}^{N^r_\theta}\sum_{j=1}^{q}
\left|r_{\phi,j}(\mathbf{x}^{r}_i;\boldsymbol{\theta})\right|,\\
R_{\mathrm{PDE}}^{\mathrm{macro}}
&=\frac{1}{|\Theta_{\mathrm{test}}|}
\sum_{\boldsymbol{\theta}\in\Theta_{\mathrm{test}}}R_{\mathrm{PDE}}(\boldsymbol{\theta}).
\end{aligned}
\label{eq:eval-res-pde}
\end{equation}

Res-PDE is an evaluation metric and should not be confused with the squared-residual training loss
$L_{\mathrm{pde}}$. For the multi-output Kovasznay flow, MSE and relative $L_2$ are evaluated
separately for $u$, $v$, and $p$; when a single macro quantity is required, the three components are
equally weighted. Aggregates across random seeds use the fixed seeds $[0,1,2]$ and are reported as
the arithmetic mean and population standard deviation.

The experiments were conducted on Ubuntu 20.04 using NVIDIA Tesla V100S-PCIE GPUs (32 GB each) and
an Intel Xeon Gold 6348 processor with 128 GB of memory. The implementation is based on Python and
PyTorch, with scikit-learn, bayesian-optimization \cite{bayesianopt}, NumPy, and Matplotlib. Core
components and selected benchmark definitions are adapted from DeepXDE \cite{lu2021deepxde}.

To facilitate reproducibility, the source code of CL-PINN is publicly available on GitHub:
\url{https://github.com/pigofmomo/CLPINN}.

Reference solutions for testing are generated either analytically or with external PDE solvers. The
one-dimensional time-dependent Burgers and Allen--Cahn problems are solved using MATLAB's
\texttt{pdepe}, whereas the two-dimensional steady linearized Poisson--Boltzmann problem is solved
using COMSOL Multiphysics through its MATLAB interface.

All primary experiments use FP32 and the fixed random seeds $[0,1,2]$, with arithmetic means and
population standard deviations reported. The requested Adam/L-BFGS limits are $20/20$k for
Schaffer-like and Burgers, $20/10$k for Allen--Cahn, and $10/10$k for Kovasznay. For the
four-parameter Poisson--Boltzmann problem, UNI and FIX request 20k Adam updates, whereas the four
dynamic methods request 40k; all methods use an L-BFGS limit of 20k iterations, with possible early
termination upon convergence. Within each benchmark, the training budget, loss weights,
Bayesian-query budget, exploration coefficient, and coverage controls are prespecified and held
fixed across the three seeds; no seed-specific tuning or result selection is performed. The
benchmark subsections below specify the network architecture, sample counts, replay capacities,
resampling period, and active-selection parameters.

Network architecture search is not an objective of this study. Each case uses a moderate network
size common in PINN literature and benchmark practice, with limited adjustments for input dimension,
output components, and solution complexity. These architectures provide comparable capacity within
each benchmark rather than a claim of equation-wise optimality. The capacity-matched
concatenation/ParamFNN control and the depth--width diagnostic further show that mean and tail
metrics can rank near-capacity-matched networks differently; we therefore do not extrapolate a
universally optimal capacity from one architecture.

\subsection{Parameterized benchmark and PDE solution results}
\label{sec:parametric-results}

\subsubsection{Continuous-function benchmark}
\label{sec:schaffer}

Schaffer-type functions \cite{schaffer2014multiple} are commonly used to test global-optimization
and metaheuristic algorithms. Their nonconvex, multiscale, and multimodal structure changes with the
parameter and is therefore nontrivial to approximate. We first use a Schaffer-like continuous
function as an illustrative example of the proposed method.

\begin{equation}
f(x,y;a)
=\frac{\sin(x^2+y^2)-0.5}
{\left[1+a(x^2+y^2)\right]^2}.
\label{eq:schaffer}
\end{equation}

The spatial domain is $\Omega = \{(x,y) \mid (x,y) \in [-5,5] \times [-5,5]\}$. The parameter domain
is $\Theta = \{a \mid a\in [0.001, 1.0] \}$. The sampling and capacity settings are
$|X_{\mathrm{phy}}|=2500$, $|X_{\mathrm{phy,sparse}}|=0.1|X_{\mathrm{phy}}|$,
$n_{\mathrm{param,active}}=15$, and $n_{\mathrm{param,replay}}=15$. ACR uses sparse replay in the
main table, whereas ACR2-arch uses full-density replay at the same parameter-task capacity. The
protocol requests $2\times10^4$ Adam updates and at most $2\times10^4$ L-BFGS iterations. The
standard FNN has four 30-neuron hidden layers with three inputs and one output; ACR2-arch replaces
the input layer with coordinate and parameter subnetworks. The current AC/ACR runs use
$N_{\mathrm{resample}}=1000$, whereas ACR2-arch uses 500. Both the training weights and BO search
scores use $f_{\mathrm{prior}}(a)=\exp[1.5(\log_{10}a+3)]$; the dynamic-weight parameters are
$\lambda_{\mathrm{static}}=2$ and $\lambda_{\mathrm{dynamic}}=-2$. Each active update uses 10
Bayesian evaluations with $\kappa=5$. Errors are evaluated at 28 logarithmically distributed
parameter values using three independent random seeds. Table~\ref{table:schaffer} reports the mean
and population standard deviation.

\begin{figure*}[!tp]
\centering
{\footnotesize\textbf{$a=0.001$}}\par\vspace{0.1em}
\includegraphics[width=\multifieldwidth]{main/04_experiments/01_benchmark_results/01_schaffer/schaffer_fields__a_0p001.png}
\par\vspace{-0.35em}
{\footnotesize\textbf{$a=0.5$}}\par\vspace{0.1em}
\includegraphics[width=\multifieldwidth]{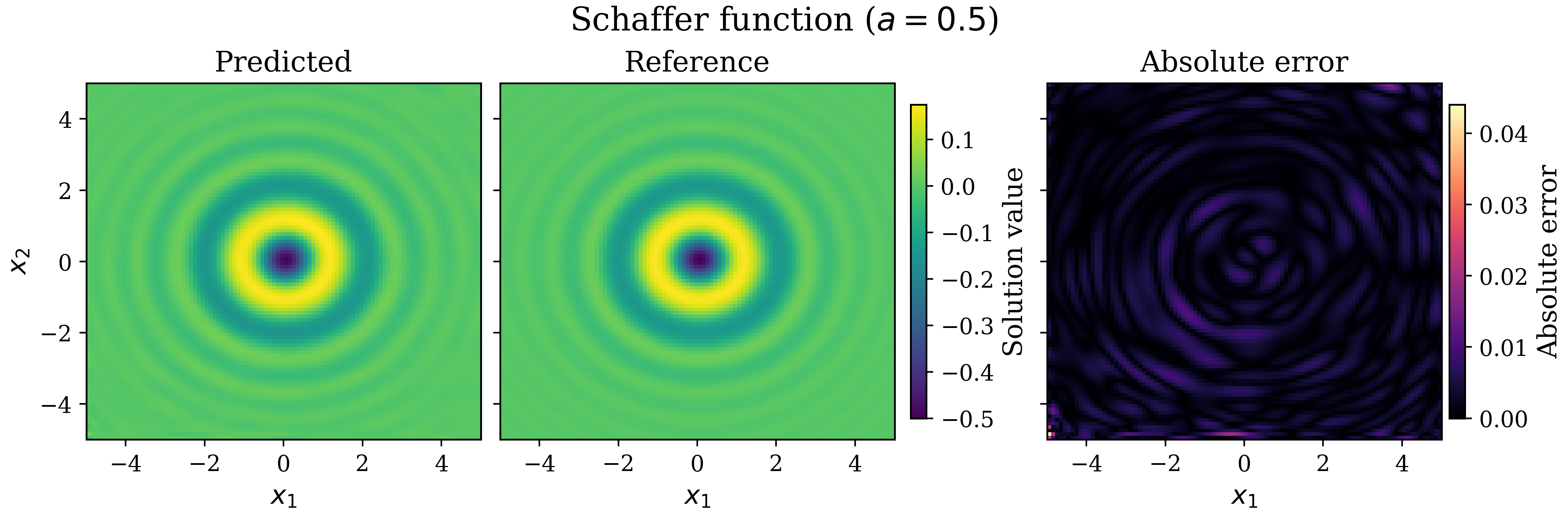}
\par\vspace{-0.35em}
{\footnotesize\textbf{$a=1.0$}}\par\vspace{0.1em}
\includegraphics[width=\multifieldwidth]{main/04_experiments/01_benchmark_results/01_schaffer/schaffer_fields__a_1p0.png}
\caption{\textbf{ACR2-arch prediction, reference, and absolute error for the Schaffer-like benchmark at $a=0.001$, $0.5$, and $1.0$, representing the lower endpoint, interior, and upper endpoint of the parameter domain.}}
\label{fig:schaffer_fields}
\end{figure*}

\begin{figure*}[!tp]
\centering
\includegraphics[width=\widebenchmarkwidth]{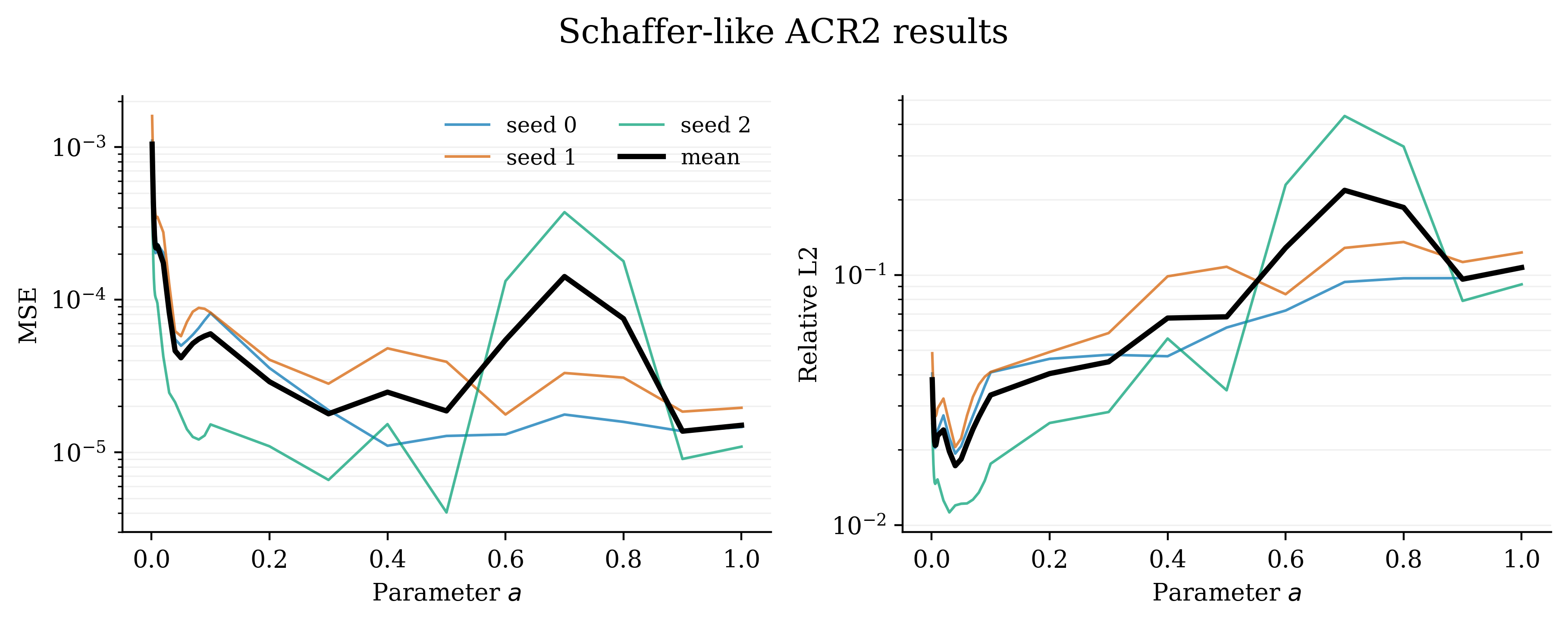}
\caption{\textbf{Parameter-wise MSE and $E_{L_2}$ of ACR2-arch over 28 Schaffer-like test parameters.} Thin curves denote independent runs, the black curve their pointwise mean, and the vertical axes are logarithmic.}
\label{fig:schaffer_err}
\end{figure*}

\begin{figure*}[!tp]
\centering
\includegraphics[width=\mechanismfigurewidth]{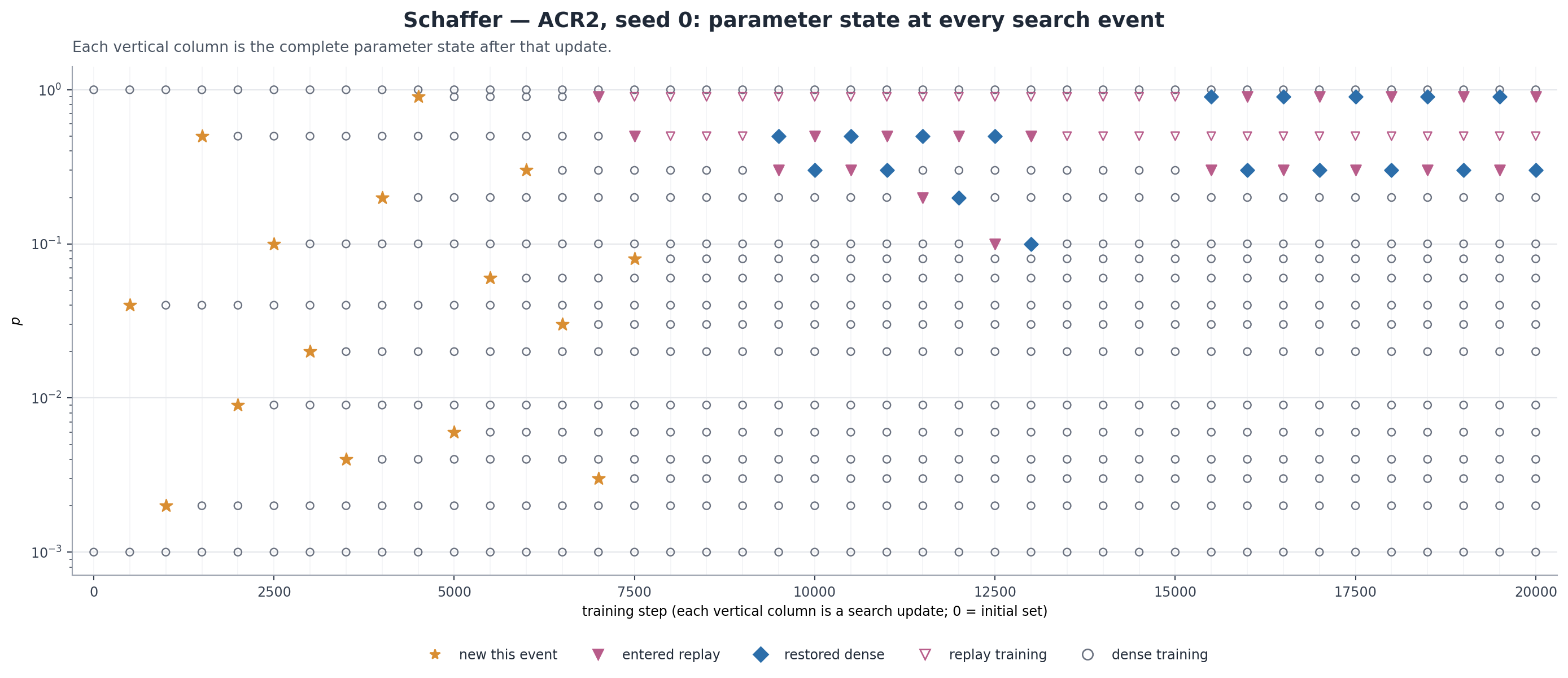}
\caption{\textbf{Parameter-state evolution of ACR2-arch in a representative Schaffer-like run, showing newly selected, densely trained, replayed, and state-transitioned tasks.}}
\label{fig:schaffer_search}
\end{figure*}

\begin{table*}[!tp]
\centering
\caption{\textbf{Test errors of the compared methods on the Schaffer-like benchmark.}}
\label{table:schaffer}
\small
\setlength{\tabcolsep}{3.5pt}
\renewcommand{\arraystretch}{1.10}
\begin{tabular}{llrrrr}
\toprule
Group & Method & $\mathrm{MSE}$ mean & $\mathrm{MSE}$ SD & $E_{L_2}$ mean & $E_{L_2}$ SD\\
\midrule
Baseline & UNI & $1.5352{\times}10^{-3}$ & $7.0547{\times}10^{-4}$ & $6.4784{\times}10^{-2}$ & $2.3278{\times}10^{-2}$\\
Baseline & FIX & $2.0351{\times}10^{-2}$ & $1.5944{\times}10^{-2}$ & $9.9870{\times}10^{-1}$ & $5.1190{\times}10^{-1}$\\
Baseline & AG & $5.0067{\times}10^{-4}$ & $1.0435{\times}10^{-4}$ & $1.2755{\times}10^{-1}$ & $1.5213{\times}10^{-2}$\\
CL-PINN & AC & $6.8660{\times}10^{-4}$ & $5.3210{\times}10^{-4}$ & $1.6010{\times}10^{-1}$ & $9.2480{\times}10^{-2}$\\
CL-PINN & ACR & $2.0392{\times}10^{-4}$ & $9.5861{\times}10^{-5}$ & $7.1950{\times}10^{-2}$ & $1.9640{\times}10^{-2}$\\
CL-PINN & ACR2-arch & $\mathbf{1.7490{\times}10^{-4}}$ & $7.0510{\times}10^{-5}$ & $\mathbf{5.0730{\times}10^{-2}}$ & $6.3910{\times}10^{-3}$\\
\bottomrule
\end{tabular}
\end{table*}

Table~\ref{table:schaffer} identifies different strongest baselines for the two metrics: AG gives
the lowest MSE, whereas UNI gives the lowest relative $L_2$ error. AG follows parameters with large
current training losses, but these losses evolve during optimization and are not equivalent to
amplitude-normalized solution errors. Its lower MSE therefore does not translate into a lower
relative error. FIX performs poorly on both metrics, indicating that its finite prescribed parameter
set does not cover the changing oscillation scales across $a$; despite not actively identifying
difficult parameters, UNI maintains better relative accuracy through more uniform coverage. AC
without replay also fails to exceed the strongest baselines and exhibits substantial between-run
variability. Under the aligned AC--ACR protocol, sparse replay reduces MSE and relative $L_2$ error
by 70.3\% and 55.0\%, respectively. ACR2-arch attains the lowest mean values, with 65.1\% lower MSE
than AG and 21.7\% lower relative $L_2$ error than UNI.

Figure~\ref{fig:schaffer_fields} shows denser, larger-amplitude concentric oscillations at small
$a$, with the main discrepancies concentrated along high-frequency rings. As $a$ increases, the
amplitude decreases, but relative error becomes more sensitive to the smaller solution magnitude.
Figure~\ref{fig:schaffer_err} quantifies this contrast: MSE is largest at the lower endpoint,
whereas the mean relative $L_2$ rises in the middle-to-high-$a$ region, where run-to-run variation
is also most pronounced. This opposite behavior of absolute and relative errors reflects the
cross-parameter scale disparity discussed above. Figure~\ref{fig:schaffer_search} shows that active
tasks cover the logarithmic parameter domain and transition between dense training and replay after
the capacity becomes active. The later ablation separates the endpoint-accuracy role of dynamic
weighting from the query-efficiency role of BO.

\subsubsection{Burgers equation}
\label{sec:burgers}

The Burgers equation \cite{bonkile2018systematic} is a standard nonlinear PDE in fluid dynamics and
a reduced model for studying the competition between nonlinear convection and viscous diffusion.
Eq.~\ref{eq:burgers} gives its one-dimensional viscous form, where $x$ and $t$ are the spatial and
temporal coordinates and $\nu$ is the viscosity parameter. The terms $u u_x$ and $\nu u_{xx}$
represent nonlinear convection and viscous diffusion, respectively.

\begin{equation}
\begin{cases}
u_t+u\,u_x=\dfrac{\nu}{\pi}u_{xx},
& (x,t)\in[-1,1]\times(0,1],\\
u(x,0)=-\sin(\pi x),
& x\in[-1,1],\\
u(-1,t)=u(1,t)=0,
& t\in[0,1].
\end{cases}
\label{eq:burgers}
\end{equation}

The spatial domain is $\Omega = \{(x,t) \mid (x,t) \in [-1,1] \times [0,1]\}$. The parameter domain
is $\Theta = \{\nu \mid \nu \in [0.01, 1.0] \}$. Each parameter uses 5000 interior points, 200
boundary points, 400 initial-condition points, and 300 near-shock anchors, for 5900 physical
samples; the initial-condition loss weight is 5. Sparse replay samples 10\% of the sparsifiable
interior-point pool, and $n_{\mathrm{param,active}}=n_{\mathrm{param,replay}}=9$. ACR uses sparse
replay in the main table, whereas ACR2-arch uses full-density replay at the same task capacity. The
protocol requests $2\times10^4$ Adam updates and at most $2\times10^4$ L-BFGS iterations. The
standard FNN has four 50-neuron hidden layers with three inputs and one output, while ACR2-arch uses
a parameter subnetwork. The current AC/ACR runs use $N_{\mathrm{resample}}=2000$, whereas ACR2-arch
uses 1000. Both the training weights and BO search scores use the uniform prior
$f_{\mathrm{prior}}(\nu)=1$, so the transformed BO score is identical to the unweighted physics
loss. The dynamic-weight parameters are $\lambda_{\mathrm{static}}=1$ and
$\lambda_{\mathrm{dynamic}}=-1$. Each active update uses 10 Bayesian evaluations with $\kappa=5$.
Errors are evaluated at 100 uniformly distributed parameter values using three independent random
seeds. Table~\ref{table:burgers} reports the results.

\begin{figure*}[!tp]
\centering
{\footnotesize\textbf{$\nu=0.01$}}\par\vspace{0.1em}
\includegraphics[width=\multifieldwidth]{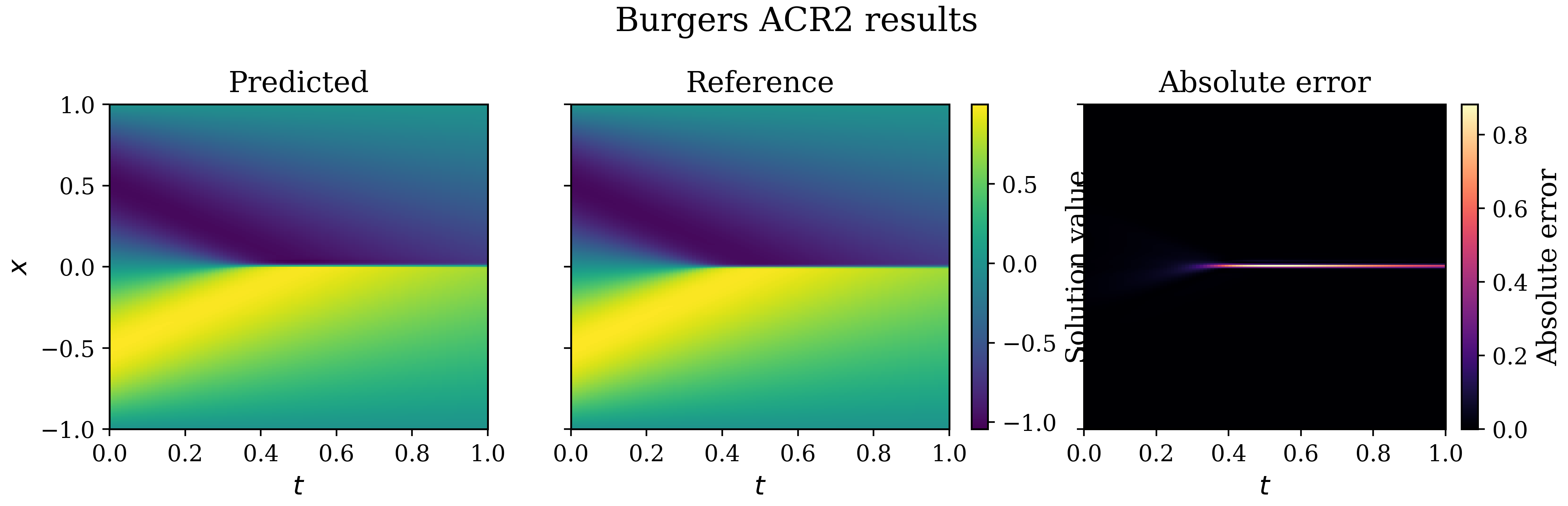}
\par\vspace{-0.35em}
{\footnotesize\textbf{$\nu=0.1$}}\par\vspace{0.1em}
\includegraphics[width=\multifieldwidth]{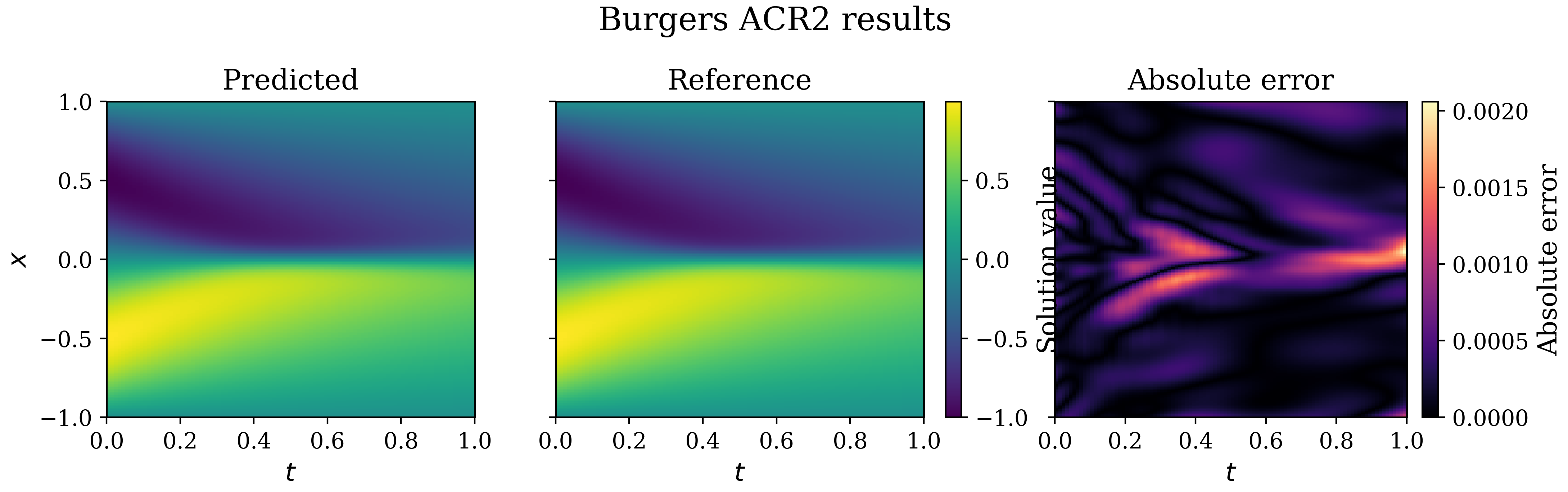}
\par\vspace{-0.35em}
{\footnotesize\textbf{$\nu=1.0$}}\par\vspace{0.1em}
\includegraphics[width=\multifieldwidth]{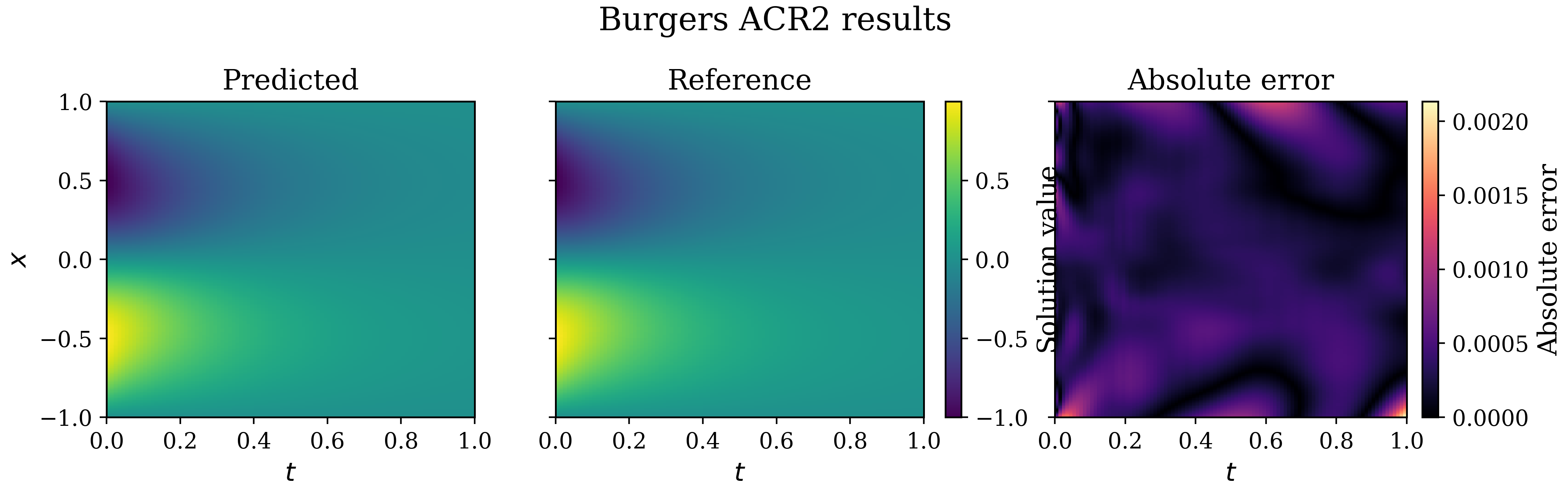}
\caption{\textbf{ACR2-arch prediction, reference, and absolute error for Burgers at $\nu=0.01$, $0.1$, and $1.0$, showing the transition from a sharp low-viscosity near-shock layer to smoother solutions.}}
\label{fig:burgers}
\end{figure*}

\begin{figure*}[!tp]
\centering
\includegraphics[width=\widebenchmarkwidth]{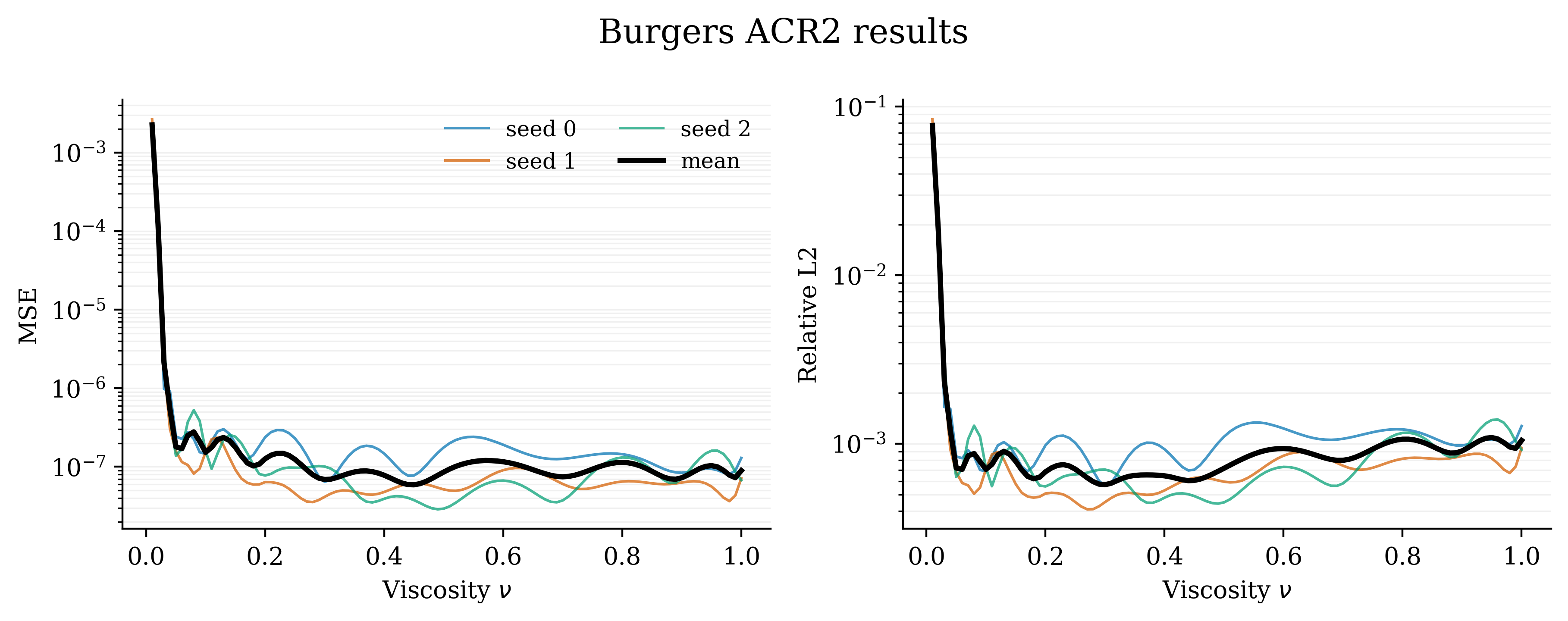}
\caption{\textbf{Parameter-wise MSE and $E_{L_2}$ of ACR2-arch over 100 Burgers test parameters.} Thin curves denote independent runs, the black curve their pointwise mean, and the vertical axes are logarithmic.}
\label{fig:burgers_err}
\end{figure*}

\begin{figure*}[!tp]
\centering
\includegraphics[width=\mechanismfigurewidth]{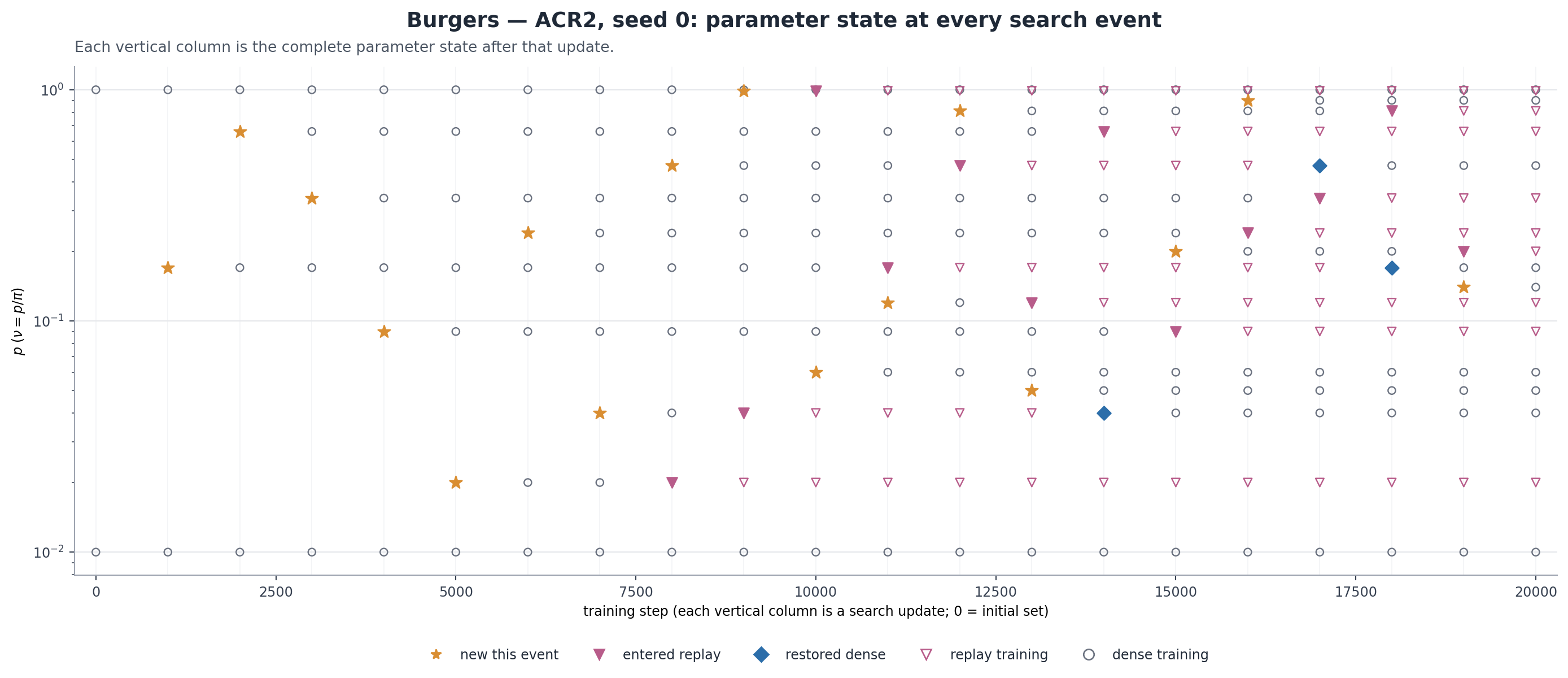}
\caption{\textbf{Parameter-state evolution of ACR2-arch in a representative Burgers run, showing newly selected, densely trained, replayed, and state-transitioned tasks.}}
\label{fig:burgers_search}
\end{figure*}

\begin{table*}[!tp]
\centering
\caption{\textbf{Test errors of the compared methods on the Burgers equation.}}
\label{table:burgers}
\small
\setlength{\tabcolsep}{3.5pt}
\renewcommand{\arraystretch}{1.10}
\begin{tabular}{llrrrr}
\toprule
Group & Method & $\mathrm{MSE}$ mean & $\mathrm{MSE}$ SD & $E_{L_2}$ mean & $E_{L_2}$ SD\\
\midrule
Baseline & UNI & $6.3431{\times}10^{-4}$ & $4.4293{\times}10^{-4}$ & $1.2153{\times}10^{-2}$ & $5.0850{\times}10^{-3}$\\
Baseline & FIX & $3.4909{\times}10^{-4}$ & $2.8515{\times}10^{-4}$ & $2.1256{\times}10^{-2}$ & $1.0633{\times}10^{-2}$\\
Baseline & AG & $4.1541{\times}10^{-4}$ & $3.0689{\times}10^{-4}$ & $1.4551{\times}10^{-2}$ & $6.2960{\times}10^{-3}$\\
CL-PINN & AC & $4.8920{\times}10^{-4}$ & $2.4840{\times}10^{-4}$ & $2.1230{\times}10^{-2}$ & $4.4080{\times}10^{-3}$\\
CL-PINN & ACR & $2.0548{\times}10^{-4}$ & $1.2219{\times}10^{-4}$ & $1.0250{\times}10^{-2}$ & $5.7900{\times}10^{-3}$\\
CL-PINN & ACR2-arch & $\mathbf{2.3900{\times}10^{-5}}$ & $3.2290{\times}10^{-6}$ & $\mathbf{1.7710{\times}10^{-3}}$ & $1.1460{\times}10^{-4}$\\
\bottomrule
\end{tabular}
\end{table*}

In Table~\ref{table:burgers}, FIX gives the lowest MSE among the three baselines, but its relative
$L_2$ error is close to that of AC and substantially higher than that of UNI; UNI instead gives the
lowest baseline relative $L_2$ error. Over this one-dimensional viscosity interval, uniform coverage
therefore maintains comparatively balanced accuracy near the low-viscosity endpoint, whereas
prescribed points or selections driven by instantaneous training loss do not necessarily improve
both absolute and normalized errors. AG lies between UNI and FIX on both metrics, further indicating
that greedy training loss is an imperfect proxy for the difficulty of the low-viscosity near-shock
regime. AC uses no replay and cannot retain dense constraints for all earlier tasks after the active
set reaches capacity; it does not exceed the strongest baselines here. Adding sparse replay in ACR
reduces MSE and relative $L_2$ error by 58.0\% and 51.7\% relative to AC. ACR2-arch further lowers
the relative $L_2$ error to $1.7710\times10^{-3}$.

Figure~\ref{fig:burgers} shows that the narrow central transition at $\nu=0.01$ contains most of the
visible error and broadens as viscosity increases. Figure~\ref{fig:burgers_err} confirms a sharp MSE
and relative-$L_2$ peak at the lowest-viscosity endpoint, followed by low and comparatively stable
mean errors over the remainder of the interval. The low-viscosity near-shock layer is therefore the
principal difficulty. In Figure~\ref{fig:burgers_search}, newly selected tasks span the viscosity
domain; replay tasks accumulate after the capacity becomes active, with some tasks later restored to
dense training. The method thus does not retain only one difficult region. For positive viscosity,
the solution remains continuous and should be described as a sharp transition rather than a true
discontinuity. Near-shock anchors improve physical-space resolution, while adaptive physical
collocation could further complement the proposed parameter-space allocation.

\subsubsection{Allen--Cahn equation}
\label{sec:allen}

The Allen--Cahn equation \cite{bartels2015allen} is a nonlinear PDE widely used in materials science
to describe phase transitions and interface motion through the interaction of diffusion and a
bistable reaction. Eq.~\ref{eq:allen} gives the reaction--diffusion form, where $\nu$ controls
diffusion and $\rho$ scales the cubic bistable reaction.

\begin{equation}
\begin{cases}
u_t=\nu u_{xx}+\rho(u-u^3),
& \begin{gathered}
x\in[-1,1],\\
t\in(0,1]
\end{gathered},\\
u(x,0)=x^2\cos(\pi x),
& x\in[-1,1],\\
u(-1,t)=u(1,t)=-1,
& t\in[0,1].
\end{cases}
\label{eq:allen}
\end{equation}

The spatial domain is $\Omega = \{(x,t) \mid (x,t) \in [-1,1] \times [0,1]\}$. The parameter domain
is $\Theta = \{(\nu,\rho) \mid (\nu,\rho) \in [0.001, 0.1] \times [3, 5] \}$. A hard constraint
enforces both the initial condition and the consistent boundary value $u(\pm1,t)=-1$. Each parameter
uses 10082 physical samples, and sparse replay samples 10\% of the sparsifiable physics-point pool.
The maximum numbers of active and replay parameter tasks are both 9. The protocol requests
$2\times10^4$ Adam updates and at most $1\times10^4$ L-BFGS iterations. The standard FNN has four
50-neuron hidden layers with four inputs and one output, while ACR2-arch uses a parameter
subnetwork. AC uses $N_{\mathrm{resample}}=2000$, whereas ACR and ACR2-arch use 1000.

The dynamic training weights use $f_{\mathrm{prior}}(\nu,\rho)=\!\rho^3\exp[-2\log_{10}(\nu)]$,
$\lambda_{\mathrm{static}}=2$, and $\lambda_{\mathrm{dynamic}}=-2$. Bayesian search operates on
physics loss without this prior weighting, using 20 evaluations per active update and $\kappa=5$.
For each method, errors are evaluated at 95 parameter points---19 logarithmically distributed values
of $\nu$ and five uniformly distributed values of $\rho$---using three independent random seeds.
Table~\ref{table:allen} reports the results.

\begin{figure*}[!tp]
\centering
{\footnotesize\textbf{$(\nu,\rho)=(0.001,3)$}}\par\vspace{0.1em}
\includegraphics[width=\multifieldwidth]{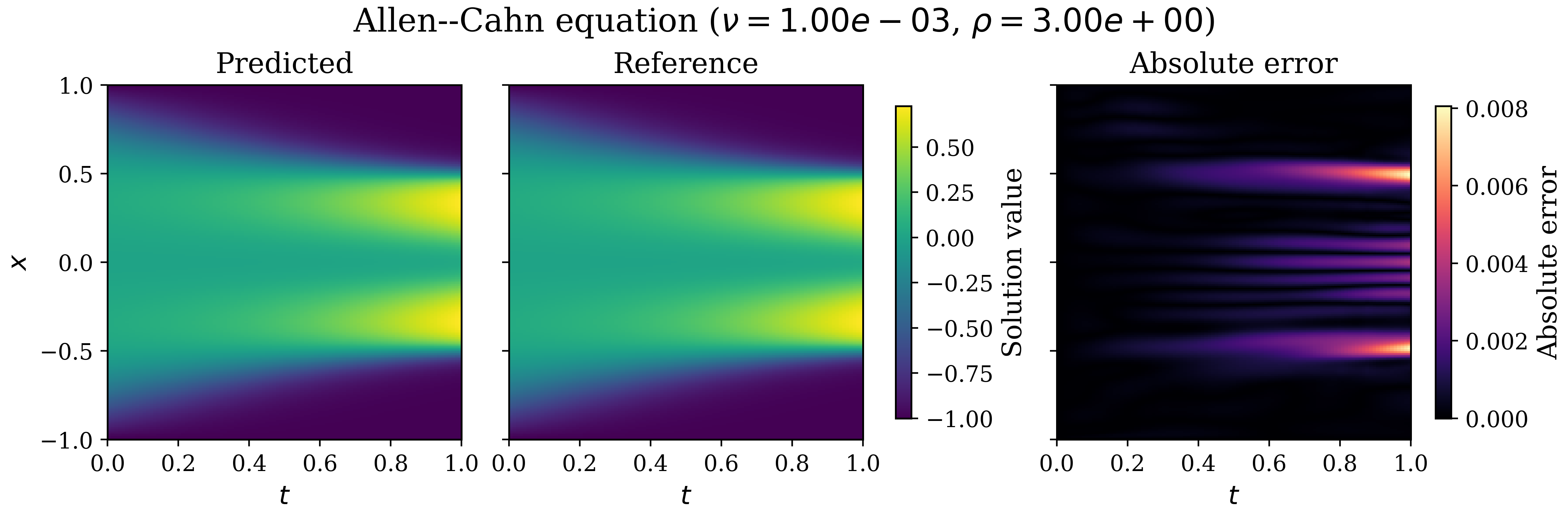}
\par\vspace{-0.35em}
{\footnotesize\textbf{$(\nu,\rho)=(0.01,4)$}}\par\vspace{0.1em}
\includegraphics[width=\multifieldwidth]{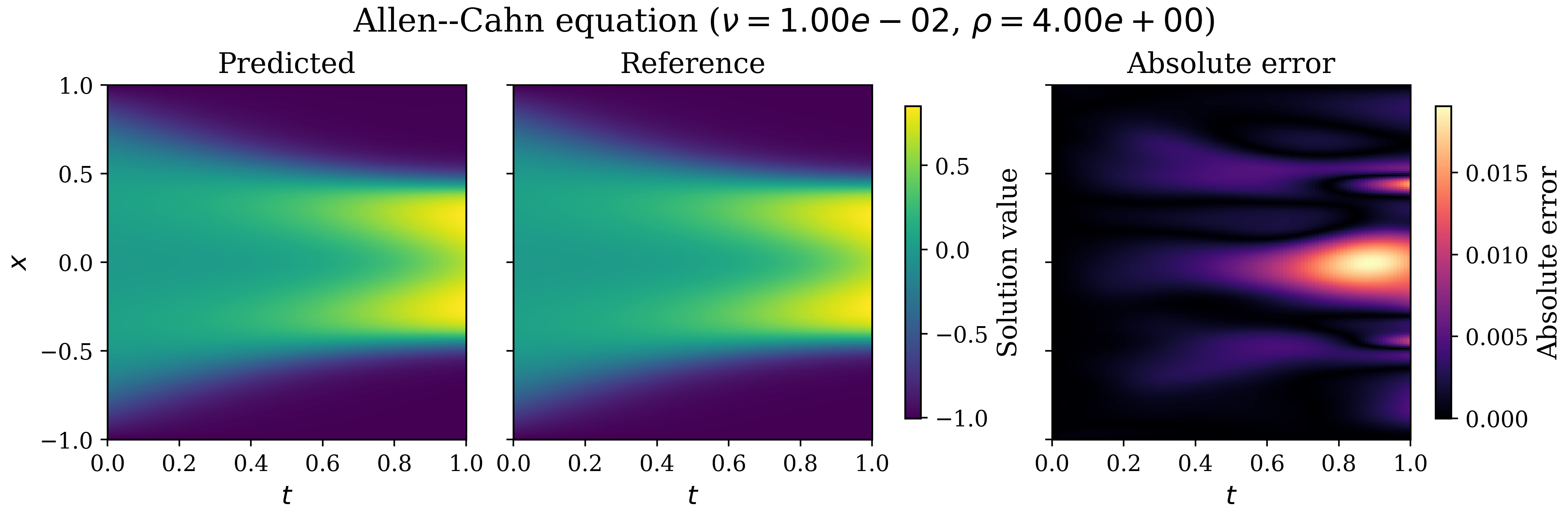}
\par\vspace{-0.35em}
{\footnotesize\textbf{$(\nu,\rho)=(0.1,5)$}}\par\vspace{0.1em}
\includegraphics[width=\multifieldwidth]{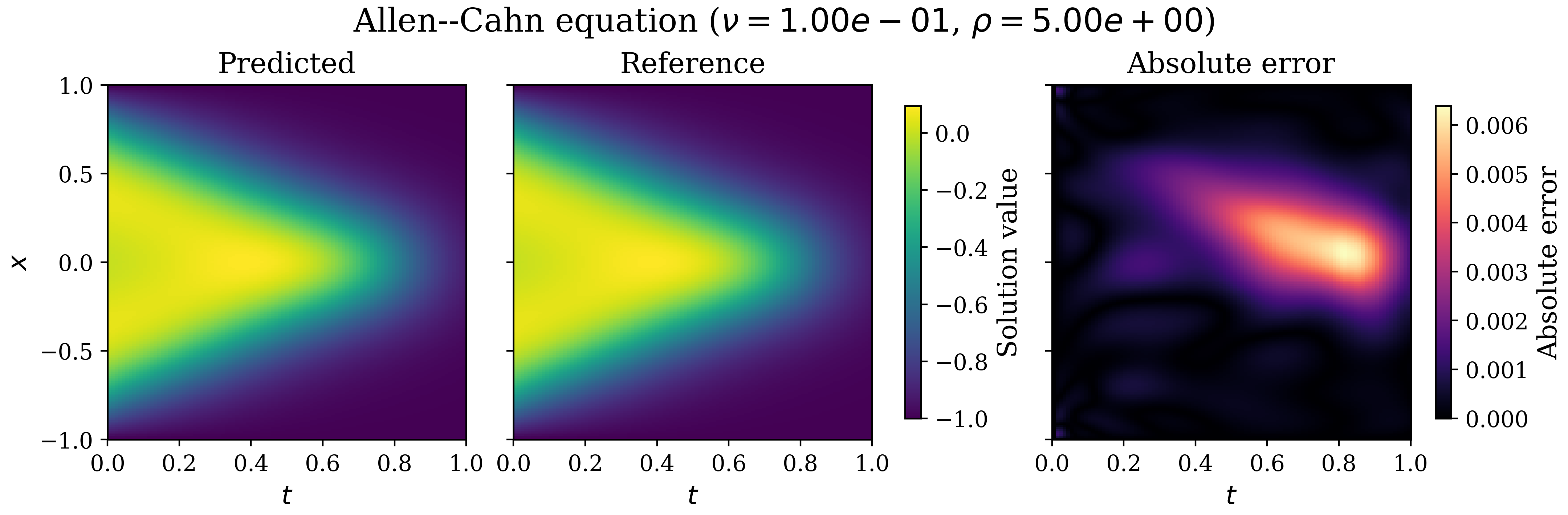}
\caption{\textbf{ACR prediction, reference, and absolute error for Allen--Cahn at $(\nu,\rho)=(0.001,3)$, $(0.01,4)$, and $(0.1,5)$, showing distinct diffusion--reaction regimes across the two-dimensional parameter domain.}}
\label{fig:allen}
\end{figure*}

\begin{figure*}[!tp]
\centering
\includegraphics[width=\wideconditionwidth]{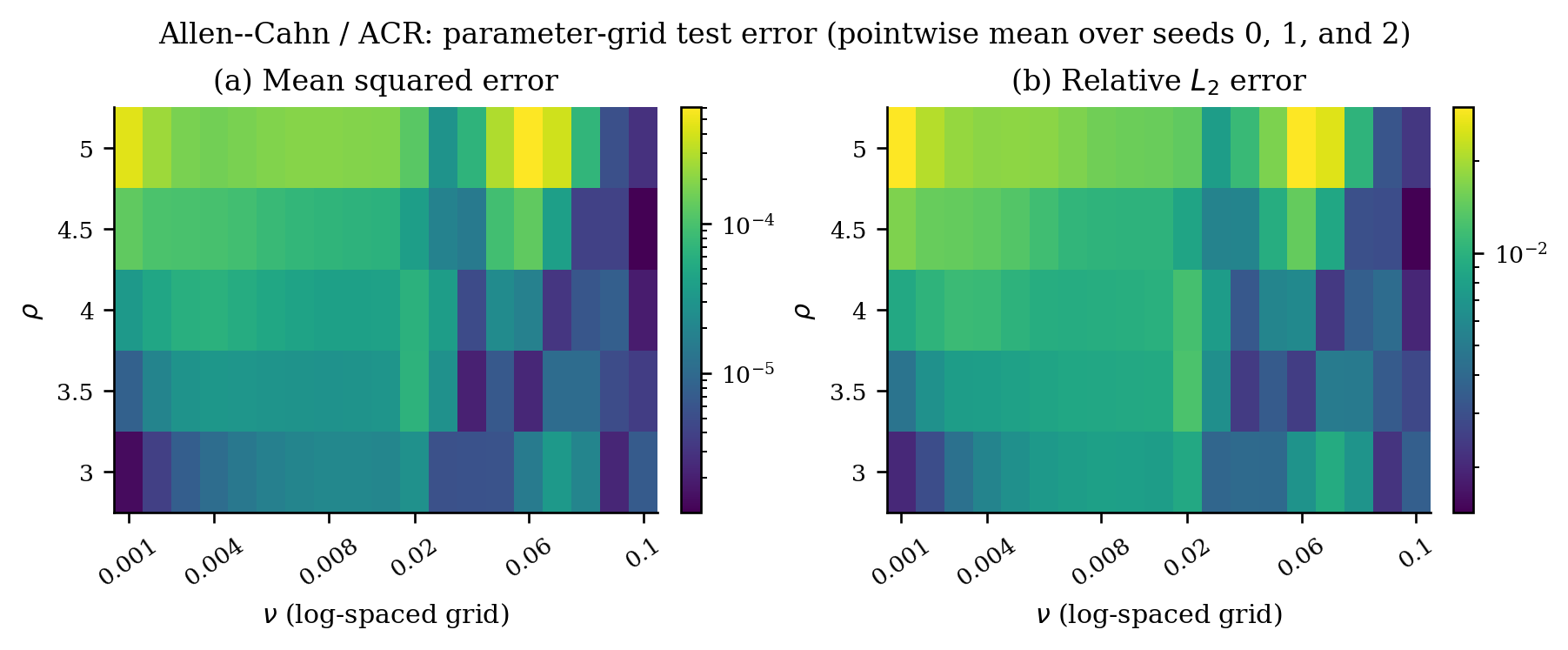}
\caption{\textbf{Mean ACR test errors over the Allen--Cahn parameter domain.} The heatmaps show MSE and $E_{L_2}$, averaged pointwise over independent runs on logarithmic color scales.}
\label{fig:allen_err}
\end{figure*}

\begin{figure*}[!tp]
\centering
\includegraphics[width=\mechanismfigurewidth]{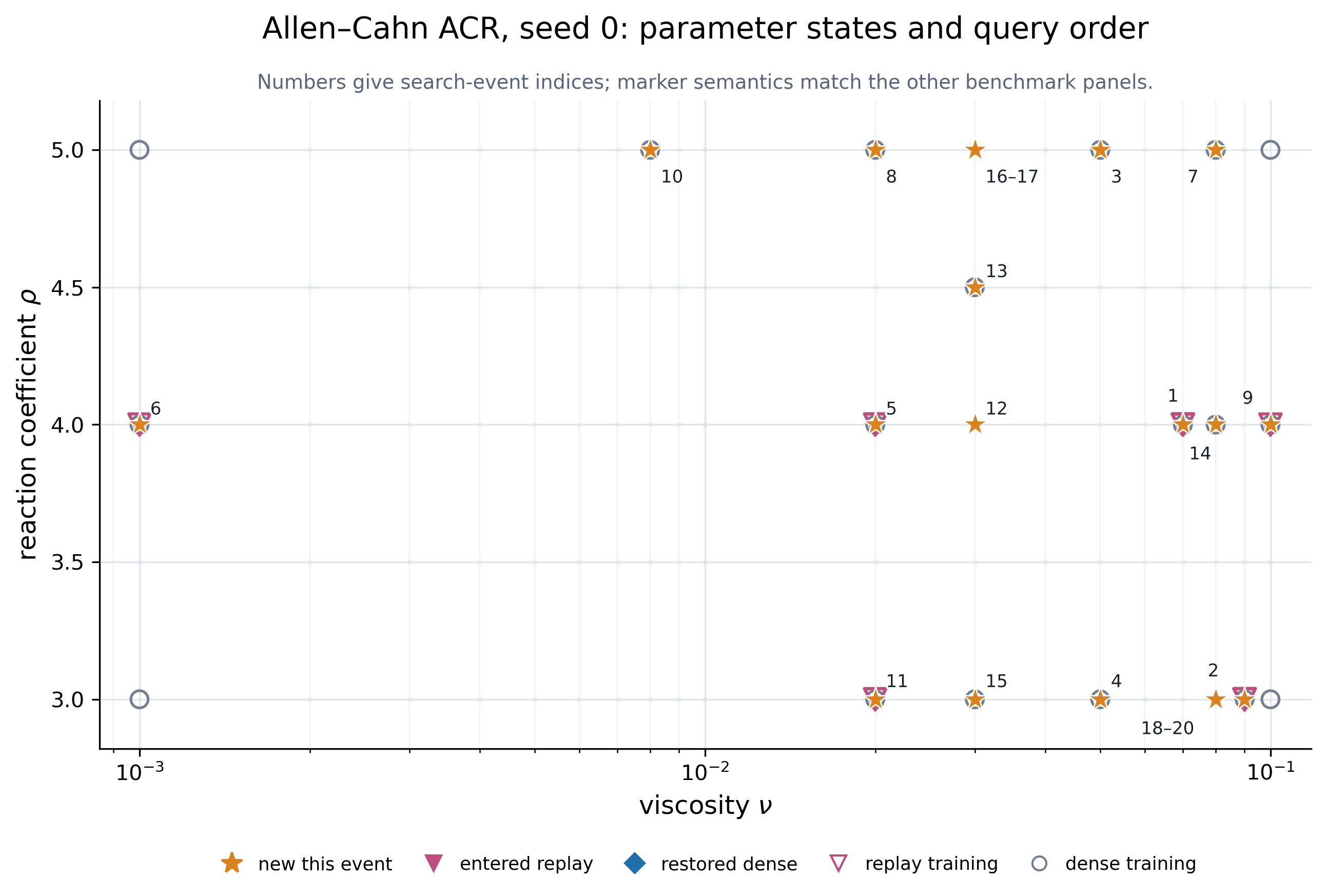}
\caption{\textbf{Parameter selection and state evolution of ACR in a representative Allen--Cahn run.} Labels give the search-event order, and markers distinguish newly selected, densely trained, and replayed tasks.}
\label{fig:allen_search}
\end{figure*}

\begin{table*}[!tp]
\centering
\caption{\textbf{Test errors of the compared methods on the Allen--Cahn equation.}}
\label{table:allen}
\small
\setlength{\tabcolsep}{3.5pt}
\renewcommand{\arraystretch}{1.10}
\begin{tabular}{llrrrr}
\toprule
Group & Method & $\mathrm{MSE}$ mean & $\mathrm{MSE}$ SD & $E_{L_2}$ mean & $E_{L_2}$ SD\\
\midrule
Baseline & UNI & $9.8805{\times}10^{-3}$ & $8.2280{\times}10^{-3}$ & $1.1619{\times}10^{-1}$ & $5.2703{\times}10^{-2}$\\
Baseline & FIX & $2.6184{\times}10^{-2}$ & $1.1203{\times}10^{-2}$ & $1.6286{\times}10^{-1}$ & $2.8756{\times}10^{-2}$\\
Baseline & AG & $1.9173{\times}10^{-3}$ & $9.5815{\times}10^{-4}$ & $5.0840{\times}10^{-2}$ & $9.3830{\times}10^{-3}$\\
CL-PINN & AC & $1.4140{\times}10^{-3}$ & $1.1410{\times}10^{-3}$ & $4.6000{\times}10^{-2}$ & $1.8330{\times}10^{-2}$\\
CL-PINN & ACR & $\mathbf{1.3280{\times}10^{-4}}$ & $3.8240{\times}10^{-5}$ & $\mathbf{1.4670{\times}10^{-2}}$ & $1.8270{\times}10^{-3}$\\
CL-PINN & ACR2-arch & $4.8440{\times}10^{-4}$ & $1.7510{\times}10^{-4}$ & $2.4390{\times}10^{-2}$ & $4.7880{\times}10^{-3}$\\
\bottomrule
\end{tabular}
\end{table*}

Table~\ref{table:allen} shows that AG is the most accurate of the three baselines and substantially
outperforms UNI and FIX, indicating that the difficult regions of this two-dimensional parameter
domain are not well represented by a small prescribed task set. AC has slightly lower mean errors
than AG but greater between-run variation. ACR attains the lowest MSE and relative $L_2$ error,
reducing them by 90.6\% and 68.1\%, respectively, relative to AC. Although ACR2-arch remains more
accurate than all three baselines and AC, it is weaker than ACR. One possible explanation is that
the parameter subnetwork and its training configuration use prespecified common hyperparameters that
may not be fully matched to the parameter scales and optimization dynamics of Allen--Cahn. In
addition, small diffusion and a strongly nonlinear reaction produce sharp phase interfaces and a
difficult PINN loss landscape; changing the parameter representation alone cannot eliminate this
optimization difficulty. The result therefore shows that the benefit of the parameter subnetwork
depends on both the equation and the optimization configuration, rather than implying that a more
elaborate parameter representation must always reduce error.

Figure~\ref{fig:allen} shows sharper phase interfaces at small $\nu$, whereas increasing $\rho$
strengthens the nonlinear reaction; the three cases therefore exhibit distinct diffusion--reaction
balances. Figure~\ref{fig:allen_err} shows that the error is not monotone in either parameter and is
instead governed jointly by $\nu$ and $\rho$. In the representative ACR run shown in
Figure~\ref{fig:allen_search}, tasks continue to be selected along both $\rho=3$ and $\rho=5$; most
new tasks lie in the middle-to-high-$\nu$ range, while several low-$\nu$ tasks are retained. Active
selection thus allocates training across multiple diffusion scales and reaction strengths rather
than tracking a single boundary.

\subsubsection{Kovasznay flow}
\label{sec:kovasznay}

Kovasznay flow \cite{kovasznay1948laminar} is a classical two-dimensional incompressible steady flow
with an analytical Navier--Stokes solution. It is commonly used to assess the accuracy and stability
of numerical methods.

The Navier--Stokes (NS) equations \cite{temam2024navier} govern fluid motion. The incompressible
velocity--pressure form in Eq.~\ref{eq:ns} includes both inertial and viscous effects. The Reynolds
number $Re$, defined as the ratio of inertial to viscous forces, is a key dimensionless flow
parameter. The strong nonlinearity of the NS equations makes their numerical solution
computationally demanding.

\begin{equation}
\begin{aligned}
\frac{\partial\mathbf{u}}{\partial t}
+(\mathbf{u}\cdot\nabla)\mathbf{u}
&=-\nabla p+\frac{1}{Re}\nabla^2\mathbf{u},\\
\nabla\cdot\mathbf{u}&=0.
\end{aligned}
\label{eq:ns}
\end{equation}

The Kovasznay solution used here is given in Eq.~\ref{eq:kov}. It is a two-dimensional
incompressible steady flow with exponential spatial decay and analytical velocity and pressure
fields. The only varying physical parameter is the Reynolds number $Re$.

\begin{equation}
\begin{aligned}
u(x,y)&=1-e^{\zeta x}\cos(2\pi y),\\
v(x,y)&=\frac{\zeta}{2\pi}e^{\zeta x}\sin(2\pi y),\\
p(x,y)&=\frac{1}{2}\left(1-e^{2\zeta x}\right),\\
\zeta&=\frac{1}{2\nu}-\sqrt{\frac{1}{4\nu^2}+4\pi^2},\\
\nu&=\frac{1}{Re}.
\end{aligned}
\label{eq:kov}
\end{equation}

The spatial domain is $\Omega = \{(x,y) \mid (x,y) \in [-0.5,1] \times [-0.5,1.5]\}$. The parameter
domain is $\Theta = \{Re \mid Re \in [5,500]\}$. Each parameter uses 3474 physical samples across
the interior and boundary sets. Sparse replay samples 10\% of the sparsifiable physics-point pool,
and the maximum numbers of active and replay parameter tasks are both 10. The protocol requests
$1\times10^4$ Adam updates and at most $1\times10^4$ L-BFGS iterations. The standard FNN has four
50-neuron hidden layers with three inputs and three outputs, while ACR2-arch uses a parameter
subnetwork. AC uses $N_{\mathrm{resample}}=1000$, whereas ACR and ACR2-arch use 500.

The dynamic training weights use $f_{\mathrm{prior}}(Re)=\log_{10}(Re)$,
$\lambda_{\mathrm{static}}=1$, and $\lambda_{\mathrm{dynamic}}=-1$. Bayesian search operates on
physics loss without this prior weighting, using 15 evaluations per active update and $\kappa=5$.
Each method is evaluated with three independent random seeds at 100 parameter values uniformly
distributed over $\Theta$; Table~\ref{table:kov} reports the results.

\begin{figure*}[!tp]
\centering
{\footnotesize\textbf{$u$}}\par\vspace{0.1em}
\includegraphics[width=\kovasfieldwidth]{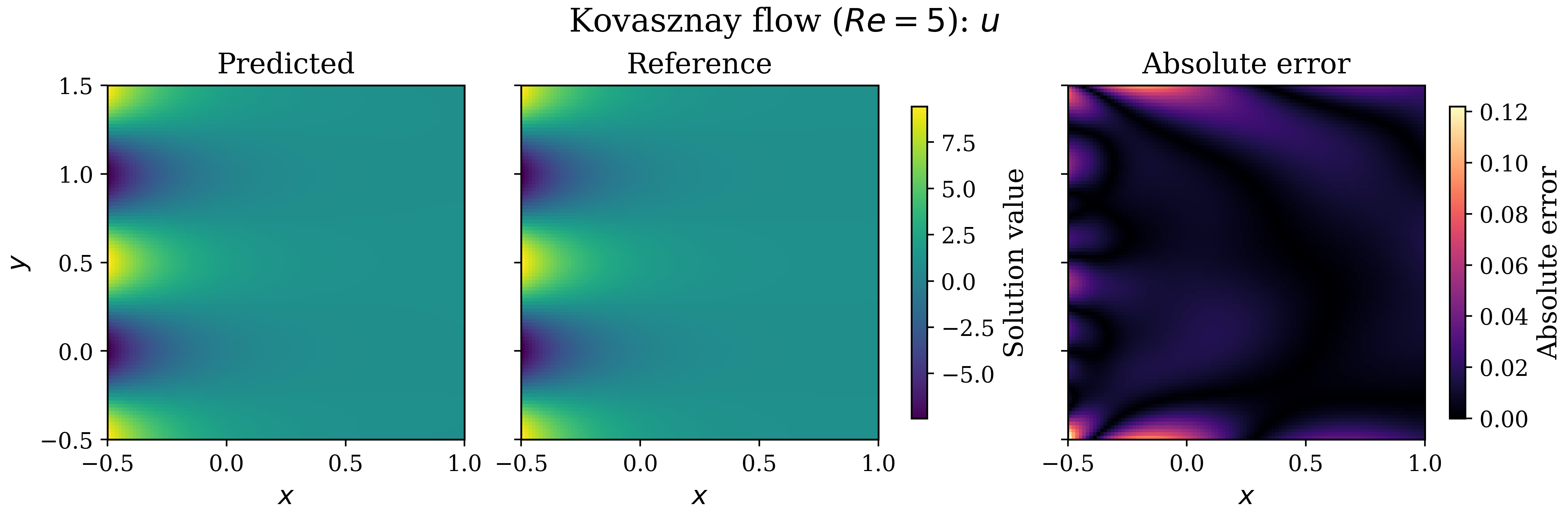}
\par\vspace{-0.35em}
{\footnotesize\textbf{$v$}}\par\vspace{0.1em}
\includegraphics[width=\kovasfieldwidth]{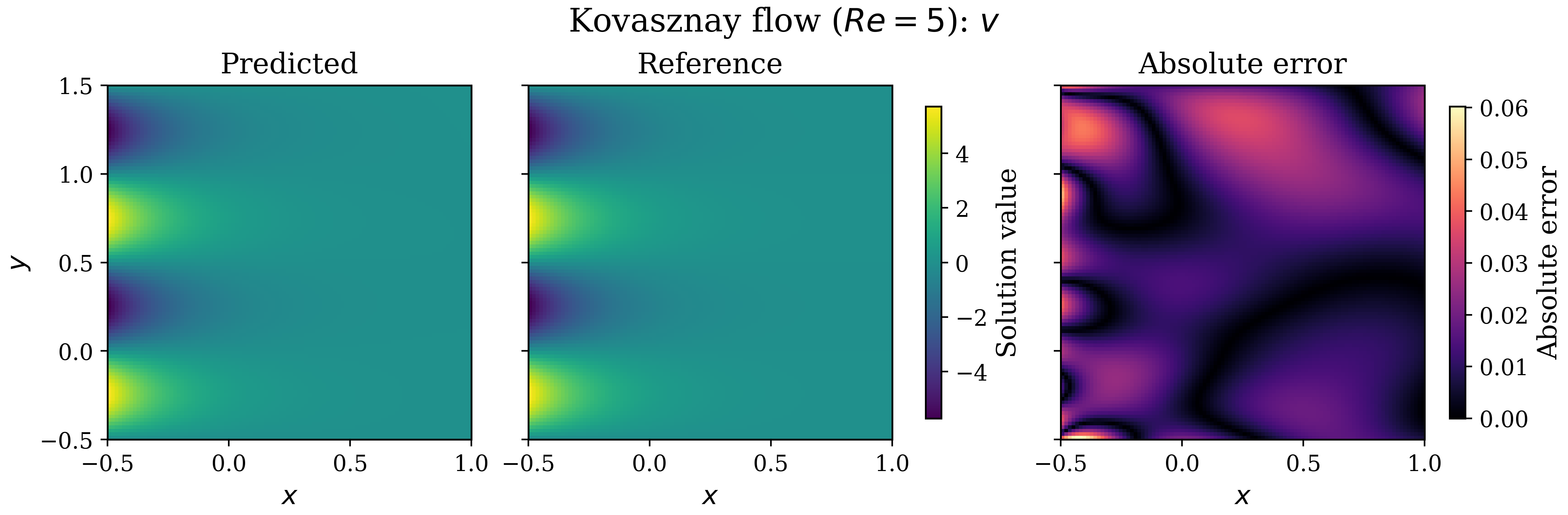}
\par\vspace{-0.35em}
{\footnotesize\textbf{$p$}}\par\vspace{0.1em}
\includegraphics[width=\kovasfieldwidth]{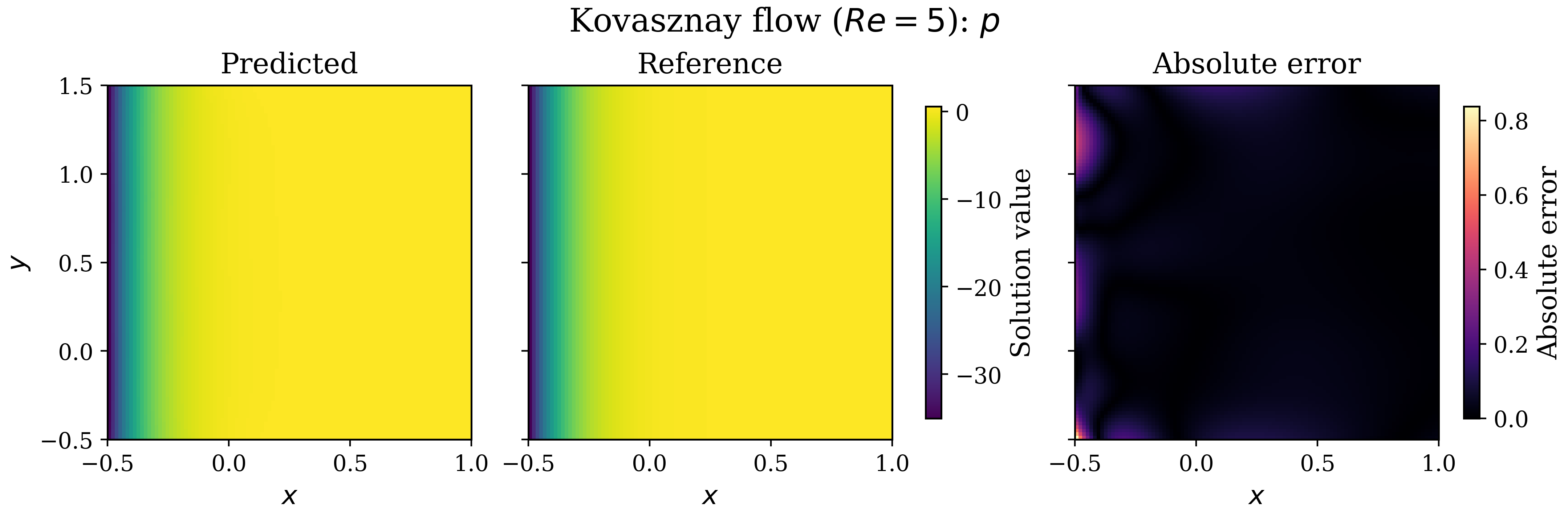}
\caption{\textbf{ACR2-arch prediction, analytical reference, and absolute error for Kovasznay $u$, $v$, and $p$ at $Re=5$.} Component-wise color scales account for their different magnitudes.}
\label{fig:kov}
\end{figure*}

\begin{figure*}[!tp]
\centering
{\footnotesize\textbf{$u$}}\par\vspace{0.1em}
\includegraphics[width=\kovasfieldwidth]{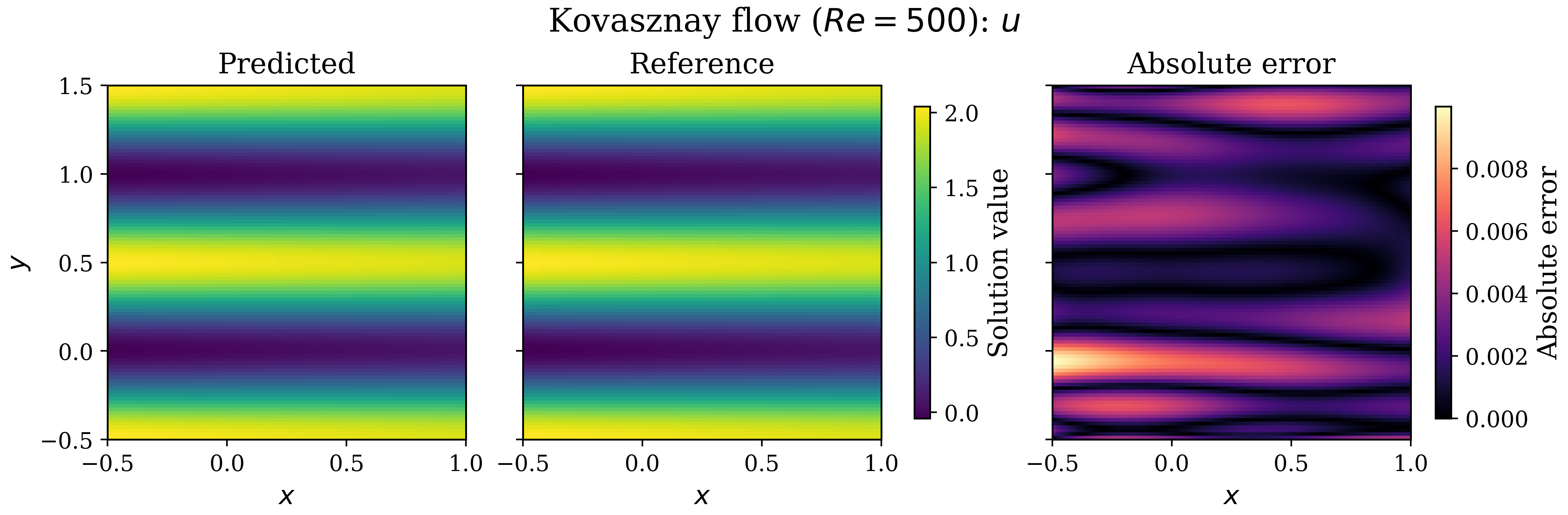}
\par\vspace{-0.35em}
{\footnotesize\textbf{$v$}}\par\vspace{0.1em}
\includegraphics[width=\kovasfieldwidth]{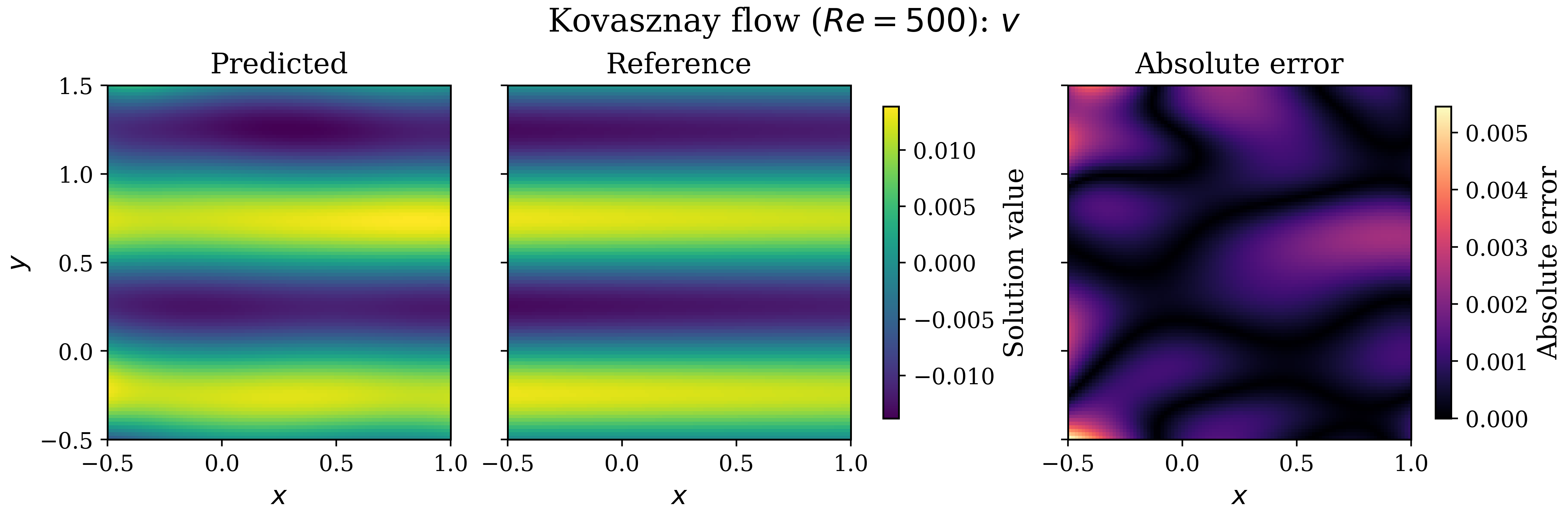}
\par\vspace{-0.35em}
{\footnotesize\textbf{$p$}}\par\vspace{0.1em}
\includegraphics[width=\kovasfieldwidth]{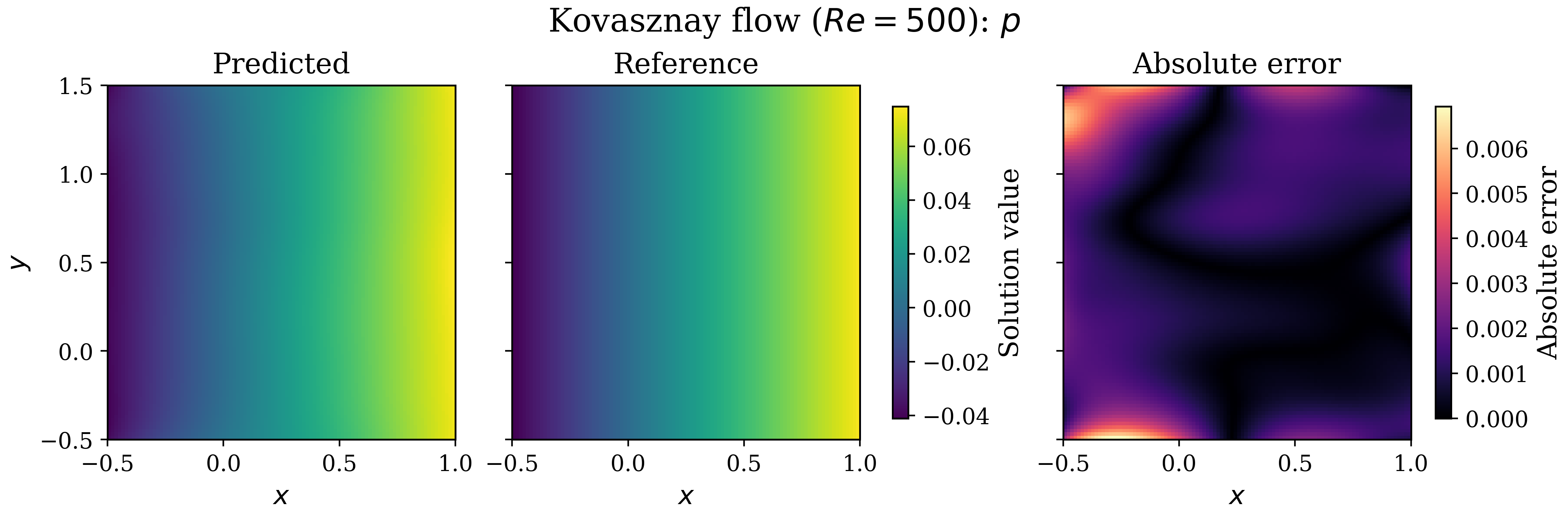}
\caption{\textbf{ACR2-arch prediction, analytical reference, and absolute error for Kovasznay $u$, $v$, and $p$ at $Re=500$.}}
\label{fig:kov_highre}
\end{figure*}

\begin{figure*}[!tp]
\centering
{\footnotesize\textbf{$Re=5$}}\par\vspace{0.1em}
\includegraphics[width=\kovasfieldwidth]{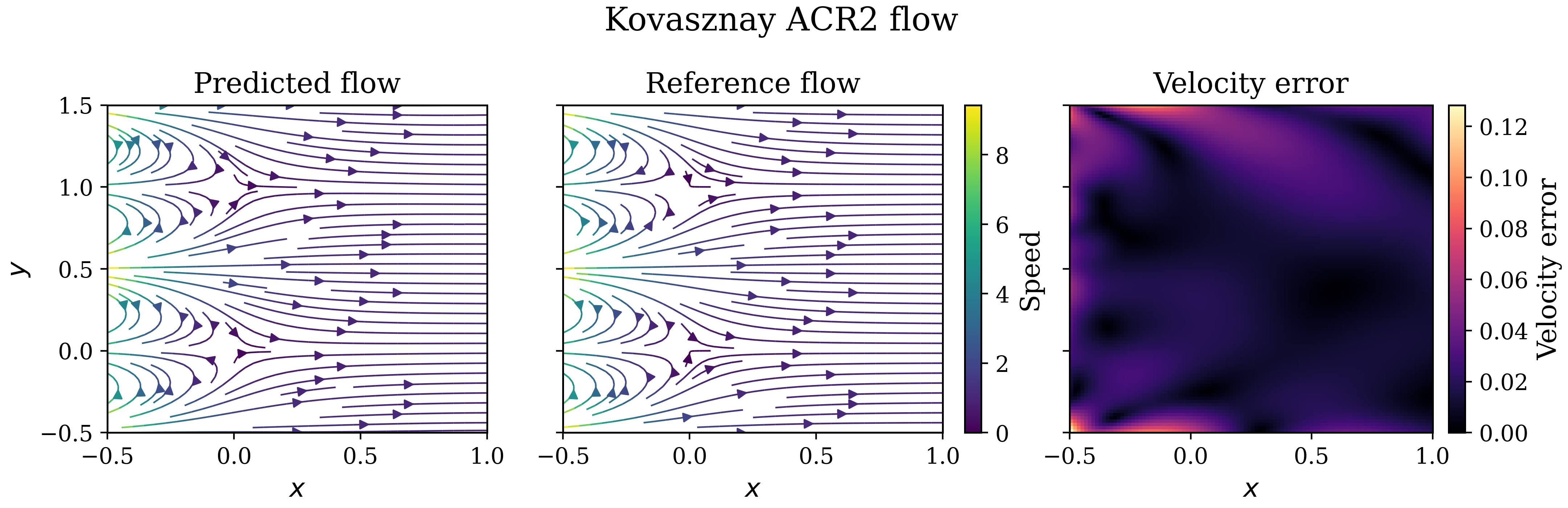}
\par\vspace{-0.35em}
{\footnotesize\textbf{$Re=500$}}\par\vspace{0.1em}
\includegraphics[width=\kovasfieldwidth]{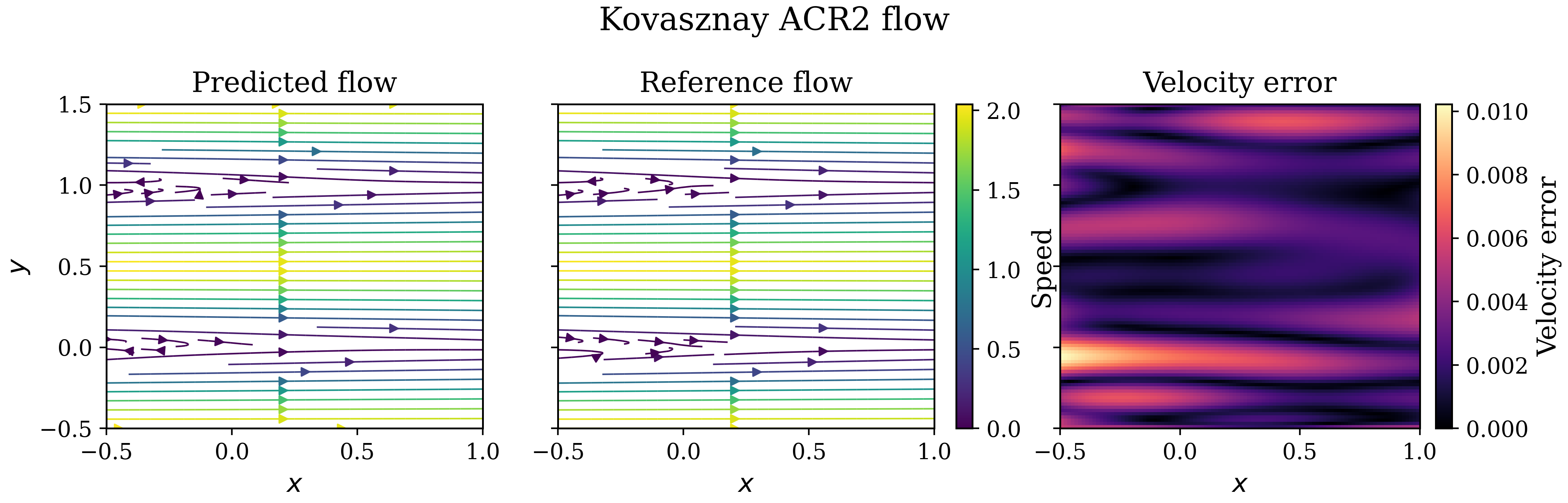}
\caption{\textbf{Predicted Kovasznay velocity magnitude and streamlines at $Re=5$ and $Re=500$, showing the overall flow structure beyond the component fields.}}
\label{fig:kov_streamline}
\end{figure*}

\begin{figure*}[!tp]
\centering
\includegraphics[width=\mechanismfigurewidth]{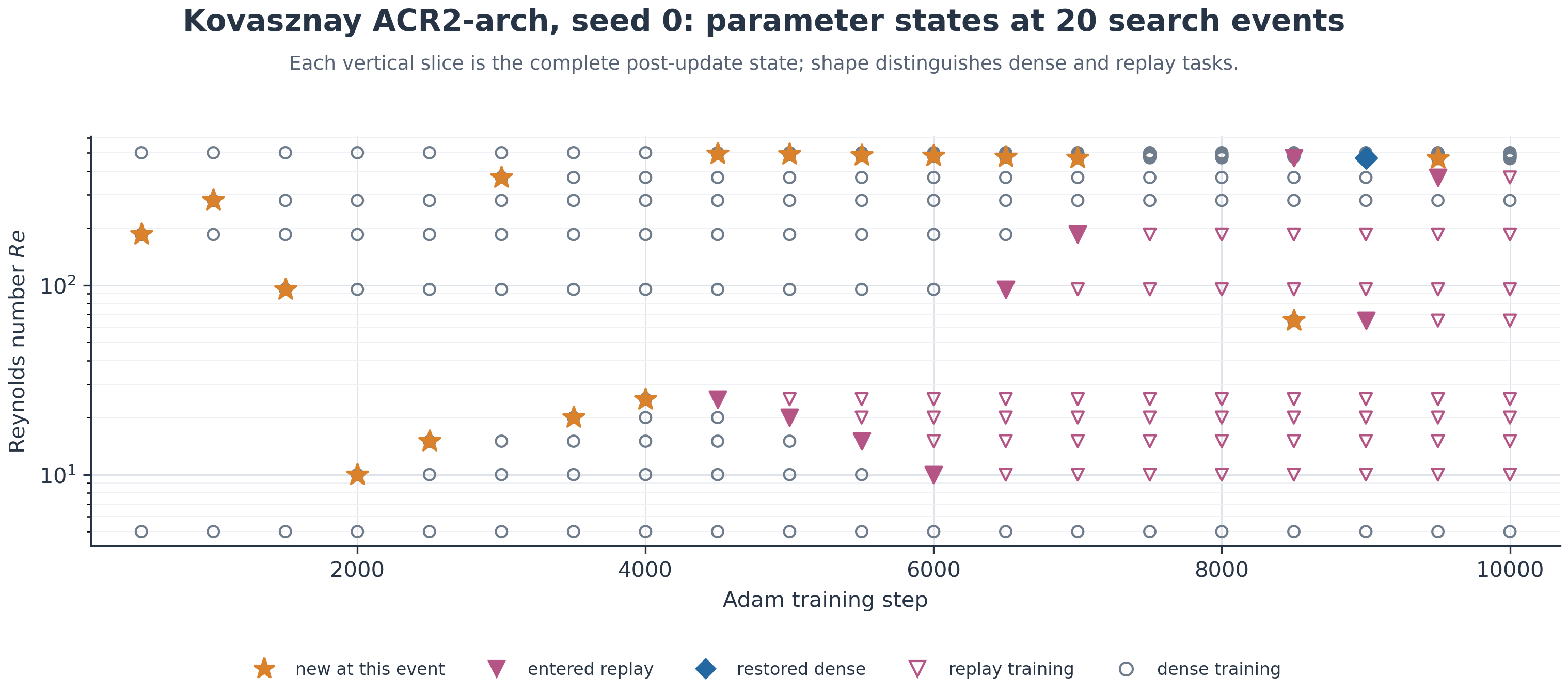}
\caption{\textbf{Reynolds-number selection and parameter-state evolution of ACR2-arch in a representative Kovasznay run, distinguishing newly selected, densely trained, and replayed tasks.}}
\label{fig:kov_search}
\end{figure*}

\begin{figure*}[!tp]
\centering
\includegraphics[width=\widebenchmarkwidth]{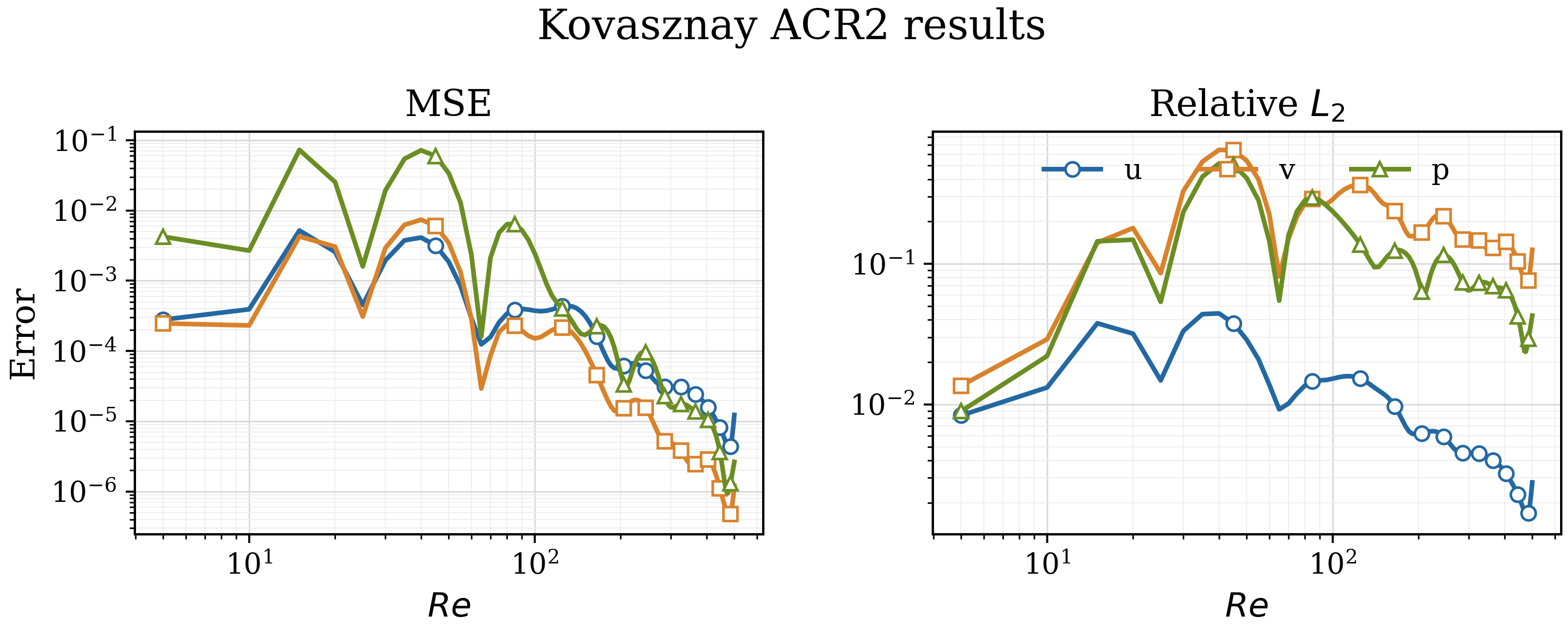}
\caption{\textbf{Mean component-wise MSE and $E_{L_2}$ of ACR2-arch over 100 Reynolds numbers, averaged pointwise over independent runs.}}
\label{fig:kov_err}
\end{figure*}

\begin{table*}[!tp]
\centering
\caption{\textbf{Component-wise Kovasznay MSE for the compared methods.}}
\label{table:kov}
\small
\setlength{\tabcolsep}{3.5pt}
\renewcommand{\arraystretch}{1.10}
\begin{adjustbox}{max width=\textwidth,center}
\begin{tabular}{lrrr}
\toprule
Method & $\mathrm{MSE}(u)$ & $\mathrm{MSE}(v)$ & $\mathrm{MSE}(p)$\\
\midrule
UNI & $6.288\times10^{-3}\pm6.373\times10^{-4}$ & $6.223\times10^{-3}\pm5.408\times10^{-4}$ & $1.165\times10^{-1}\pm2.228\times10^{-2}$\\
FIX & $5.798\times10^{-4}\pm5.664\times10^{-4}$ & $1.358\times10^{-4}\pm4.368\times10^{-5}$ & $\mathbf{8.551\times10^{-4}}\pm7.688\times10^{-4}$\\
AG & $4.074\times10^{-3}\pm3.725\times10^{-3}$ & $1.111\times10^{-2}\pm1.478\times10^{-2}$ & $1.882\times10^{-1}\pm2.504\times10^{-1}$\\
AC & $3.754\times10^{-2}\pm2.565\times10^{-2}$ & $3.445\times10^{-2}\pm4.710\times10^{-2}$ & $2.956\times10^{-1}\pm4.064\times10^{-1}$\\
ACR & $7.523\times10^{-4}\pm4.852\times10^{-4}$ & $\mathbf{1.266\times10^{-4}}\pm8.922\times10^{-5}$ & $1.373\times10^{-3}\pm9.927\times10^{-4}$\\
ACR2-arch & $\mathbf{3.408\times10^{-4}}\pm2.594\times10^{-4}$ & $3.950\times10^{-4}\pm4.384\times10^{-4}$ & $4.017\times10^{-3}\pm5.263\times10^{-3}$\\
\bottomrule
\end{tabular}
\end{adjustbox}
\end{table*}

\begin{table*}[!tp]
\centering
\caption{\textbf{Component-wise Kovasznay relative $L_2$ for the compared methods.}}
\label{table:kov_l2}
\small
\setlength{\tabcolsep}{3.5pt}
\renewcommand{\arraystretch}{1.10}
\begin{adjustbox}{max width=\textwidth,center}
\begin{tabular}{lrrr}
\toprule
Method & $E_{L_2}(u)$ & $E_{L_2}(v)$ & $E_{L_2}(p)$\\
\midrule
UNI & $1.810\times10^{-2}\pm1.246\times10^{-3}$ & $1.534\times10^{-1}\pm1.352\times10^{-2}$ & $9.144\times10^{-2}\pm1.228\times10^{-2}$\\
FIX & $1.233\times10^{-2}\pm3.558\times10^{-3}$ & $5.080\times10^{-1}\pm1.476\times10^{-1}$ & $2.574\times10^{-1}\pm1.698\times10^{-1}$\\
AG & $2.477\times10^{-2}\pm1.074\times10^{-2}$ & $5.345\times10^{-1}\pm4.721\times10^{-1}$ & $5.421\times10^{-1}\pm5.444\times10^{-1}$\\
AC & $5.250\times10^{-2}\pm1.826\times10^{-2}$ & $6.523\times10^{-1}\pm4.070\times10^{-1}$ & $3.873\times10^{-1}\pm2.600\times10^{-1}$\\
ACR & $1.113\times10^{-2}\pm4.656\times10^{-3}$ & $\mathbf{1.380\times10^{-1}}\pm6.591\times10^{-2}$ & $\mathbf{8.864\times10^{-2}}\pm6.145\times10^{-2}$\\
ACR2-arch & $\mathbf{8.956\times10^{-3}}\pm1.606\times10^{-3}$ & $1.946\times10^{-1}\pm2.964\times10^{-2}$ & $1.121\times10^{-1}\pm4.236\times10^{-2}$\\
\bottomrule
\end{tabular}
\end{adjustbox}
\end{table*}

Tables~\ref{table:kov} and~\ref{table:kov_l2} show strong component dependence. The amplitude of $u$
is comparatively large and dominates the velocity field; ACR2-arch gives the lowest MSE and relative
$L_2$ error for $u$, indicating that the separate parameter branch improves the cross-parameter
representation of this dominant velocity component. By contrast, ACR gives the lowest $v$ MSE and
the lowest relative $L_2$ errors for $v$ and $p$, and is therefore better balanced across the three
components. FIX gives the lowest $p$ MSE, but its relative $L_2$ error for $p$ is substantially
higher than that of ACR. Because the amplitude of $p$ is small and varies with $Re$, even a small
absolute discrepancy can produce a large relative error at some parameters. AG and AC without replay
also exhibit large errors and variability for several components, showing that parameter selection
based on an aggregate task loss still requires careful balancing between large- and small-amplitude
outputs.

Figures~\ref{fig:kov} and~\ref{fig:kov_highre} show the component fields at the two endpoints of the
Reynolds-number interval, while Figure~\ref{fig:kov_streamline} confirms that the corresponding
overall streamline structures are retained. As $Re$ increases from 5 to 500, the amplitudes of $v$
and $p$ decrease markedly. Figure~\ref{fig:kov_err} shows local MSE peaks in the
low-to-intermediate-$Re$ range, whereas the relative $L_2$ errors of $v$ and $p$ remain
substantially higher than that of $u$ over a broader interval, consistent with the within-equation
scale disparity discussed above. In Figure~\ref{fig:kov_search}, newly selected tasks cover the full
parameter domain along the logarithmic Reynolds-number axis, and later replay continues to constrain
earlier tasks. The training prior $f_{\mathrm{prior}}(Re)=\log_{10}(Re)$ moderately upweights
large-$Re$ tasks to compensate for the relative-error risk of the smaller-amplitude components in
that region; this choice follows the amplitude variation of the present problem and is not proposed
as a universal prior for other equations.

\subsubsection{Poisson--Boltzmann equation}
\label{sec:poisson}

The Poisson equation is widely used in mathematical physics to describe a potential field induced by
a source such as charge or mass density, with applications in electrostatics, gravitation, and heat
conduction. Here we study the two-dimensional steady linearized Poisson--Boltzmann equation
\cite{hao2023pinnacle} in Eq.~\ref{eq:poi}, with four independent parameters $\mu_1$, $\mu_2$, $k$,
and $A$. Because the nonlinear ionic source in the full Poisson--Boltzmann equation is replaced by
the linear reaction term $k^2u$, this benchmark can also be viewed as a screened-Poisson problem.
The forcing, boundary condition, and parameter domain stated in Eq.~\ref{eq:poi} and below match
those used in the implementation and reference-data generation.

\begin{equation}
\begin{cases}
-\Delta u+k^2u=f(x,y), & (x,y)\in\Omega,\\
\begin{aligned}
f(x,y)={}&A(\mu_1^2+x^2+\mu_2^2+y^2)\\
&{}\times\sin(\mu_1\pi x)\sin(\mu_2\pi y)
\end{aligned}
& (x,y)\in\Omega,\\
u(x,y)=0.2, & (x,y)\in\partial\Omega.
\end{cases}
\label{eq:poi}
\end{equation}

\begin{figure*}[!tp]
\centering
{\footnotesize\textbf{$(\mu_1,\mu_2,k,A)=(1,3,5,5)$}}\par\vspace{0.1em}
\includegraphics[width=\poissonfieldwidth]{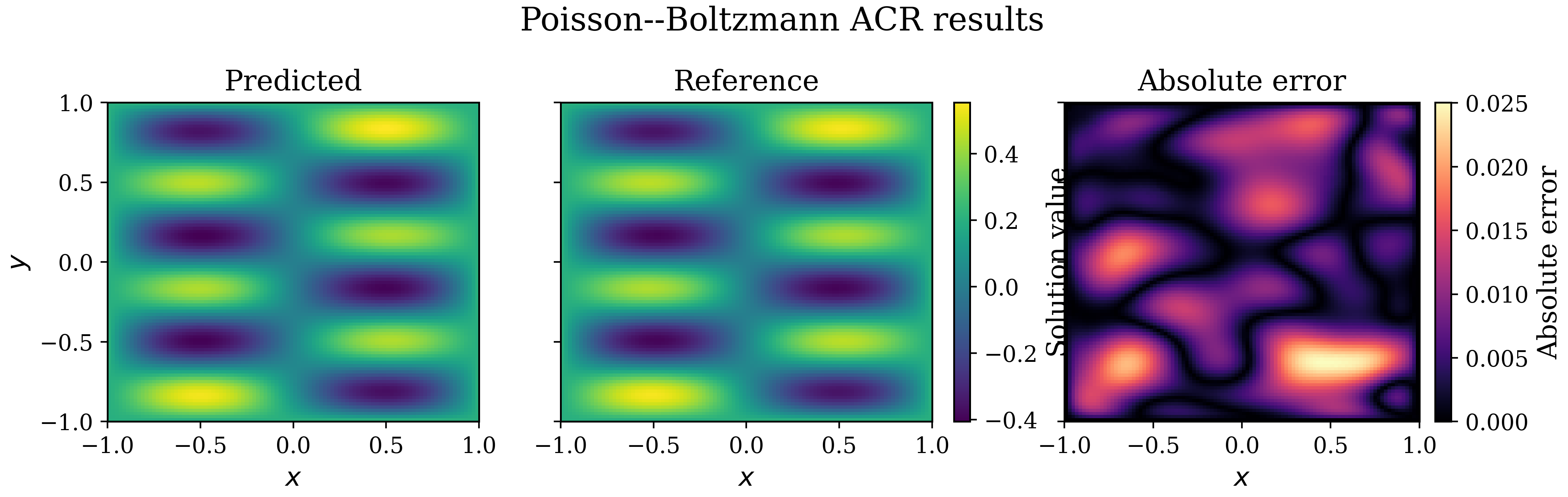}
\par\vspace{-0.35em}
{\footnotesize\textbf{$(2,3,5,5)$}}\par\vspace{0.1em}
\includegraphics[width=\poissonfieldwidth]{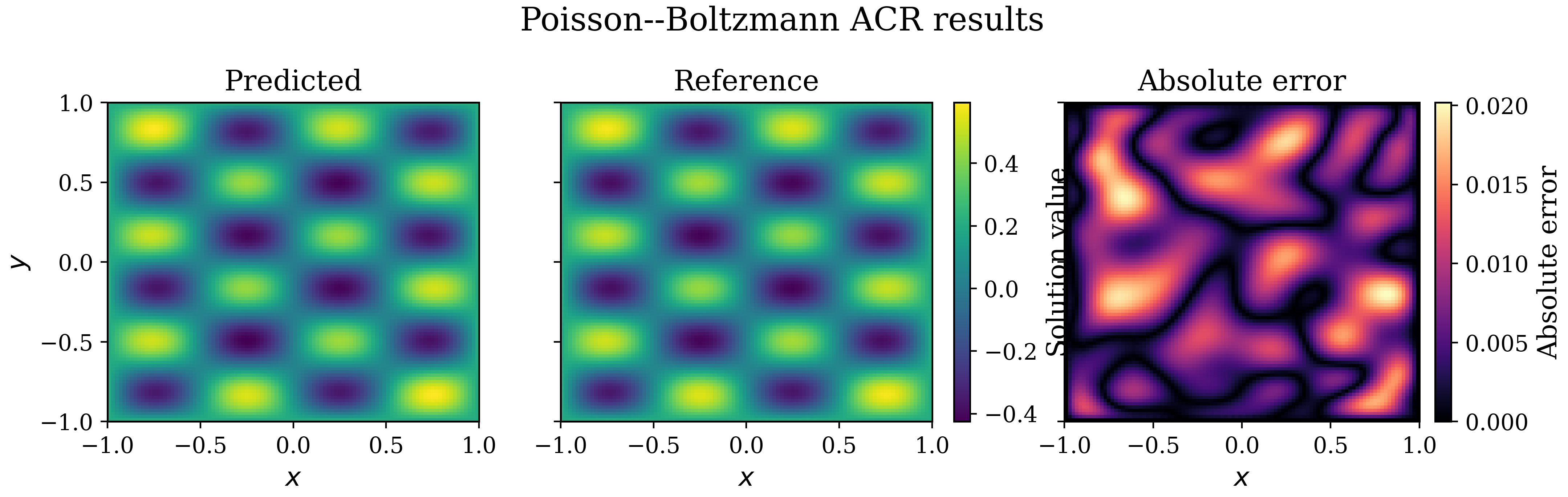}
\par\vspace{-0.35em}
{\footnotesize\textbf{$(1,4,10,15)$}}\par\vspace{0.1em}
\includegraphics[width=\poissonfieldwidth]{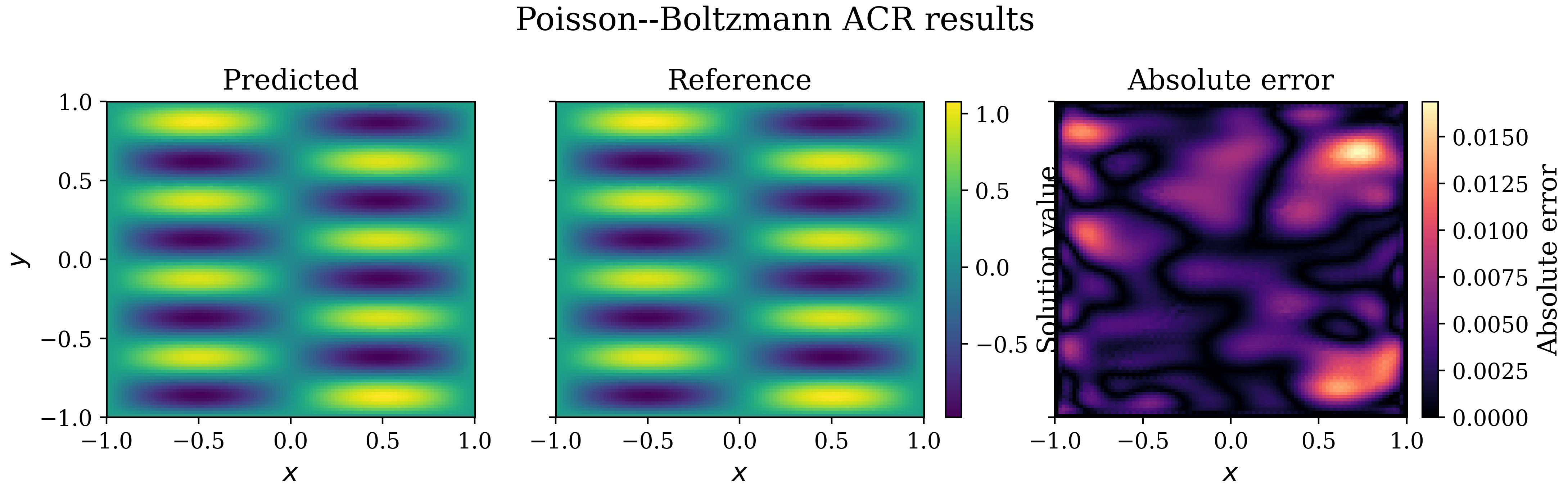}
\par\vspace{-0.35em}
{\footnotesize\textbf{$(2,4,10,15)$}}\par\vspace{0.1em}
\includegraphics[width=\poissonfieldwidth]{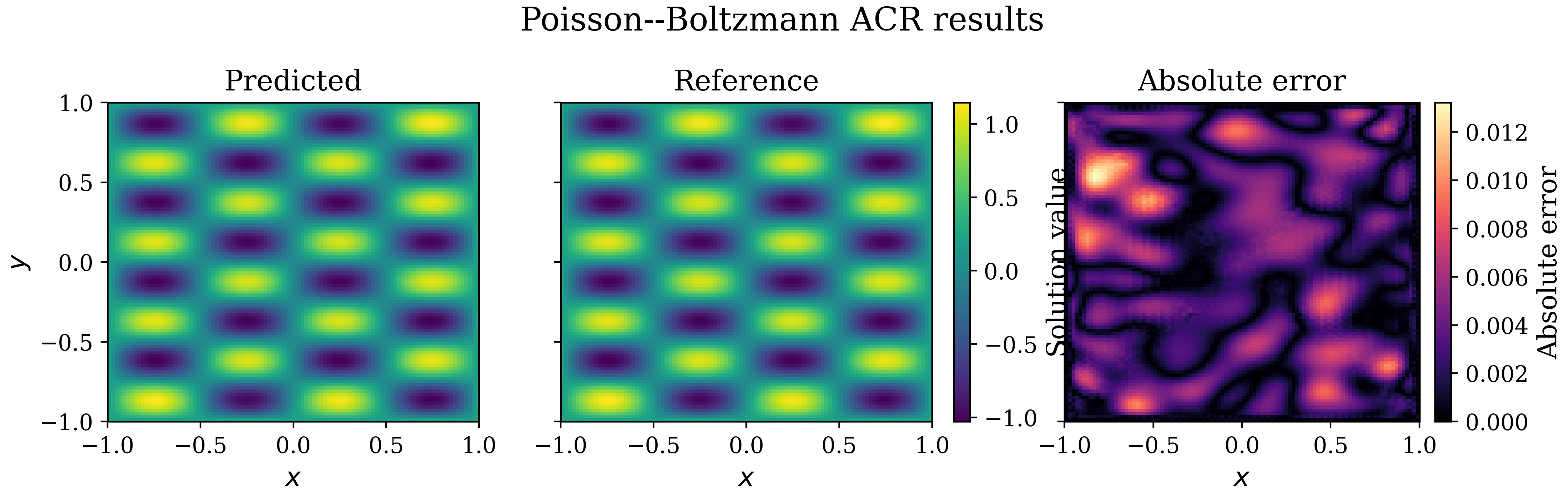}
\caption{\textbf{ACR predictions, reference solutions, and absolute errors at four representative parameter tuples of the linearized Poisson--Boltzmann benchmark.}}
\label{fig:poi}
\end{figure*}

\begin{figure*}[!tp]
\centering
\includegraphics[width=\wideconditionwidth]{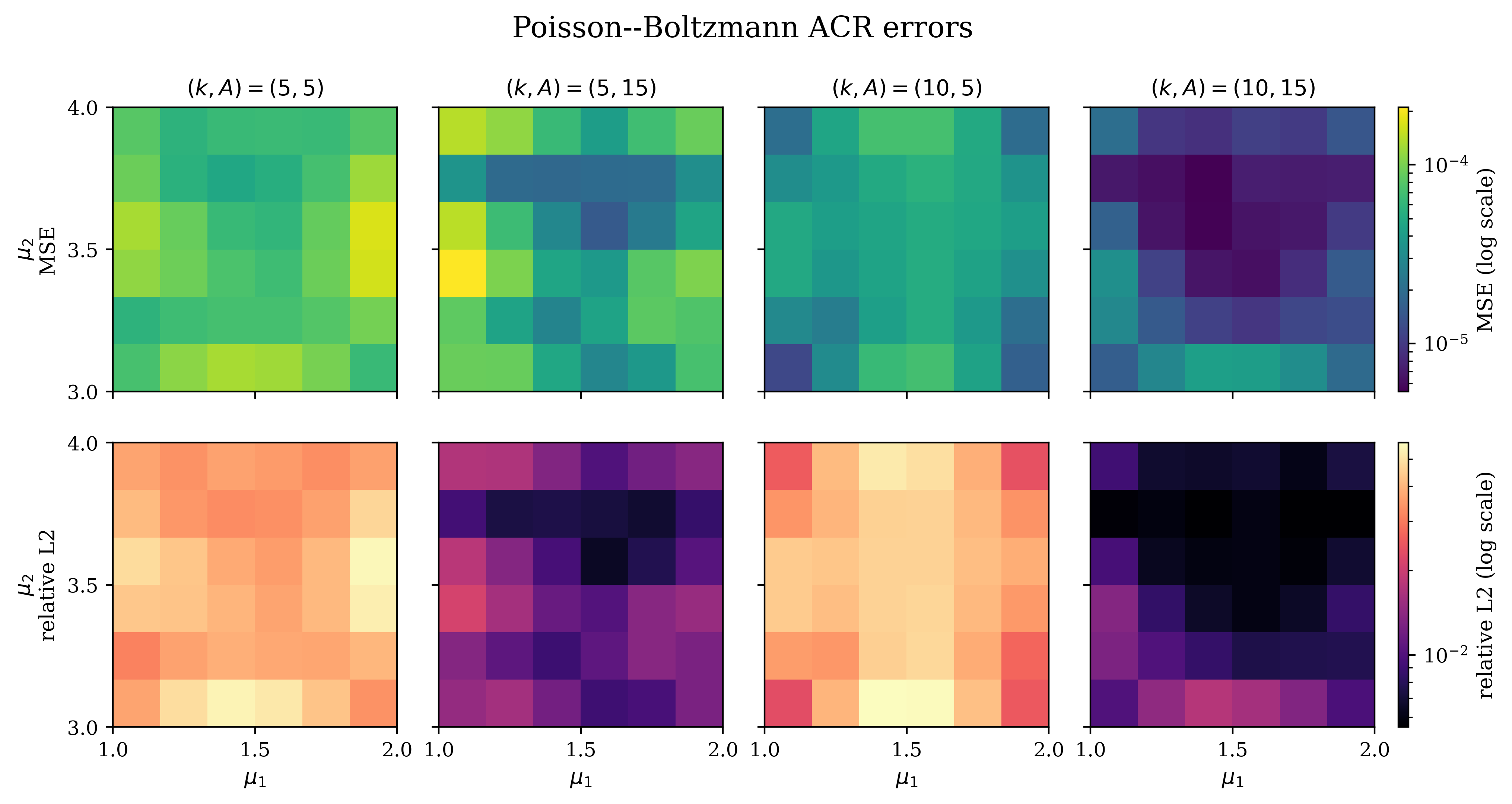}
\caption{\textbf{Representative conditional error slices of ACR on the four-dimensional Poisson--Boltzmann test grid.} Each column fixes one $(k,A)$ pair, with $\mu_1$ and $\mu_2$ on the horizontal and vertical axes; the upper and lower rows show $\mathrm{MSE}$ and $E_{L_2}$, respectively, on logarithmic color scales.}
\label{fig:poi_err}
\end{figure*}

Parameters are ordered as $(\mu_1,\mu_2,k,A)$. The initial set contains the 16 corners of the
parameter domain, and each dense and sparse replay task uses 5041 and 529 samples, respectively.
Only key events, including training initialization, the onset of replay, and the optimizer
transition, are listed.

\begin{table*}[!tp]
\centering
\caption{\textbf{Four-dimensional parameter initialization and key selection events in a representative Poisson--Boltzmann run.}}
\label{table:poi_search}
\small
\setlength{\tabcolsep}{4pt}
\renewcommand{\arraystretch}{1.12}
\begin{tabularx}{\textwidth}{>{\raggedright\arraybackslash}p{0.10\textwidth}>{\raggedright\arraybackslash}p{0.28\textwidth}>{\raggedright\arraybackslash}X>{\centering\arraybackslash}p{0.18\textwidth}}
\toprule
Step & Initial corner tasks & New this event & Dense / replay count\\
\midrule
0 & $\{1,2\}\times\{3,4\}\times\{5,10\}\times\{5,15\}$ (16 in total, retained throughout) & --- & 16 / 0\\
250 & --- & $(2,3.8,10,15)$ & 17 / 0\\
500 & --- & $(1.8,4,10,15)$ & 18 / 0\\
750 & --- & $(1,3.6,7,13)$ & 19 / 0\\
1000 & --- & $(1.6,4,5,15)$ & 20 / 0\\
1250 & --- & $(2,4,7,15)$ & 21 / 0\\
1500--16000 & --- & \emph{The intervening 59 search events are omitted for brevity} & \emph{Gradually increased from 21 / 0 to 80 / 0}\\
16250 & --- & $(1.8,3.8,10,15)$ & 80 / 1\\
16500 & --- & $(1.6,3.6,10,15)$ & 80 / 2\\
16750 & --- & $(1.4,3.6,5,15)$ & 80 / 3\\
17000--17750 & --- & \emph{The intervening four search events are omitted for brevity} & \emph{Changed gradually from 80 / 4 to 80 / 7}\\
18000 & --- & $(1.8,3.6,10,15)$ & 80 / 8\\
18250 & --- & $(1.8,3.4,5,15)$ & 80 / 9\\
18500--39750 & --- & \emph{The subsequent 86 search events are omitted for brevity} & \emph{The dense-task count remained at 80}\\
40000--60000 & --- & --- (all 130 selected parameters fixed) & 80 / 50 (search ended and L-BFGS began)\\
\bottomrule
\end{tabularx}
\end{table*}

\begin{table*}[!tp]
\centering
\addtocounter{table}{-1}
\caption{\textbf{Four-dimensional parameter initialization and key state transitions (continued): replay-state transitions from the onset of replay at step 16250.}}
\small
\setlength{\tabcolsep}{4pt}
\renewcommand{\arraystretch}{1.12}
\begin{tabularx}{\textwidth}{>{\raggedright\arraybackslash}p{0.12\textwidth}>{\raggedright\arraybackslash}X>{\raggedright\arraybackslash}X>{\raggedright\arraybackslash}X>{\centering\arraybackslash}p{0.17\textwidth}}
\toprule
Step & Replay training & Entered replay & Restored to dense & Dense / replay count\\
\midrule
16250 & 1 task & $(1,3.4,8,5)$ (first entry) & --- & 80 / 1\\
16500 & 2 tasks; current focus: $(1,3.4,8,5)$ & $(1,3.6,6,7)$ & --- & 80 / 2\\
16750 & 3 tasks; current focus: $(1,3.4,8,5)$ & $(1.2,3.8,6,9)$ & --- & 80 / 3\\
17000--17750 & \emph{Gradually increased from 4 to 7 tasks} & \emph{One replay task added at each event} & --- & \emph{Changed gradually from 80 / 4 to 80 / 7}\\
18000 & 8 tasks; current focus: $(1,3.4,8,5)$ & $(1,3,10,9)$ & --- & 80 / 8\\
18250 & 9 tasks; current focus: $(1,3.4,8,5)$ & $(1.4,3.4,7,13)$ & --- & 80 / 9\\
18500--39750 & \emph{Replay training continued throughout} & \emph{State transitions occurred during this interval} & \emph{State transitions occurred during this interval} & \emph{The dense-task count remained at 80}\\
40000--60000 & 50 tasks & --- & --- & 80 / 50 (search ended and L-BFGS began)\\
\bottomrule
\end{tabularx}
\end{table*}

\begin{table*}[!tp]
\centering
\caption{\textbf{Mean and population standard deviation of the test errors on the four-parameter linearized Poisson--Boltzmann equation.}}
\label{table:poi}
\footnotesize
\setlength{\tabcolsep}{3.5pt}
\renewcommand{\arraystretch}{1.10}
\begin{tabularx}{\textwidth}{l*{6}{>{\centering\arraybackslash}X}}
\toprule
Method & $\mathrm{MSE}$ mean & $\mathrm{MSE}$ SD & $E_{L_2}$ mean & $E_{L_2}$ SD & $R_{\mathrm{PDE}}$ mean & $R_{\mathrm{PDE}}$ SD\\
\midrule
UNI & $7.675\times10^{-2}$ & $9.958\times10^{-2}$ & $0.4189$ & $0.3654$ & $17.56$ & $14.25$\\
FIX & $22.50$ & $7.704$ & $12.67$ & $2.151$ & $248.49$ & $29.66$\\
AG & $1.556\times10^{-4}$ & $1.452\times10^{-4}$ & $0.03125$ & $0.01521$ & $1.549$ & $0.581$\\
AC & $9.644\times10^{-4}$ & $1.287\times10^{-3}$ & $0.06065$ & $0.05663$ & $2.349$ & $1.770$\\
\textbf{ACR} & \textbf{$1.348\times10^{-4}$} & $1.068\times10^{-4}$ & \textbf{$0.03046$} & $0.01071$ & \textbf{$1.545$} & $0.499$\\
ACR2-arch & $2.909\times10^{-4}$ & $3.040\times10^{-4}$ & $0.04397$ & $0.02487$ & $1.867$ & $0.870$\\
\bottomrule
\end{tabularx}
\end{table*}

The spatial domain is $\Omega = \{(x,y) \mid (x,y) \in [-1,1] \times [-1,1]\}$, and the parameter
domain is $\Theta = \{(\mu_1,\mu_2,k,A) \mid (\mu_1,\mu_2,k,A) \in
[1,2]\times[3,4]\times[5,10]\times[5,15]\}$. The model uses a hard boundary constraint. Each
parameter task uses 5041 physical samples, and sparse replay retains 10\% of the sparsifiable
physics-point pool. The dense active set and replay set each contain at most 80 parameter tasks. UNI
and FIX request $2\times10^4$ Adam updates, whereas the four dynamic methods request $4\times10^4$;
all methods have an L-BFGS limit of $2\times10^4$ iterations. The dynamic methods add tasks
gradually, and their larger Adam limit compensates for the smaller average active-set size early in
training. Query counts and online work are nevertheless reported separately and are not presented as
strictly equal computational cost. The standard FNN has six inputs, one output, and five hidden
layers of width 30, whereas ACR2-arch uses a parameter subnetwork. AC uses
$N_{\mathrm{resample}}=500$, while ACR and ACR2-arch use 250. Training uses the constant prior
$f_{\mathrm{prior}}(\mu_1,\mu_2,k,A)=1$, with $\lambda_{\mathrm{static}}=1$ and
$\lambda_{\mathrm{dynamic}}=-1$. Bayesian search uses the physics-informed loss without an
additional prior multiplier, performs 50 evaluations per active update, and sets $\kappa=5$.

Each method is evaluated with three independent random seeds on 1296 uniformly distributed parameter
values (a $6^4$ grid) in the parameter domain. Table~\ref{table:poi} reports the results.

Table~\ref{table:poi} shows large errors and variability for UNI and FIX. FIX fails markedly in
$\mathrm{MSE}$, $E_{L_2}$, and $R_{\mathrm{PDE}}$, indicating that a small set of predetermined
tasks is insufficient to cover this four-dimensional parameter domain. Grid-greedy active selection
in AG substantially improves on fixed sampling and approaches ACR on all three mean metrics, but
every selection event requires many objective-loss evaluations over a dense candidate grid. AC
without replay is less accurate than AG. ACR attains the lowest mean $\mathrm{MSE}$, $E_{L_2}$, and
$R_{\mathrm{PDE}}$, showing that retaining sparse physical constraints from earlier tasks helps
preserve accuracy when active-task capacity is limited. ACR2-arch does not outperform ACR, further
indicating that the benefit of the parameter subnetwork depends on the equation and optimization
conditions.

Figure~\ref{fig:poi} shows that increasing $\mu_1$ or $\mu_2$ raises the oscillation frequency in
the corresponding spatial direction, whereas $k$ and $A$ change the screening strength and source
amplitude, respectively. In Figure~\ref{fig:poi_err}, the high-error region moves across the
$\mu_1$--$\mu_2$ plane as $(k,A)$ changes, and the $\mathrm{MSE}$ and $E_{L_2}$ patterns do not
coincide. The four parameters therefore affect solution difficulty jointly through frequency,
reaction strength, and amplitude; the error is not a simple monotone function of any single
parameter.

A regular candidate grid grows rapidly with parameter dimension: with six candidate values per
dimension, one exhaustive search over this four-dimensional domain requires evaluating $6^4=1296$
parameter values, whereas BO performs only 50 sequential queries at each active update.
Table~\ref{table:poi_search} further shows that the algorithm starts from the 16 corners of the
parameter domain and gradually adds interior parameters. Once the dense-task set reaches its
capacity of 80, displaced tasks enter sparse replay; training ultimately uses 80 dense tasks and 50
replay tasks. Bayesian selection thus limits expensive loss queries in the high-dimensional
candidate space, while sparse replay expands the parameter coverage that remains physics constrained
at fixed dense capacity. This comparison concerns the objective-loss queries required for parameter
selection and does not imply identical end-to-end training time across methods.

\FloatBarrier
\subsection{ACR2-finetune: rapid single-parameter adaptation}
\label{sec:rapid-adaptation}

The pretrained global parameterized model provides direct predictions throughout the parameter
domain, but a few strictly unseen parameters may still lie in local high-error regions. We use
\textbf{ACR2-finetune} to denote an optional single-parameter residual adaptation: the global model
is frozen, a one-hidden-layer residual head with zero-output initialization is appended, and only
this head is updated using newly sampled PDE, boundary-condition, and initial-condition residuals at
the target parameter. The two names should be distinguished: \textbf{ACR2-arch} denotes the global
ParamFNN architecture used in the primary experiments, whereas \textbf{ACR2-finetune} denotes a
post-training adaptation procedure that can, in principle, be applied to an AC, ACR, or ACR2-arch
model. The experiments in this section use a frozen ACR model as the base. Because Schaffer-like has
no corresponding physical residual, this section evaluates only the four PDEs. All targets, learning
rates, and checkpoints are fixed in advance. Online time includes only CUDA-synchronized Adam
updates and excludes model restoration and reference-error evaluation.

To distinguish one-time offline pretraining from per-target online adaptation, we performed a
separate exclusive single-GPU timing audit. ACR pretraining took approximately 19 min for Burgers,
17 min for Allen--Cahn, 20 min for Kovasznay, and 136 min for Poisson--Boltzmann. The longer
Poisson--Boltzmann run mainly reflects the doubled Adam-update budget of the dynamic methods and the
larger number of parameter searches. After deployment, only the compact residual head is updated at
each unseen target; the fixed 500-step online times in Table~\ref{tab:finetune-diagnostic} range
from 11.26 to 34.47 s per target. Offline pretraining is therefore a one-time cost that can be
amortized across many targets, whereas adaptation of an individual target remains on the order of
tens of seconds.

The study uses three independently pretrained models with random seeds $[0,1,2]$. For each seed,
five prespecified strictly unseen Burgers targets and three such targets for each of the other PDEs
are evaluated, yielding 15, 9, 9, and 9 seed--target pairs, respectively. The errors and times in
the table are aggregated over these fixed pairs.

\begin{table*}[!tp]
\centering
\caption{\textbf{ACR2-finetune results and online time after 500 adaptation steps at fixed unseen targets.}}
\label{tab:finetune-diagnostic}
\footnotesize
\setlength{\tabcolsep}{3.5pt}
\renewcommand{\arraystretch}{1.10}
\begin{tabularx}{\textwidth}{>{\raggedright\arraybackslash}p{0.105\textwidth}>{\centering\arraybackslash}X>{\centering\arraybackslash}X>{\centering\arraybackslash}p{0.12\textwidth}>{\centering\arraybackslash}p{0.14\textwidth}}
\toprule
PDE & $\mathrm{MSE}$: step 0 $\rightarrow$ 500 & $E_{L_2}$: step 0 $\rightarrow$ 500 & Improved seed--target pairs & Time per target for 500 steps\\
\midrule
Burgers & $1.966\times10^{-7}\rightarrow$ \textbf{$6.279\times10^{-8}$} & $9.522\times10^{-4}\pm3.135\times10^{-4}\rightarrow$ \textbf{$5.651\times10^{-4}\pm8.489\times10^{-5}$} (40.7\% reduction) & 14/15 & $11.26\pm0.84$ s\\
Allen--Cahn & $9.996\times10^{-5}\rightarrow1.502\times10^{-4}$ & $1.156\times10^{-2}\pm6.812\times10^{-3}\rightarrow1.284\times10^{-2}\pm7.018\times10^{-3}$ (11.1\% increase) & 4/9 & $16.38\pm1.71$ s\\
Kovasznay & $1.681\times10^{-5}\rightarrow$ \textbf{$4.979\times10^{-6}$} & $6.959\times10^{-2}\pm4.176\times10^{-2}\rightarrow$ \textbf{$4.249\times10^{-2}\pm2.104\times10^{-2}$} (38.9\% reduction) & 9/9 & $34.47\pm1.09$ s\\
Poisson--Boltzmann & $9.703\times10^{-5}\rightarrow$ \textbf{$3.379\times10^{-6}$} & $2.686\times10^{-2}\pm3.232\times10^{-3}\rightarrow$ \textbf{$5.667\times10^{-3}\pm1.445\times10^{-3}$} (78.9\% reduction) & 9/9 & $26.41\pm1.36$ s\\
\bottomrule
\end{tabularx}
\end{table*}

Table~\ref{tab:finetune-diagnostic} shows that, after 500 steps, the mean relative $L_2$ errors of
Burgers, Kovasznay, and Poisson--Boltzmann decrease by 40.7\%, 38.9\%, and 78.9\%, respectively,
with improvements in 14/15, 9/9, and 9/9 seed--target pairs. Updating only the compact residual head
can therefore reduce the error of most targets within tens of seconds. Allen--Cahn is a clear
boundary case: its mean relative $L_2$ error increases by 11.1\%, and only 4/9 pairs improve,
showing that a fixed short adaptation budget does not guarantee a lower solution error for every
PDE.

We next use the complete strictly unseen test grid from the seed-0 run. Burgers, Allen--Cahn,
Kovasznay, and Poisson--Boltzmann contain 83, 77, 80, and 1166 parameter values, respectively, and
each target undergoes one fixed 500-step adaptation. The difficult subset is defined solely as the
highest quartile of the pre-adaptation $E_{L_2}$ values, giving 21, 20, 20, and 292 targets; it is
not reselected according to the post-adaptation improvement.

\begin{table*}[!tp]
\centering
\caption{\textbf{ACR2-finetune results after 500 steps over all strictly unseen parameters and their difficult quartiles.}}
\label{tab:finetune-full-grid}
\footnotesize
\setlength{\tabcolsep}{3.5pt}
\renewcommand{\arraystretch}{1.10}
\begin{tabularx}{\textwidth}{>{\raggedright\arraybackslash}p{0.10\textwidth}*{6}{>{\centering\arraybackslash}X}}
\toprule
PDE & Full grid: improved / total & Full-grid mean $E_{L_2}$ reduction & Difficult quartile: improved / total & Difficult-quartile mean $E_{L_2}$ change & Difficult-quartile median improvement factor & Time per target\\
\midrule
Burgers & 77/83 & 63.1\% & \textbf{21/21} & \textbf{78.2\% reduction} & \textbf{$6.23\times$} & $8.45\pm0.56$ s\\
Allen--Cahn & 60/77 & 5.3\% & 9/20 & 41.3\% increase & $0.88\times$ & $12.31\pm0.57$ s\\
Kovasznay & 79/80 & 28.5\% & \textbf{20/20} & \textbf{36.7\% reduction} & \textbf{$1.58\times$} & $23.77\pm0.81$ s\\
Poisson--Boltzmann & \textbf{1,166/1,166} & \textbf{81.7\%} & \textbf{292/292} & \textbf{83.6\% reduction} & \textbf{$6.19\times$} & $19.69\pm0.87$ s\\
\bottomrule
\end{tabularx}
\end{table*}

Table~\ref{tab:finetune-full-grid} extends the analysis to all strictly unseen parameters in the
representative run. Burgers, Kovasznay, and Poisson--Boltzmann improve at 77/83, 79/80, and
1166/1166 targets, respectively. Within the difficult quartile selected only by the zero-shot error,
all 21/21, 20/20, and 292/292 targets improve, with mean relative-$L_2$ reductions of 78.2\%,
36.7\%, and 83.6\%. ACR2-finetune is therefore particularly effective at reducing errors in local
high-error regions left by the pretrained model. For Allen--Cahn, only 9/20 difficult targets
improve and the subset mean increases by 41.3\%; this conclusion cannot be extended unconditionally
to all equations.

\begin{figure*}[!tp]
\centering
\includegraphics[width=\fullgridadaptwidth]{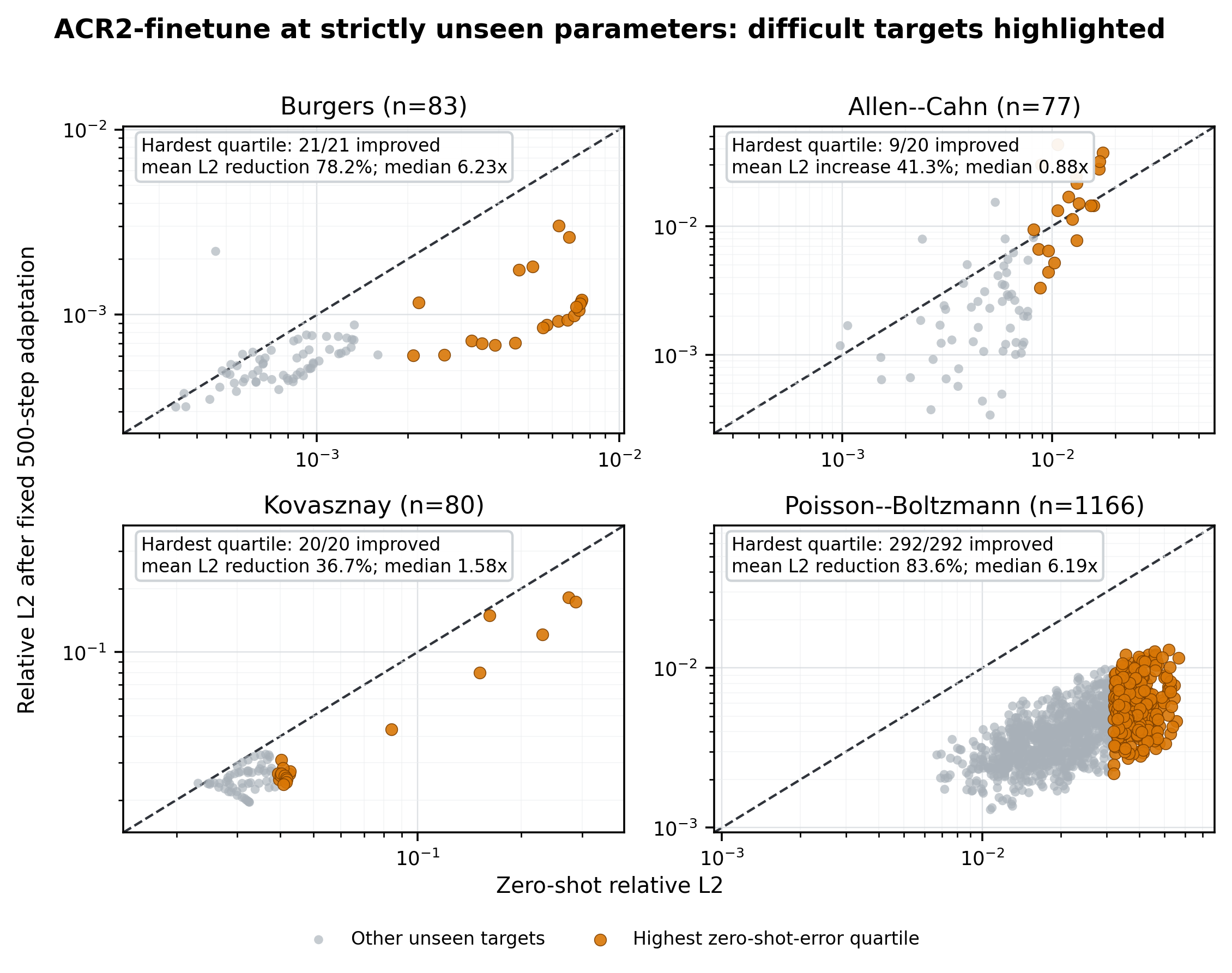}
\caption{\textbf{Zero-shot versus post-adaptation $E_{L_2}$ after 500 ACR2-finetune steps over all strictly unseen parameters.} Points below the identity line improve; orange markers denote the highest zero-shot-error quartile, and all degraded cases are retained.}
\label{fig:finetune-full-grid}
\end{figure*}

Figure~\ref{fig:finetune-full-grid} shows that most Burgers, Kovasznay, and Poisson--Boltzmann
points lie below the identity line, with the largest concentrated reductions in the difficult
quartiles. This pattern indicates that bounded-capacity global training can leave some parameter
regions insufficiently resolved and that additional physical constraints at a target parameter can
reduce these local errors. Allen--Cahn contains substantial numbers of both improved and degraded
targets. Its sharp phase interfaces and strongly nonlinear reaction make the single-parameter
optimization itself difficult, so adding a residual head does not necessarily lower the solution
error under a short budget. The benefit of fine-tuning thus depends jointly on the error left by the
global model and the local optimization difficulty of the PDE.

\begin{figure*}[!tp]
\centering
{\footnotesize\textbf{Burgers, $\nu=0.91$}}\par\vspace{0.1em}
\includegraphics[width=\adaptfieldwidth]{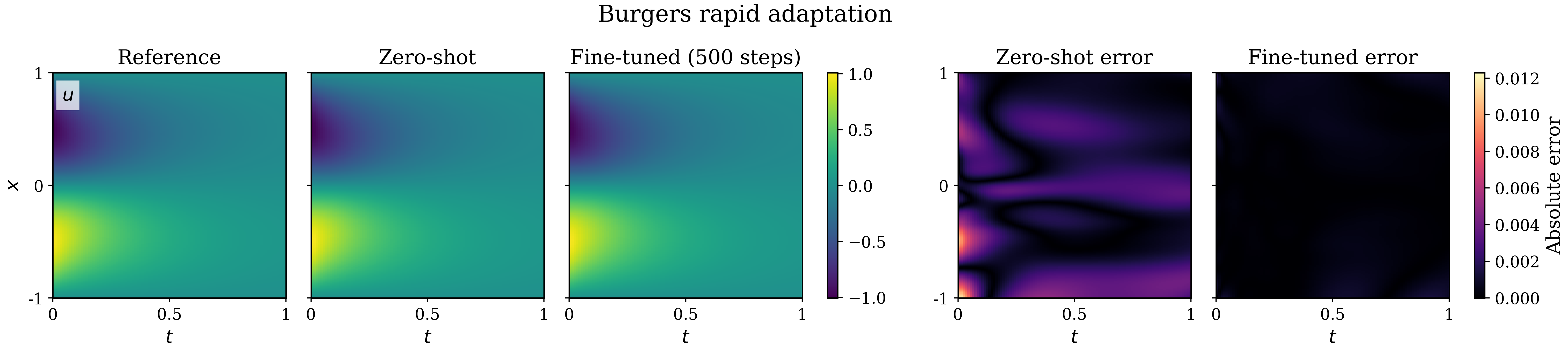}
\par\vspace{-0.35em}
{\footnotesize\textbf{Allen--Cahn, $(\nu,\rho)=(0.004,5.0)$}}\par\vspace{0.1em}
\includegraphics[width=\adaptfieldwidth]{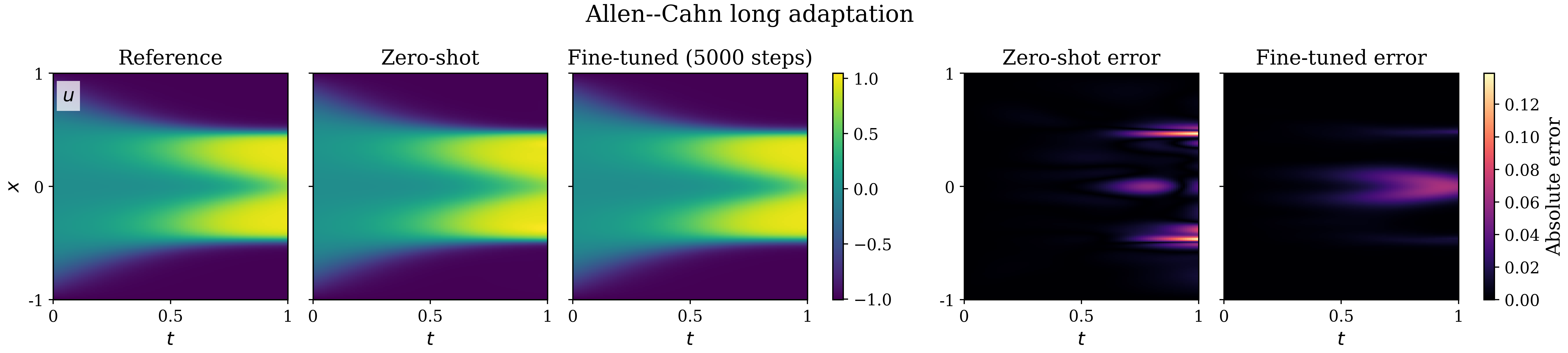}
\par\vspace{-0.35em}
{\footnotesize\textbf{Kovasznay, $Re=25$}}\par\vspace{0.1em}
\includegraphics[width=\adaptfieldwidth]{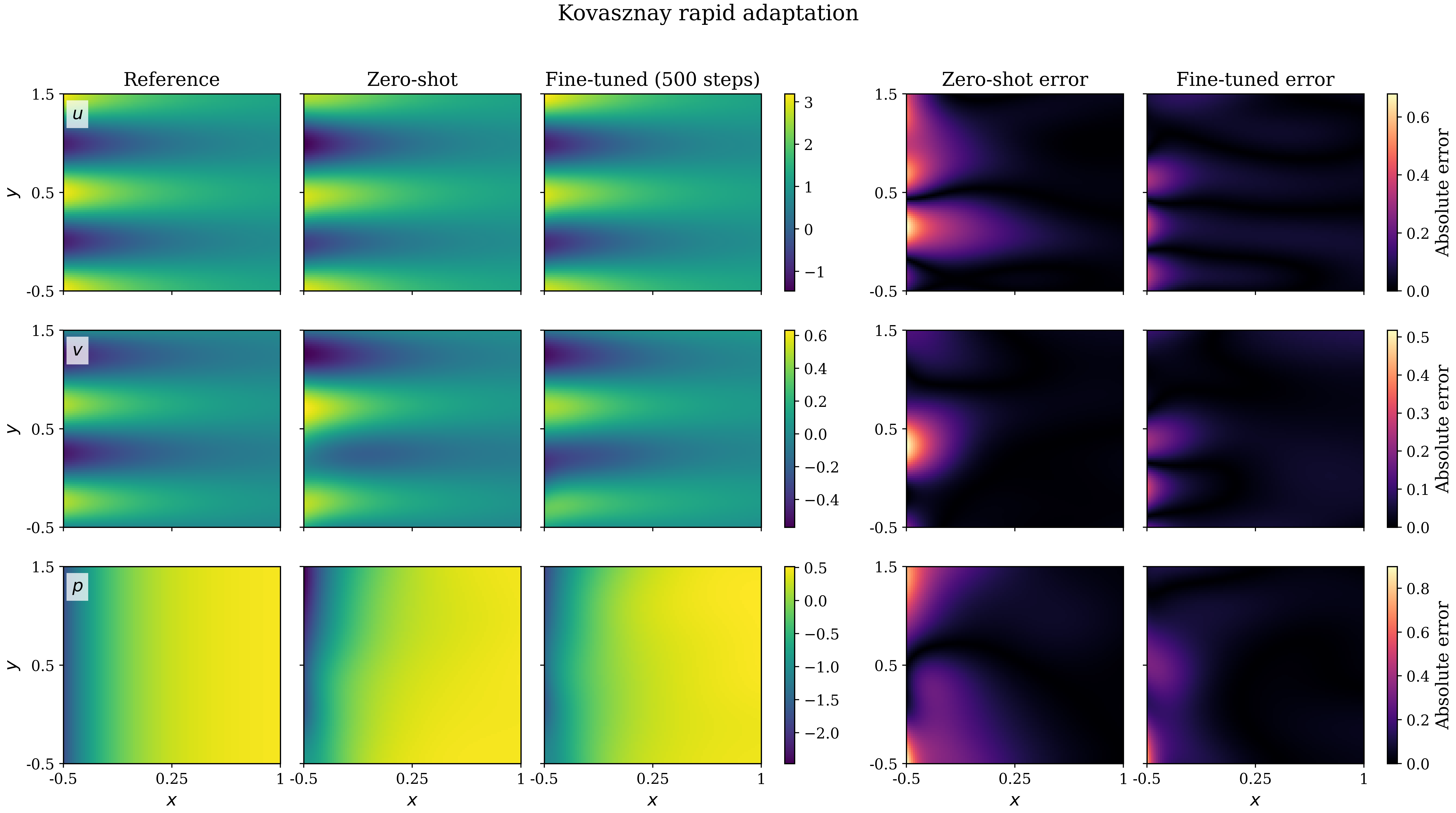}
\par\vspace{-0.35em}
{\footnotesize\textbf{Poisson--Boltzmann, $(\mu_1,\mu_2,k,A)=(1.6,3.0,10,5)$}}\par\vspace{0.1em}
\includegraphics[width=\adaptfieldwidth]{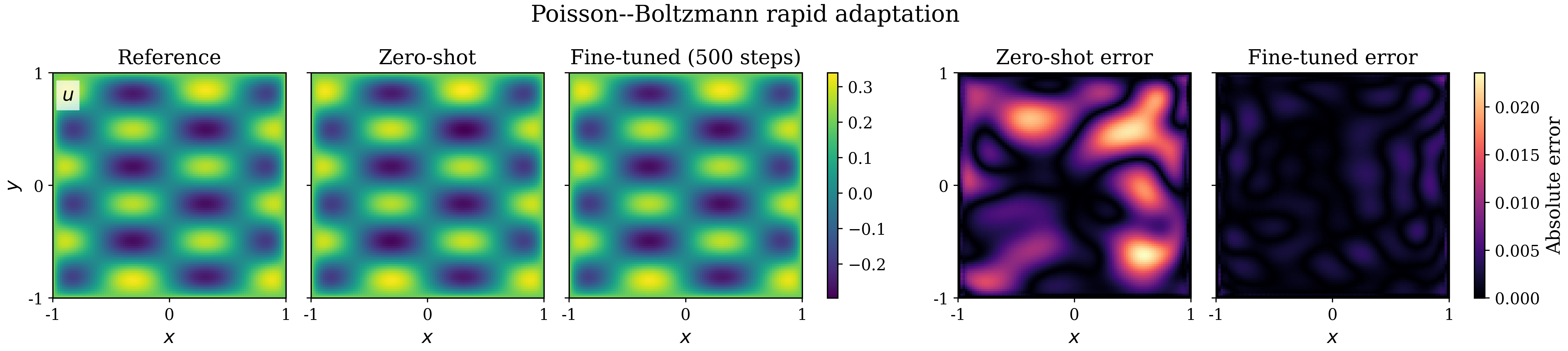}
\caption{\textbf{Field comparisons at four difficult parameters selected by the largest zero-shot error.} Each panel shows the reference solution, zero-shot prediction, post-adaptation prediction, and the absolute errors before and after adaptation. Burgers, Kovasznay, and Poisson--Boltzmann use the 500-step endpoint; Allen--Cahn uses the prespecified 5000-step endpoint, while its degraded 500-step result is retained in Table~\ref{tab:finetune-diagnostic}.}
\label{fig:finetune-hard-fields}
\end{figure*}

Figure~\ref{fig:finetune-hard-fields} provides direct field-level evidence. The space--time banded
error in Burgers and the dispersed local errors in Poisson--Boltzmann decrease substantially after
500 steps, while all three Kovasznay components, $u$, $v$, and $p$, improve. Allen--Cahn requires a
longer budget to reduce the error around its phase interfaces, consistent with its degraded mean
result at 500 steps. The residual head can therefore reduce several forms of local representation
error, but the required budget and benefit remain equation dependent.

Overall, ACR2-finetune can improve parameter values that are insufficiently resolved by the
pretrained model at modest online cost, but its benefit depends on the PDE and adaptation budget and
is not guaranteed to be monotonic. The complete set of degraded cases retained in
Table~\ref{tab:finetune-full-grid} delineates this boundary.

\FloatBarrier
\subsubsection{Comparison with standard PINNs trained from scratch}
\label{sec:standard-pinn}

To evaluate the role of a shared pretrained representation in single-parameter adaptation, we
compare ACR2-finetune with standard single-parameter PINNs trained from random initialization. Both
use the same targets, reference solutions, evaluation grids, multiple independent runs, and fixed
500-step Adam endpoint. The standard PINN updates a complete network for each target, whereas
ACR2-finetune freezes the pretrained model and updates only the residual head. The comparison
therefore measures adaptation quality at matched online steps rather than total cost including
offline pretraining.

\begin{strip}
\centering
\captionof{table}{\textbf{Error and online trainable parameter counts of ACR2-finetune and standard PINNs trained from scratch at the 500-step endpoint.}}
\footnotesize
\setlength{\tabcolsep}{3.5pt}
\renewcommand{\arraystretch}{1.10}
\begin{tabularx}{\textwidth}{>{\raggedright\arraybackslash}p{0.11\textwidth}*{4}{>{\centering\arraybackslash}X}}
\toprule
PDE & ACR2-finetune mean relative $L_2$ & Standard PINN mean relative $L_2$ & Error ratio: standard PINN / ACR2-finetune & Online trainable parameters: ACR2-finetune / standard PINN\\
\midrule
Burgers & \textbf{$5.651\times10^{-4}\pm8.489\times10^{-5}$} & $0.3937\pm0.0027$ & $696.7\times$ & $1{,}301/7{,}851$\\
Allen--Cahn & \textbf{$0.01284\pm0.00702$} & $0.5245\pm0.0714$ & $40.8\times$ & $1{,}301/7{,}851$\\
Kovasznay & \textbf{$0.04249\pm0.02104$} & $18.585\pm4.647$ & $437.4\times$ & $1{,}353/7{,}953$\\
Poisson--Boltzmann & \textbf{$0.005667\pm0.001445$} & $0.9459\pm0.0017$ & $166.9\times$ & $481/3{,}841$\\
\bottomrule
\end{tabularx}
\end{strip}

Table 10 shows that, at the common 500-step endpoint, the mean
target relative $L_2$ errors of standard PINNs trained from scratch are $40.8$--$696.7\times$ those
of ACR2-finetune. ACR2-finetune also updates only about $1/5.88$--$1/7.99$ as many parameters as the
complete standard PINN. The representation learned during offline parameterized training therefore
provides a substantially better starting point for short-budget adaptation. This comparison does not
establish lower total training cost, because the standard PINN has no shared offline-pretraining
stage. Although ACR2-finetune slightly degrades from its zero-shot prediction on Allen--Cahn, its
500-step endpoint remains substantially more accurate than that of the standard PINN trained from
scratch.

\FloatBarrier
\subsection{External-method comparisons}
\label{sec:external}

\subsubsection{External methods and fair-comparison protocol}
\label{sec:external-protocol}

We compare three representative parameterized PINN methods. P2INN \cite{cho2024p2inn} separately
encodes PDE parameters and space--time coordinates and modulates them in a shared latent space; its
native SVD adaptation updates only a small number of low-rank variables. HyperPINN
\cite{de2021hyperpinn} uses a hypernetwork to map PDE parameters to the weights of a compact target
PINN. Meta-PINN \cite{penwarden2023metalearning} pretrains task networks at support parameters,
learns a surrogate mapping from PDE parameters to network weights, and then performs task-level
optimization at a new parameter. P2INN and HyperPINN are compared under the common protocol on all
four PDEs, whereas Meta-PINN is evaluated only on Burgers and is not extrapolated to the other
equations.

The common protocol uses the same PDEs, parameter domains, reference solutions, strictly unseen
targets, and multiple independent offline runs. It reports both direct prediction after offline
training (step 0) and the endpoint after 500 online Adam steps. P2INN and HyperPINN are
independently implemented, capacity-matched adaptations of the respective mechanisms rather than
official four-PDE reproductions provided by the original authors. A capacity audit shows that their
total parameter counts differ from that of the standard parameterized FNN by at most 3.03\% and
0.40\%, respectively. The total capacities are therefore comparable, although the methods distribute
that capacity differently among parameter encoders, modulation modules, hypernetworks, and target
networks.

The offline costs are not identical. Our global model is trained with both Adam and L-BFGS and has
already reached relatively high accuracy before online adaptation. P2INN and HyperPINN use different
parameter representations and offline optimization procedures, and under the present budgets their
direct predictions do not reach the same error level. The results below therefore compare direct
prediction and short-budget adaptation on the same unseen targets after the stated offline training,
rather than ranking methods under strictly equal offline cost.

The common elements are the PDEs, unseen targets, reference solutions, independent random seeds, and
500 online Adam steps for ACR2-finetune, P2INN, and HyperPINN. The methods differ in their
underlying architectures, offline optimizers and budgets, and native deployment modes. P2INN's SVD
modulation emphasizes a small online state, HyperPINN typically performs direct inference with a
target PINN generated by its hypernetwork, and Meta-PINN uses a capped endpoint of 500 Adam plus at
most 100 L-BFGS steps on Burgers. Thus, this experiment provides a numerical comparison on common
targets under a fixed online budget, but it is neither a reproduction of every method's complete
original protocol nor a comparison under strictly equal offline cost.

\subsubsection{Main results}
\label{sec:external-results}

\begin{table*}[!tp]
\centering
\caption{\textbf{External-method comparison on common unseen targets.} ACR2-finetune, P2INN, and HyperPINN use the 500-step endpoint; Meta-PINN uses a capped endpoint of 500 Adam plus 100 L-BFGS steps on Burgers. N/R denotes an unavailable protocol-matched result, and a dash an unevaluated case.}
\label{tab:external-primary}
\footnotesize
\setlength{\tabcolsep}{3.5pt}
\renewcommand{\arraystretch}{1.10}
\begin{tabularx}{\textwidth}{>{\raggedright\arraybackslash}p{0.12\textwidth}>{\raggedright\arraybackslash}p{0.12\textwidth}*{5}{>{\centering\arraybackslash}X}}
\toprule
PDE & Metric & ACR2-finetune & P2INN full & P2INN SVD & HyperPINN & Meta-PINN\\
\midrule
Burgers & MSE & \textbf{$6.279\times10^{-8}$} & $1.636\times10^{-2}$ & $1.750\times10^{-2}$ & $2.072\times10^{-5}$ & N/R\\
 & Mean $E_{L_2}$ & \textbf{$5.651\times10^{-4}$} & $0.3112$ & $0.3155$ & $0.007333$ & $0.003036$\\
Allen--Cahn & MSE & \textbf{$1.502\times10^{-4}$} & $0.04657$ & $0.03667$ & $0.03303$ & ---\\
 & Mean $E_{L_2}$ & \textbf{$0.01284$} & $0.3068$ & $0.2770$ & $0.2545$ & ---\\
Kovasznay & Macro-averaged MSE & \textbf{$4.979\times10^{-6}$} & $0.001376$ & $0.002955$ & $0.002604$ & ---\\
 & Macro-averaged $E_{L_2}$ & \textbf{$0.04249$} & $0.5892$ & $0.8945$ & $0.8919$ & ---\\
Poisson--Boltzmann & MSE & \textbf{$3.379\times10^{-6}$} & $0.07011$ & $0.08207$ & $3.921\times10^{-4}$ & ---\\
 & Mean $E_{L_2}$ & \textbf{$0.005667$} & $0.5838$ & $0.6320$ & $0.04493$ & ---\\
\bottomrule
\end{tabularx}
\end{table*}

\begin{figure*}[!tp]
\centering
\includegraphics[width=\externalendpointwidth]{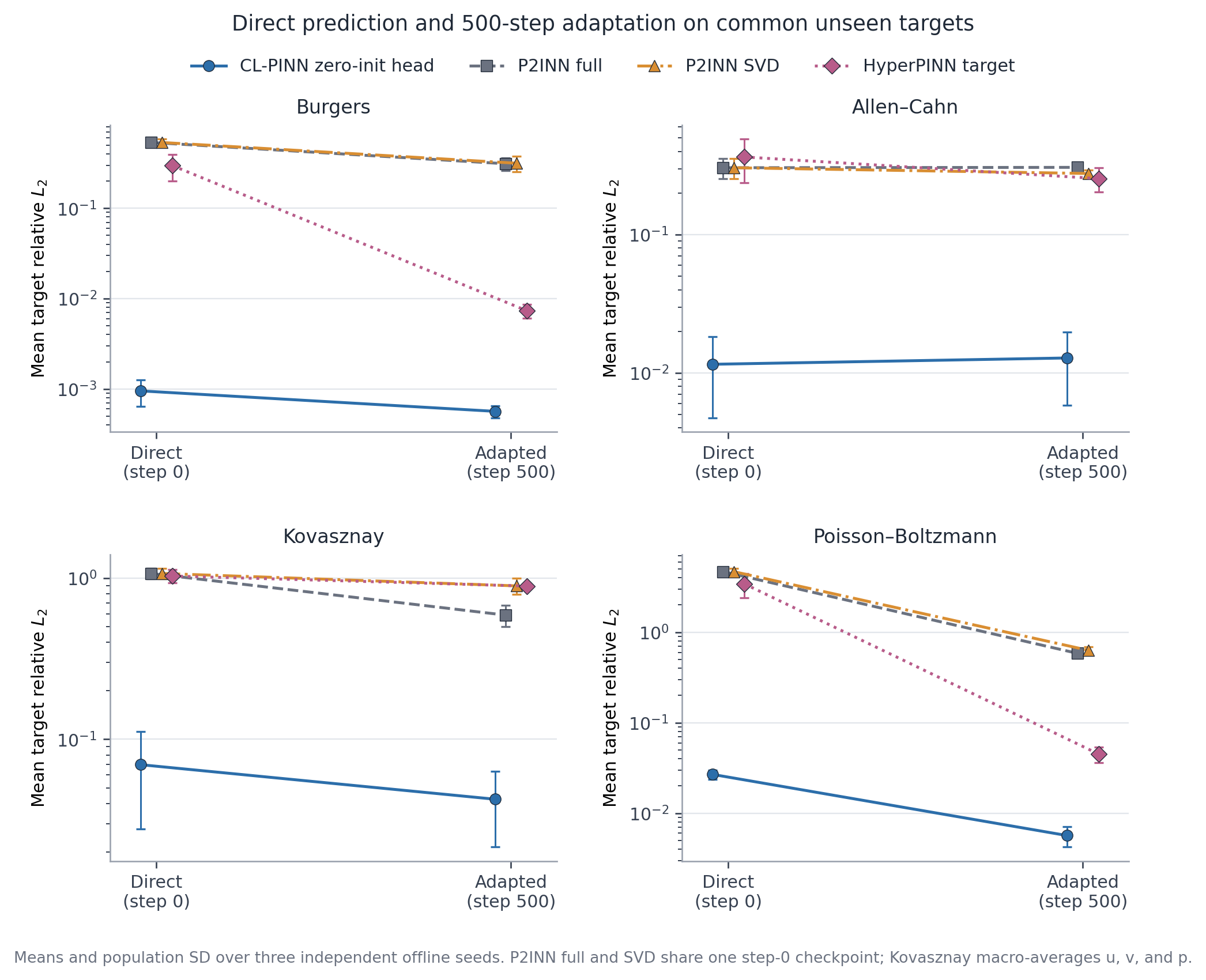}
\caption{\textbf{Mean target $E_{L_2}$ from direct prediction (step 0) to 500-step adaptation under the common four-PDE protocol.} Markers and intervals show the mean and population standard deviation on a logarithmic scale; Kovasznay is macro-averaged over $u$, $v$, and $p$. Meta-PINN uses a different capped endpoint and appears only in Table~\ref{tab:external-primary}.}
\label{fig:external-four-pde-endpoint}
\end{figure*}

Table~\ref{tab:external-primary} and Figure~\ref{fig:external-four-pde-endpoint} show that the
frozen ACR model already has lower direct-prediction errors at step 0. After 500 steps,
ACR2-finetune still gives the lowest MSE and mean $E_{L_2}$ on all four PDEs, with mean $E_{L_2}$
values approximately $5.37$--$19.8\times$ lower than the most accurate external endpoint for each
equation. A more accurate parameterized pretrained representation therefore provides a favorable
starting point for single-parameter adaptation, enabling the compact residual head to reach high
accuracy within a limited online budget. Although Allen--Cahn degrades slightly over the 500
adaptation steps, its fixed endpoint remains more accurate than those of the external methods.

This result does not imply that the external methods are intrinsically inferior. P2INN, HyperPINN,
and Meta-PINN use different representations and deployment modes and retain their own resource
advantages: P2INN SVD updates only 192--230 variables, HyperPINN and Meta-PINN have shorter recorded
times, and their offline costs may be lower than our global pretraining with L-BFGS. The present
evidence therefore supports higher accuracy and effective short-budget adaptation for ACR2-finetune
on the common targets and fixed online endpoints, but not a comprehensive advantage in offline cost,
online state size, or wall-clock time.

\FloatBarrier
\subsection{Ablation and sensitivity studies}
\label{sec:ablations}

The preceding experiments establish overall accuracy and online cost. This section analyzes three
separable questions on Schaffer-like, Burgers, Allen--Cahn, Kovasznay, and Poisson--Boltzmann: (1)
the respective roles of Bayesian parameter selection and dynamic weighting; (2) whether
fixed-capacity sparse replay mitigates forgetting of earlier parameter tasks; and (3) whether the
representation effect of the parameter subnetwork can be separated from parameter-branch Adam decay
and L-BFGS-stage freezing. Each comparison uses prespecified test sets and aggregation rules.
Incomplete repeated cells, non-finite outcomes, and unequal-capacity references are labeled
explicitly and are not given the same causal weight as complete controlled results.

\subsubsection{Bayesian selection and dynamic weighting within AC}
\label{sec:ablation-active}

This ablation independently enables or disables Bayesian parameter selection (BO) and task-wise loss
weighting while holding the network architecture, active-task capacity, and optimization budget
fixed. BO constructs a surrogate from queried physics losses and selects new parameter tasks; when a
case uses a search prior, $f_{\mathrm{prior}}$ shapes the BO observation score $S_{\mathrm{BO}}$.
For tasks already admitted to training, the loss weights are jointly controlled by
$f_{\mathrm{prior}}$ and the static and dynamic terms. This $2\times2$ design separately examines
the roles of parameter selection and training weights; the case-specific choices are stated in the
benchmark settings above.

\begin{table*}[!tp]
\centering
\caption{\textbf{Macro/worst relative $L_2$ in the within-case controlled $2\times2$ ablation; each cell gives the macro mean followed by the worst-task mean.}}
\label{tab:ablation-active-five}
\small
\setlength{\tabcolsep}{3.5pt}
\renewcommand{\arraystretch}{1.10}
\begin{tabularx}{\textwidth}{>{\raggedright\arraybackslash}p{0.13\textwidth}*{4}{>{\centering\arraybackslash}X}}
\toprule
Case & Fixed selection, equal weighting & BO only, equal weighting & Fixed selection + training weighting & BO + case-specific weighting\\
\midrule
Schaffer-like & 0.9987 / 10.572 & 2.5745 / 27.672 & 0.1671 / \textbf{1.2314} & \textbf{0.16007} / 1.2987\\
Burgers & 0.02126 / 0.2002 & \textbf{0.01961 / 0.1027} & 0.02578 / 0.2689 & 0.02123 / 0.2846\\
Allen--Cahn & 0.1629 / 0.5877 & \textbf{0.03796 / 0.1036} & 0.1642 / 0.5978 & 0.04600 / 0.1228\\
Kovasznay & 0.2592 / 2.256 & \textbf{0.1284 / 1.826} & 0.3344 / 3.331 & 0.3641 / 5.286\\
Poisson--Boltzmann & 12.667 / 55.01 & 0.09499 / 0.6146 & 11.862 / 64.73 & \textbf{0.06065 / 0.3749}\\
\bottomrule
\end{tabularx}
\end{table*}

Table~\ref{tab:ablation-active-five} shows that training weighting has the strongest effect on
Schaffer-like, where enabling both BO and weighting gives the lowest macro-average error. BO alone
performs best on Burgers, Allen--Cahn, and Kovasznay, indicating that locating difficult parameter
tasks is more important than further adjusting their training weights for these PDEs.
Poisson--Boltzmann performs best when BO and weighting are both enabled. Overall, the benefits of
Bayesian selection and loss weighting depend on the error distribution and optimization properties
of the equation; no fixed combination is optimal for every problem.

This controlled result separates the effects of the two mechanisms: BO changes which tasks are
queried and admitted, whereas the training weights redistribute optimization effort among admitted
tasks; neither mechanism is uniformly beneficial across all five cases.

\begin{table*}[!tp]
\centering
\caption{\textbf{Complete objective-loss queries and parameter coverage of AG and AC at the same final active-task capacity.}}
\label{tab:ablation-active-query}
\small
\setlength{\tabcolsep}{3.5pt}
\renewcommand{\arraystretch}{1.10}
\begin{tabularx}{\textwidth}{
>{\raggedright\arraybackslash}p{0.22\textwidth}
>{\centering\arraybackslash}p{0.12\textwidth}
>{\centering\arraybackslash}p{0.12\textwidth}
>{\centering\arraybackslash}p{0.14\textwidth}
>{\centering\arraybackslash}X}
\toprule
Case & \makecell[c]{AG\\queries} & \makecell[c]{AC\\queries} & Reduction & \makecell[c]{Maximum coverage\\hole, AG/AC}\\
\midrule
Schaffer-like & 580 & 429 & 26.0\% & 0.100 / 0.167\\
Burgers & 1010 & 172 & 83.0\% & 0.135 / 0.185\\
Allen--Cahn & 960 & 285 & 70.3\% & 0.495 / 0.500\\
Kovasznay & 1010 & 224 & 77.8\% & 0.094 / 0.101\\
Poisson--Boltzmann & 103760 & 8400 & 91.9\% & 0.682 / 0.630\\
\bottomrule
\end{tabularx}
\end{table*}

Table~\ref{tab:ablation-active-query} shows that, at the same final active-task capacity, AC reduces
complete objective-loss queries by $26.0\%$--$91.9\%$ relative to grid-greedy AG, with the largest
reduction on the four-parameter Poisson--Boltzmann problem. Although Bayesian search iteratively
updates its surrogate, it does not repeatedly query every candidate at each selection event. Its
advantage therefore becomes more pronounced as the parameter dimension and per-query cost increase.
Query reduction need not produce the same proportional reduction in wall-clock time because fitting
the surrogate also incurs computation. Nevertheless, the same principle naturally extends to
supervised active learning: when each query requires an expensive external solver or experiment to
provide a reference solution, fewer queries directly improve data-acquisition efficiency.

\subsubsection{Experience-replay ablation}
\label{sec:ablation-replay}

This ablation compares three treatments of earlier tasks: no replay, fixed-capacity sparse replay,
and uncapped full replay. The primary controlled contrast is no replay versus sparse replay. Sparse
replay stores physical collocation coordinates rather than solution labels and recomputes the
governing-equation and condition residuals during later training. Approximately $10\%$ of the
sparsifiable physics-point pool is retained, while non-sparsifiable boundary or initial-condition
points are handled according to the benchmark; the total retained fraction therefore need not be
exactly $10\%$. Uncapped full replay retains all earlier tasks and therefore uses more parameter
tasks and computation; it is a capacity reference rather than an equal-budget competitor.

\begin{figure*}[!tp]
\centering
\includegraphics[width=\compactdiagnosticwidth]{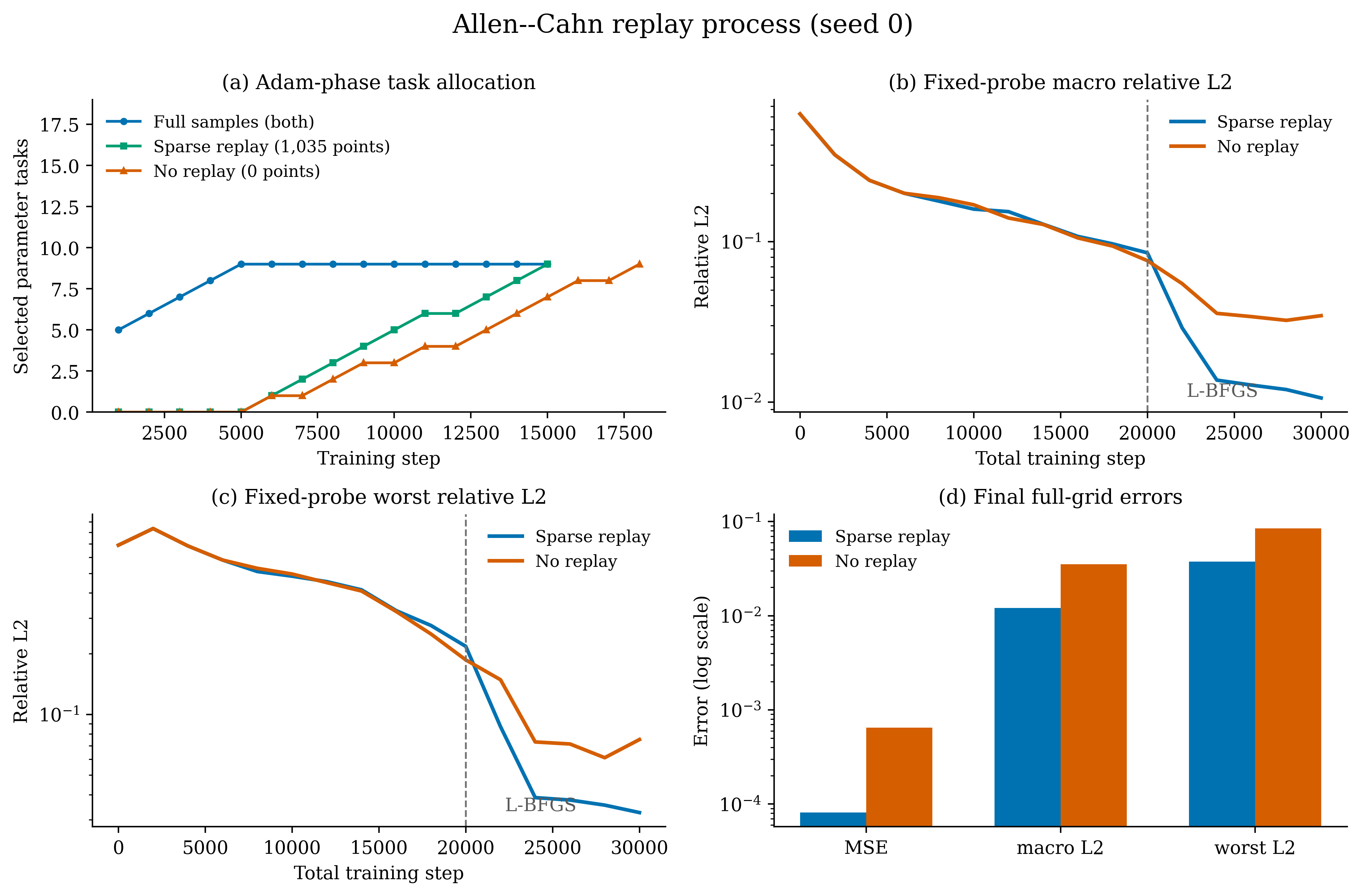}
\caption{\textbf{Process diagnostic of sparse versus no replay on Allen--Cahn.} Panel (a) shows task sample states, panels (b,c) fixed-probe macro/worst relative $L_2$, and panel (d) the final parameter grid. The dashed line marks the Adam/L-BFGS transition; this single run illustrates the mechanism.}
\label{fig:allen_replay_process}
\end{figure*}

Macro relative $L_2$ is the primary metric. It first evaluates each parameter--output task
separately and then assigns equal weight to all tasks, preventing severe degradation of a few
earlier tasks from being obscured by tasks with more spatial points or larger numerical amplitudes.
Tail forgetting is further examined through the worst relative $L_2$ in
Figure~\ref{fig:allen_replay_process}.

\begin{table*}[!tp]
\centering
\caption{\textbf{Macro relative $L_2$ for the five-case experience-replay ablation.} Uncapped full replay is not budget matched to the first two columns.}
\label{tab:ablation-replay-five}
\small
\setlength{\tabcolsep}{3.5pt}
\renewcommand{\arraystretch}{1.10}
\begin{tabular}{lrrr}
\toprule
Case & No replay & Fixed-capacity sparse replay & Uncapped full replay\\
\midrule
Schaffer-like & 0.16007 & \textbf{0.07195} & 0.12642\\
Burgers & 0.02123 & 0.01025 & \textbf{0.00751}\\
Allen--Cahn & 0.04600 & 0.01467 & \textbf{0.00908}\\
Kovasznay & 0.5555 & \textbf{0.07925} & 0.1148\\
Poisson--Boltzmann & 0.03449 & 0.03046 & \textbf{0.02730}\\
\bottomrule
\end{tabular}
\end{table*}

Table~\ref{tab:ablation-replay-five} shows that fixed-capacity sparse replay outperforms no replay
on all five cases, reducing macro $L_2$ by 55.0\%, 51.7\%, 68.1\%, 85.7\%, and 11.7\%, respectively.
The Schaffer-like and Burgers contrasts are strictly protocol matched. The Allen--Cahn runs also
differ in resampling period (2000 versus 1000), so this case supports improvement under the combined
protocol but not a strict single-factor attribution. The consistent direction across all five cases
indicates that retaining a small set of physics constraints from earlier tasks improves overall
retention after sequential training, although the magnitude is equation dependent.

Fixed-capacity sparse replay retains only about $10\%$ of the sparsifiable physics points, yet it
clearly outperforms no replay on all five cases. It even outperforms uncapped full replay on
Schaffer-like and Kovasznay. More replay is therefore not necessarily better: continuously retaining
many tasks that have already been learned well can consume subsequent optimization capacity and
interfere with learning new tasks. By selectively preserving earlier-task constraints, sparse replay
provides a better balance between knowledge retention and adaptation to new tasks.

\begin{figure*}[!tp]
\centering
\includegraphics[width=\compactdiagnosticwidth]{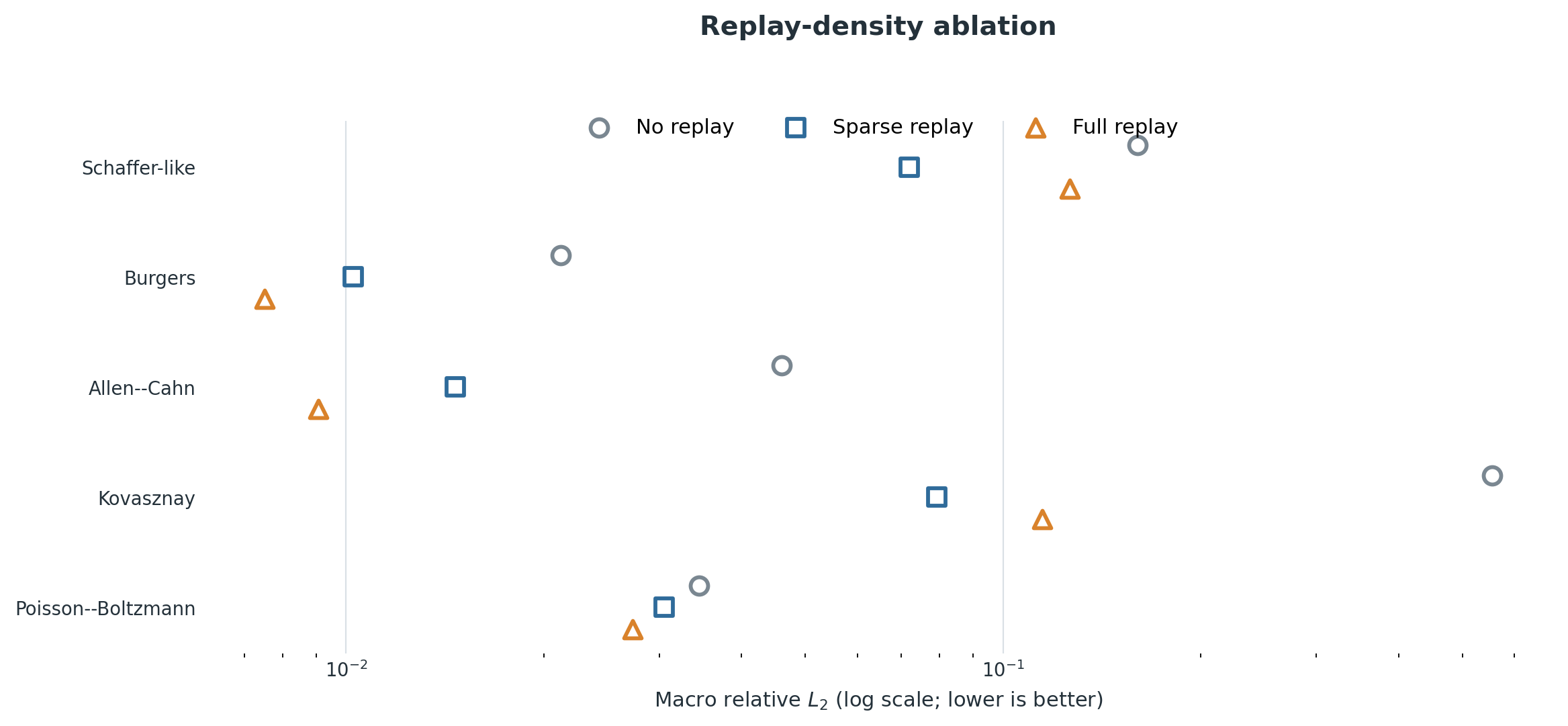}
\caption{\textbf{Macro relative $L_2$ for no replay, fixed-capacity sparse replay, and uncapped full replay across five cases.} The horizontal axis is logarithmic, and the three marker types denote the replay settings; uncapped full replay is a capacity reference.}
\label{fig:ablation-replay-five}
\end{figure*}

Figure~\ref{fig:ablation-replay-five} makes this equation dependence explicit. The shift from no
replay to sparse replay approaches one order of magnitude on Kovasznay, whereas the three
Poisson--Boltzmann settings are much closer. Sparse replay also lies to the left of uncapped replay
on Schaffer-like and Kovasznay. The cross-equation result is therefore that sparse replay
consistently improves on no replay, not that it provides a fixed improvement factor or that denser
replay is always better.

Accordingly, the supported replay conclusion is limited to improved retention relative to no
replay under the stated protocols; uncapped full replay is an extra-capacity reference rather than an
equal-budget baseline.

To illustrate the replay mechanism more directly, Figure~\ref{fig:allen_replay_process} examines one
representative Allen--Cahn run. Under the same active-task capacity, sparse replay retains a small
number of physics constraints for tasks that leave the active set, whereas no replay stops using
samples from those tasks. The two error trajectories remain similar during Adam but gradually
separate after the transition to L-BFGS; sparse replay ultimately gives both lower macro-average and
lower worst-task errors. This case illustrates how a small number of earlier-task constraints can
preserve learned knowledge during later optimization, but the process curves are used only to
explain the mechanism and not as an independent single-factor statistical conclusion.

\subsubsection{Parameter subnetwork and regularization-control ablation}
\label{sec:ablation-subnetwork}

This section contains two independent ablations. The first compares a direct-concatenation FNN with
ParamFNN under approximately matched parameter counts and otherwise unchanged training settings,
isolating the representation effect of the parameter subnetwork. The second holds ParamFNN fixed and
independently enables or disables parameter-branch decay during Adam and parameter-branch freezing
during L-BFGS, thereby examining the two optimizer controls. Because the two experiments answer
different questions, they are interpreted separately. In the main results, ACR2-arch denotes only
the global model using ParamFNN; the effects of decay and freezing are evaluated in the second
ablation.

\paragraph{Capacity-matched architecture control}
\label{sec:ablation-architecture}

\begin{table*}[!tp]
\centering
\caption{\textbf{MSE / macro relative $L_2$ / worst relative $L_2$ in the five-case capacity-matched architecture control; bold identifies the lower value for each case and metric.}}
\label{tab:ablation-subnetwork-architecture}
\small
\setlength{\tabcolsep}{3.5pt}
\renewcommand{\arraystretch}{1.10}
\begin{tabularx}{\textwidth}{>{\raggedright\arraybackslash}p{0.14\textwidth}*{2}{>{\centering\arraybackslash}X}}
\toprule
Case & Capacity-matched direct-concatenation FNN & Capacity-matched ParamFNN control\\
\midrule
Schaffer-like & $3.3731\times10^{-4} / 0.10424 / 0.68699$ & $\mathbf{2.9073\times10^{-4} / 0.05831 / 0.22469}$\\
Burgers & $1.722\times10^{-4} / 0.004277 / 0.1697$ & $\mathbf{2.390\times10^{-5} / 0.001771 / 0.07735}$\\
Allen--Cahn & $\mathbf{3.246\times10^{-4} / 0.02203 / 0.08343}$ & $4.844\times10^{-4} / 0.02439 / 0.09159$\\
Kovasznay & $0.001908 / \mathbf{0.05211} / \mathbf{0.4697}$ & $\mathbf{0.001584} / 0.1052 / 0.8636$\\
Poisson--Boltzmann & $1.788\times10^{-3} / 0.04433 / 0.7910$ & $\mathbf{2.739\times10^{-4} / 0.04240 / 0.2039}$\\
\bottomrule
\end{tabularx}
\end{table*}

Table~\ref{tab:ablation-subnetwork-architecture} shows that ParamFNN simultaneously reduces MSE,
macro-average relative $L_2$, and worst-task relative $L_2$ on Schaffer-like, Burgers, and
Poisson--Boltzmann, with particularly clear improvements in tail error for some cases. The benefit
is not universal: the direct-concatenation network is lower on all three Allen--Cahn metrics, while
the MSE and relative-error rankings disagree on Kovasznay. The parameter subnetwork should therefore
be interpreted as an equation-dependent conditional representation bias. It can substantially
improve overall accuracy or tail stability for some problems, but it is not a universally optimal
replacement for direct concatenation.

\paragraph{Parameter-branch optimizer controls}
\label{sec:ablation-optimizer}

A complete finite three-seed $2\times2$ control under a common protocol is available for the
four cases reported below. Burgers is excluded from this strict table because its historical control
cells changed additional protocol factors and the freeze-containing settings included non-finite
runs; we therefore do not form a partial finite-seed mean or attribute a single-factor effect.

\begin{table*}[!tp]
\centering
\caption{\textbf{Strict $2\times2$ ablation of parameter-branch Adam decay and L-BFGS-stage freezing conditional on ParamFNN; bold denotes the row minimum.}}
\label{tab:ablation-optimizer-control}
\small
\setlength{\tabcolsep}{3.5pt}
\renewcommand{\arraystretch}{1.10}
\begin{tabularx}{\textwidth}{>{\raggedright\arraybackslash}p{0.13\textwidth}*{4}{>{\centering\arraybackslash}X}}
\toprule
Case & ACR2-arch (main configuration) & + Adam decay & + L-BFGS freeze & + decay and freeze\\
\midrule
Schaffer-like & $0.05073\pm0.00639$ & $\mathbf{0.04173\pm0.01422}$ & $0.05304\pm0.00200$ & $0.10746\pm0.00265$\\
Allen--Cahn & $0.02439\pm0.00479$ & $0.02717\pm0.00532$ & $0.02371\pm0.00645$ & $\mathbf{0.01828\pm0.00160}$\\
Kovasznay & $0.1052\pm0.0163$ & $\mathbf{0.07383\pm0.01655}$ & $0.4152\pm0.3824$ & $0.1194\pm0.0120$\\
Poisson--Boltzmann & $0.04397\pm0.02487$ & $\mathbf{0.02926\pm0.00174}$ & $0.05285\pm0.02788$ & $0.03350\pm0.00274$\\
\bottomrule
\end{tabularx}
\end{table*}

Table~\ref{tab:ablation-optimizer-control} shows that Adam decay alone gives the lowest error on
Schaffer-like, Kovasznay, and Poisson--Boltzmann, indicating that gradual regularization of the
parameter branch provides most of the benefit in the majority of controlled cases. Freezing the
parameter branch alone does not yield a consistent advantage and causes clear degradation on
Kovasznay. Allen--Cahn is an exception, for which decay and freezing together perform best. In
general, freezing during L-BFGS can prevent the parameter branch from continuing to adapt and should
therefore be validated for each equation. The current parameter-subnetwork capacity and decay
settings were chosen empirically, and the limited configuration study does not establish globally
optimal hyperparameters.

The two controls therefore support an equation-dependent representation and regularization
effect, not universal superiority of ParamFNN, decay, freezing, or their combination.

\subsubsection{Additional sensitivity diagnostics}
\label{sec:additional-sensitivity}

Beyond the three component ablations, we conducted additional diagnostics of the prior function,
ParamFNN depth/width, and the mini-batch resource trade-off.
Table~\ref{tab:additional-sensitivity-summary} reports the representative evidence needed to delimit
the conclusions. These diagnostics characterize robustness and configuration boundaries rather than
extending the primary accuracy claims.

\begin{table*}[!tp]
\centering
\caption{\textbf{Representative prior, ParamFNN depth/width, and mini-batch resource sensitivities; comparisons are restricted to each diagnostic.}}
\label{tab:additional-sensitivity-summary}
\footnotesize
\setlength{\tabcolsep}{3.5pt}
\renewcommand{\arraystretch}{1.10}
\begin{tabularx}{\textwidth}{>{\raggedright\arraybackslash}p{0.13\textwidth}>{\raggedright\arraybackslash}X>{\raggedright\arraybackslash}X}
\toprule
Diagnostic & Representative evidence & Concise conclusion\\
\midrule
Prior function & Separate Burgers diagnostic, original / unit / reversed prior: $0.06724/0.02069/0.01292$; Kovasznay configured prior: $0.1194$, versus $0.1751$--$0.7738$ for the other variants & The best prior direction changes with the equation; both prior direction and injection location require problem-specific validation\\
ParamFNN depth--width & The three structures differ in parameter count by at most about $2.2\%$; $4\times27$ has the lowest macro/worst $L_2$, whereas $3\times30$ has lower MSE and mean maximum absolute error & Even at nearly equal capacity, mean and tail errors trade off; no universal structure can be selected from a single metric\\
Mini-batch resources & At fixed point exposure, moving from full batch to batch $=2500$ changes peak training memory from $999.1$ to $64.6$ MiB, wall time from $26.6$ to $295.5$ s, and relative $L_2$ from $0.01374$ to $0.001341$ & Mini-batching greatly reduces training memory but increases wall time; the unbatched evaluation peak remains about $923$ MiB\\
\bottomrule
\end{tabularx}
\end{table*}

The Burgers values come from a separate prior-direction diagnostic rather than the unit-prior
primary configuration; that diagnostic changes both the training prior and BO score shaping, whereas
the Kovasznay diagnostic changes only the training prior. Burgers favors the reversed diagnostic
prior, while Kovasznay favors its configured prior. The defensible conclusion is therefore that both
prior direction and injection location require equation-specific validation. The ParamFNN diagnostic
likewise shows that near-capacity-matched depth and width choices redistribute average and tail
errors. In the mini-batch diagnostic, reducing the batch from full to 2,500 lowers training-phase
peak memory by approximately $93.5\%$ but increases wall time by about $11.1\times$. This
fixed-exposure, single-parameter Burgers experiment does not establish an equal-time accuracy
advantage for end-to-end ParamPINN training.

Taken together, the five-case ablations show that the CL-PINN components play complementary but
non-equivalent roles. Bayesian selection primarily improves the efficiency of finding difficult
parameter tasks, dynamic weighting regulates training imbalance across tasks, sparse replay
mitigates forgetting during sequential training, and the parameter subnetwork and its optimizer
controls affect conditional representation. Their benefits vary with the equation and evaluation
metric, so the components should be selected and validated for the problem at hand rather than
assumed to improve performance monotonically when combined.

In particular, the neutral prior remains a valid operating point, mini-batching is compatible
with the framework, and component and architecture choices should be validated for the target PDE
rather than treated as universal defaults.

\FloatBarrier
\section{Conclusions}
\label{sec:conclusions}

We propose CL-PINN, a continual-learning framework for training parameterized physics-informed
neural networks without observational data and under limited computational resources. Bayesian
parameter selection improves the efficiency of acquiring training tasks, dynamic weighting mitigates
optimization imbalance among parameter tasks, and sparse physics-constrained replay preserves
previously learned tasks. A parameter subnetwork further enhances conditional representation of
spatial variables and equation parameters. For a specified new parameter, a lightweight residual
head can also be trained while the global model remains frozen, enabling rapid single-parameter
adaptation.

Experiments on one continuous-function benchmark and four parameterized PDEs show that CL-PINN can
locate difficult parameter tasks with fewer expensive queries and improve cross-parameter accuracy
and task retention through dynamic weighting and sparse replay. Sparse replay retains only a small
number of physics points from earlier tasks yet outperforms no replay in all five cases. The
parameter subnetwork also improves overall accuracy or tail stability on several problems. The
ablations further show that these effects are equation dependent and that no fixed combination is
optimal for every problem.

The single-parameter adaptation results show that an accurate parameterized pretrained model
provides a favorable starting point for short-budget adaptation and substantially reduces prediction
error at many target parameters. Under the common protocol, ACR2-finetune attains lower
reference-solution error at the fixed adaptation endpoint on all four PDEs, whereas the external
methods retain advantages in online trainable state, recorded online time, or offline cost. The
results therefore support high adaptation accuracy under a limited online budget rather than
superiority across all computational-resource measures.

CL-PINN is a parameter-task-level training-resource allocation framework and can be combined with
hard initial or boundary constraints, adaptive spatial collocation, loss balancing, and alternative
PINN architectures. Its current limitations include the imperfect correspondence between physics
loss and reference-solution error, the equation dependence of component and hyperparameter effects,
and validation primarily on fixed continuous parameter domains. Future work will investigate more
reliable task-uncertainty measures and reuse of historical information, and extend parameter
selection and physics-constrained replay to higher-dimensional parameter domains, tasks with
observational data, and physics-informed operator-learning models.

\section{Acknowledgement}

This research did not receive any specific grant from funding agencies in the public, commercial, or
not-for-profit sectors.

\section*{Declaration of generative AI and AI-assisted technologies in the manuscript preparation process}

During the preparation and revision of this work, the authors used ChatGPT and Codex (OpenAI) to
assist with drafting and checking code, organizing and summarizing author-generated experimental
outputs, and improving language, readability, and bilingual translation. The research questions,
methodology, experimental design, scientific interpretation, and conclusions were developed and
decided by the authors. After using these tools, the authors reviewed, edited, and verified the
outputs against the source code and underlying data as appropriate, and take full responsibility for
the content of the published article.

\bibliographystyle{elsarticle-num}
\bibliography{references}
\end{document}


\renewcommand{\thesection}{S.\arabic{section}}
\renewcommand{\thesubsection}{S.\arabic{section}.\arabic{subsection}}
\renewcommand{\thesubsubsection}{S.\arabic{section}.\arabic{subsection}.\arabic{subsubsection}}
\renewcommand{\thefigure}{S.\arabic{figure}}
\renewcommand{\thetable}{S.\arabic{table}}
\renewcommand{\theequation}{S.\arabic{equation}}
\numberwithin{equation}{section}
\numberwithin{table}{section}
\numberwithin{figure}{section}

\begin{frontmatter}

\title{Supplementary Material for ``Continual-Learning Physics-Informed Neural Networks for Parameterized Partial Differential Equations''}

\author[thu]{Xujia Chen\corref{cor1}}
\ead{chenxj20@mails.tsinghua.edu.cn}
\author[thu]{Xinyue Hu}
\ead{hu-xy25@mails.tsinghua.edu.cn}
\author[thu]{Letian Chen}
\ead{clt21@mails.tsinghua.edu.cn}
\author[thu]{Yi Liu}
\ead{yiliu@tsinghua.edu.cn}
\author[thu]{Wenhui Fan}
\ead{fanwenhui@tsinghua.edu.cn}
\affiliation[thu]{organization={Department of Automation, Tsinghua University},
  city={Beijing},
  postcode={100084},
  country={China}}
\cortext[cor1]{Corresponding author}

\begin{abstract}

This supplement is organized into four parts. Part I reports the reproducibility protocol, case-specific implementation settings, reference solutions and evaluation grids, and the source-code entry point. Part II illustrates three practical challenges of parameterized PINNs: computational cost, parameter-wise accuracy imbalance, and parameter-space generalization. Following the experimental sections of the main text, Part III provides additional evidence for single-parameter adaptation, external-method comparisons, ablations and sensitivity analyses, training cost, and search overhead. Part IV consolidates the complete parameter-wise error atlas and optimization histories for all six methods. Unless otherwise stated, formal aggregate results use FP32, the fixed random seeds $[0,1,2]$, and the population standard deviation.

\end{abstract}

\end{frontmatter}

\tableofcontents

\clearpage

\section{Reproducibility and implementation details}

\label{sec:supp-part-1}

This part consolidates the method labels, randomness controls, software and hardware, network and sampling settings, search parameters, reference solutions, and evaluation grids required to reproduce the experiments.

\FloatBarrier

\subsection{Method labels, statistics, and common protocol}

\label{sec:supp-1-1}

UNI denotes uniform parameter sampling, FIX a prespecified parameter set, and AG grid-greedy parameter selection. AC combines Bayesian parameter-task selection with task-wise dynamic loss weighting, and ACR adds fixed-capacity sparse physics-constrained replay. \textbf{ACR2-arch} refers specifically to the ParamFNN configuration used in the primary multi-parameter results. The full ACR2 configuration also includes parameter-branch weight decay during Adam and parameter-branch freezing during L-BFGS; these optimizer controls are discussed only in the ablation study. \textbf{ACR2-finetune} is an optional downstream single-parameter adaptation applied after global training: the global model is frozen and only an added residual head is optimized. It can, in principle, be applied after AC, ACR, or ACR2-arch; the experiments in this study use a frozen ACR model. ACR2-finetune and ACR2-arch therefore have distinct meanings.

Unless otherwise stated, formal comparisons use FP32 and fixed random seeds $[0,1,2]$, and report the arithmetic mean and population standard deviation. All valid seeds are included in the aggregate results; non-finite runs are reported separately as stability diagnostics rather than mixed into finite-sample means.

\begin{table}[!htbp]
\centering
\caption{Requested optimizer-step limits and training samples per parameter for each case. L-BFGS may converge and terminate before reaching the requested limit.}
\label{tab:supp-protocol}
\small
\setlength{\tabcolsep}{4pt}
\renewcommand{\arraystretch}{1.16}
\begin{adjustbox}{max width=\linewidth,center}
\begin{tabular}{lrrr}
\toprule
\textbf{Case} & \textbf{Adam limit} & \textbf{L-BFGS limit} & \textbf{Training samples/parameter}\\
\midrule
Schaffer-like & 20,000 & 20,000 & 2,500\\
Burgers & 20,000 & 20,000 & 5,900\\
Allen--Cahn & 20,000 & 10,000 & 10,082\\
Kovasznay & 10,000 & 10,000 & 3,474\\
Linearized Poisson--Boltzmann & 20,000 (UNI/FIX) or 40,000 (dynamic methods) & 20,000 & 5,041\\
\bottomrule
\end{tabular}
\end{adjustbox}
\end{table}

\begin{table}[!htbp]
\centering
\caption{Software and hardware environment, optimizers, and deterministic settings used in the formal experiments. Wall times in ordinary run logs are not treated as exclusive-GPU timing measurements.}
\label{tab:supp-environment}
\footnotesize
\setlength{\tabcolsep}{4pt}
\renewcommand{\arraystretch}{1.16}
\begin{tabularx}{\linewidth}{@{}p{0.23\linewidth}X@{}}
\toprule
\textbf{Item} & \textbf{Setting}\\
\midrule
Software & Python 3.10.13; NumPy 1.26.4; scikit-learn 1.4.1.post1\\
Backend & PyTorch 2.4.1+cu121; CUDA runtime 12.1; DeepXDE PyTorch backend\\
DeepXDE provenance & Read-only snapshot \texttt{\detokenize{v1.15.0-7-g0ffcbed}}\\
Hardware and precision & Tesla V100S-PCIE-32GB; FP32; active TF32 disabled\\
Randomness & Seeds $[0,1,2]$ applied to Python, NumPy, PyTorch, CUDA, DeepXDE, and BO\\
Determinism & cuDNN deterministic enabled; cuDNN benchmark disabled\\
Adam & Learning rate $10^{-3}$\\
PyTorch L-BFGS & history size 100; gradient tolerance $10^{-8}$; change tolerance 0\\
L-BFGS execution & Function-evaluation budget $1.25\times$ requested steps; chunks up to 1000; no line search; actual steps stored separately\\
\bottomrule
\end{tabularx}
\end{table}

Tables~\ref{tab:supp-protocol} and \ref{tab:supp-environment} define the comparison scope. All five cases share the software, precision, and randomness controls, while optimizer budgets are fixed by equation. For Poisson--Boltzmann, UNI/FIX use 20k Adam updates and the dynamic methods use 40k. Dynamic methods begin with only a few parameter tasks and add later tasks progressively, so the additional updates give those later tasks sufficient optimization. This difference is included in the timing results, which are therefore not interpreted as a strictly equal-budget comparison.

\FloatBarrier

\subsection{Case-specific implementation settings}

\label{sec:supp-1-2}

The following settings are taken from the locked formal runs. Network notation is input dimension--(number of hidden layers $\times$ width)--output dimension. Sparse replay retains approximately $10\%$ of the sparsifiable dense physics-point pool; initial-condition, boundary-condition, and other samples that cannot be sparsified at the same ratio are retained according to the case-specific protocol. The replay and resampling settings of ACR2-arch are stated in the main text and the table below, so this configuration must not be interpreted simply as ACR with only the network architecture replaced.

\begin{table}[!htbp]
\centering
\caption{Parameter domains, network architectures, training samples per parameter, and active/replay task capacities for the five cases.}
\label{tab:supp-case-settings}
\footnotesize
\setlength{\tabcolsep}{4pt}
\renewcommand{\arraystretch}{1.16}
\begin{adjustbox}{max width=\linewidth,center}
\begin{tabular}{llllr}
\toprule
\textbf{Case} & \textbf{Parameter domain} & \textbf{Network} & \textbf{Samples per parameter} & \textbf{Active/replay capacity}\\
\midrule
Schaffer-like & $a\in[0.001,1]$ & $3$--$(4\times30)$--$1$ & 2,500 & 15/15\\
Burgers & $\nu\in[0.01,1]$ & $3$--$(4\times50)$--$1$ & 5,900 & 9/9\\
Allen--Cahn & $(\nu,\rho)\in[0.001,0.1]\times[3,5]$ & $4$--$(4\times50)$--$1$ & 10,082 & 9/9\\
Kovasznay & $Re\in[5,500]$ & $3$--$(4\times50)$--$3$ & 3,474 & 10/10\\
Linearized Poisson--Boltzmann & $(\mu_1,\mu_2,k,A)\in[1,2]\times[3,4]\times[5,10]\times[5,15]$ & $6$--$(5\times30)$--$1$ & 5,041 & 80/80\\
\bottomrule
\end{tabular}
\end{adjustbox}
\end{table}

\begin{table}[!htbp]
\centering
\caption{Resampling periods, Bayesian-query settings, and dynamic-weighting configurations for the five cases. The resampling periods correspond to AC, ACR, and ACR2-arch, respectively; $f_{\mathrm{prior}}$ denotes the training prior, and only Schaffer-like additionally uses a nonuniform search prior.}
\label{tab:supp-search-settings}
\scriptsize
\setlength{\tabcolsep}{4pt}
\renewcommand{\arraystretch}{1.16}
\begin{adjustbox}{max width=\linewidth,center}
\begin{tabular}{lrrrrr}
\toprule
\textbf{Case} & \textbf{$N_{\mathrm{resample}}$ (AC / ACR / ACR2-arch)} & \textbf{BO evaluations per active update} & \textbf{$\kappa$} & \textbf{Training prior $f_{\mathrm{prior}}$} & \textbf{$(\lambda_{\mathrm{static}},\lambda_{\mathrm{dynamic}})$}\\
\midrule
Schaffer-like & 1,000 / 1,000 / 500 & 10 & 5 & $\exp\{1.5[\log_{10}(a)+3]\}$ & $(2,-2)$\\
Burgers & 2,000 / 2,000 / 1,000 & 10 & 5 & $1$ & $(1,-1)$\\
Allen--Cahn & 2,000 / 1,000 / 1,000 & 20 & 5 & $\rho^3\exp[-2\log_{10}(\nu)]$ & $(2,-2)$\\
Kovasznay & 1,000 / 500 / 500 & 15 & 5 & $\log_{10}(Re)$ & $(1,-1)$\\
Linearized Poisson--Boltzmann & 500 / 250 / 250 & 50 & 5 & $1$ & $(1,-1)$\\
\bottomrule
\end{tabular}
\end{adjustbox}
\end{table}

Tables~\ref{tab:supp-case-settings} and \ref{tab:supp-search-settings} provide the locked case-level architecture, capacity, resampling, search, and weighting settings.

The 5,900 Burgers samples comprise 5,000 interior, 200 boundary, 400 initial-condition, and 300 near-shock anchor points; the initial-condition loss weight is 5. Allen--Cahn hard-enforces its initial condition and $u(\pm1,t)=-1$, while the linearized Poisson--Boltzmann problem uses a hard boundary constraint. The Schaffer-like training prior is also used for BO score shaping. The Burgers search transform is identical to the unweighted score because its prior is one, and the Allen--Cahn, Kovasznay, and Poisson--Boltzmann BO scores do not use a prior. The Gaussian process is rebuilt at every continuous-BO event, and its length scale is optimized during fitting; $\kappa$, the query budget, and the coverage controls remain fixed. Reference solutions are not used by any of these procedures.

\FloatBarrier

\subsection{Reference solutions and evaluation grids}

\label{sec:supp-1-3}

Analytical solutions are used where available; otherwise, pregenerated and fixed numerical reference arrays are used. Reference solutions are used only for evaluation and never for training, parameter selection, dynamic weighting, early stopping, or model selection.

\begin{table}[!htbp]
\centering
\caption{Test-parameter sets, evaluation grids, and reference-solution sources for the five cases. ``Numerical data'' denotes pregenerated and fixed reference arrays.}
\label{tab:supp-evaluation-grids}
\footnotesize
\setlength{\tabcolsep}{4pt}
\renewcommand{\arraystretch}{1.16}
\begin{tabularx}{\linewidth}{@{}p{0.16\linewidth}p{0.25\linewidth}Xp{0.16\linewidth}@{}}
\toprule
\textbf{Case} & \textbf{Tested parameter set} & \textbf{Spatial/temporal grid} & \textbf{Reference}\\
\midrule
Schaffer-like & 28 prespecified logarithmically distributed parameter values & $100\times100$ on $[-5,5)^2$, step 0.1 & Analytical\\
Burgers & 100 equally spaced $\nu\in[0.01,1]$ & $200\times100$ on $x\in[-1,1], t\in[0,1]$ & Numerical data\\
Allen--Cahn & $19\times5=95$ parameter pairs & $201\times101$ on $x\in[-1,1], t\in[0,1]$ & Numerical data\\
Kovasznay & 100 equally spaced $Re\in[5,500]$ & $100\times100$ on $x\in[-0.5,1], y\in[-0.5,1.5]$ & Analytical\\
Linearized Poisson--Boltzmann & $6^4=1296$ parameter tuples & $100\times100$ on $x,y\in[-1,1]$ & Numerical data\\
\bottomrule
\end{tabularx}
\end{table}

Table~\ref{tab:supp-evaluation-grids} defines the fixed evaluation sets. Training-time monitoring is diagnostic only and is not used for early stopping or model selection. In particular, the Poisson--Boltzmann monitor uses 16 parameter tuples, whereas the reported final evaluation covers all 1,296 tuples.

\FloatBarrier

\subsection{Data and code availability}

\label{sec:supp-1-4}

The available CL-PINN source code and selected data are publicly accessible at \href{https://github.com/pigofmomo/CLPINN}{https://github.com/pigofmomo/CLPINN}.

\FloatBarrier

\clearpage

\section{Three practical challenges in parameterized PINNs}

\label{sec:supp-part-2}

This part uses representative cases to illustrate three practical challenges of parameterized PINNs: computational cost as the number of parameter tasks grows, uneven accuracy across parameters, and generalization error caused by finite parameter sampling. Complete results are reported in Parts III and IV.

\FloatBarrier

\subsection{Computational and task-allocation cost}

\label{sec:supp-2-1}

A parameterized PINN takes both physical coordinates and parameters as inputs. If each parameter uses $N_{\mathrm{phy}}$ physical points and $N_{\mathrm{param}}$ parameter tasks are trained together, one complete pass processes approximately $N_{\mathrm{phy}}N_{\mathrm{param}}$ residual points. With a fixed number of physical points per task, residual evaluation and graph storage therefore grow approximately linearly with the number of concurrently trained parameter tasks. Higher-dimensional parameter spaces also generally require more tasks for adequate coverage.

The four-parameter linearized Poisson--Boltzmann case provides a direct example. A single-parameter PINN solves one specified parameter instance, whereas the parameterized model in this study retains up to 80 active parameter tasks concurrently.

\begin{table}[!htbp]
\centering
\caption{Approximate interior-residual-point scales of a single-parameter PINN and a parameterized PINN in the linearized Poisson--Boltzmann case. This table illustrates the scaling relation and does not report measured memory or wall-clock time.}
\label{tab:supp-practical-cost}
\small
\setlength{\tabcolsep}{4pt}
\renewcommand{\arraystretch}{1.16}
\begin{adjustbox}{max width=\linewidth,center}
\begin{tabular}{lrr}
\toprule
\textbf{Setting} & \textbf{Single-parameter PINN} & \textbf{Parameterized PINN}\\
\midrule
Interior residual points per parameter & Approximately 5,000 & Approximately 5,000\\
Concurrently trained parameter tasks & 1 & 80\\
Approximate total residual points & Approximately 5,000 & Approximately 400,000\\
\bottomrule
\end{tabular}
\end{adjustbox}
\end{table}

As Table~\ref{tab:supp-practical-cost} illustrates, increasing the active-task count from 1 to 80 raises the approximate interior-residual-point count per complete update from 5,000 to 400,000. The full four-dimensional test grid contains $6^4=1296$ parameter tuples, making it impractical to use every test parameter as a training task. CL-PINN therefore allocates training resources through active selection and replay under bounded active-task capacity. Section~\ref{sec:supp-3-3-3} reports a separate single-parameter Burgers mini-batch memory diagnostic, and Section~\ref{sec:supp-3-4} reports training time and search overhead; none of these quantities can be inferred directly from the Poisson--Boltzmann point-count ratio.

\FloatBarrier

\subsection{Parameter-wise accuracy imbalance}

\label{sec:supp-2-2}

Solutions and PDE residuals can vary substantially in scale across parameters. Two examples illustrate the resulting imbalance. Schaffer-like shows how solution magnitude can reverse the ordering of MSE and $E_{L_2}$, while Burgers shows why similar PDE residuals need not imply similar reference-solution errors.

\begin{figure}[p]
\centering
\begin{minipage}[t]{0.96\linewidth}
\centering
\textbf{$a=0.001$}\par\smallskip
\suppinclude[width=\linewidth,height=0.245\textheight]{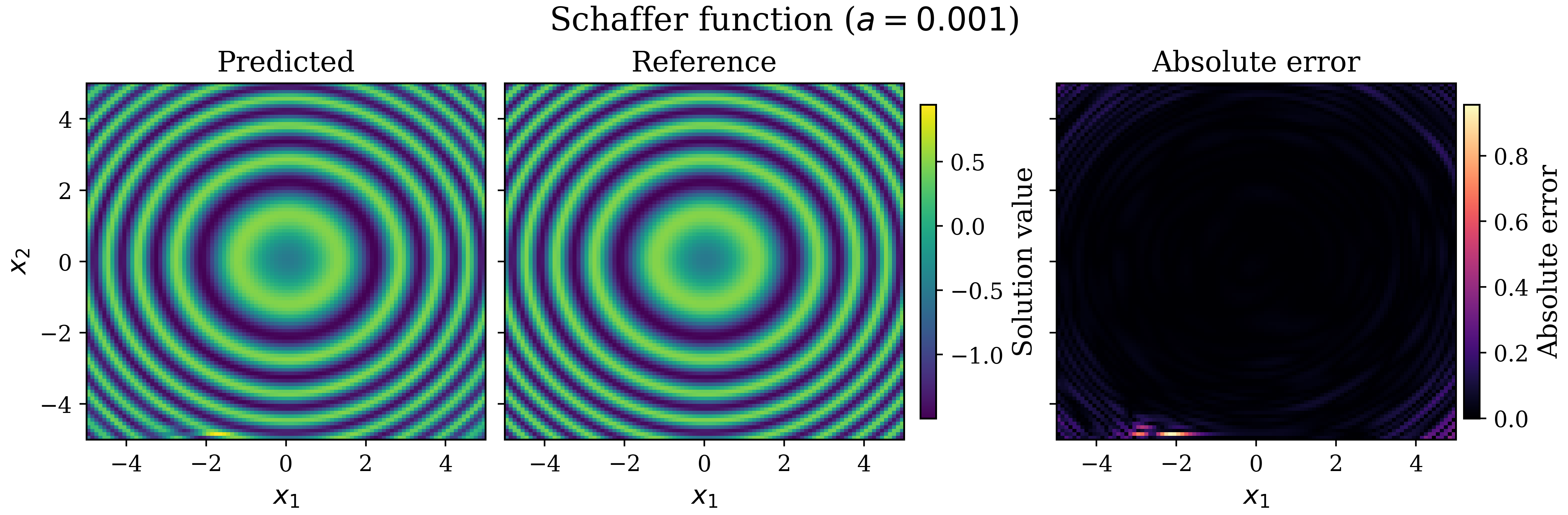}
\end{minipage}
\par\medskip
\begin{minipage}[t]{0.96\linewidth}
\centering
\textbf{$a=1$}\par\smallskip
\suppinclude[width=\linewidth,height=0.245\textheight]{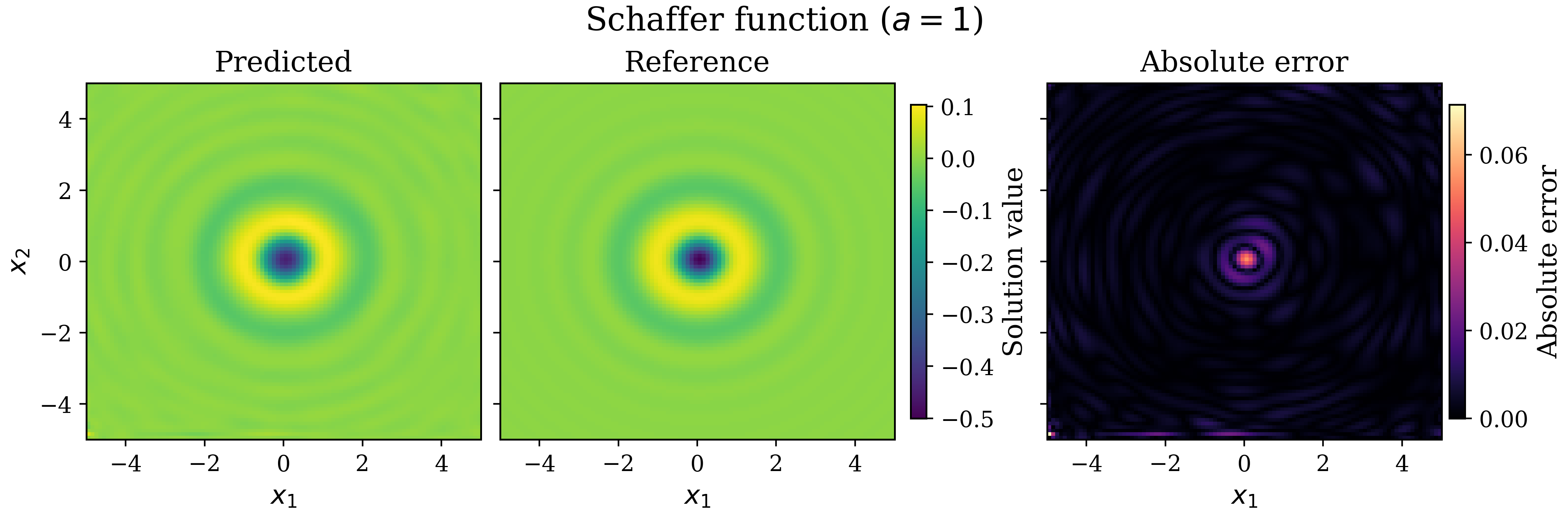}
\end{minipage}
\caption{ACR2-arch prediction, reference solution, and absolute error for Schaffer-like at the two parameter-domain endpoints. Oscillations are denser and have larger amplitude at $a=0.001$, whereas the solution magnitude is smaller at $a=1$.}
\label{fig:supp-problem-schaffer-scale}
\end{figure}

Figure~\ref{fig:supp-problem-schaffer-scale} shows that, across three seeds, the MSE values at $a=0.001$ and $a=1$ are $(1.045\pm0.477)\times10^{-3}$ and $(1.503\pm0.356)\times10^{-5}$, respectively, so the former is approximately $69.5\times$ higher. The corresponding $E_{L_2}$ values are $0.03813\pm0.00964$ and $0.10699\pm0.01277$, making the latter approximately $2.81\times$ higher. Changes in solution scale can therefore reverse the ordering of absolute and relative errors.

\begin{figure}[p]
\centering
\begin{minipage}[t]{0.96\linewidth}
\centering
\textbf{$\nu=0.05$}\par\smallskip
\suppinclude[width=\linewidth,height=0.245\textheight]{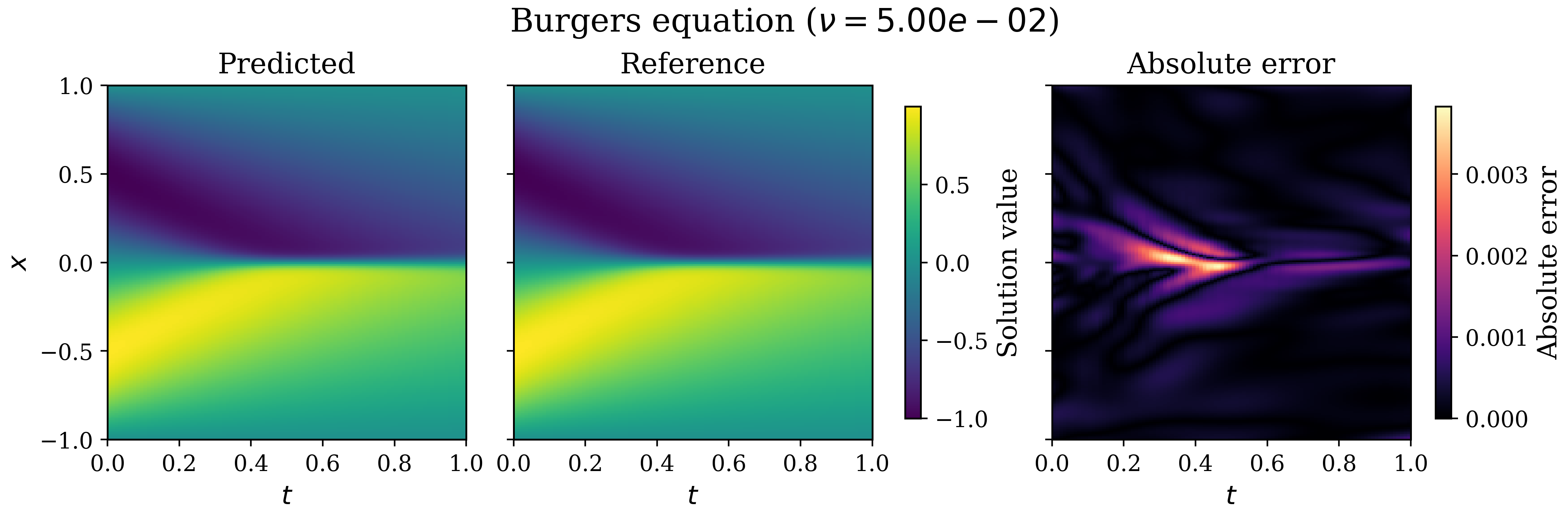}
\end{minipage}
\par\medskip
\begin{minipage}[t]{0.96\linewidth}
\centering
\textbf{$\nu=1$}\par\smallskip
\suppinclude[width=\linewidth,height=0.245\textheight]{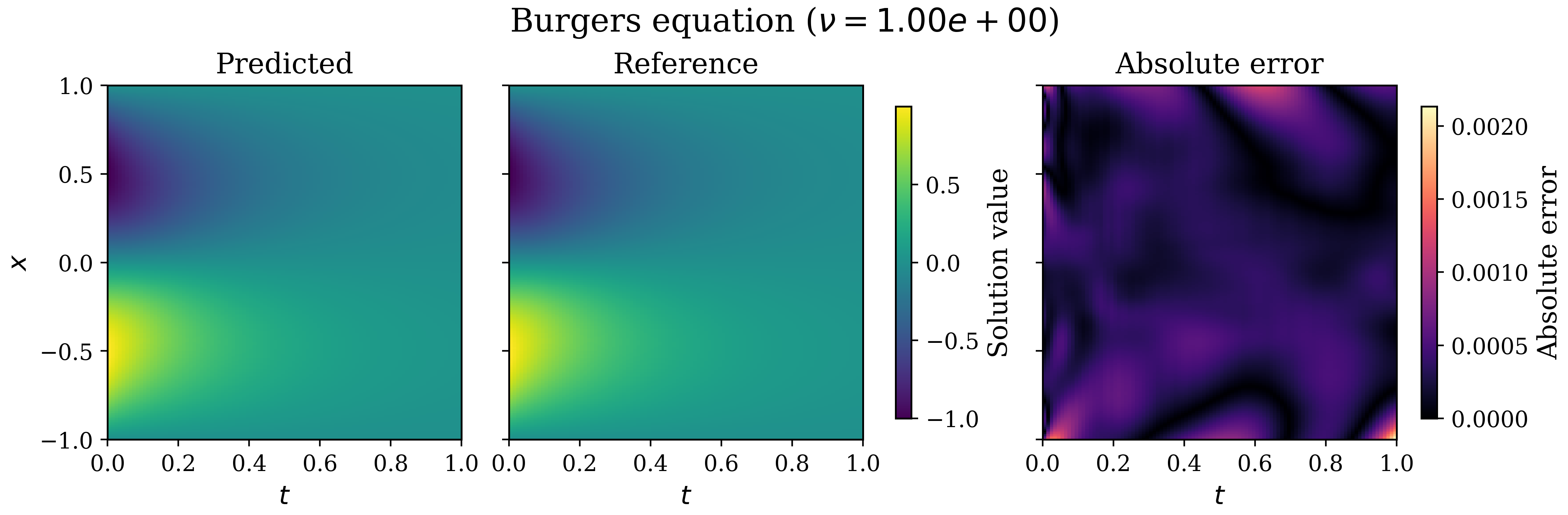}
\end{minipage}
\caption{Seed-0 ACR2-arch prediction, reference solution, and absolute error for Burgers at $\nu=0.05$ and $\nu=1$. The fields illustrate differences in solution structure; quantitative conclusions use all three fixed seeds.}
\label{fig:supp-problem-burgers-residual-scale}
\end{figure}

For the three-seed results illustrated by Figure~\ref{fig:supp-problem-burgers-residual-scale}, the mean PDE residuals at $\nu=0.05$ and $\nu=1$ are $0.002024\pm0.000318$ and $0.002083\pm0.000231$, a difference of only approximately $2.9\%$. The corresponding $E_{L_2}$ values are $(7.183\pm0.886)\times10^{-4}$ and $(1.048\pm0.161)\times10^{-3}$, a difference of approximately $45.9\%$. The physics-informed training loss is therefore only a proxy for task difficulty and cannot replace reference-solution error over the parameter domain.

\FloatBarrier

\subsection{Parameter-space generalization and overfitting}

\label{sec:supp-2-3}

\begin{figure}[p]
\centering
\begin{minipage}[t]{0.90\linewidth}
\centering
\suppinclude[width=\linewidth,height=0.64\textheight]{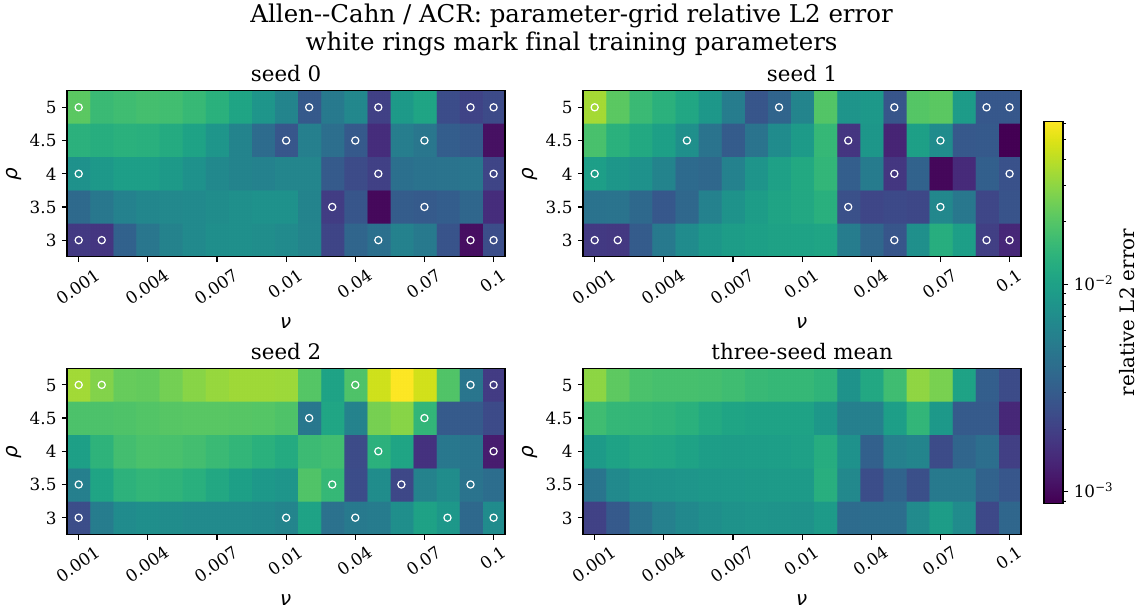}
\end{minipage}
\caption{ACR $E_{L_2}$ over the two-dimensional Allen--Cahn parameter plane. The first three panels correspond to fixed seeds $[0,1,2]$, and the summary panel shows their mean; white rings mark the retained training parameters.}
\label{fig:supp-allen-errors}
\end{figure}

A parameterized PINN can be trained on only finitely many parameter tasks, so good fitting around the sampled tasks does not guarantee uniform accuracy throughout a continuous parameter domain. Figure~\ref{fig:supp-allen-errors} illustrates this issue for Allen--Cahn.

Figure~\ref{fig:supp-allen-errors} shows a strongly nonuniform error distribution. Larger $\rho$ is generally more difficult, and seed 2 contains a local high-error band at large $\rho$ and intermediate $\nu$; even cells near retained training parameters are not necessarily the most accurate. Active selection can prioritize difficult regions and replay can retain constraints from earlier tasks, but parameter-space generalization must still be evaluated on the complete test grid.

\FloatBarrier

\clearpage

\section{Supplementary evidence for the main-text experimental conclusions}

\label{sec:supp-part-3}

This part follows the order of the experimental sections in the main text and retains only figures and tables that add evidence omitted for length. Main-result tables, representative solution fields, difficult-parameter fields, and replay figures already presented in the main text are not repeated. The complete parameter-wise error atlas and optimization histories for all six methods are collected in Part IV.

The correspondence between the main text and this part is summarized in Table~\ref{tab:supp-part3-map}.

\begin{table}[!htbp]
\centering
\caption{Organization of the additional evidence in Part III.}
\label{tab:supp-part3-map}
\footnotesize
\setlength{\tabcolsep}{4pt}
\renewcommand{\arraystretch}{1.16}
\begin{tabularx}{\linewidth}{@{}p{0.30\linewidth}X@{}}
\toprule
\textbf{Main-text location} & \textbf{Additional evidence in this part}\\
\midrule
``ACR2-finetune: rapid single-parameter adaptation'' & Residual-head width check, fixed long-horizon checkpoints, distributions over all strictly unseen parameters, and a long-horizon diagnostic for standard PINNs trained from scratch\\
``External-method comparisons'' & Native deployment differences, target-wise error distributions, online state size, and representative long-horizon results\\
``Ablation and sensitivity studies'' & Additional metrics for BO and dynamic weighting, together with prior, query-budget, ParamFNN depth--width, and mini-batch diagnostics\\
Cost statements in the experimental setup and result sections & Exclusive-GPU training times for six methods on five cases and the search-time decomposition for AG and AC\\
\bottomrule
\end{tabularx}
\end{table}

\FloatBarrier

\subsection{Additional diagnostics for ACR2-finetune}

\label{sec:supp-3-1}

ACR2-finetune is an independent downstream procedure applied after global training; it is distinct from both ACR2-arch and the full ACR2 configuration. In the present experiments, the ACR model is frozen and only a zero-output-initialized residual head with one hidden layer is trained at the requested parameter. The same adaptation mechanism can, in principle, be attached to an AC, ACR, or ACR2-arch model. The main text reports the fixed 500-step results; this section adds the architecture check, long-horizon behavior, and parameter-local distributions over the complete strictly unseen domain.

\FloatBarrier

\subsubsection{Residual-head width and fixed long-horizon checkpoints}

\label{sec:supp-3-1-1}

The width check uses seed 0 and three predeclared diagnostic targets per PDE, with a fixed 500-step endpoint. It assesses residual-head capacity but does not select the formal configuration using either the full test grid or the lowest post-adaptation reference-solution error. The three-seed experiments retain the prespecified widths: 25 for Burgers, Allen--Cahn, and Kovasznay, and 15 for linearized Poisson--Boltzmann.

\begin{table}[!htbp]
\centering
\caption{Seed-0 residual-head width check at the fixed 500-step endpoint. Each value is the mean $E_{L_2}$ over three predeclared diagnostic targets.}
\label{tab:supp-finetune-width}
\small
\setlength{\tabcolsep}{4pt}
\renewcommand{\arraystretch}{1.16}
\begin{adjustbox}{max width=\linewidth,center}
\begin{tabular}{lrrrr}
\toprule
\textbf{PDE} & \textbf{Width setting 1} & \textbf{Width setting 2} & \textbf{Width setting 3} & \textbf{Width setting 4}\\
\midrule
Burgers & 12: 0.000733 & 20: 0.000681 & 25: 0.000552 & 40: 0.000465\\
Allen--Cahn & 12: 0.00842 & 20: 0.01106 & 25: 0.01032 & 40: 0.00928\\
Kovasznay & 12: 0.02472 & 20: 0.02520 & 25: 0.02478 & 40: 0.02494\\
Linearized Poisson--Boltzmann & 8: 0.00492 & 12: 0.00425 & 15: 0.00428 & 20: 0.00417\\
\bottomrule
\end{tabular}
\end{adjustbox}
\end{table}

Table~\ref{tab:supp-finetune-width} shows that the width ranking varies by equation. Three diagnostic targets are insufficient to justify equation-wise selection from reference-solution error, so the minimum value in each row was not used to choose the formal width.

\begin{table}[!htbp]
\centering
\caption{Long-horizon ACR2-finetune diagnostic on the fixed targets. Values are macro-averaged $E_{L_2}$ over all prespecified seed--target pairs. Every column is a fixed checkpoint rather than the best point selected from a curve. Times are descriptive records from the corresponding long-horizon runs and are not used for wall-time ranking.}
\label{tab:supp-finetune-long-horizon}
\small
\setlength{\tabcolsep}{4pt}
\renewcommand{\arraystretch}{1.16}
\begin{adjustbox}{max width=\linewidth,center}
\begin{tabular}{lrrrr}
\toprule
\textbf{PDE} & \textbf{1,000 steps} & \textbf{2,000 steps} & \textbf{5,000 steps} & \textbf{Time per target at 5,000 steps}\\
\midrule
Burgers & 0.0006075 & 0.0004634 & 0.0004586 & $116.51\pm29.89$ s\\
Allen--Cahn & 0.01031 & 0.007529 & 0.005512 & $149.29\pm16.36$ s\\
Kovasznay & 0.03631 & 0.03095 & 0.02656 & $278.29\pm18.52$ s\\
Linearized Poisson--Boltzmann & 0.004783 & 0.003873 & 0.003997 & $195.18\pm3.48$ s\\
\bottomrule
\end{tabular}
\end{adjustbox}
\end{table}

Table~\ref{tab:supp-finetune-long-horizon} shows that the fixed 500-step Allen--Cahn result in the main text is a short-budget counterexample. Its mean degradation disappears by 1,000 steps, and all nine seed--target pairs improve at 5,000 steps. The other equations also benefit overall, but reference-solution error is not strictly monotone in the number of updates: Burgers first worsens between 500 and 1,000 steps before improving, and Poisson--Boltzmann rebounds slightly from 2,000 to 5,000 steps. A larger online budget can therefore reduce under-optimization, but reference-solution error must not be used retrospectively to choose the stopping point.

\FloatBarrier

\subsubsection{Adaptation distributions over all strictly unseen parameters}

\label{sec:supp-3-1-2}

\begin{figure}[p]
\centering
\begin{minipage}[t]{0.485\linewidth}
\centering
\textbf{Burgers: 83 unseen parameters}\par\smallskip
\suppinclude[width=\linewidth,height=0.225\textheight]{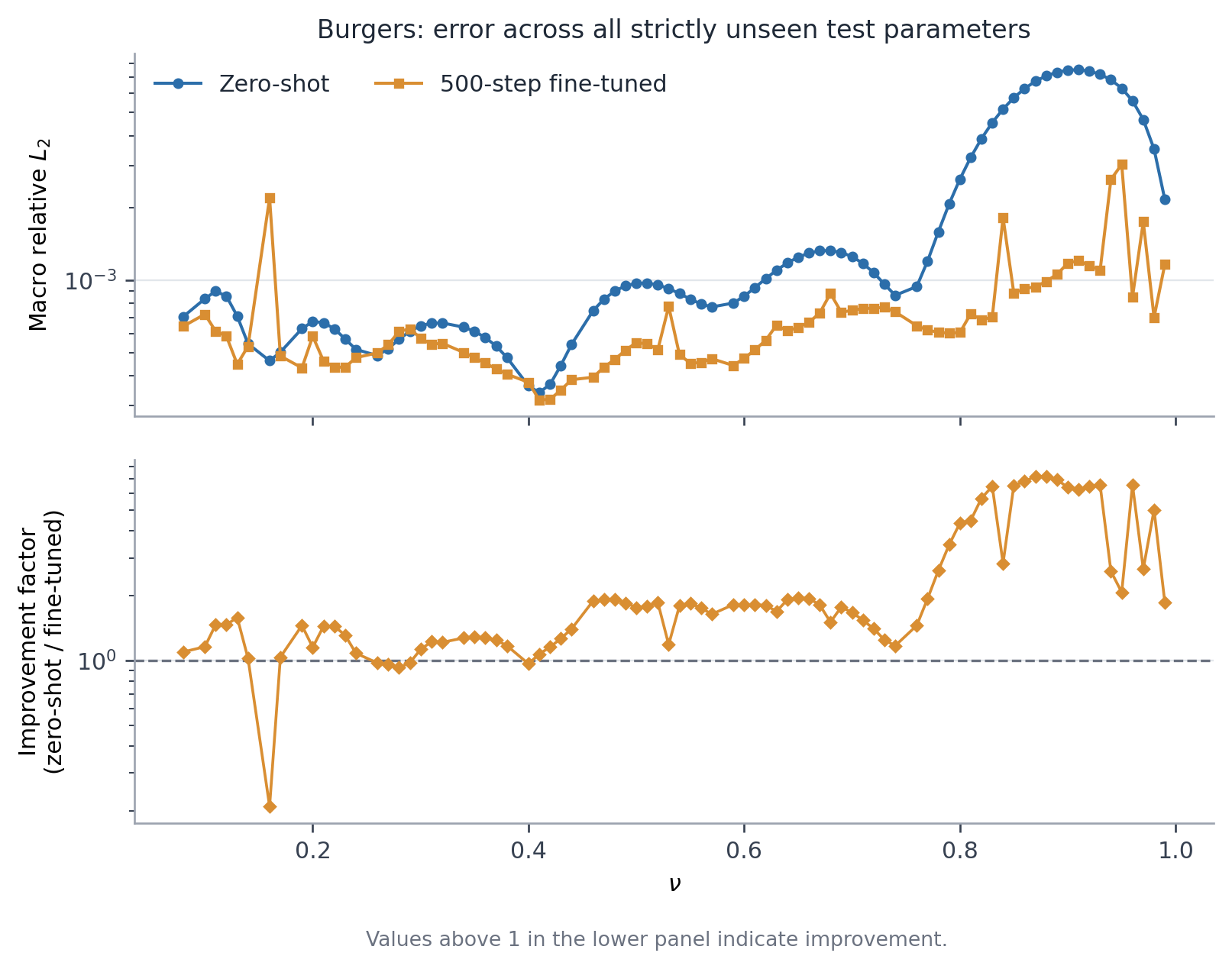}
\end{minipage}
\hfill
\begin{minipage}[t]{0.485\linewidth}
\centering
\textbf{Allen--Cahn: 77 unseen parameters}\par\smallskip
\suppinclude[width=\linewidth,height=0.225\textheight]{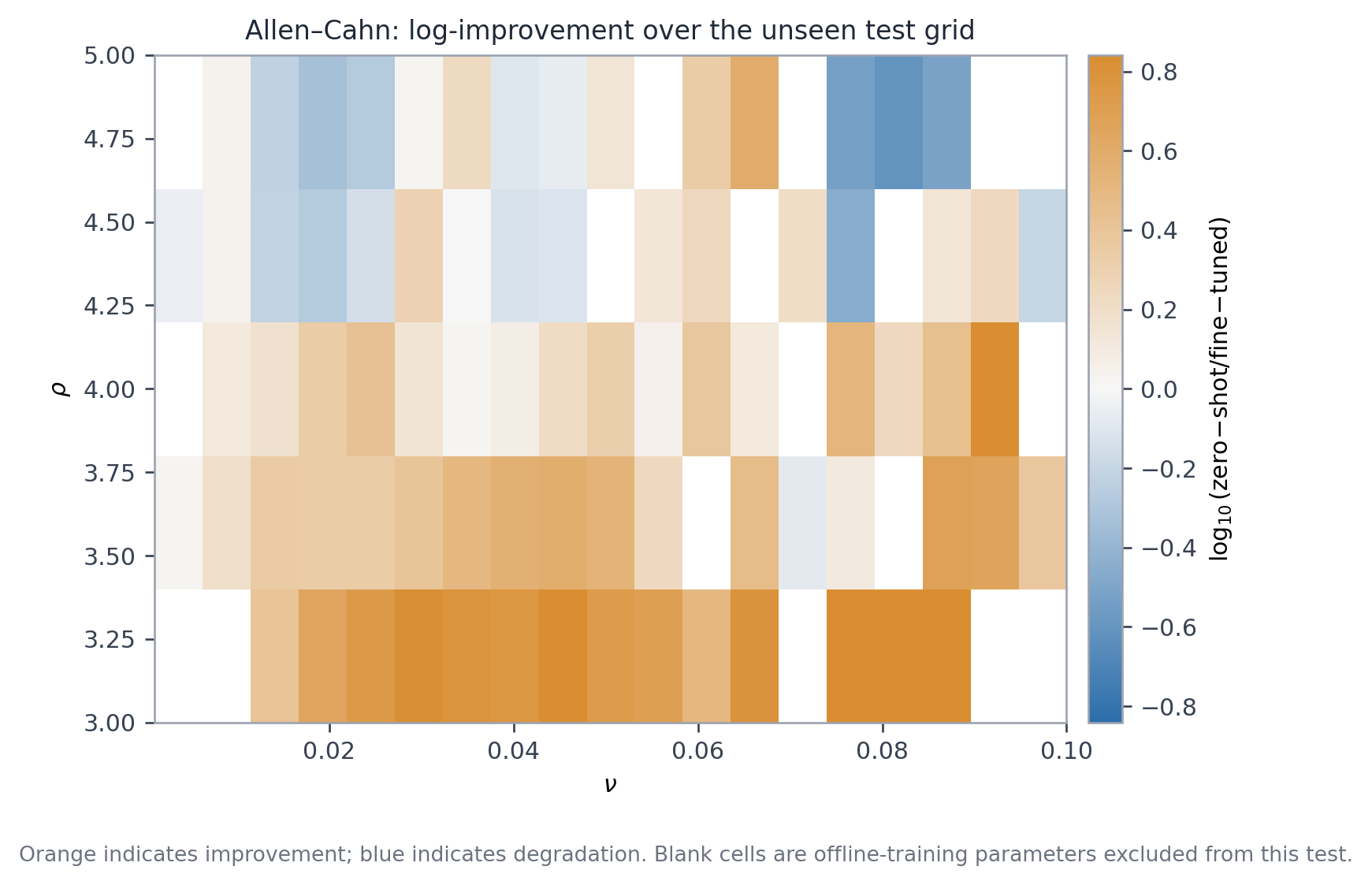}
\end{minipage}
\par\medskip
\begin{minipage}[t]{0.485\linewidth}
\centering
\textbf{Kovasznay: 80 unseen Reynolds numbers}\par\smallskip
\suppinclude[width=\linewidth,height=0.225\textheight]{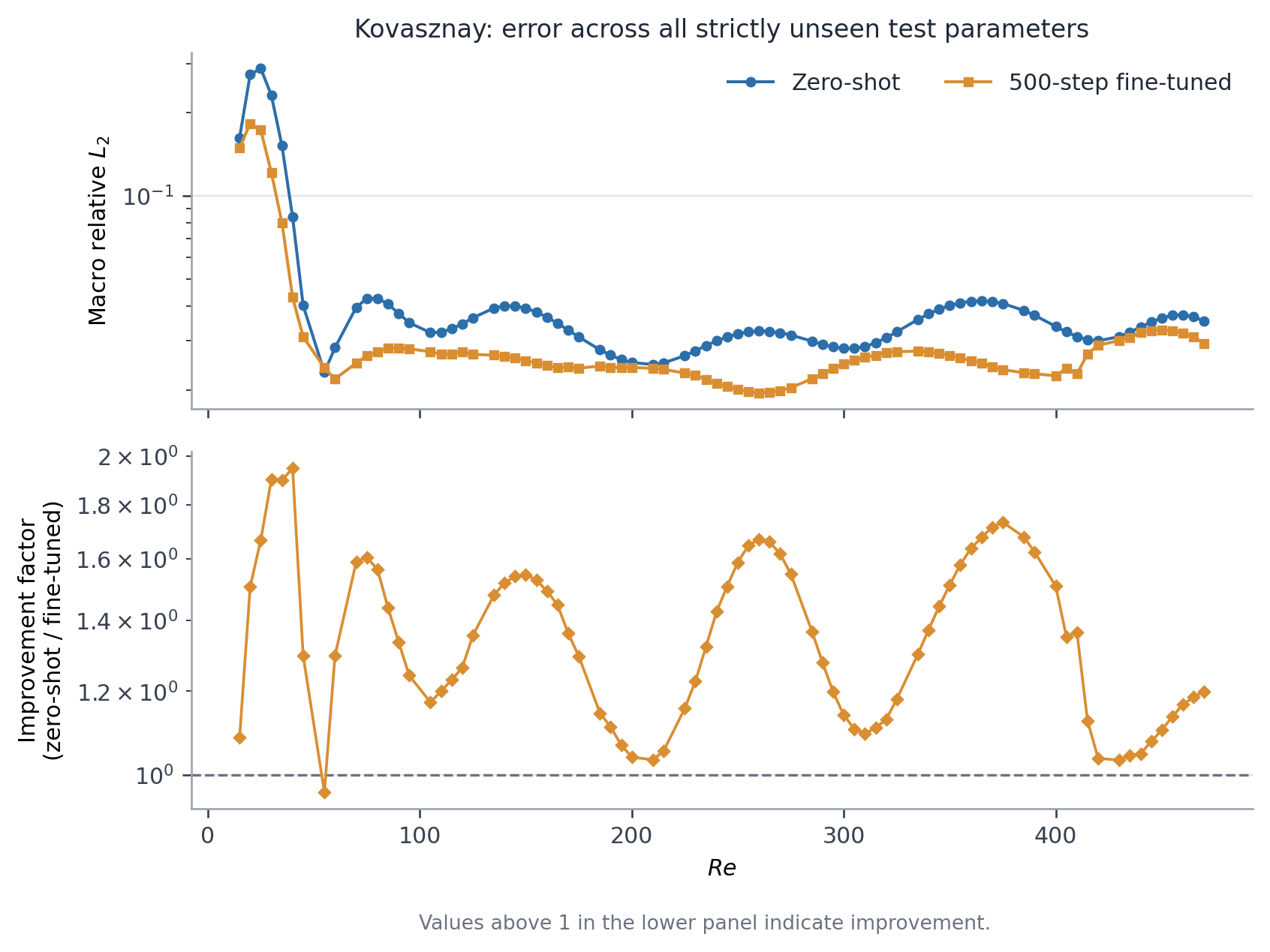}
\end{minipage}
\hfill
\begin{minipage}[t]{0.485\linewidth}
\centering
\textbf{Linearized Poisson--Boltzmann: 1,166 unseen parameters}\par\smallskip
\suppinclude[width=\linewidth,height=0.225\textheight]{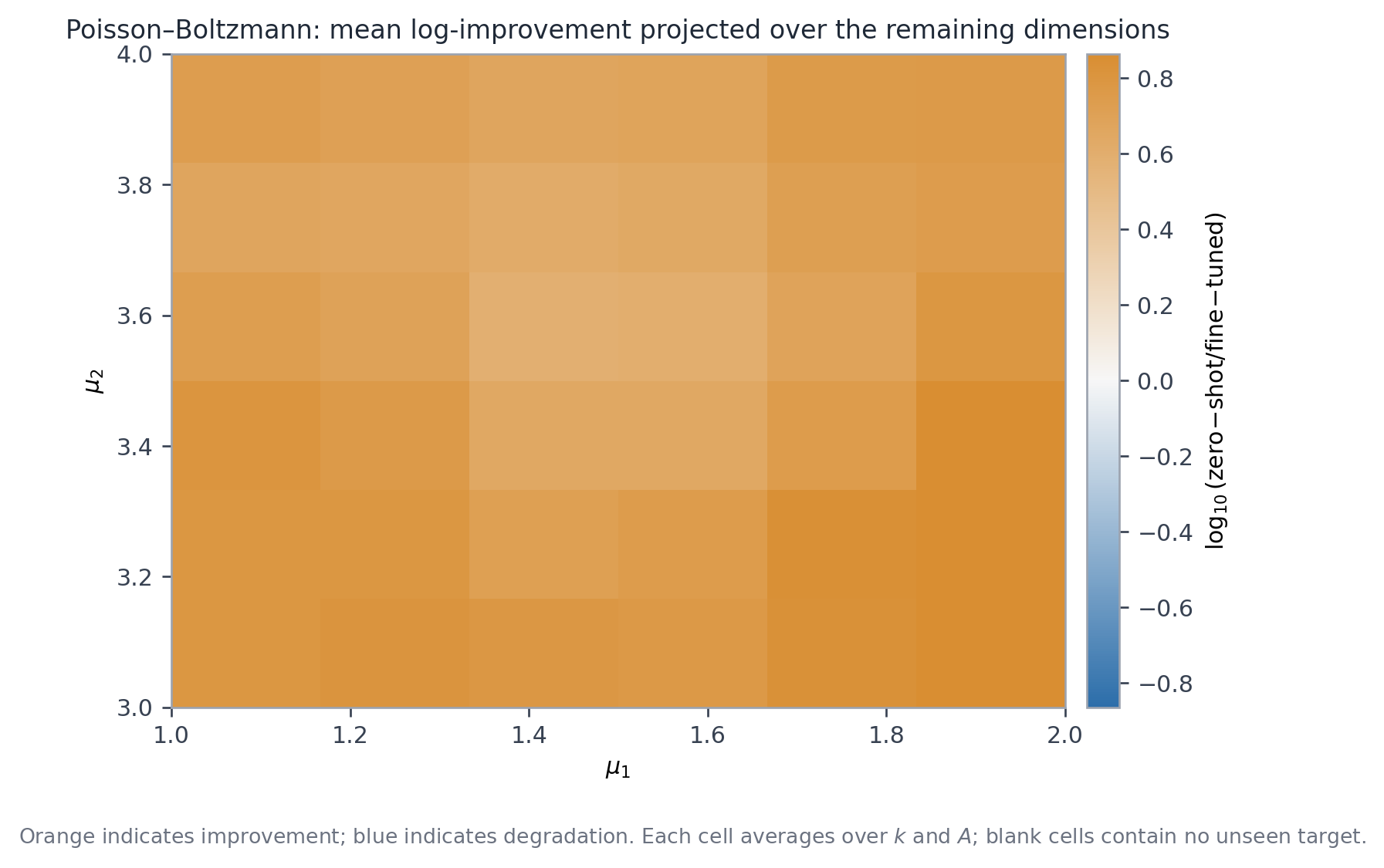}
\end{minipage}
\caption{Fixed 500-step ACR2-finetune distributions over all strictly unseen parameters of the four PDEs. Burgers and Kovasznay show parameter-wise results directly; Allen--Cahn uses pointwise log improvement; and Poisson--Boltzmann averages log improvement over the remaining $(k,A)$ coordinates in the $(\mu_1,\mu_2)$ projection. Blank cells denote offline-training parameters or locations without a strictly unseen target, not missing outcomes.}
\label{fig:supp-finetune-all-unseen}
\end{figure}

The main text already reports the aggregate full-grid results, the difficult quartiles, and the common pre/post-adaptation scatter plot. Figure~\ref{fig:supp-finetune-all-unseen} instead shows where improvements and degradations occur in each parameter domain. One-dimensional cases are plotted directly against the parameter, Allen--Cahn is shown as a two-dimensional map, and the four-dimensional Poisson--Boltzmann result is projected onto two coordinates. Each panel covers every strictly unseen target of the corresponding case.

Figure~\ref{fig:supp-finetune-all-unseen} supports the main-text conclusion at the parameter level. Most Burgers, Kovasznay, and Poisson--Boltzmann targets improve. Allen--Cahn degradations are concentrated mainly at larger $\rho$, consistent with the greater local optimization difficulty introduced by its sharp phase interfaces and strongly nonlinear reaction.

\FloatBarrier

\subsubsection{Long-horizon diagnostic for standard PINNs trained from scratch}

\label{sec:supp-3-1-3}

The main text compares ACR2-finetune and standard PINNs at the common 500-step endpoint. Table~\ref{tab:supp-standard-pinn-long-horizon} adds only the fixed 5,000-step result for a complete standard PINN trained from random initialization. This endpoint tests whether a longer Adam run closes the gap to the pretrained residual head; it is not claimed to represent full convergence.

\begin{table}[!htbp]
\centering
\caption{Errors, descriptive runtime, and complete-network parameter count for standard PINNs trained from scratch for 5,000 fixed Adam steps. Errors are the mean $\pm$ population standard deviation over three random seeds. Times are descriptive records from the corresponding long-horizon runs and are not used for wall-time ranking.}
\label{tab:supp-standard-pinn-long-horizon}
\scriptsize
\setlength{\tabcolsep}{4pt}
\renewcommand{\arraystretch}{1.16}
\begin{adjustbox}{max width=\linewidth,center}
\begin{tabular}{lrrrrr}
\toprule
\textbf{PDE} & \textbf{$\mathrm{MSE}$} & \textbf{Macro-averaged $E_{L_2}$} & \textbf{Worst $E_{L_2}$} & \textbf{Time per target} & \textbf{Complete-network parameters}\\
\midrule
Burgers & $0.07699\pm0.006865$ & $0.6932\pm0.0327$ & $1.2207\pm0.0427$ & 70.87 s & 7,851\\
Allen--Cahn & $0.05001\pm0.0008560$ & $0.3324\pm0.0034$ & $0.3979\pm0.0135$ & 108.96 s & 7,851\\
Kovasznay & $13.444\pm0.154$ & $75.063\pm0.274$ & $79.022\pm0.288$ & 210.35 s & 7,953\\
Linearized Poisson--Boltzmann & $0.1413\pm0.00005707$ & $0.9478\pm0.0004298$ & $1.1773\pm0.0001797$ & 168.76 s & 3,841\\
\bottomrule
\end{tabular}
\end{adjustbox}
\end{table}

\begin{figure}[p]
\centering
\begin{minipage}[t]{0.90\linewidth}
\centering
\suppinclude[width=\linewidth,height=0.64\textheight]{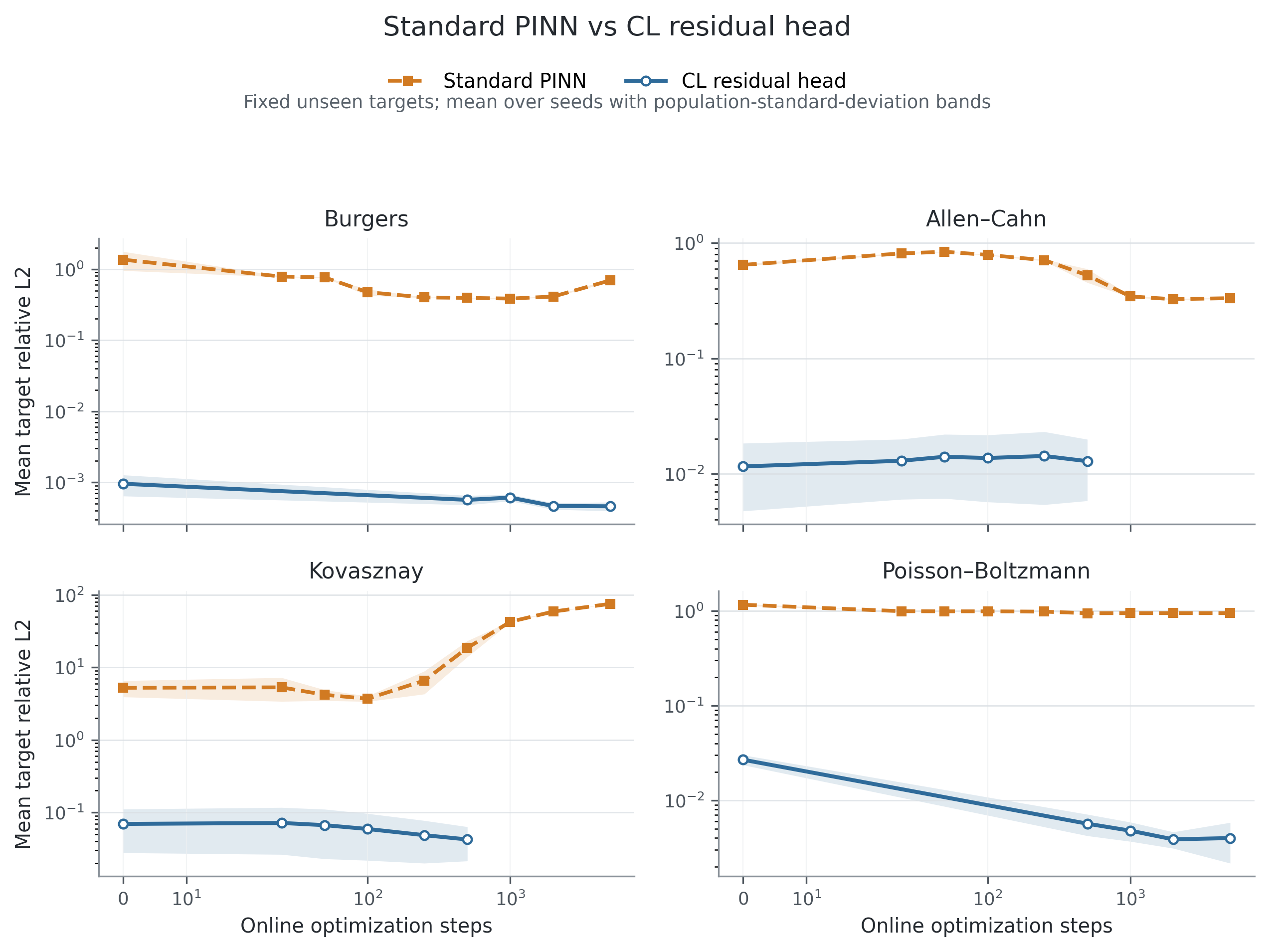}
\end{minipage}
\caption{Fixed-checkpoint trajectories for complete standard PINNs trained from scratch and ACR2-finetune, which updates only the residual head. Curves and bands show the mean and population standard deviation of target-averaged $E_{L_2}$ over three seeds. The comparison matches online update counts and does not include the offline pretraining cost of ACR2-finetune.}
\label{fig:supp-standard-pinn-long-horizon}
\end{figure}

As shown in Table~\ref{tab:supp-standard-pinn-long-horizon} and Figure~\ref{fig:supp-standard-pinn-long-horizon}, only the Allen--Cahn standard PINN continues to improve between 500 and 5,000 steps; reference-solution error increases for the other three PDEs. Native L-BFGS was also applied to one prespecified seed-0 target per equation and produced no improving case, with the Burgers run becoming non-finite. Simply extending optimization from random initialization therefore does not reliably remove the initial advantage provided by a pretrained parameterized representation.

\FloatBarrier

\subsection{Additional comparisons with external methods}

\label{sec:supp-3-2}

The main text reports the fixed-endpoint comparison and the four-PDE summary figure on the common target sets. This section adds the native deployment differences among the external methods, target-wise error distributions, online state sizes, and representative long-horizon diagnostics. P2INN and HyperPINN are independently implemented, capacity-matched mechanism adaptations rather than official four-PDE reproductions supplied by the original authors. Formal Meta-PINN results are available only for Burgers.

\FloatBarrier

\subsubsection{Native deployment and comparison scope}

\label{sec:supp-3-2-1}

\begin{table}[!htbp]
\centering
\caption{Native deployment of each external method and interpretation of the fixed endpoint used in this study. An epoch in the original work cannot be converted directly into one online update in the present protocol.}
\label{tab:supp-external-native-protocol}
\scriptsize
\setlength{\tabcolsep}{4pt}
\renewcommand{\arraystretch}{1.16}
\begin{tabularx}{\linewidth}{@{}p{0.13\linewidth}p{0.26\linewidth}p{0.27\linewidth}X@{}}
\toprule
\textbf{Method} & \textbf{Native deployment or adaptation} & \textbf{Endpoint used here} & \textbf{Interpretation boundary}\\
\midrule
P2INN \cite{cho2024p2inn} & Direct prediction by a parameterized network; optional SVD modulation updates only low-rank variables & Direct prediction and 500 fixed Adam updates, reported for both full-network and SVD adaptation & The 500-step endpoint is a common short-budget diagnostic, not a stepwise reproduction of the original epoch protocol\\
HyperPINN \cite{de2021hyperpinn} & A hypernetwork generates a compact target PINN from the PDE parameters, normally followed by direct inference & Direct prediction and an additional 500-step target-network adaptation introduced in this study & The added adaptation gives HyperPINN an extra opportunity to adjust and is not required by its native deployment\\
Meta-PINN \cite{penwarden2023metalearning} & A parameter-to-weight surrogate initializes a complete task network, which is then optimized & On Burgers, 500 Adam steps followed by at most 100 L-BFGS steps & The fixed L-BFGS cap does not represent the fully converged, tolerance-driven endpoint of the original protocol\\
\bottomrule
\end{tabularx}
\end{table}

As summarized in Table~\ref{tab:supp-external-native-protocol}, the methods do not have equal offline budgets. CL-PINN uses Adam followed by L-BFGS, whereas the capacity-matched P2INN and HyperPINN runs use Adam only; their online representations and native deployment modes also differ. The comparisons therefore concern reference-solution error and online state size on common targets at fixed online endpoints, not a strict ranking of offline training cost.

\FloatBarrier

\subsubsection{Target-wise error distributions}

\label{sec:supp-3-2-2}

The primary table in the main text reports MSE and mean $E_{L_2}$ at the fixed 500-step endpoint. Table~\ref{tab:supp-external-tail-distribution} adds the mean direct-prediction error at step 0 and the median and worst target errors before and after adaptation. For each offline seed, the statistics are first computed over the fixed target set and then aggregated across seeds $[0,1,2]$.

\begin{sidewaystable}[p]
\centering
\caption{Target-wise $E_{L_2}$ distributions under the common four-PDE protocol. Arrows denote pre-adaptation prediction (step 0) to the fixed 500-step endpoint. Kovasznay is first macro-averaged over $u$, $v$, and $p$.}
\label{tab:supp-external-tail-distribution}
\footnotesize
\setlength{\tabcolsep}{4pt}
\renewcommand{\arraystretch}{1.16}
\begin{tabularx}{\textheight}{@{}p{0.10\textheight}p{0.20\textheight}XXX@{}}
\toprule
\textbf{PDE} & \textbf{Method} & \textbf{Mean $E_{L_2}$ at step 0} & \textbf{Median $E_{L_2}$: 0 $\rightarrow$ 500} & \textbf{Worst $E_{L_2}$: 0 $\rightarrow$ 500}\\
\midrule
Burgers & ACR2-finetune & $0.0009522\pm0.0003135$ & $0.0006961\pm0.0001695\rightarrow0.0005059\pm0.00001598$ & $0.002294\pm0.001179\rightarrow0.0008329\pm0.0002244$\\
 & P2INN full-network adaptation & $0.5355\pm0.05049$ & $0.5073\pm0.05797\rightarrow0.2764\pm0.04382$ & $0.8699\pm0.06715\rightarrow0.5752\pm0.08297$\\
 & P2INN SVD adaptation & $0.5355\pm0.05050$ & $0.5073\pm0.05798\rightarrow0.2928\pm0.05654$ & $0.8699\pm0.06715\rightarrow0.6300\pm0.1275$\\
 & HyperPINN target-network adaptation & $0.2968\pm0.09582$ & $0.3074\pm0.09833\rightarrow0.005932\pm0.001054$ & $0.3811\pm0.1344\rightarrow0.01663\pm0.002964$\\
\addlinespace[2pt]
Allen--Cahn & ACR2-finetune & $0.01156\pm0.006812$ & $0.01059\pm0.006427\rightarrow0.01056\pm0.006723$ & $0.01854\pm0.01024\rightarrow0.02453\pm0.01394$\\
 & P2INN full-network adaptation & $0.3049\pm0.05118$ & $0.3118\pm0.05364\rightarrow0.2887\pm0.03138$ & $0.3555\pm0.05652\rightarrow0.4117\pm0.01440$\\
 & P2INN SVD adaptation & $0.3049\pm0.05118$ & $0.3118\pm0.05363\rightarrow0.2902\pm0.03080$ & $0.3555\pm0.05652\rightarrow0.3393\pm0.001282$\\
 & HyperPINN target-network adaptation & $0.3647\pm0.1279$ & $0.3627\pm0.1300\rightarrow0.2665\pm0.02859$ & $0.5194\pm0.2330\rightarrow0.3267\pm0.04606$\\
\addlinespace[2pt]
Kovasznay & ACR2-finetune & $0.06959\pm0.04176$ & $0.06664\pm0.03935\rightarrow0.04168\pm0.01929$ & $0.08497\pm0.05691\rightarrow0.05010\pm0.02940$\\
 & P2INN full-network adaptation & $1.068\pm0.08213$ & $1.078\pm0.08752\rightarrow0.5953\pm0.09246$ & $1.100\pm0.1036\rightarrow0.6094\pm0.08853$\\
 & P2INN SVD adaptation & $1.068\pm0.08208$ & $1.078\pm0.08746\rightarrow0.8963\pm0.1098$ & $1.100\pm0.1036\rightarrow0.9202\pm0.1019$\\
 & HyperPINN target-network adaptation & $1.030\pm0.09855$ & $1.016\pm0.08232\rightarrow0.8934\pm0.02723$ & $1.114\pm0.1775\rightarrow0.9069\pm0.02290$\\
\addlinespace[2pt]
Linearized Poisson--Boltzmann & ACR2-finetune & $0.02686\pm0.003232$ & $0.02628\pm0.005429\rightarrow0.004641\pm0.0009361$ & $0.03914\pm0.005634\rightarrow0.009164\pm0.002821$\\
 & P2INN full-network adaptation & $4.681\pm0.4101$ & $4.510\pm0.1695\rightarrow0.4422\pm0.02417$ & $5.947\pm0.6923\rightarrow0.9346\pm0.1431$\\
 & P2INN SVD adaptation & $4.681\pm0.4101$ & $4.510\pm0.1695\rightarrow0.4542\pm0.04313$ & $5.947\pm0.6923\rightarrow1.016\pm0.1196$\\
 & HyperPINN target-network adaptation & $3.429\pm1.015$ & $3.214\pm1.161\rightarrow0.03996\pm0.01228$ & $4.492\pm0.7299\rightarrow0.06769\pm0.008854$\\
\bottomrule
\end{tabularx}
\end{sidewaystable}

Table~\ref{tab:supp-external-tail-distribution} complements the mean-error result in the main text. ACR2-finetune reduces both median and worst-target errors on Burgers, Kovasznay, and Poisson--Boltzmann. On Allen--Cahn, the median is nearly unchanged while the worst target degrades, preserving the short-budget counterexample. Most external adaptations also improve within 500 steps, but their fixed-endpoint errors remain higher.

\FloatBarrier

\subsubsection{Online state size and representative long-horizon results}

\label{sec:supp-3-2-3}

\begin{table}[!htbp]
\centering
\caption{Trainable online state per target under the common protocol. For P2INN SVD, only the low-rank variables actually updated online are counted; for HyperPINN, the generated target-network parameters are counted. Bold denotes the smallest state for each equation.}
\label{tab:supp-external-online-state}
\scriptsize
\setlength{\tabcolsep}{4pt}
\renewcommand{\arraystretch}{1.16}
\begin{adjustbox}{max width=\linewidth,center}
\begin{tabular}{lrrrr}
\toprule
\textbf{PDE} & \textbf{ACR2-finetune residual head} & \textbf{P2INN full-network adaptation} & \textbf{P2INN SVD adaptation} & \textbf{HyperPINN target network}\\
\midrule
Burgers & 1,301 & 7,913 & 207 & \textbf{151}\\
Allen--Cahn & 1,301 & 7,982 & 230 & \textbf{151}\\
Kovasznay & 1,353 & 7,961 & 207 & \textbf{165}\\
Linearized Poisson--Boltzmann & 481 & 4,081 & \textbf{192} & 781\\
\bottomrule
\end{tabular}
\end{adjustbox}
\end{table}

Table~\ref{tab:supp-external-online-state} shows that the ACR2-finetune residual head is substantially smaller than the full P2INN network. P2INN SVD is nevertheless more compact on all four PDEs, and the HyperPINN target network is smaller except on Poisson--Boltzmann. The supported advantage of ACR2-finetune is therefore fixed-endpoint accuracy, not minimum online state in every case. Formal Meta-PINN comparison is limited to Burgers and updates all 7,851 parameters of the task network; its capped endpoint and accuracy are already reported in the main text and are not repeated here.

Available long-horizon external diagnostics cover only Burgers and linearized Poisson--Boltzmann, representing a one-dimensional single-parameter case and a four-dimensional multi-parameter case. They test whether 500 online steps are particularly restrictive for the external methods and do not constitute a complete four-PDE long-horizon comparison.

\begin{table}[!htbp]
\centering
\caption{Representative external-method diagnostics after 5,000 fixed online Adam steps on the common unseen targets. The corresponding ACR2-finetune results are reported in Section~\ref{sec:supp-3-1-1}. Times are recorded run summaries rather than a synchronized exclusive-GPU latency benchmark.}
\label{tab:supp-external-long-horizon}
\footnotesize
\setlength{\tabcolsep}{4pt}
\renewcommand{\arraystretch}{1.16}
\begin{adjustbox}{max width=\linewidth,center}
\begin{tabular}{llrr}
\toprule
\textbf{PDE} & \textbf{Method} & \textbf{Mean $E_{L_2}$ at 5,000 steps} & \textbf{Time per target}\\
\midrule
Burgers & P2INN full-network adaptation & $0.2924\pm0.0483$ & 169.5 s\\
 & HyperPINN capacity-matched target network & $0.003276\pm0.0004615$ & 69.4 s\\
 & HyperPINN specialized generated-target configuration (mechanism diagnostic only) & $0.002700\pm0.0003906$ & 43.3 s\\
Linearized Poisson--Boltzmann & HyperPINN capacity-matched target network & $0.03749\pm0.01288$ & 167.5 s\\
 & P2INN full-network adaptation & $0.4416\pm0.0715$ & 261.3 s\\
\bottomrule
\end{tabular}
\end{adjustbox}
\end{table}

\begin{figure}[p]
\centering
\begin{minipage}[t]{0.76\linewidth}
\centering
\textbf{Burgers}\par\smallskip
\suppinclude[width=\linewidth,height=0.31\textheight]{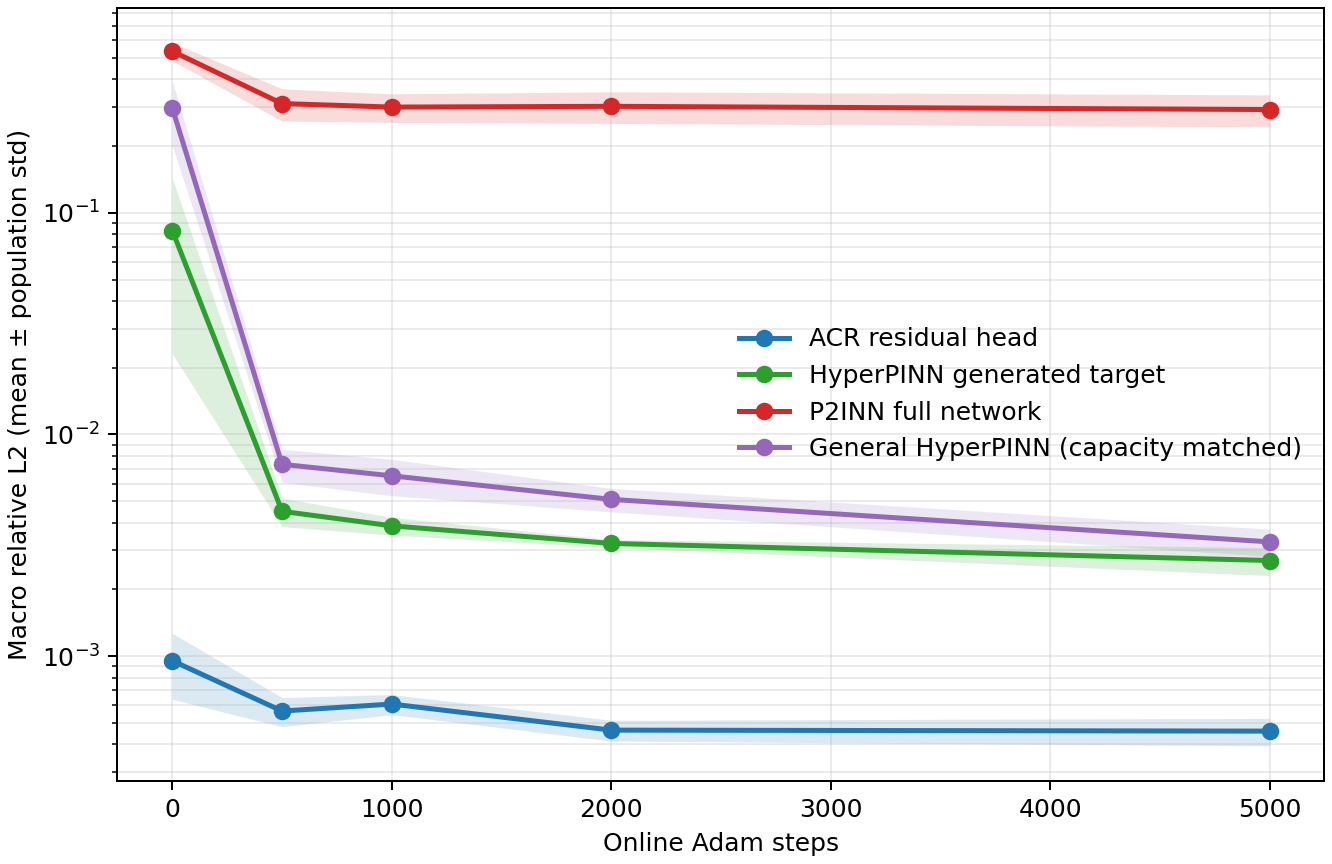}
\end{minipage}
\par\medskip
\begin{minipage}[t]{0.76\linewidth}
\centering
\textbf{Linearized Poisson--Boltzmann}\par\smallskip
\suppinclude[width=\linewidth,height=0.31\textheight]{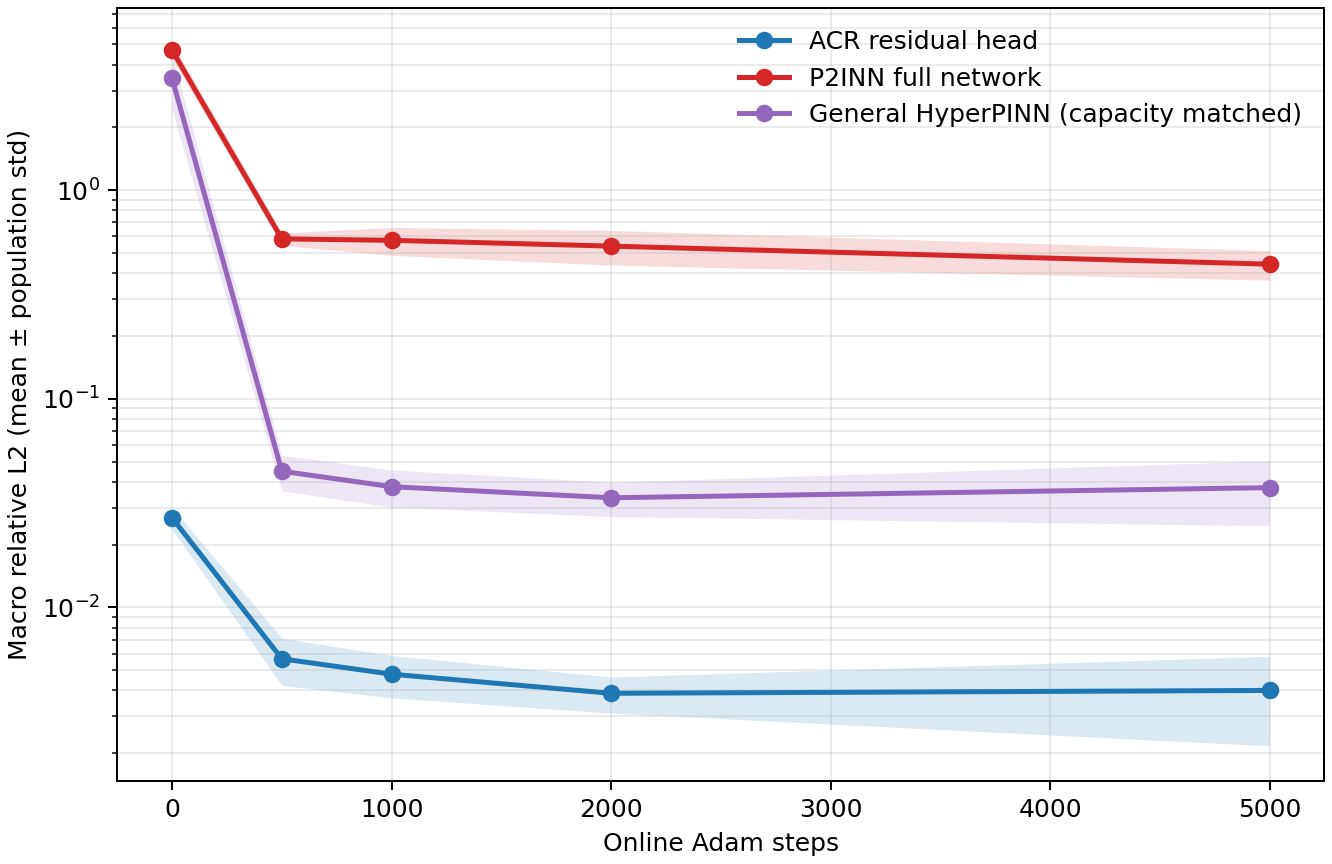}
\end{minipage}
\caption{Online adaptation trajectories over 5,000 fixed steps on Burgers and linearized Poisson--Boltzmann. Curves and bands show the mean and population standard deviation over three independent offline seeds. Every method is reported at prespecified checkpoints rather than at the lowest point of its curve.}
\label{fig:supp-external-long-horizon}
\end{figure}

Table~\ref{tab:supp-external-long-horizon} and Figure~\ref{fig:supp-external-long-horizon} show that the external methods continue to improve at 5,000 steps but do not reach the corresponding 5,000-step ACR2-finetune accuracy reported in Section~\ref{sec:supp-3-1-1}. HyperPINN has shorter recorded times in both cases, indicating a method-dependent trade-off between accuracy and online time. This representative result must not be extrapolated to a uniform wall-time conclusion across all four PDEs.

\FloatBarrier

\subsection{Additional ablation and sensitivity results}

\label{sec:supp-3-3}

The main text already reports the primary controls for BO and dynamic weighting, fixed-capacity sparse replay, the capacity-matched parameter subnetwork, and parameter-branch decay/freezing. This section does not repeat those primary tables or replay figures. It retains only additional metrics and diagnostics that clarify configuration boundaries. Single-case sensitivities are identified explicitly and are not extrapolated as common behavior across all five cases.

\FloatBarrier

\subsubsection{Bayesian selection, dynamic weighting, and query budget}

\label{sec:supp-3-3-1}

The main text evaluates cross-task accuracy using macro-averaged and worst $E_{L_2}$. Table~\ref{tab:supp-active-mse} adds pointwise absolute error from the same controlled $2\times2$ runs, allowing the relative-error conclusions to be checked against MSE.

\begin{table}[!htbp]
\centering
\caption{Test MSE in the within-case $2\times2$ ablation of Bayesian selection and task-wise training weights. Values are the mean $\pm$ population standard deviation over fixed seeds $[0,1,2]$.}
\label{tab:supp-active-mse}
\footnotesize
\setlength{\tabcolsep}{4pt}
\renewcommand{\arraystretch}{1.16}
\begin{tabularx}{\linewidth}{@{}p{0.17\linewidth}XXXX@{}}
\toprule
\textbf{Case} & \makecell{\textbf{Fixed}\\\textbf{selection,}\\\textbf{equal training}\\\textbf{weights}} & \makecell{\textbf{BO only,}\\\textbf{equal training}\\\textbf{weights}} & \makecell{\textbf{Fixed}\\\textbf{selection}\\\textbf{+ training}\\\textbf{weights}} & \makecell{\textbf{BO + case-}\\\textbf{specific}\\\textbf{weighting}}\\
\midrule
Schaffer-like & $2.0351\times10^{-2}\pm1.5944\times10^{-2}$ & $1.822\times10^{-1}\pm1.805\times10^{-1}$ & $9.0496\times10^{-4}\pm5.2076\times10^{-4}$ & \textbf{$6.8658\times10^{-4}\pm5.3208\times10^{-4}$}\\
Burgers & \textbf{$3.491\times10^{-4}\pm2.852\times10^{-4}$} & $4.626\times10^{-4}\pm3.218\times10^{-4}$ & $4.437\times10^{-4}\pm1.542\times10^{-4}$ & $4.892\times10^{-4}\pm2.484\times10^{-4}$\\
Allen--Cahn & $2.618\times10^{-2}\pm1.120\times10^{-2}$ & \textbf{$9.116\times10^{-4}\pm1.982\times10^{-4}$} & $2.610\times10^{-2}\pm1.279\times10^{-2}$ & $1.414\times10^{-3}\pm1.141\times10^{-3}$\\
Kovasznay & \textbf{$5.236\times10^{-4}\pm4.562\times10^{-4}$} & $5.612\times10^{-2}\pm3.351\times10^{-2}$ & $6.486\times10^{-4}\pm5.826\times10^{-4}$ & $1.225\times10^{-1}\pm1.579\times10^{-1}$\\
Linearized Poisson--Boltzmann & $22.50\pm7.704$ & $2.788\times10^{-3}\pm3.833\times10^{-3}$ & $22.74\pm8.853$ & \textbf{$9.644\times10^{-4}\pm1.287\times10^{-3}$}\\
\bottomrule
\end{tabularx}
\end{table}

\begin{figure}[p]
\centering
\begin{minipage}[t]{0.90\linewidth}
\centering
\suppinclude[width=\linewidth,height=0.64\textheight]{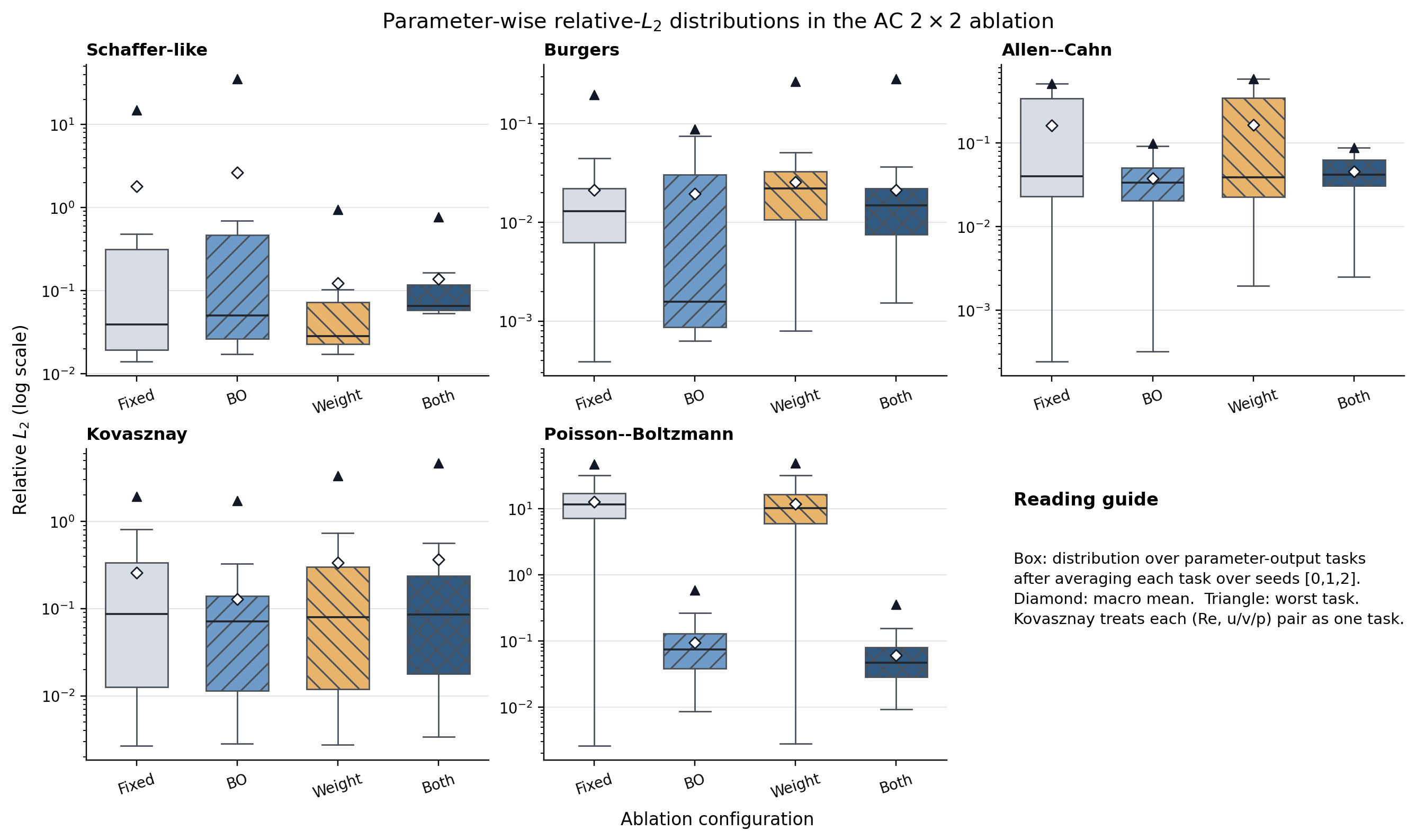}
\end{minipage}
\caption{Task-wise $E_{L_2}$ distributions in the five-case $2\times2$ ablation. Each box plot first averages the three seeds for the same parameter--output task. Open diamonds denote the task macro-average, filled triangles denote the worst task, and the vertical axis is logarithmic. For Kovasznay, each $(Re,u/v/p)$ pair is treated as one task.}
\label{fig:supp-active-weight-balance}
\end{figure}

Table~\ref{tab:supp-active-mse} and Figure~\ref{fig:supp-active-weight-balance} further demonstrate equation- and metric-dependent component effects. The BO-only configuration improves both MSE and relative error on Allen--Cahn, while BO combined with task-wise training weights is best across the reported metrics on Poisson--Boltzmann. Kovasznay shows the opposite metric ranking: the BO-only configuration has lower relative error, whereas fixed selection has lower MSE. Task selection and loss weighting must therefore not be judged from a single aggregate metric.

\paragraph{Prior-function sensitivity}

Systematic prior variants were completed only for Burgers and Kovasznay. In the separate Burgers prior-direction diagnostic, which is distinct from the unit-prior primary configuration, the diagnostic prior enters both the training weights and BO utility adjustment. On Kovasznay, the prior enters the training weights only. The two studies establish equation dependence but do not form a common-protocol causal test of a search prior.

\begin{table}[!htbp]
\centering
\caption{Prior-function sensitivity on Burgers and Kovasznay. Values are macro-averaged $E_{L_2}$ and are compared only within a case.}
\label{tab:supp-prior-sensitivity}
\small
\setlength{\tabcolsep}{4pt}
\renewcommand{\arraystretch}{1.16}
\begin{adjustbox}{max width=\linewidth,center}
\begin{tabular}{llr}
\toprule
\textbf{Case} & \textbf{Prior variant} & \textbf{Macro-averaged $E_{L_2}$}\\
\midrule
Burgers & Configured diagnostic prior & 0.06724\\
 & Uniform & 0.02069\\
 & Reversed & \textbf{0.01292}\\
Kovasznay & Locked configuration (exponent 1) & \textbf{$0.1194\pm0.0120$}\\
 & Reversed (exponent $-1$) & $0.1751\pm0.0359$\\
 & Uniform (exponent 0) & $0.5391\pm0.6415$\\
 & Weak (exponent 0.5) & $0.7738\pm0.5150$\\
\bottomrule
\end{tabular}
\end{adjustbox}
\end{table}

Table~\ref{tab:supp-prior-sensitivity} shows that Burgers favors the reversed prior, whereas Kovasznay favors its locked configuration. Both prior direction and injection location therefore require problem-specific validation.

\paragraph{Query-budget sensitivity}

\begin{figure}[p]
\centering
\begin{minipage}[t]{0.98\linewidth}
\centering
\suppinclude[width=\linewidth,height=0.64\textheight]{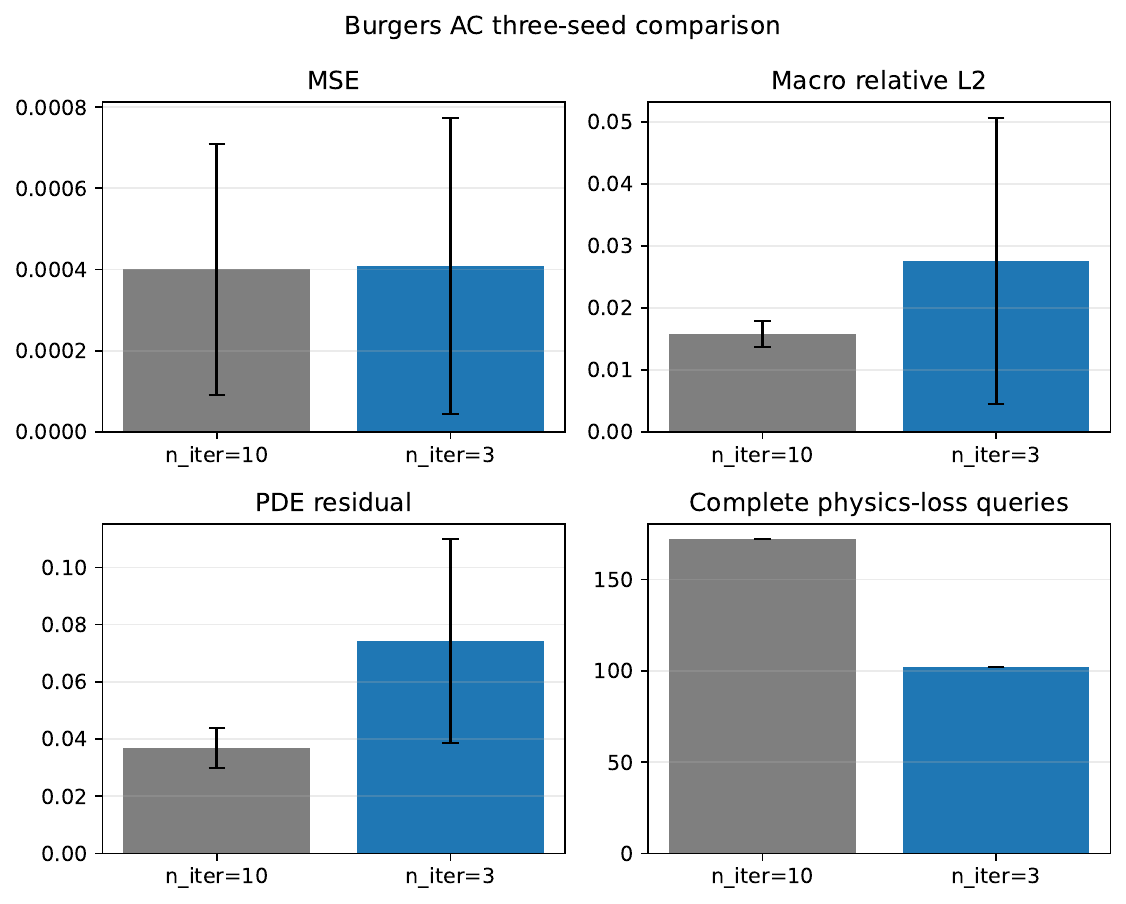}
\end{minipage}
\caption{Burgers single-case sensitivity to the number of BO evaluations per active update. Reducing the budget from 10 to 3 lowers complete objective-loss queries from approximately 172 to 102 but increases macro-averaged $E_{L_2}$ from approximately 0.0157 to 0.0277 and increases seed variation.}
\label{fig:supp-burgers-query}
\end{figure}

Figure~\ref{fig:supp-burgers-query} shows that fewer queries reduce search work but need not preserve accuracy or stability. The ordering is specific to this Burgers diagnostic and is not extrapolated to other equations.

\FloatBarrier

\subsubsection{ParamFNN depth--width sensitivity and regularization boundary}

\label{sec:supp-3-3-2}

\begin{figure}[p]
\centering
\begin{minipage}[t]{0.98\linewidth}
\centering
\suppinclude[width=\linewidth,height=0.64\textheight]{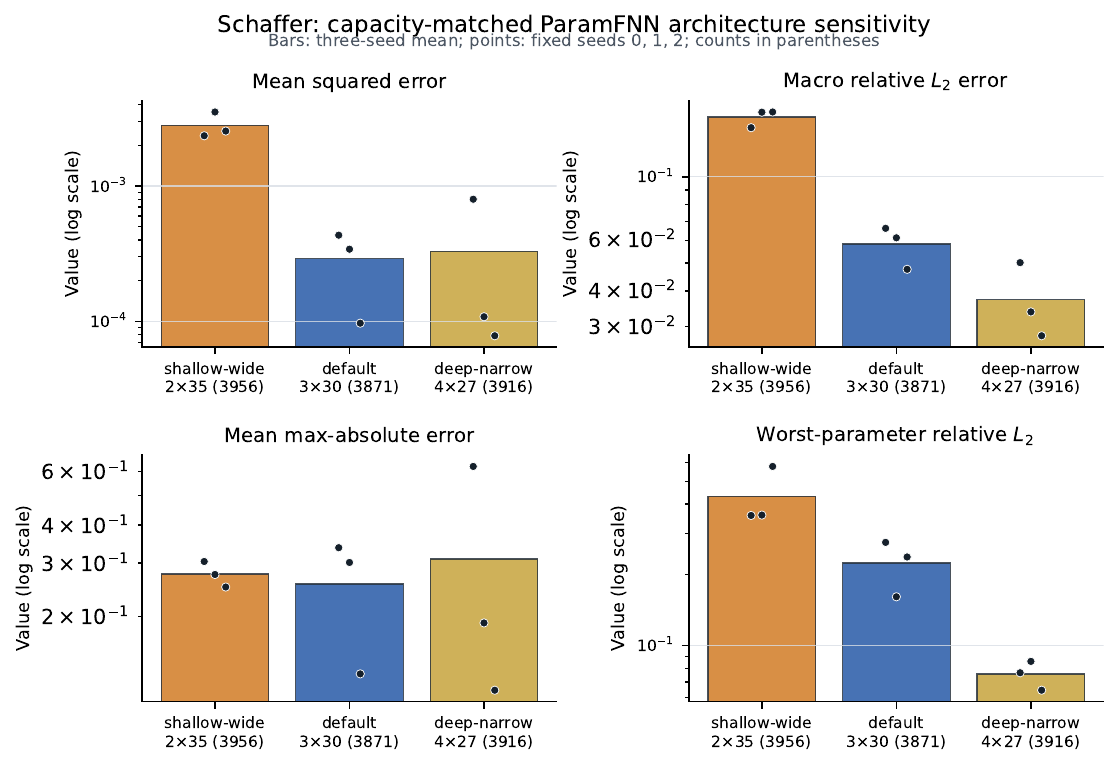}
\end{minipage}
\caption{Near-capacity-matched ParamFNN depth--width sensitivity on the Schaffer-like case. The largest parameter-count difference among the three architectures is approximately $2.2\%$. Parameter-branch decay, L-BFGS-stage freezing, and replay are unchanged in this diagnostic.}
\label{fig:supp-schaffer-paramfnn-shape}
\end{figure}

The main text separately reports the capacity-matched direct-concatenation FNN/ParamFNN comparison and the decay/freezing control conditional on ParamFNN. Figure~\ref{fig:supp-schaffer-paramfnn-shape} adds only the depth--width sensitivity of near-capacity-matched ParamFNN architectures.

The deep--narrow $4\times27$ architecture has the lowest macro-averaged and worst $E_{L_2}$, while the default $3\times30$ architecture has lower mean MSE and mean maximum absolute error; the shallow--wide architecture is weaker overall. Even at near-equal capacity, average and tail errors therefore trade off, and no single metric identifies a universally optimal depth and width.

An additional global-decay diagnostic on the ordinary direct-concatenation FNN also shows an interaction between architecture and regularization. Under that setting, Poisson--Boltzmann can outperform the unregularized ParamFNN, and Burgers improves but remains less accurate than ParamFNN. Allen--Cahn degrades substantially, while Schaffer-like and Kovasznay contain non-finite or extreme outcomes. Because the decay strengths were selected empirically, these results delimit stability and hyperparameter dependence rather than defining a new universal default.

\FloatBarrier

\subsubsection{Resource sensitivity of mini-batch training}

\label{sec:supp-3-3-3}

This single-parameter Burgers diagnostic fixes the total point exposure at $10^8$ and changes only the batch size. It characterizes a training-stage memory--time--accuracy trade-off and is not an end-to-end multi-parameter ParamPINN comparison.

\begin{table}[!htbp]
\centering
\caption{Single-parameter Burgers mini-batch diagnostic at a fixed point exposure of $10^8$. Error and wall time are reported as the mean $\pm$ population standard deviation. ``Training peak'' is the per-update peak during training; evaluation still uses the unbatched full point pool.}
\label{tab:supp-minibatch-fixed-exposure}
\footnotesize
\setlength{\tabcolsep}{4pt}
\renewcommand{\arraystretch}{1.16}
\begin{adjustbox}{max width=\linewidth,center}
\begin{tabular}{lrrrr}
\toprule
\textbf{Batch size} & \textbf{$E_{L_2}$} & \textbf{Training peak (MiB)} & \textbf{Evaluation peak (MiB)} & \textbf{Wall time (s)}\\
\midrule
Full batch (50,000) & $0.01374\pm0.0144$ & 999.1 & 923.9 & $26.6\pm7.6$\\
20,000 & $0.01215\pm0.00559$ & 392.2 & 923.9 & $37.0\pm1.5$\\
10,000 & $0.008621\pm0.00450$ & 214.2 & 923.3 & $86.9\pm13.4$\\
5,000 & $0.002574\pm0.00172$ & 115.9 & 922.8 & $120.5\pm4.0$\\
2,500 & $0.001341\pm0.000973$ & 64.6 & 923.2 & $295.5\pm1.7$\\
1,000 & $0.008593\pm0.0116$ & 36.5 & 923.3 & $698.2\pm51.9$\\
\bottomrule
\end{tabular}
\end{adjustbox}
\end{table}

\begin{figure}[p]
\centering
\begin{minipage}[t]{0.98\linewidth}
\centering
\suppinclude[width=\linewidth,height=0.64\textheight]{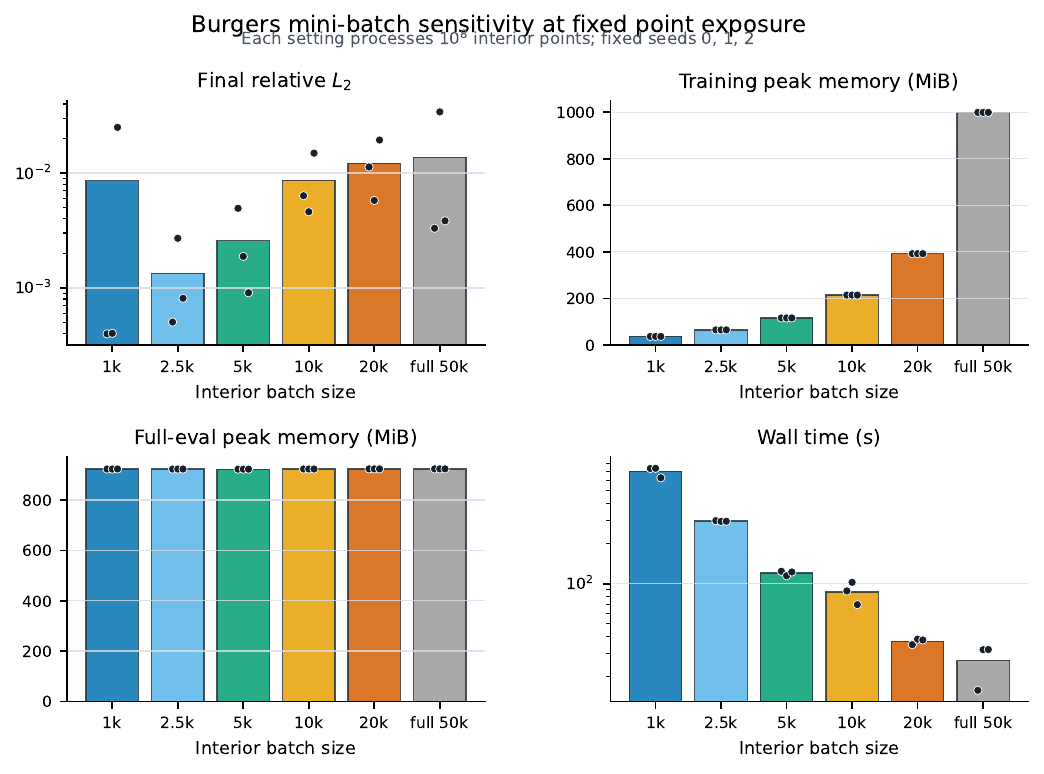}
\end{minipage}
\caption{Accuracy, training-stage per-update peak memory, and wall-time trade-off in the single-parameter Burgers diagnostic. Relative to full batch, a 2,500-point batch lowers the training peak by approximately $93.5\%$ but increases wall time by approximately $11.1\times$. The 1,000-point setting retains the poor but finite seed-2 outcome.}
\label{fig:supp-minibatch}
\end{figure}

Table~\ref{tab:supp-minibatch-fixed-exposure} and Figure~\ref{fig:supp-minibatch} show that smaller batches can substantially reduce the peak memory of a training update, but they do not reduce the peak of unbatched evaluation and do not guarantee monotonically lower error. The evidence supports only the three-way trade-off under the fixed-exposure, single-parameter protocol; it does not support an end-to-end ParamPINN memory advantage or an equal-wall-time accuracy advantage.

\FloatBarrier

\subsection{Training cost and search overhead}

\label{sec:supp-3-4}

The main text reports approximate times and query efficiency only where they directly support the principal conclusions. This section provides the complete exclusive-GPU timing audit for the locked configurations and separately decomposes the search cost of AG and AC. The audit uses seed 0, FP32, and identical Tesla V100S-PCIE-32GB GPUs. All $30/30$ case--method runs completed, and search time is already included in Adam time.

\FloatBarrier

\subsubsection{Training times for six methods on five cases}

\label{sec:supp-3-4-1}

\begin{sidewaystable}[p]
\centering
\caption{Adam, L-BFGS, and total training wall-clock times for six methods on five cases (minutes). Each three-row block gives the two optimizer stages and their total. The \texttt{\detokenize{--no-test}} timing audit excludes evaluation on the reference-solution grid.}
\label{tab:supp-fair-timing-stages}
\small
\setlength{\tabcolsep}{4pt}
\renewcommand{\arraystretch}{1.16}
\begin{adjustbox}{max width=\linewidth,center}
\begin{tabular}{llrrrrrr}
\toprule
\textbf{Case} & \textbf{Stage} & \textbf{UNI} & \textbf{FIX} & \textbf{AG} & \textbf{AC} & \textbf{ACR} & \textbf{ACR2-arch}\\
\midrule
Schaffer-like & Adam & 1.77 & 1.78 & 1.78 & 2.18 & 2.17 & 2.62\\
 & L-BFGS & 11.92 & 11.98 & 11.93 & 12.60 & 12.42 & 12.43\\
 & Total & 13.69 & 13.76 & 13.71 & 14.78 & 14.59 & 15.05\\
Burgers & Adam & 4.60 & 4.67 & 4.78 & 4.78 & 4.76 & 6.64\\
 & L-BFGS & 14.19 & 14.45 & 14.28 & 14.26 & 14.47 & 16.71\\
 & Total & 18.80 & 19.12 & 19.06 & 19.04 & 19.23 & 23.35\\
Allen--Cahn & Adam & 8.03 & 7.54 & 7.92 & 8.40 & 8.10 & 10.33\\
 & L-BFGS & 8.30 & 8.19 & 8.17 & 8.13 & 8.40 & 9.18\\
 & Total & 16.33 & 15.73 & 16.09 & 16.54 & 16.50 & 19.51\\
Kovasznay & Adam & 7.06 & 6.80 & 7.12 & 7.34 & 7.65 & 8.80\\
 & L-BFGS & 11.45 & 11.19 & 11.36 & 11.76 & 12.00 & 12.80\\
 & Total & 18.51 & 17.99 & 18.48 & 19.10 & 19.64 & 21.59\\
Linearized Poisson--Boltzmann & Adam & 37.13 & 34.90 & 77.39 & 75.96 & 90.49 & 104.55\\
 & L-BFGS & 42.52 & 36.96 & 40.89 & 50.06 & 45.06 & 47.44\\
 & Total & 79.65 & 71.86 & 118.28 & 126.02 & 135.56 & 151.99\\
\bottomrule
\end{tabular}
\end{adjustbox}
\end{sidewaystable}

Table~\ref{tab:supp-fair-timing-stages} gives within-equation longest/shortest ratios of $1.10$--$1.24\times$ for the first four cases. On Poisson--Boltzmann, UNI and FIX request 20k Adam updates, whereas the dynamic methods request 40k. Its larger time differences therefore arise mainly from the doubled update budget plus additional search events; this row is not a strict equal-update comparison.

\FloatBarrier

\subsubsection{Search-time decomposition for AG and AC}

\label{sec:supp-3-4-2}

\begin{table}[!htbp]
\centering
\caption{Search-time decomposition for AG and AC in the exclusive-GPU timing audit. Loss-evaluation time covers complete objective-loss queries, whereas GP/selection time covers surrogate fitting and candidate selection. Total search time is already included in Adam time.}
\label{tab:supp-fair-timing-search}
\footnotesize
\setlength{\tabcolsep}{3pt}
\renewcommand{\arraystretch}{1.16}
\begin{tabularx}{\linewidth}{@{}p{0.13\linewidth}*{6}{>{\centering\arraybackslash}X}@{}}
\toprule
\textbf{Case} & \scalebox{0.97}{\makecell{\textbf{AG: loss-}\\\textbf{evaluation}\\\textbf{time}}} & \makecell{\textbf{AG: GP/}\\\textbf{selection}\\\textbf{time}} & \textbf{AG: total search} & \scalebox{0.97}{\makecell{\textbf{AC: loss-}\\\textbf{evaluation}\\\textbf{time}}} & \makecell{\textbf{AC: GP/}\\\textbf{selection}\\\textbf{time}} & \textbf{AC: total search}\\
\midrule
Schaffer-like & 0.83 s & 0.01 s & 0.85 s & 0.67 s & 13.35 s & 14.02 s\\
Burgers & 5.31 s & 0.03 s & 5.34 s & 1.01 s & 6.48 s & 7.49 s\\
Allen--Cahn & 9.57 s & 0.04 s & 9.61 s & 2.80 s & 12.56 s & 15.36 s\\
Kovasznay & 20.16 s & 0.05 s & 20.21 s & 4.60 s & 6.56 s & 11.16 s\\
Linearized Poisson--Boltzmann & 28.97 min & 0.08 min & 29.05 min & 2.38 min & 8.21 min & 10.58 min\\
\bottomrule
\end{tabularx}
\end{table}

The main text reports that AC requires fewer complete objective-loss queries than AG. Table~\ref{tab:supp-fair-timing-search} shows why fewer queries do not necessarily imply shorter search wall time. On Schaffer-like, Burgers, and Allen--Cahn, GP fitting and selection cost more than the saved loss evaluations. Loss evaluation accounts for a larger share of the search cost on Kovasznay and four-parameter Poisson--Boltzmann, for which AC has the shorter total search time. Fewer objective-loss queries are therefore supported across cases, whereas faster search is supported only for the latter two cases.

\FloatBarrier

\clearpage

\section{Full error atlas}

\label{sec:supp-part-4}

This part collects the parameter-wise errors and optimization histories of all six methods. Every error panel uses the same fixed test parameters and evaluation grid as the corresponding aggregate table in the main text, supplementing the aggregate results with localized difficult regions, seed variation, and multi-parameter interactions.

\FloatBarrier

\subsection{Six-method parameter-space error atlas and optimization histories}

\label{sec:supp-4-1}

All results in this section use FP32. Schaffer-like and Burgers show the three fixed seeds and their summary curves; Allen--Cahn uses the two-dimensional $(\nu,\rho)$ parameter grid; Kovasznay reports $u$, $v$, and $p$ separately; and four-parameter linearized Poisson--Boltzmann uses conditional two-dimensional slices with the held coordinates stated explicitly.

Curves labeled ``training loss'' and ``validation loss'' are physics/constraint losses recorded by the training framework; the validation loss is evaluated on independent physical points and is not a reference-solution error. Only Schaffer-like additionally includes a reference-solution $E_{L_2}$ history. That history is diagnostic and is not used for early stopping, parameter selection, or model selection.

\FloatBarrier

\subsubsection{Schaffer-like continuous function}

\label{sec:supp-4-1-1}

\begin{figure}[p]
\centering
\begin{minipage}[t]{0.485\linewidth}
\centering
\textbf{UNI}\par\smallskip
\suppinclude[width=\linewidth,height=0.225\textheight]{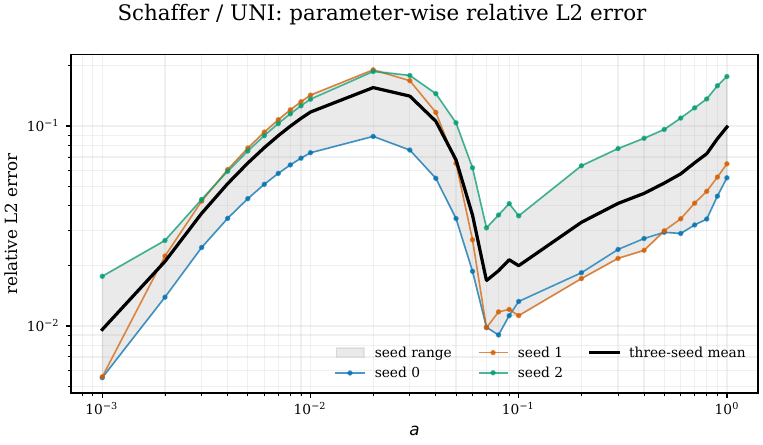}
\end{minipage}
\hfill
\begin{minipage}[t]{0.485\linewidth}
\centering
\textbf{FIX}\par\smallskip
\suppinclude[width=\linewidth,height=0.225\textheight]{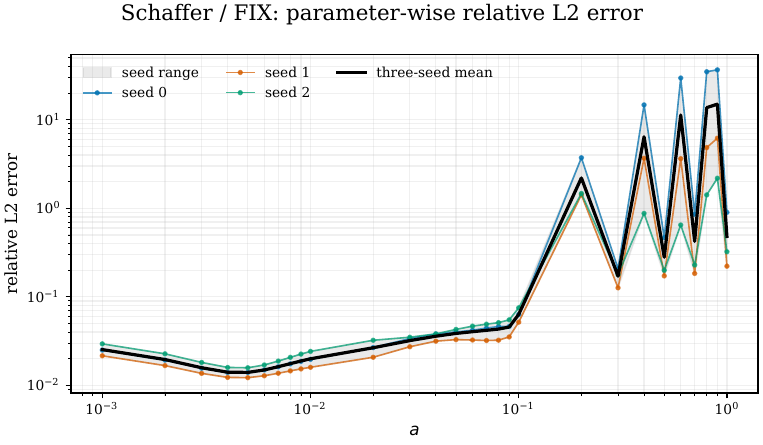}
\end{minipage}
\par\medskip
\begin{minipage}[t]{0.485\linewidth}
\centering
\textbf{AG}\par\smallskip
\suppinclude[width=\linewidth,height=0.225\textheight]{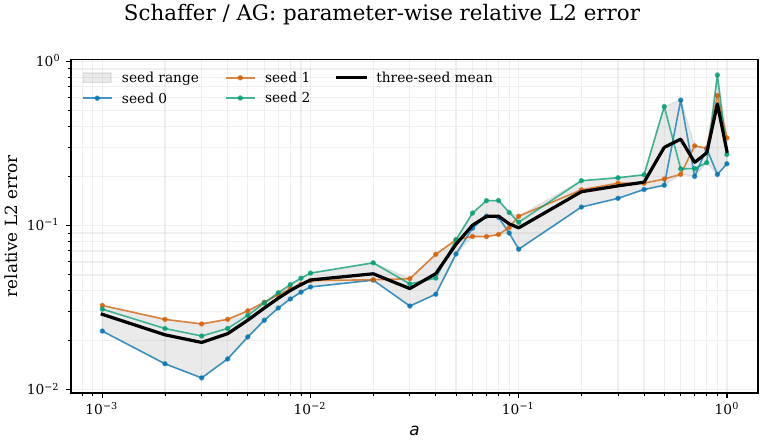}
\end{minipage}
\hfill
\begin{minipage}[t]{0.485\linewidth}
\centering
\textbf{AC}\par\smallskip
\suppinclude[width=\linewidth,height=0.225\textheight]{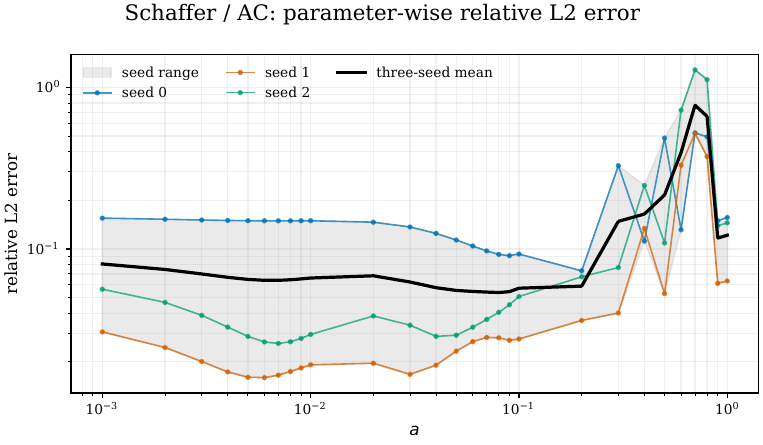}
\end{minipage}
\par\medskip
\begin{minipage}[t]{0.485\linewidth}
\centering
\textbf{ACR}\par\smallskip
\suppinclude[width=\linewidth,height=0.225\textheight]{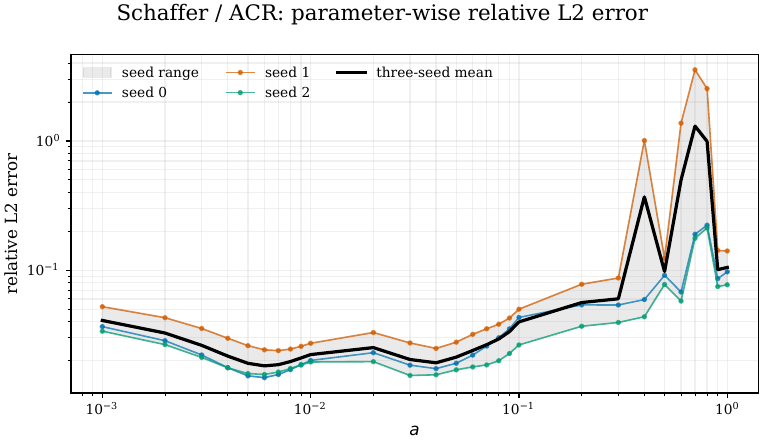}
\end{minipage}
\hfill
\begin{minipage}[t]{0.485\linewidth}
\centering
\textbf{ACR2-arch}\par\smallskip
\suppinclude[width=\linewidth,height=0.225\textheight]{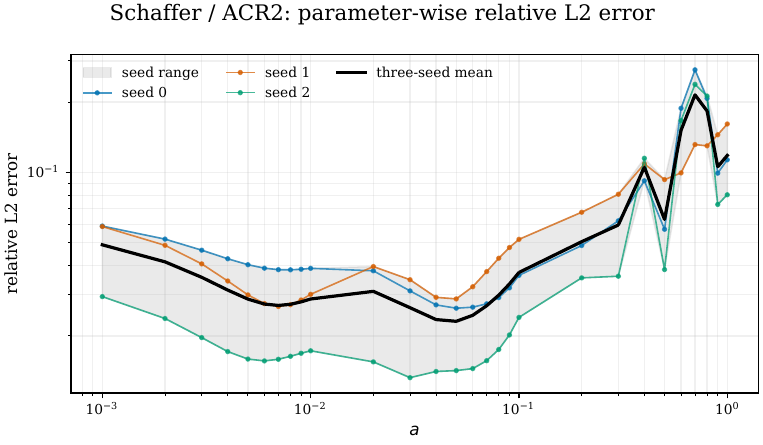}
\end{minipage}
\caption{$E_{L_2}$ distributions of the six methods over the Schaffer-like test parameters. Each method shows the three fixed seeds and their summary curve; ACR2-arch denotes the global ParamFNN configuration used in the main results table.}
\label{fig:supp-expanded-schaffer-l2}
\end{figure}

\begin{figure}[p]
\centering
\begin{minipage}[t]{0.76\linewidth}
\centering
\textbf{Training physics/constraint loss}\par\smallskip
\suppinclude[width=\linewidth,height=0.31\textheight]{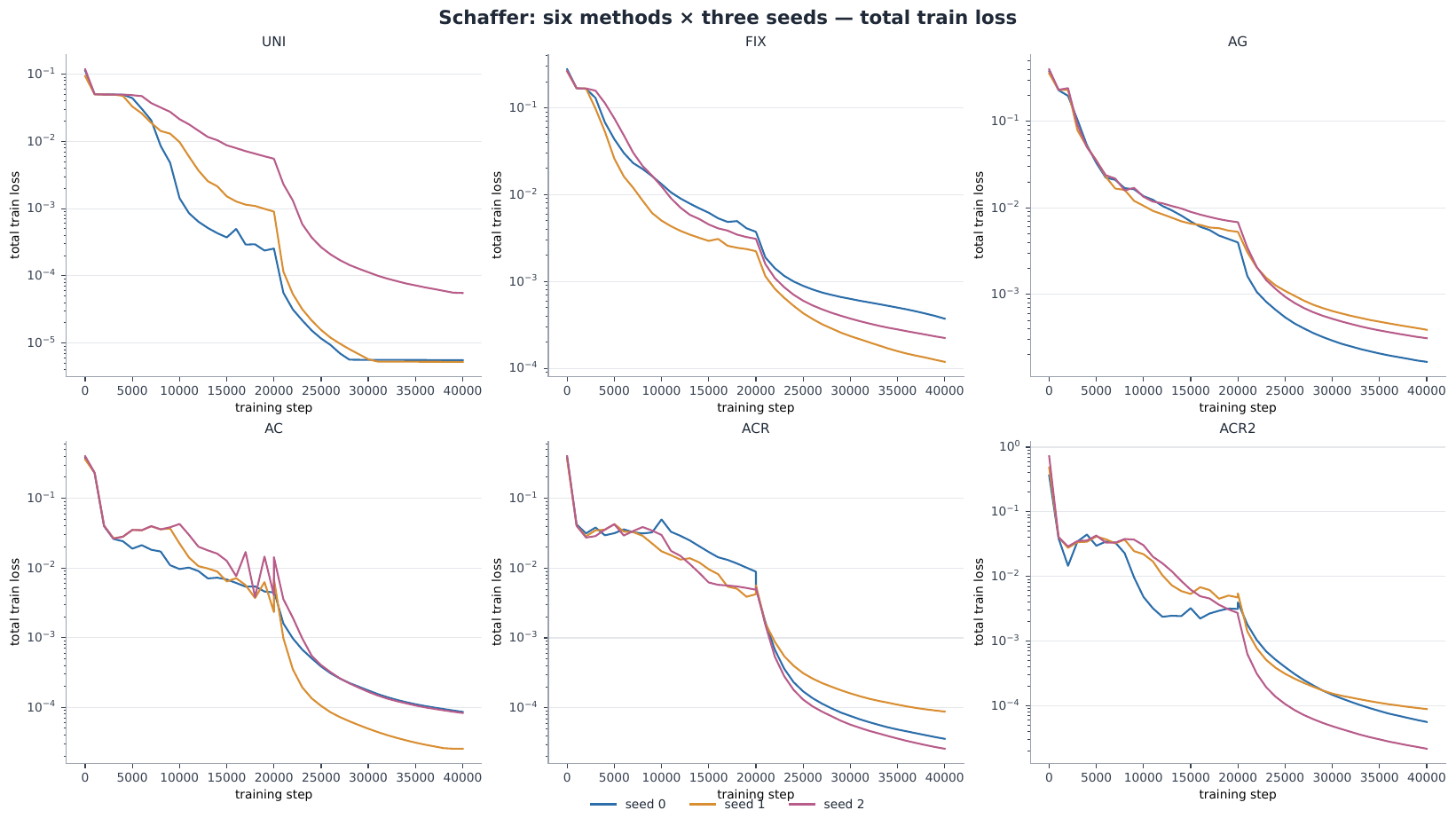}
\end{minipage}
\par\medskip
\begin{minipage}[t]{0.76\linewidth}
\centering
\textbf{Validation physics/constraint loss}\par\smallskip
\suppinclude[width=\linewidth,height=0.31\textheight]{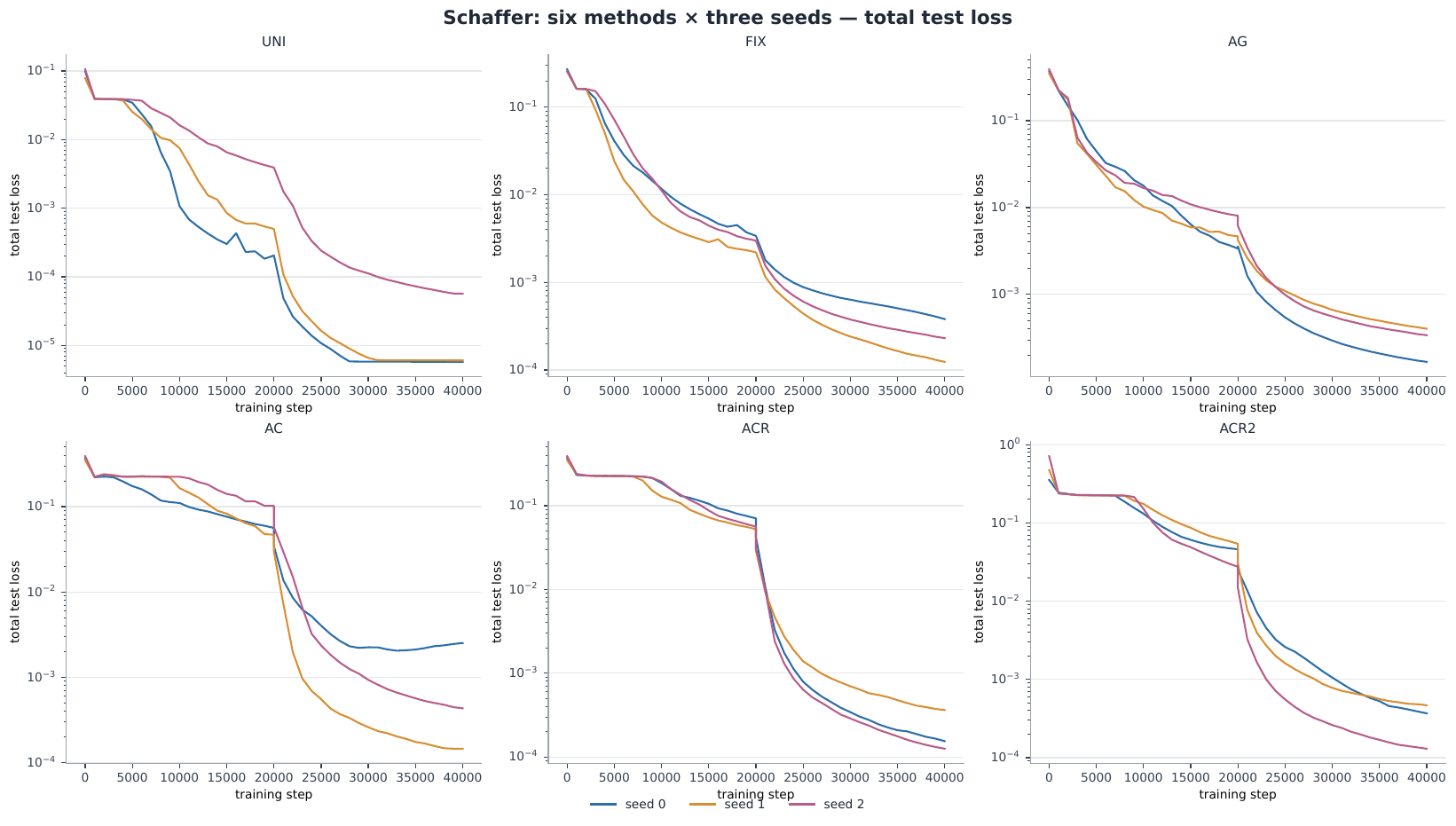}
\end{minipage}
\caption{Training and validation physics/constraint losses of the six methods on Schaffer-like over three seeds. The curves retain the Adam/L-BFGS transition and the actual recorded step counts and are not normalized by wall-clock time; the validation panel is not a reference-solution error.}
\label{fig:supp-expanded-schaffer-loss}
\end{figure}

\begin{figure}[p]
\centering
\begin{minipage}[t]{0.90\linewidth}
\centering
\suppinclude[width=\linewidth,height=0.64\textheight]{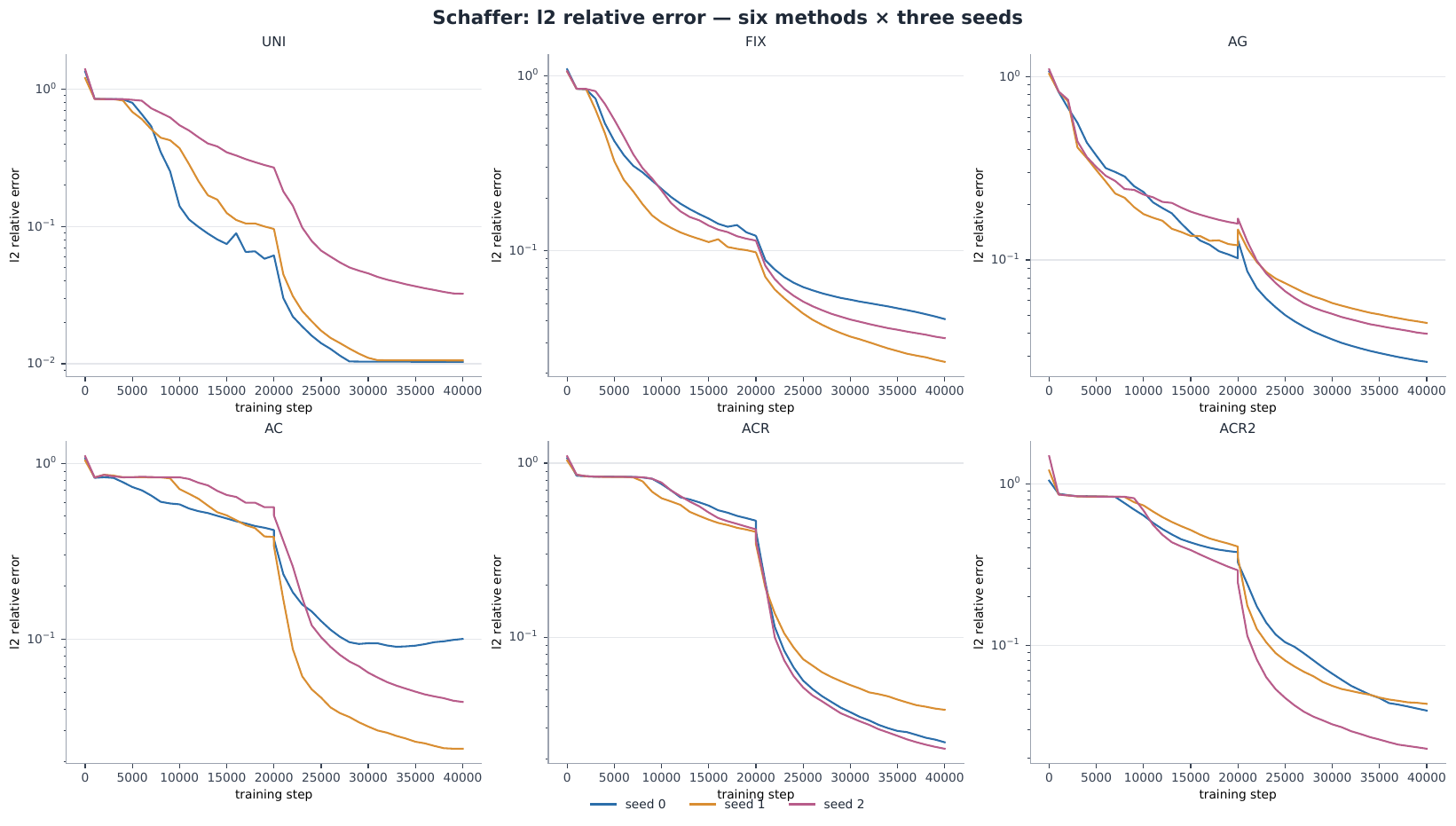}
\end{minipage}
\caption{Reference-solution $E_{L_2}$ histories of the six methods on Schaffer-like over the three fixed seeds. This metric is not used for training, Bayesian selection, dynamic weighting, early stopping, or model selection.}
\label{fig:supp-expanded-schaffer-reference}
\end{figure}

Figures~\ref{fig:supp-expanded-schaffer-l2}, \ref{fig:supp-expanded-schaffer-loss}, and \ref{fig:supp-expanded-schaffer-reference} show that error peaks and seed variation are concentrated in selected parameter regions and cannot be represented by one aggregate value. ACR and ACR2-arch reduce error over most parameters but retain a few local peaks. Together with the controlled ablation in the main text, the evidence indicates that dynamic weighting provides the dominant endpoint improvement in this case, while replay and the parameter subnetwork further improve the aggregate result. Physics/constraint loss and $E_{L_2}$ do not always have the same ordering, so final accuracy is assessed using reference-solution error.

\FloatBarrier

\subsubsection{Burgers equation}

\label{sec:supp-4-1-2}

\begin{figure}[p]
\centering
\begin{minipage}[t]{0.485\linewidth}
\centering
\textbf{UNI}\par\smallskip
\suppinclude[width=\linewidth,height=0.225\textheight]{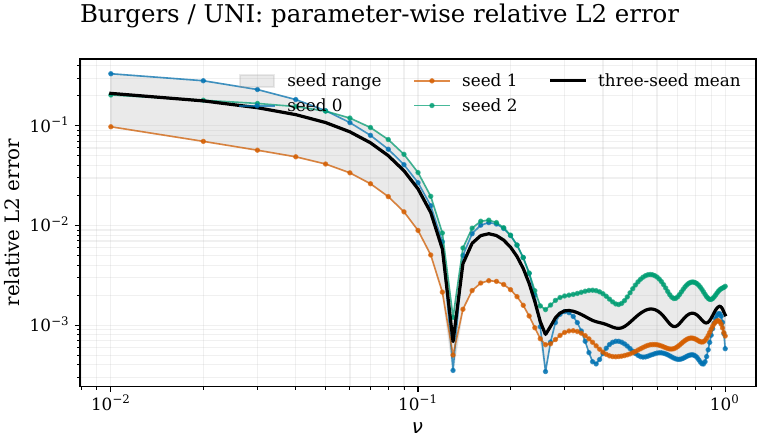}
\end{minipage}
\hfill
\begin{minipage}[t]{0.485\linewidth}
\centering
\textbf{FIX}\par\smallskip
\suppinclude[width=\linewidth,height=0.225\textheight]{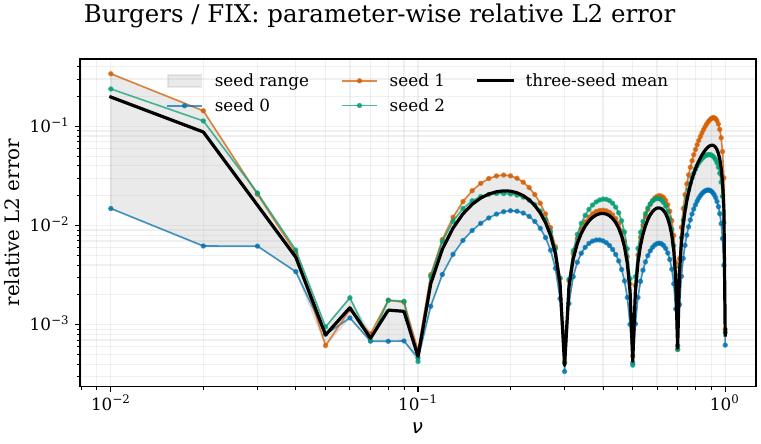}
\end{minipage}
\par\medskip
\begin{minipage}[t]{0.485\linewidth}
\centering
\textbf{AG}\par\smallskip
\suppinclude[width=\linewidth,height=0.225\textheight]{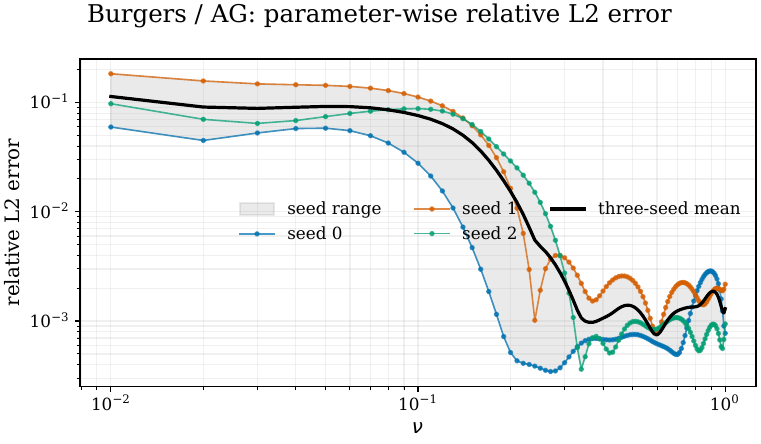}
\end{minipage}
\hfill
\begin{minipage}[t]{0.485\linewidth}
\centering
\textbf{AC}\par\smallskip
\suppinclude[width=\linewidth,height=0.225\textheight]{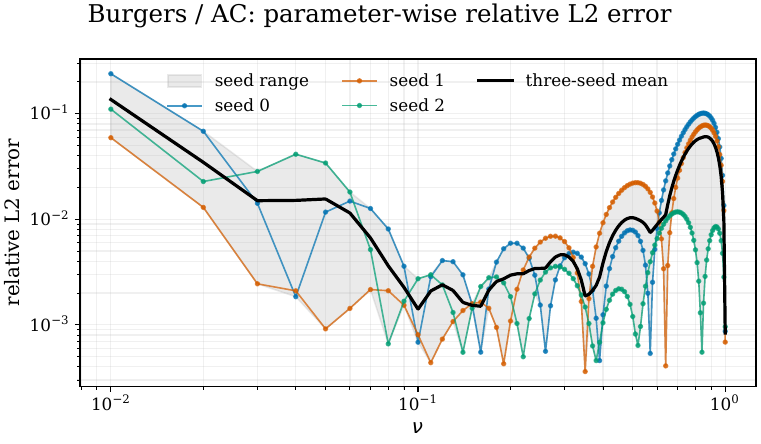}
\end{minipage}
\par\medskip
\begin{minipage}[t]{0.485\linewidth}
\centering
\textbf{ACR}\par\smallskip
\suppinclude[width=\linewidth,height=0.225\textheight]{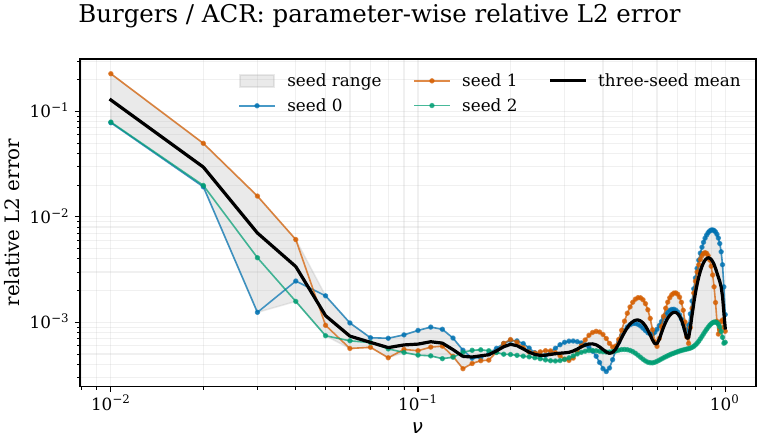}
\end{minipage}
\hfill
\begin{minipage}[t]{0.485\linewidth}
\centering
\textbf{ACR2-arch}\par\smallskip
\suppinclude[width=\linewidth,height=0.225\textheight]{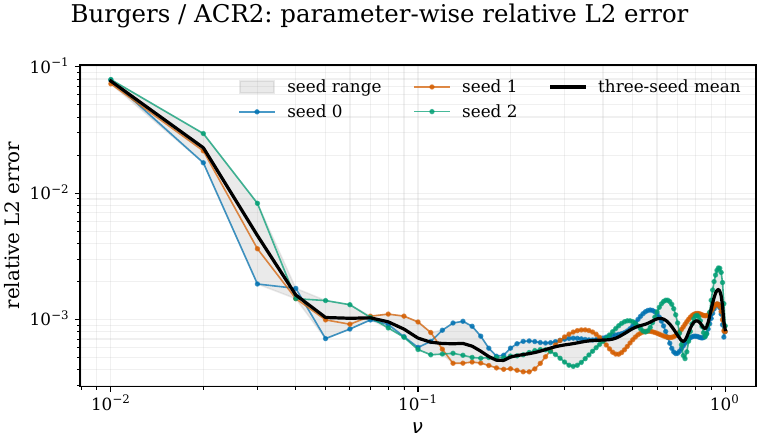}
\end{minipage}
\caption{$E_{L_2}$ distributions of the six methods over the Burgers viscosity interval $\nu\in[0.01,1]$. Each method shows the three fixed seeds and their summary curve.}
\label{fig:supp-expanded-burgers-l2}
\end{figure}

\begin{figure}[p]
\centering
\begin{minipage}[t]{0.76\linewidth}
\centering
\textbf{Training physics/constraint loss}\par\smallskip
\suppinclude[width=\linewidth,height=0.31\textheight]{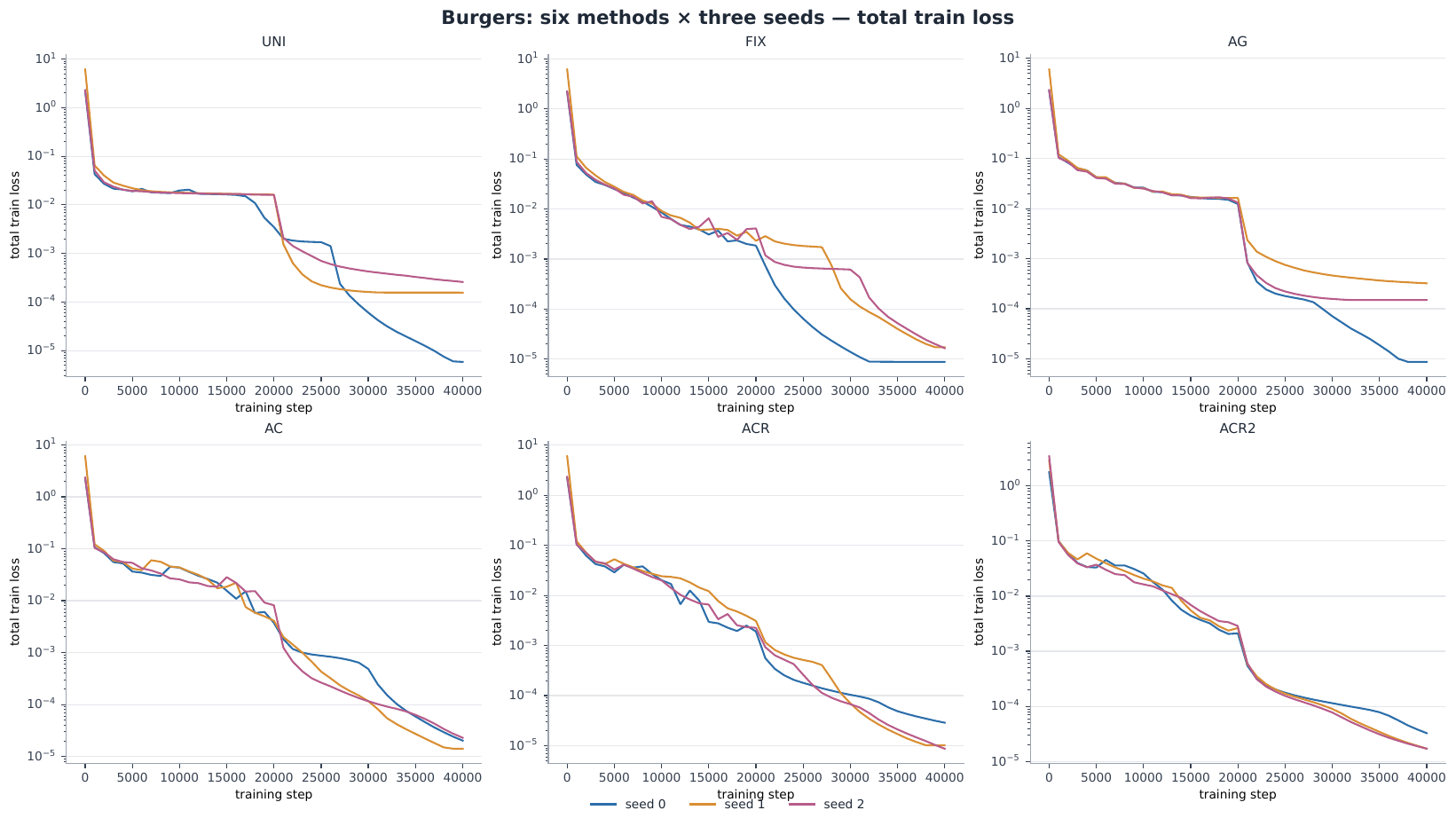}
\end{minipage}
\par\medskip
\begin{minipage}[t]{0.76\linewidth}
\centering
\textbf{Validation physics/constraint loss}\par\smallskip
\suppinclude[width=\linewidth,height=0.31\textheight]{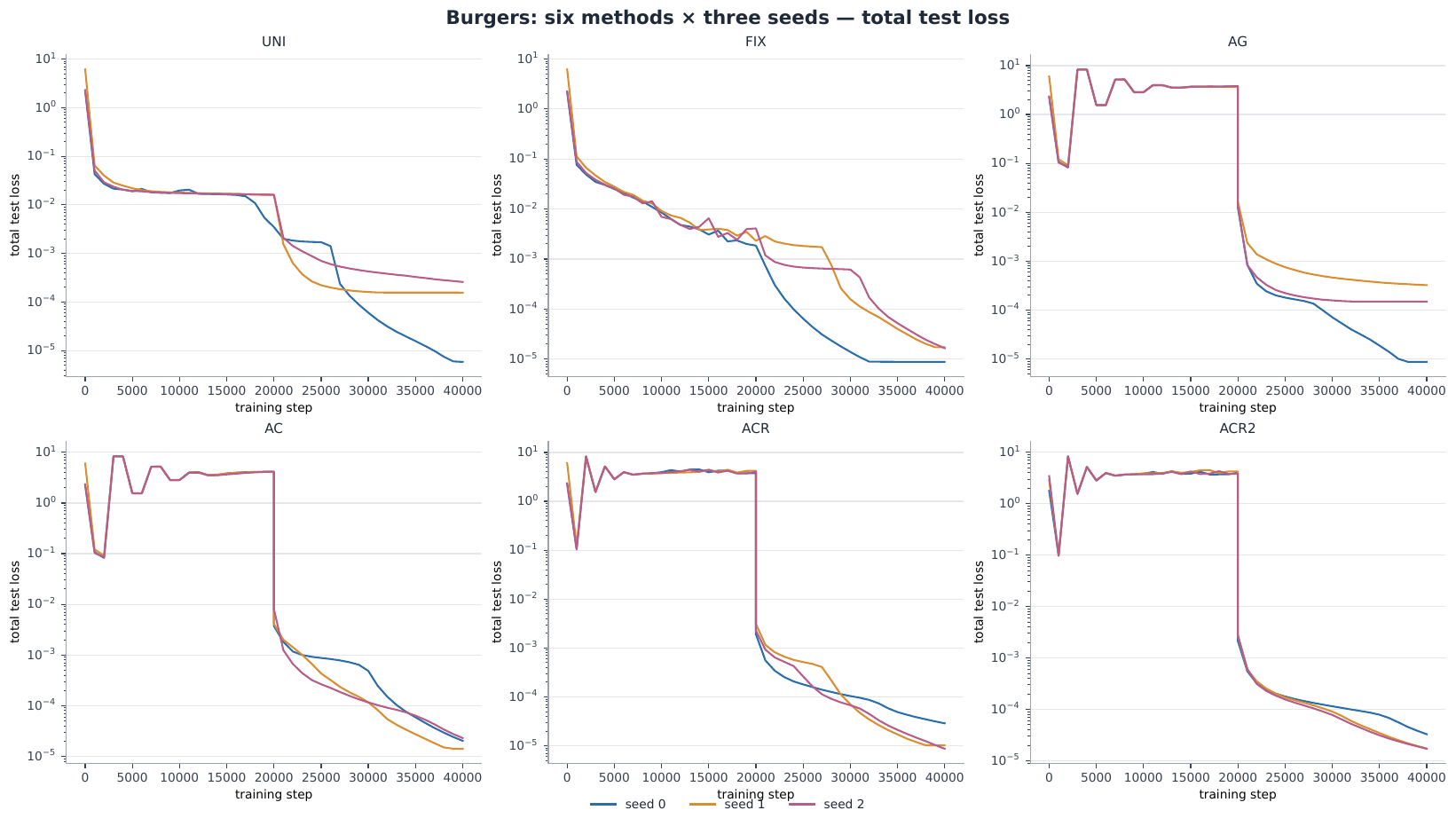}
\end{minipage}
\caption{Training and validation physics/constraint losses of the six methods on Burgers over three seeds. Validation is evaluated on an independent set of physical points and is not a reference-solution error.}
\label{fig:supp-expanded-burgers-loss}
\end{figure}

Figures~\ref{fig:supp-expanded-burgers-l2} and \ref{fig:supp-expanded-burgers-loss} show that all six methods are less accurate in the low-viscosity region, where the near-shock structure is most difficult. ACR2-arch maintains lower error and smaller seed variation over most of the $\nu$ interval, followed by ACR, whereas AC retains more pronounced local peaks. Training and validation physics losses reveal optimizer-stage behavior, but the method comparison is based on $E_{L_2}$ over the complete $\nu$ grid.

\FloatBarrier

\subsubsection{Allen--Cahn equation}

\label{sec:supp-4-1-3}

\begin{figure}[p]
\centering
\begin{minipage}[t]{0.76\linewidth}
\centering
\textbf{UNI}\par\smallskip
\suppinclude[width=\linewidth,height=0.31\textheight]{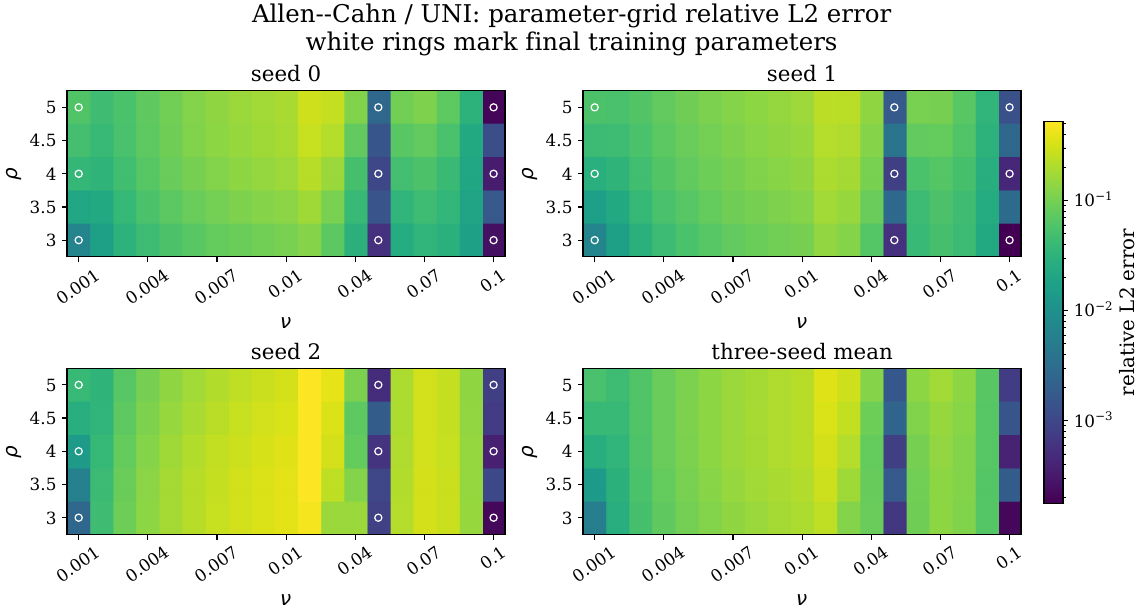}
\end{minipage}
\par\medskip
\begin{minipage}[t]{0.76\linewidth}
\centering
\textbf{FIX}\par\smallskip
\suppinclude[width=\linewidth,height=0.31\textheight]{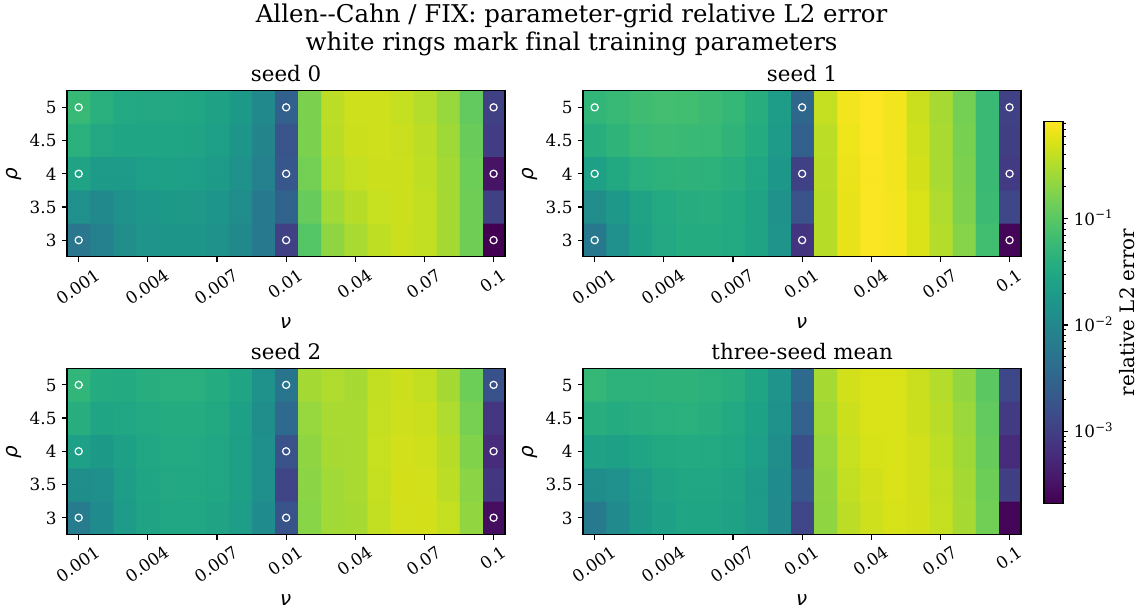}
\end{minipage}
\caption{$E_{L_2}$ of UNI and FIX over the Allen--Cahn $(\nu,\rho)$ test grid. Each method shows the three fixed seeds and their mean; white rings mark the retained training parameters.}
\label{fig:supp-expanded-allen-l2-baselines}
\end{figure}

\begin{figure}[p]
\centering
\begin{minipage}[t]{0.76\linewidth}
\centering
\textbf{AG}\par\smallskip
\suppinclude[width=\linewidth,height=0.31\textheight]{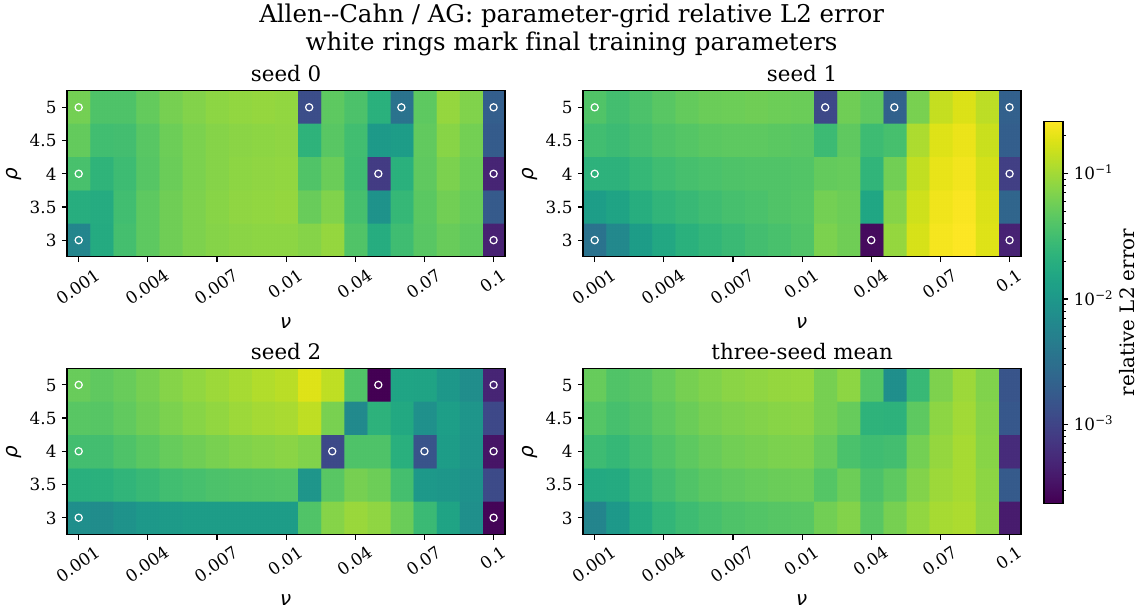}
\end{minipage}
\par\medskip
\begin{minipage}[t]{0.76\linewidth}
\centering
\textbf{AC}\par\smallskip
\suppinclude[width=\linewidth,height=0.31\textheight]{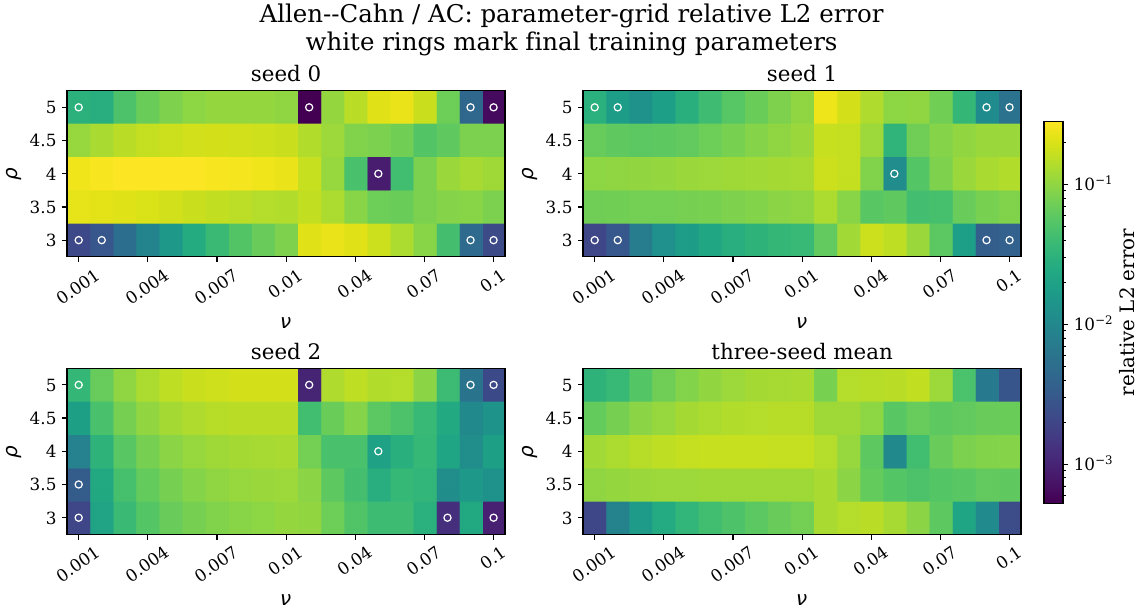}
\end{minipage}
\caption{$E_{L_2}$ of AG and AC over the Allen--Cahn $(\nu,\rho)$ test grid. Seed-specific local high-error regions are retained in full.}
\label{fig:supp-expanded-allen-l2-active}
\end{figure}

\begin{figure}[p]
\centering
\begin{minipage}[t]{0.76\linewidth}
\centering
\textbf{ACR}\par\smallskip
\suppinclude[width=\linewidth,height=0.31\textheight]{figures/supplement/02_practical_challenges/03_parameter_space_generalization_and_overfitting/supp_allen_errors.png}
\end{minipage}
\par\medskip
\begin{minipage}[t]{0.76\linewidth}
\centering
\textbf{ACR2-arch}\par\smallskip
\suppinclude[width=\linewidth,height=0.31\textheight]{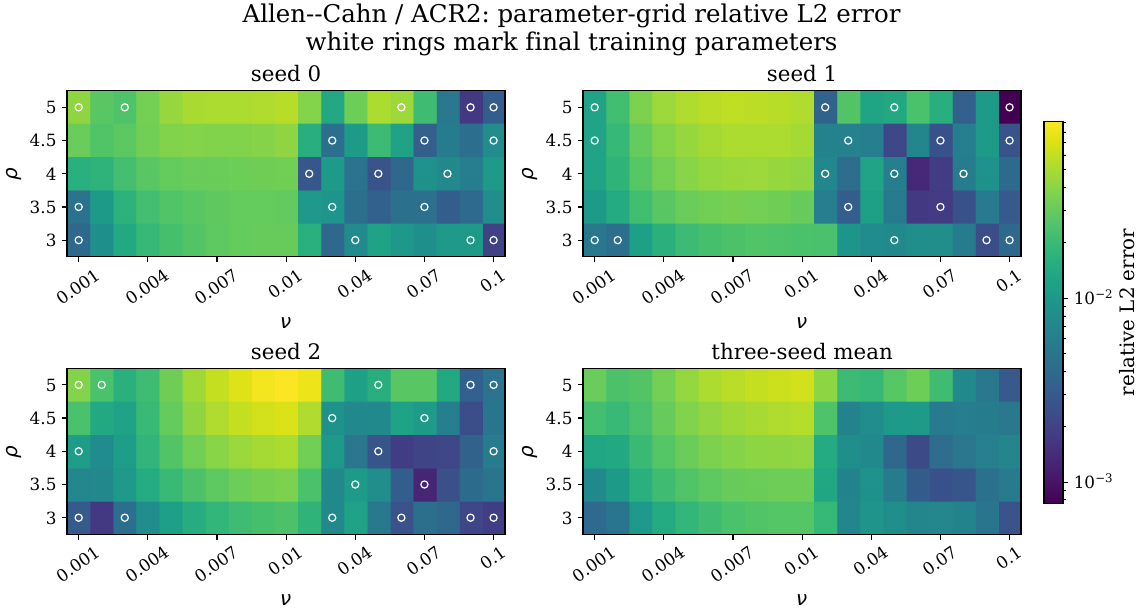}
\end{minipage}
\caption{$E_{L_2}$ of ACR and ACR2-arch over the Allen--Cahn $(\nu,\rho)$ test grid. ACR2-arch denotes the global ParamFNN configuration used in the main results table.}
\label{fig:supp-expanded-allen-l2-replay}
\end{figure}

\begin{figure}[p]
\centering
\begin{minipage}[t]{0.76\linewidth}
\centering
\textbf{Training physics/constraint loss}\par\smallskip
\suppinclude[width=\linewidth,height=0.31\textheight]{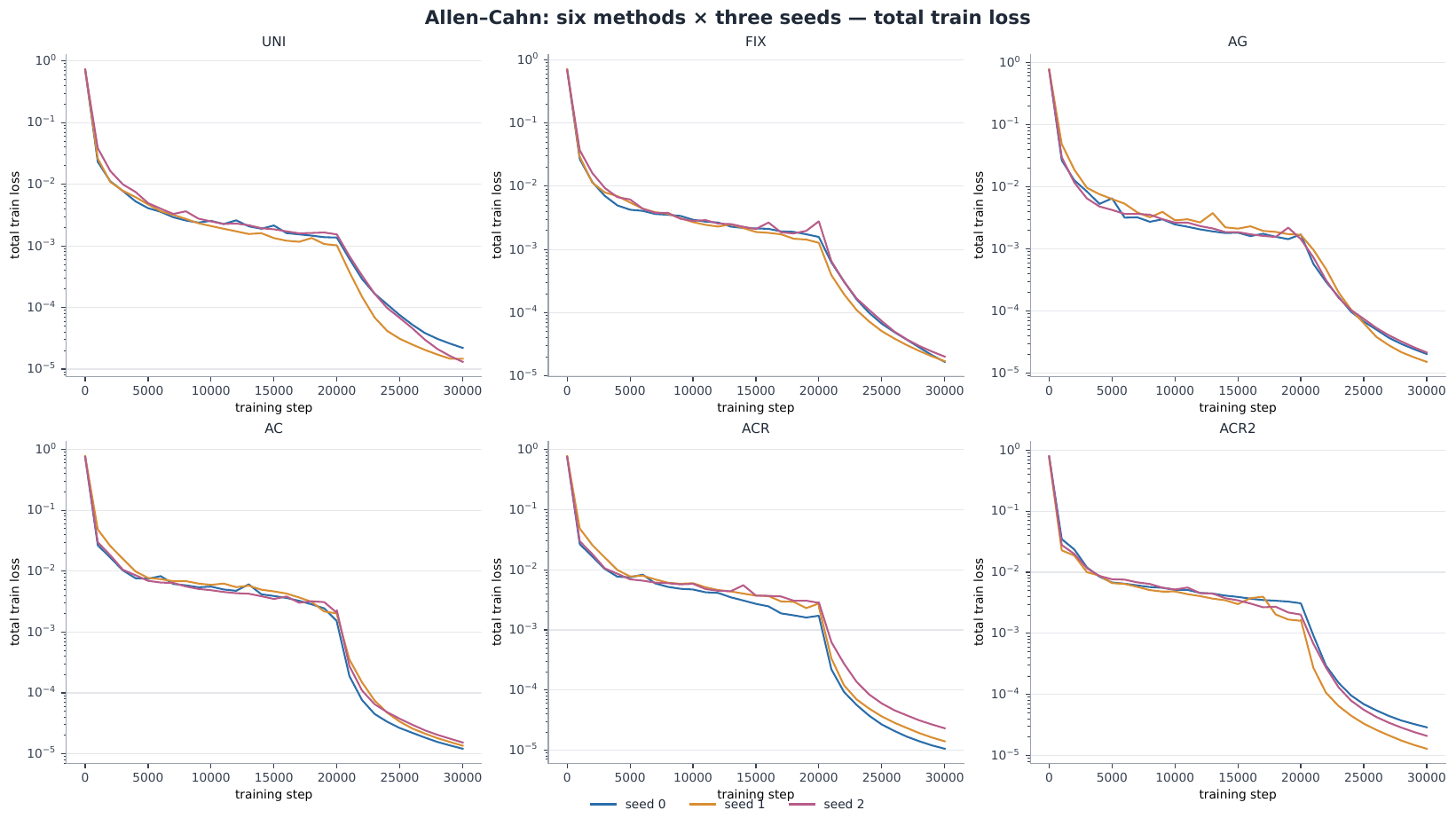}
\end{minipage}
\par\medskip
\begin{minipage}[t]{0.76\linewidth}
\centering
\textbf{Validation physics/constraint loss}\par\smallskip
\suppinclude[width=\linewidth,height=0.31\textheight]{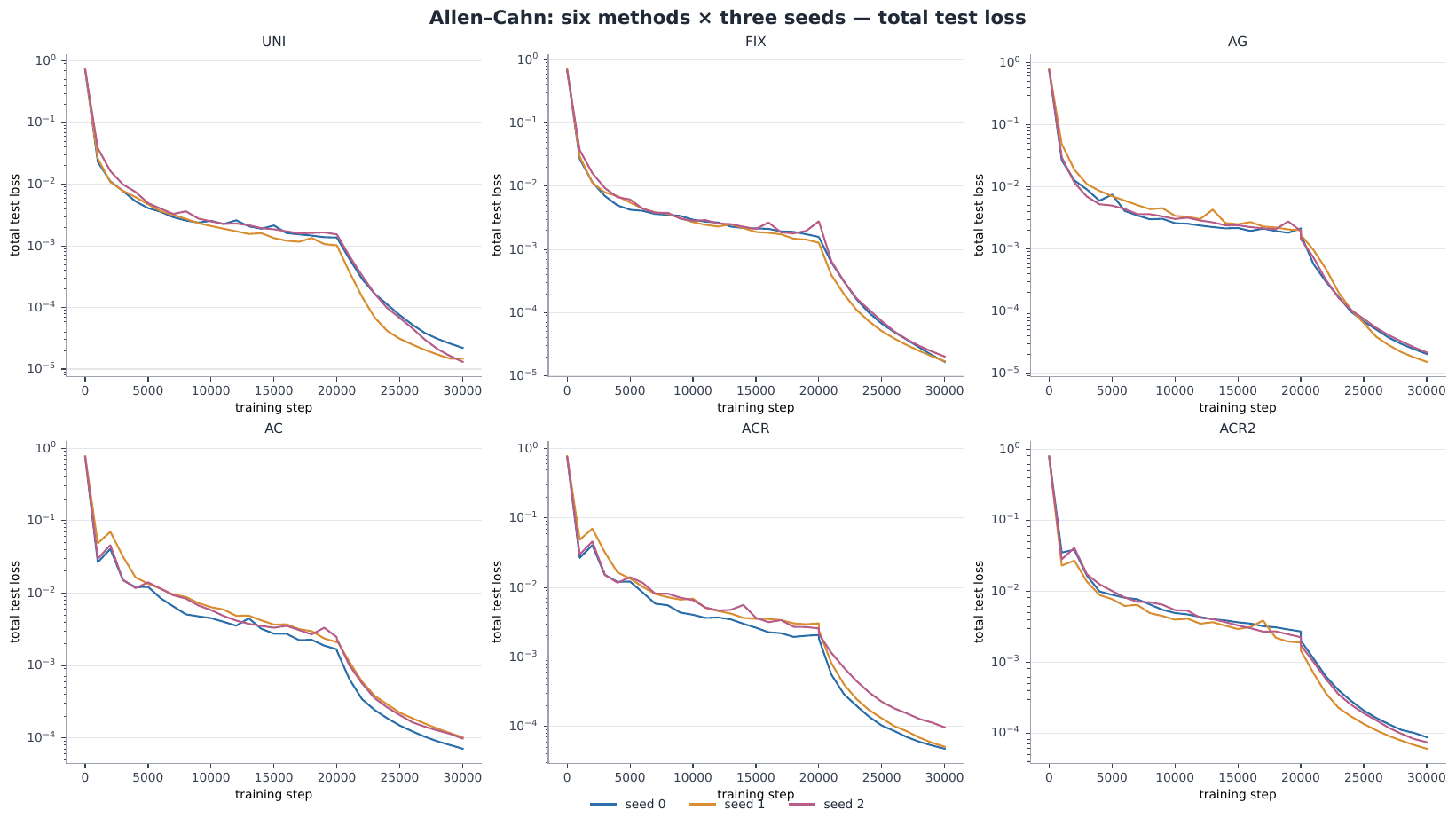}
\end{minipage}
\caption{Training and validation physics/constraint losses of the six methods on Allen--Cahn over three seeds. All results use the boundary condition $u(\pm1,t)=-1$; neither panel is a reference-solution error history.}
\label{fig:supp-expanded-allen-loss}
\end{figure}

Figures~\ref{fig:supp-expanded-allen-l2-baselines}, \ref{fig:supp-expanded-allen-l2-active}, \ref{fig:supp-expanded-allen-l2-replay}, and \ref{fig:supp-expanded-allen-loss} show that local high-error regions vary across seeds, although larger $\rho$ and some intermediate $\nu$ values are repeatedly difficult. ACR is more accurate overall than AC and gives the lowest mean error in the main table, whereas ACR2-arch does not improve further; the benefit of the parameter subnetwork is therefore equation dependent. White rings indicate retained training parameters but do not imply minimum error in their neighborhoods.

\FloatBarrier

\subsubsection{Kovasznay flow}

\label{sec:supp-4-1-4}

\begin{figure}[p]
\centering
\begin{minipage}[t]{0.485\linewidth}
\centering
\textbf{UNI}\par\smallskip
\suppinclude[width=\linewidth,height=0.225\textheight]{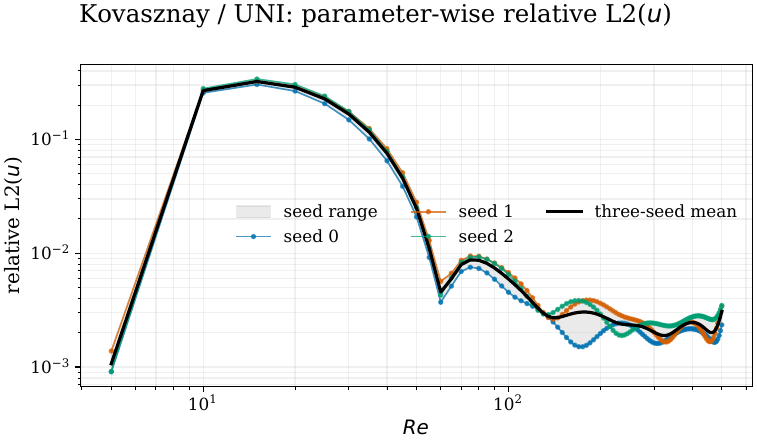}
\end{minipage}
\hfill
\begin{minipage}[t]{0.485\linewidth}
\centering
\textbf{FIX}\par\smallskip
\suppinclude[width=\linewidth,height=0.225\textheight]{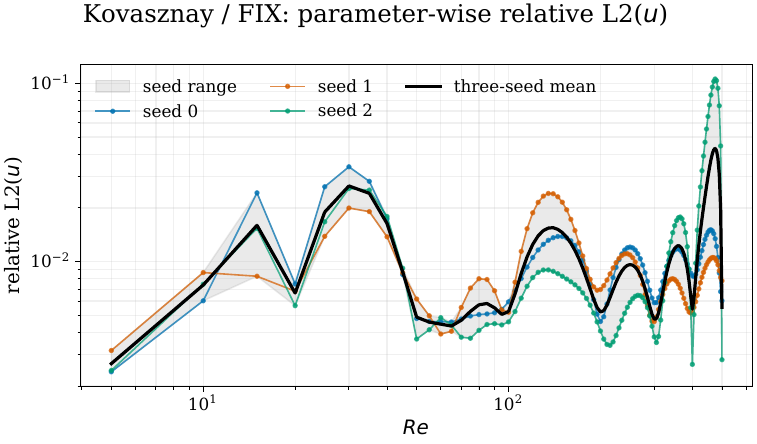}
\end{minipage}
\par\medskip
\begin{minipage}[t]{0.485\linewidth}
\centering
\textbf{AG}\par\smallskip
\suppinclude[width=\linewidth,height=0.225\textheight]{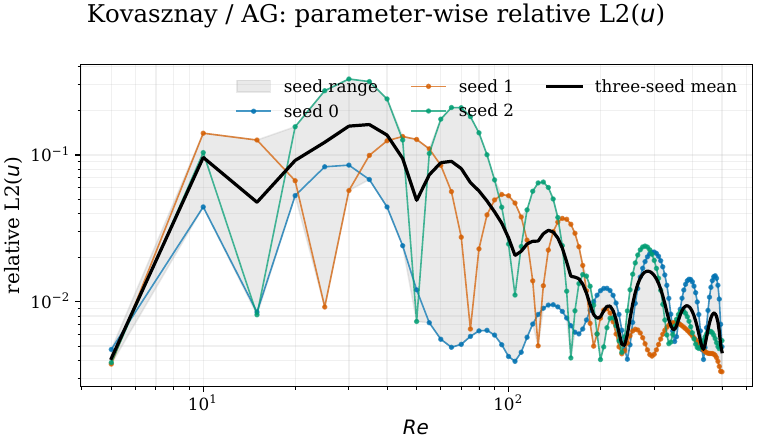}
\end{minipage}
\hfill
\begin{minipage}[t]{0.485\linewidth}
\centering
\textbf{AC}\par\smallskip
\suppinclude[width=\linewidth,height=0.225\textheight]{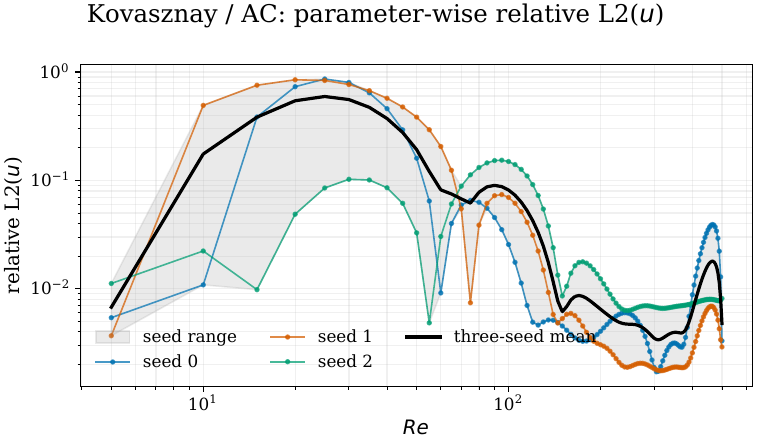}
\end{minipage}
\par\medskip
\begin{minipage}[t]{0.485\linewidth}
\centering
\textbf{ACR}\par\smallskip
\suppinclude[width=\linewidth,height=0.225\textheight]{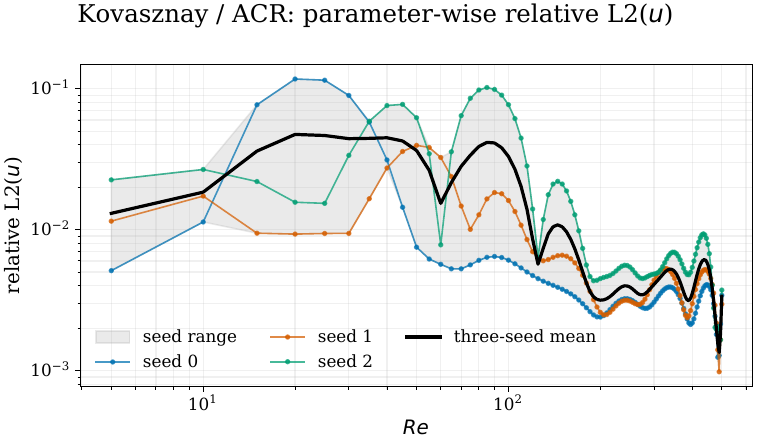}
\end{minipage}
\hfill
\begin{minipage}[t]{0.485\linewidth}
\centering
\textbf{ACR2-arch}\par\smallskip
\suppinclude[width=\linewidth,height=0.225\textheight]{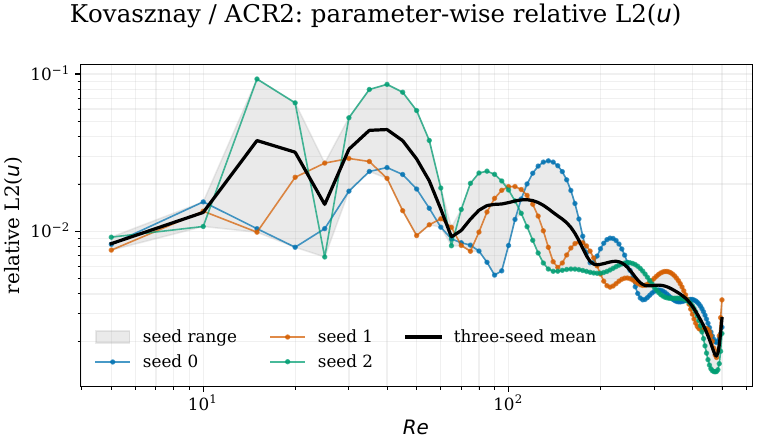}
\end{minipage}
\caption{$E_{L_2}$ distributions of the Kovasznay velocity component $u$ for the six methods over $Re\in[5,500]$. Each method shows the three fixed seeds and their summary curve.}
\label{fig:supp-expanded-kovas-u}
\end{figure}

\begin{figure}[p]
\centering
\begin{minipage}[t]{0.485\linewidth}
\centering
\textbf{UNI}\par\smallskip
\suppinclude[width=\linewidth,height=0.225\textheight]{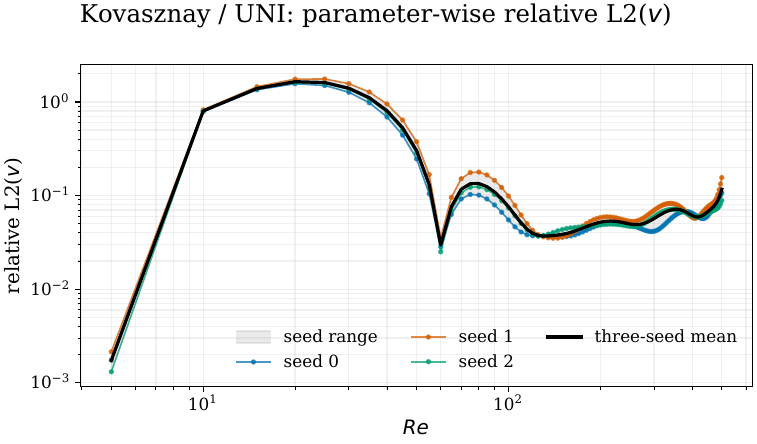}
\end{minipage}
\hfill
\begin{minipage}[t]{0.485\linewidth}
\centering
\textbf{FIX}\par\smallskip
\suppinclude[width=\linewidth,height=0.225\textheight]{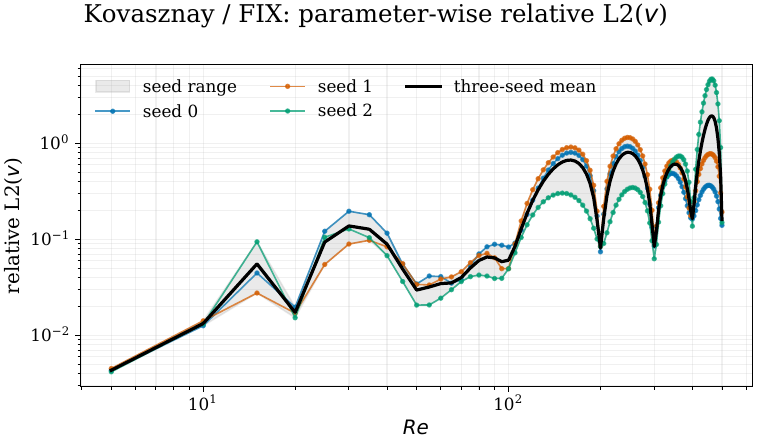}
\end{minipage}
\par\medskip
\begin{minipage}[t]{0.485\linewidth}
\centering
\textbf{AG}\par\smallskip
\suppinclude[width=\linewidth,height=0.225\textheight]{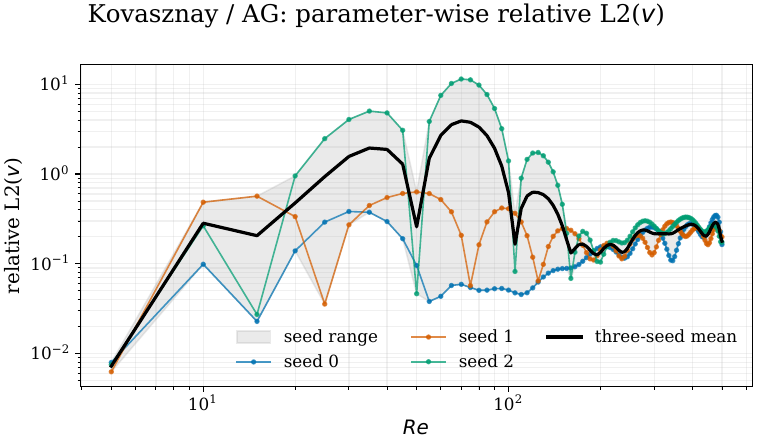}
\end{minipage}
\hfill
\begin{minipage}[t]{0.485\linewidth}
\centering
\textbf{AC}\par\smallskip
\suppinclude[width=\linewidth,height=0.225\textheight]{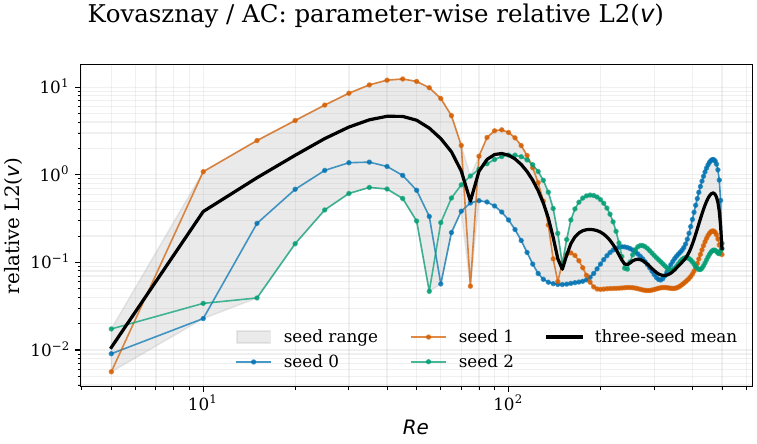}
\end{minipage}
\par\medskip
\begin{minipage}[t]{0.485\linewidth}
\centering
\textbf{ACR}\par\smallskip
\suppinclude[width=\linewidth,height=0.225\textheight]{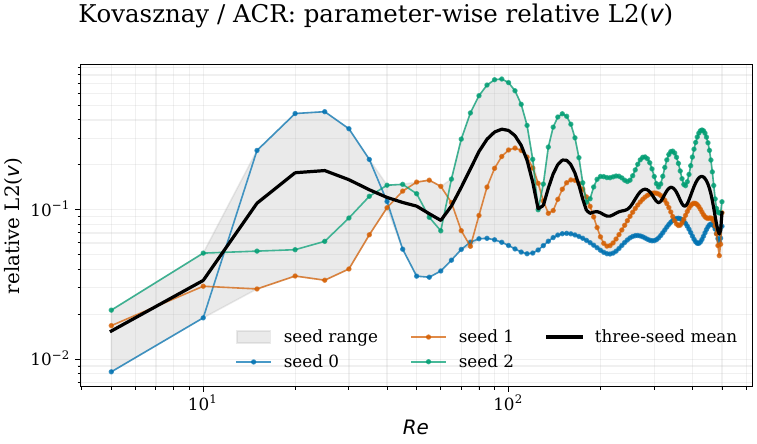}
\end{minipage}
\hfill
\begin{minipage}[t]{0.485\linewidth}
\centering
\textbf{ACR2-arch}\par\smallskip
\suppinclude[width=\linewidth,height=0.225\textheight]{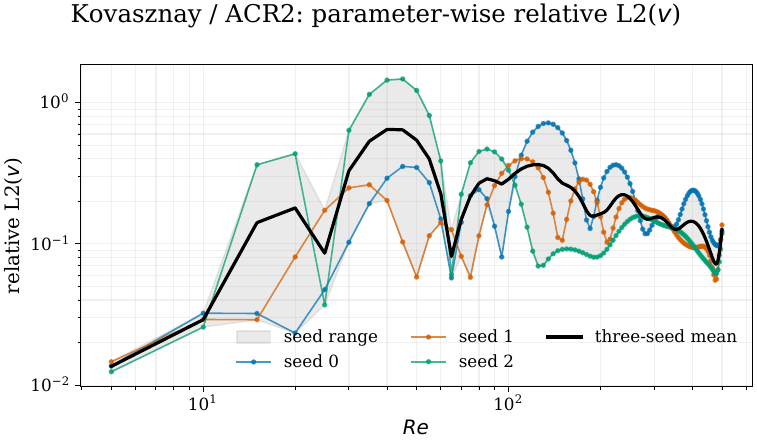}
\end{minipage}
\caption{$E_{L_2}$ distributions of the Kovasznay velocity component $v$ for the six methods over $Re\in[5,500]$. Components are evaluated separately so that differences in scale are not hidden by cross-component averaging.}
\label{fig:supp-expanded-kovas-v}
\end{figure}

\begin{figure}[p]
\centering
\begin{minipage}[t]{0.485\linewidth}
\centering
\textbf{UNI}\par\smallskip
\suppinclude[width=\linewidth,height=0.225\textheight]{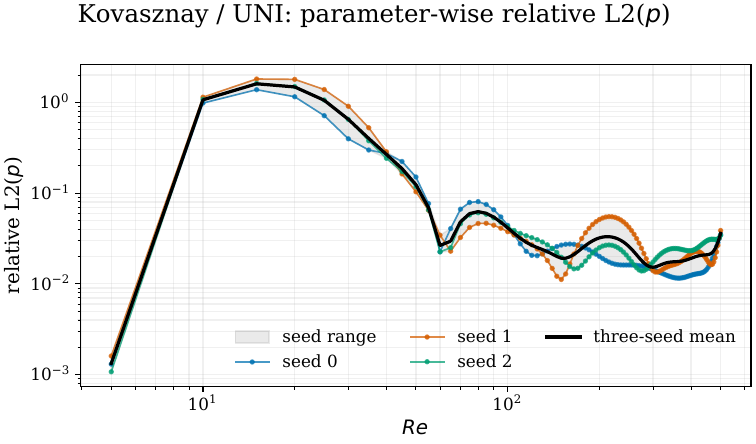}
\end{minipage}
\hfill
\begin{minipage}[t]{0.485\linewidth}
\centering
\textbf{FIX}\par\smallskip
\suppinclude[width=\linewidth,height=0.225\textheight]{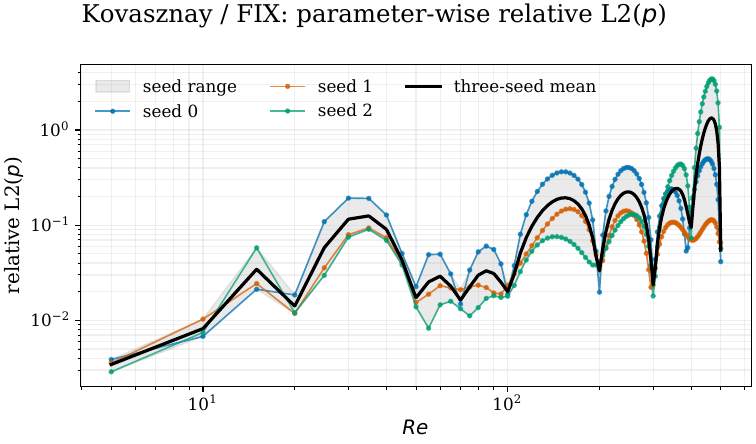}
\end{minipage}
\par\medskip
\begin{minipage}[t]{0.485\linewidth}
\centering
\textbf{AG}\par\smallskip
\suppinclude[width=\linewidth,height=0.225\textheight]{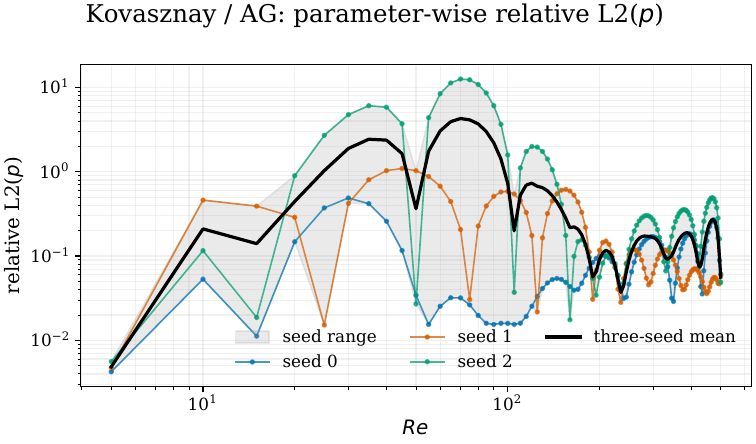}
\end{minipage}
\hfill
\begin{minipage}[t]{0.485\linewidth}
\centering
\textbf{AC}\par\smallskip
\suppinclude[width=\linewidth,height=0.225\textheight]{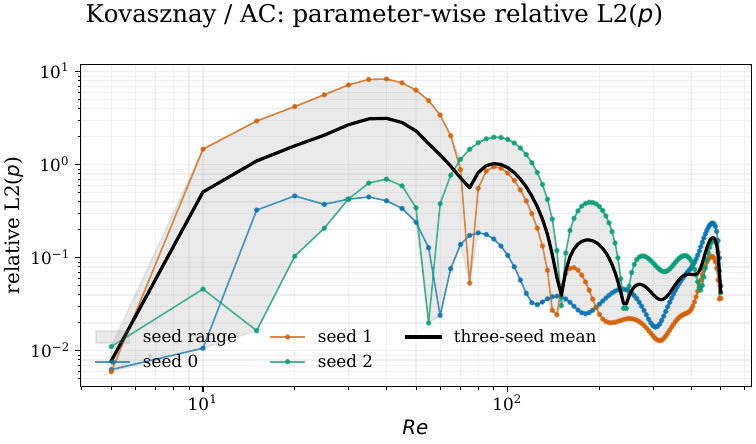}
\end{minipage}
\par\medskip
\begin{minipage}[t]{0.485\linewidth}
\centering
\textbf{ACR}\par\smallskip
\suppinclude[width=\linewidth,height=0.225\textheight]{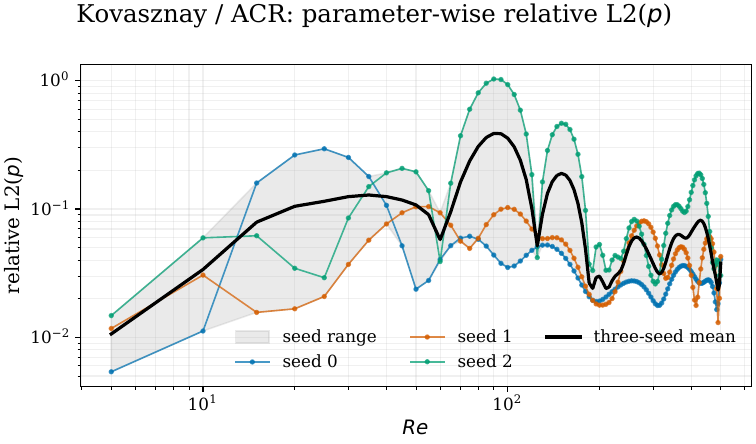}
\end{minipage}
\hfill
\begin{minipage}[t]{0.485\linewidth}
\centering
\textbf{ACR2-arch}\par\smallskip
\suppinclude[width=\linewidth,height=0.225\textheight]{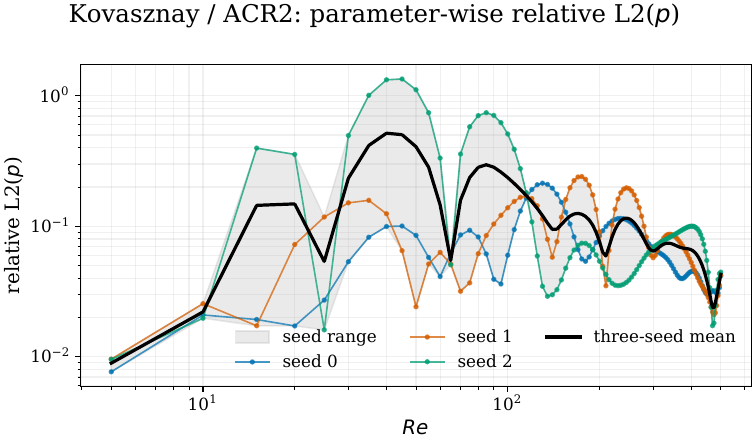}
\end{minipage}
\caption{$E_{L_2}$ distributions of the Kovasznay pressure $p$ for the six methods over $Re\in[5,500]$. Together with the $u$ and $v$ panels, this completes the comparison for the three-output problem.}
\label{fig:supp-expanded-kovas-p}
\end{figure}

\begin{figure}[p]
\centering
\begin{minipage}[t]{0.76\linewidth}
\centering
\textbf{Training physics/constraint loss}\par\smallskip
\suppinclude[width=\linewidth,height=0.31\textheight]{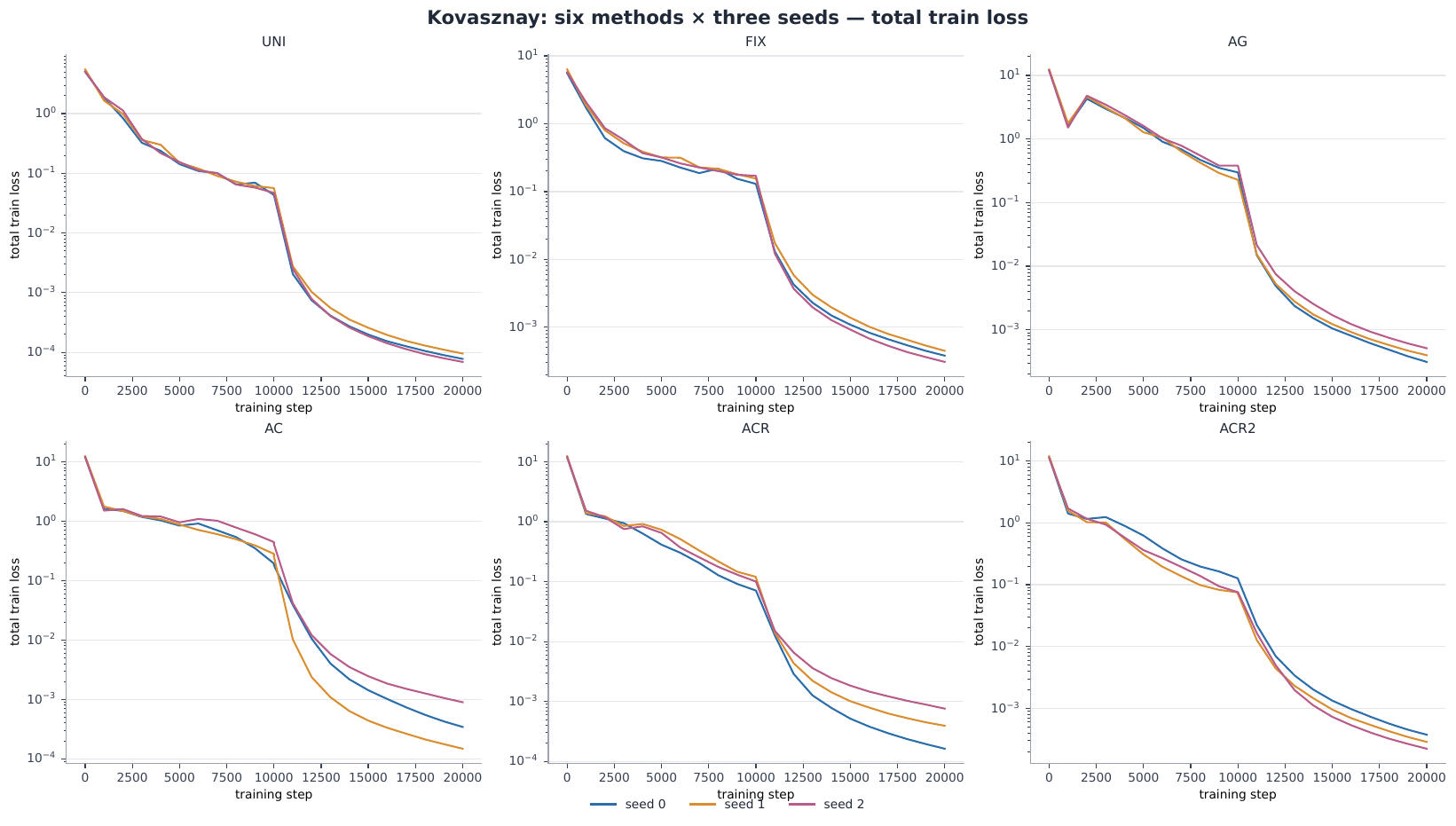}
\end{minipage}
\par\medskip
\begin{minipage}[t]{0.76\linewidth}
\centering
\textbf{Validation physics/constraint loss}\par\smallskip
\suppinclude[width=\linewidth,height=0.31\textheight]{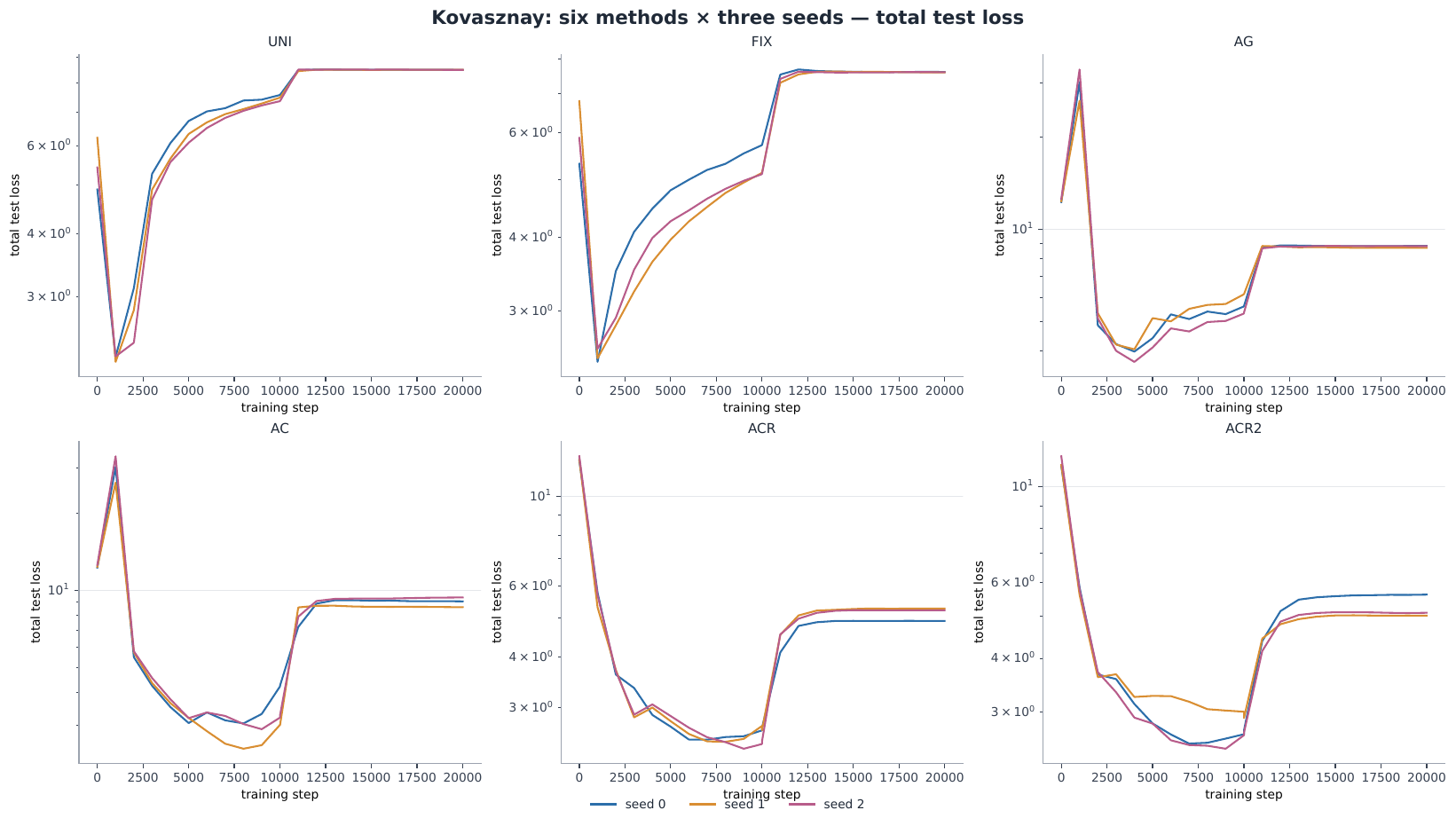}
\end{minipage}
\caption{Training and validation physics/constraint losses of the six methods on Kovasznay over three seeds. The validation quantity is not an aggregate of the reference-solution errors for $u$, $v$, and $p$.}
\label{fig:supp-expanded-kovas-loss}
\end{figure}

Figures~\ref{fig:supp-expanded-kovas-u}, \ref{fig:supp-expanded-kovas-v}, \ref{fig:supp-expanded-kovas-p}, and \ref{fig:supp-expanded-kovas-loss} explain the metric differences in the aggregate results. The $u$ component generally has smaller relative error, while the lower-amplitude $v$ and $p$ components are more sensitive to normalization. ACR2-arch is most favorable for $u$, whereas ACR is more favorable for $v$ and $p$. Because the component-wise and seed-wise rankings are not identical, the three-output problem cannot be summarized by one component or one seed.

\FloatBarrier

\subsubsection{Linearized Poisson--Boltzmann equation}

\label{sec:supp-4-1-5}

\begin{figure}[p]
\centering
\begin{minipage}[t]{0.96\linewidth}
\centering
\textbf{UNI}\par\smallskip
\suppinclude[width=\linewidth,height=0.70\textheight]{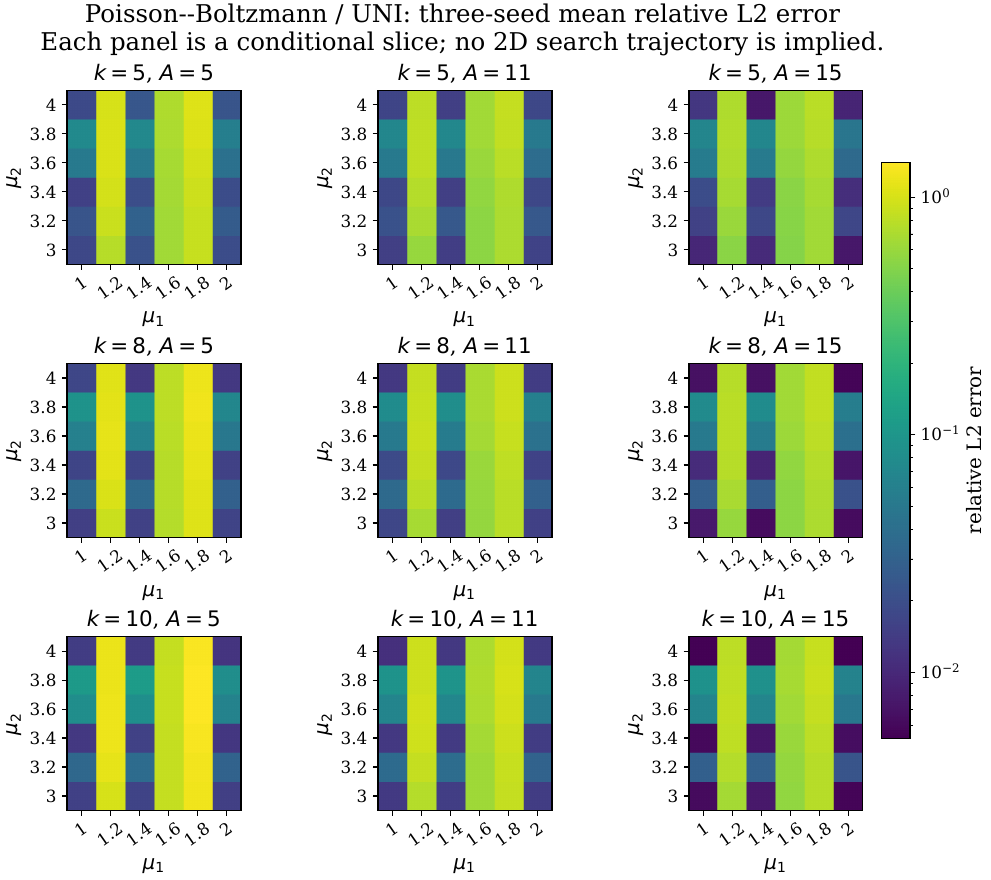}
\end{minipage}
\caption{Conditional slices of the three-seed mean $E_{L_2}$ over the four-dimensional linearized Poisson--Boltzmann parameter domain for UNI and FIX. Each panel fixes two parameters and shows the error over the other two; no two-dimensional search trajectory is implied.}
\label{fig:supp-expanded-poisson-l2-baselines}
\end{figure}

\begin{figure}[p]
\ContinuedFloat
\centering
\begin{minipage}[t]{0.96\linewidth}
\centering
\textbf{FIX}\par\smallskip
\suppinclude[width=\linewidth,height=0.70\textheight]{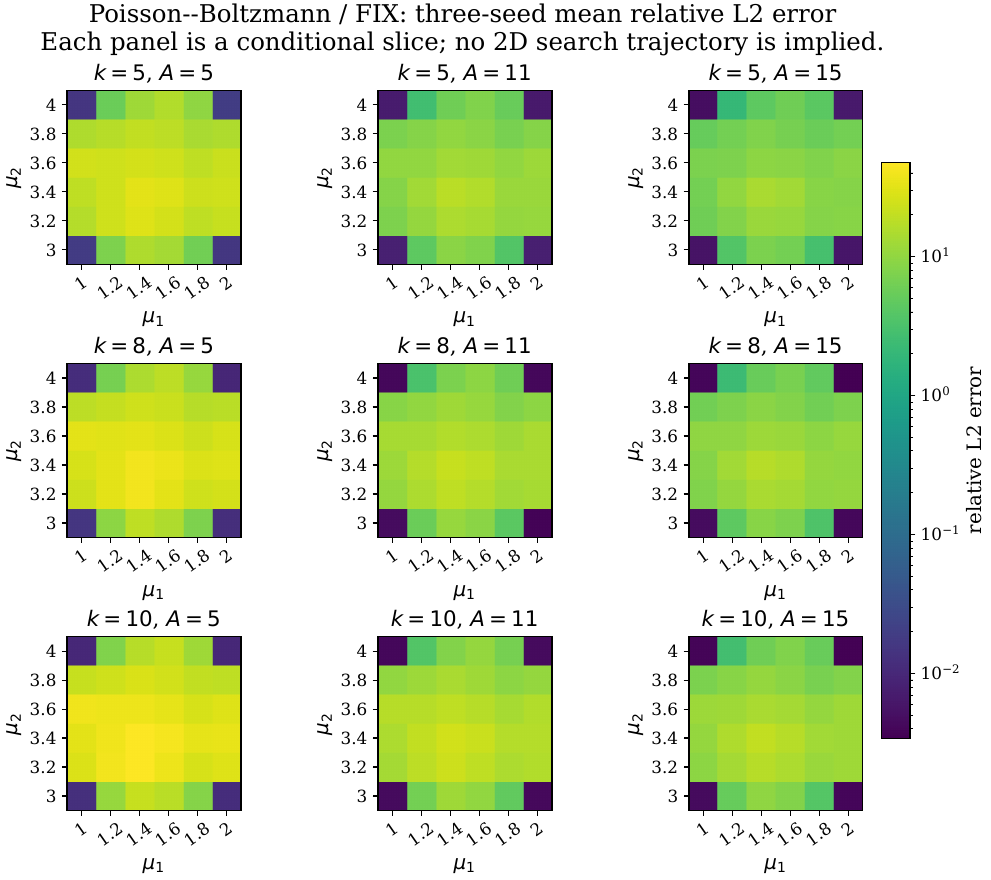}
\end{minipage}
\caption[]{Conditional slices for FIX, continued from Figure~\ref{fig:supp-expanded-poisson-l2-baselines}. The fixed-parameter and displayed-parameter conventions are unchanged.}
\end{figure}

\begin{figure}[p]
\centering
\begin{minipage}[t]{0.96\linewidth}
\centering
\textbf{AG}\par\smallskip
\suppinclude[width=\linewidth,height=0.70\textheight]{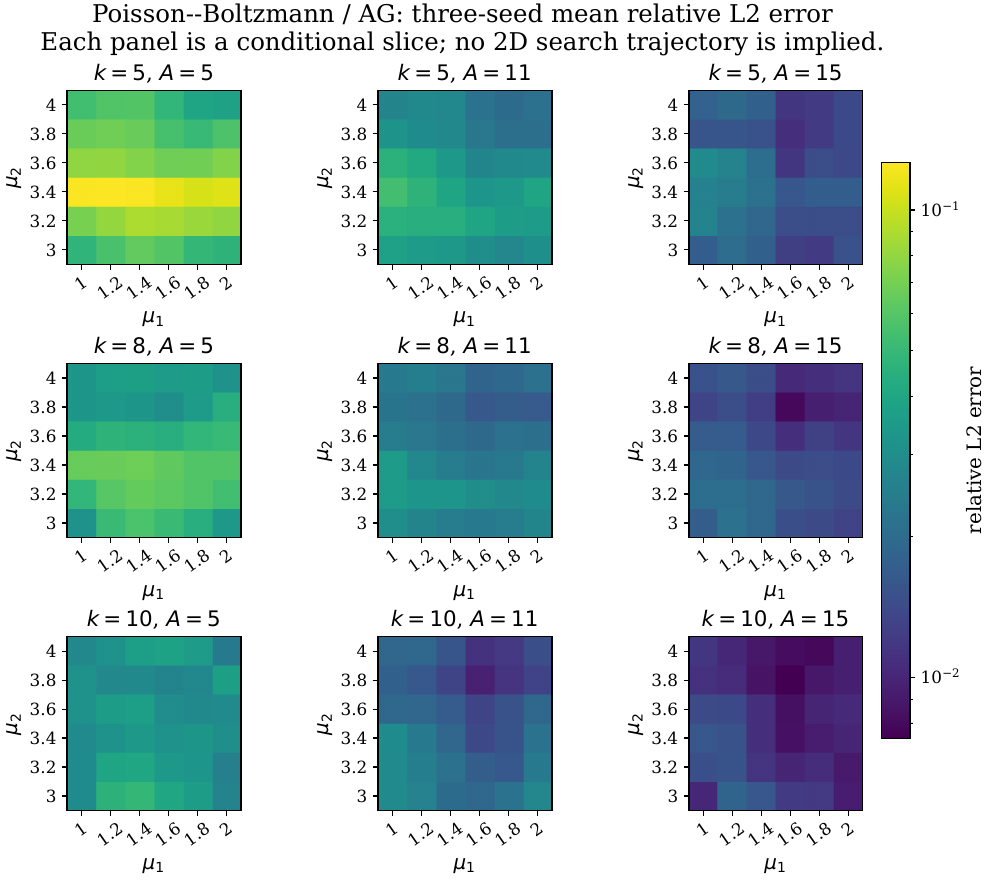}
\end{minipage}
\caption{Conditional slices of the three-seed mean $E_{L_2}$ over the four-dimensional linearized Poisson--Boltzmann parameter domain for AG and AC. The fixed parameter values are stated in the panel titles.}
\label{fig:supp-expanded-poisson-l2-active}
\end{figure}

\begin{figure}[p]
\ContinuedFloat
\centering
\begin{minipage}[t]{0.96\linewidth}
\centering
\textbf{AC}\par\smallskip
\suppinclude[width=\linewidth,height=0.70\textheight]{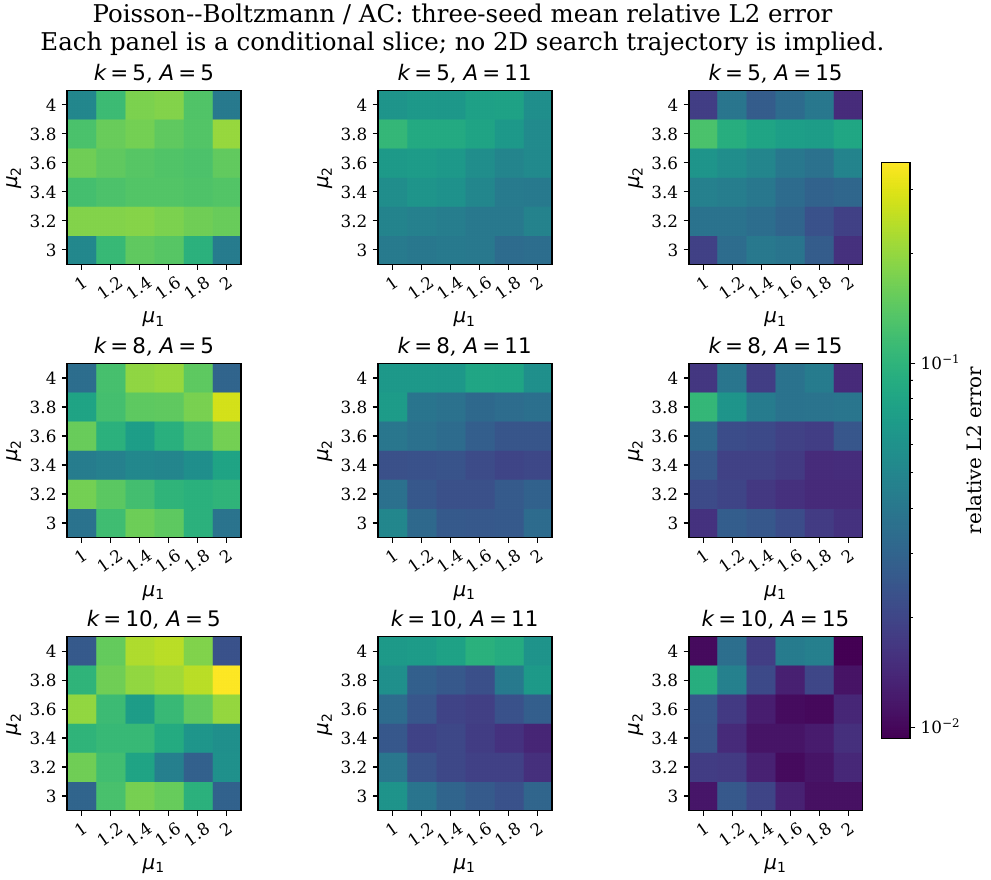}
\end{minipage}
\caption[]{Conditional slices for AC, continued from Figure~\ref{fig:supp-expanded-poisson-l2-active}. The fixed parameter values are stated in the panel titles.}
\end{figure}

\begin{figure}[p]
\centering
\begin{minipage}[t]{0.96\linewidth}
\centering
\textbf{ACR}\par\smallskip
\suppinclude[width=\linewidth,height=0.70\textheight]{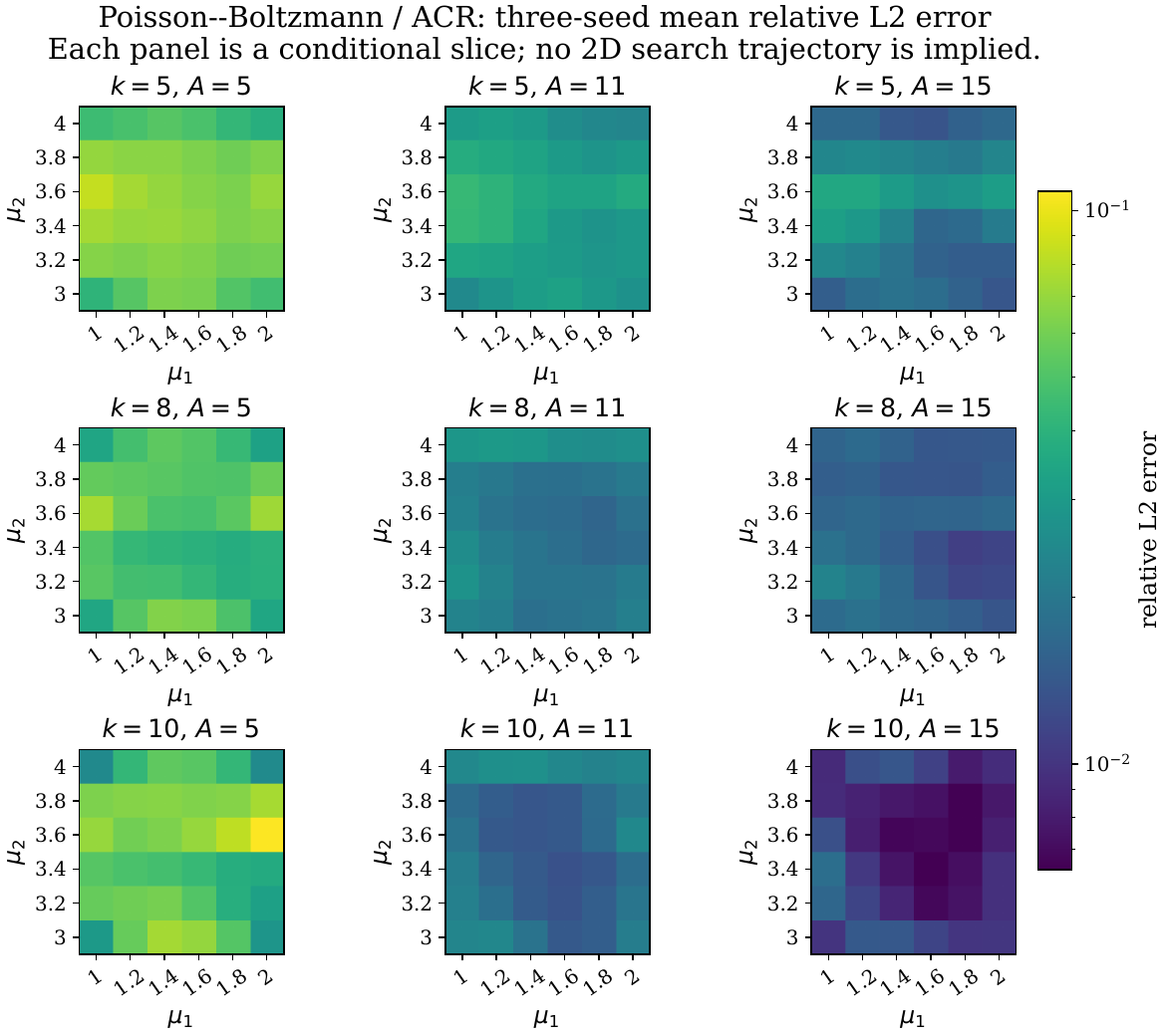}
\end{minipage}
\caption{Conditional slices of the three-seed mean $E_{L_2}$ over the four-dimensional linearized Poisson--Boltzmann parameter domain for ACR and ACR2-arch. These slices expose localized difficult regions that are not visible in the global mean alone.}
\label{fig:supp-expanded-poisson-l2-replay}
\end{figure}

\begin{figure}[p]
\ContinuedFloat
\centering
\begin{minipage}[t]{0.96\linewidth}
\centering
\textbf{ACR2-arch}\par\smallskip
\suppinclude[width=\linewidth,height=0.70\textheight]{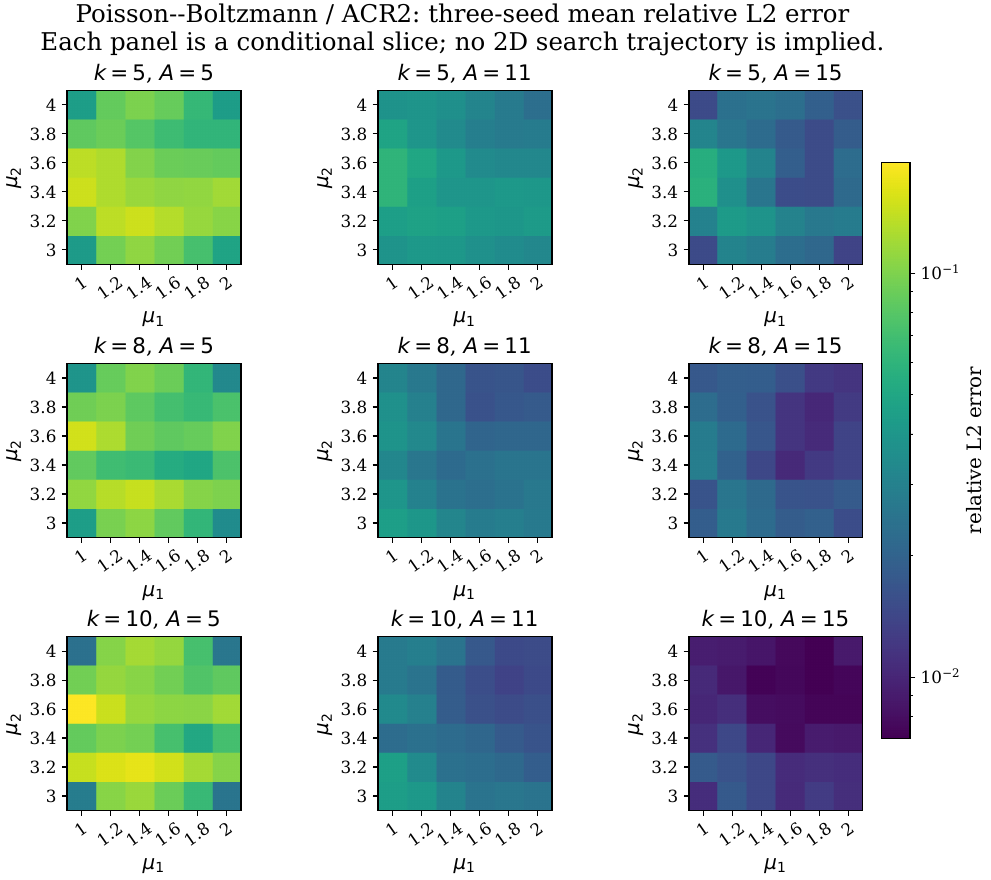}
\end{minipage}
\caption[]{Conditional slices for ACR2-arch, continued from Figure~\ref{fig:supp-expanded-poisson-l2-replay}. These slices retain the same parameter conditions and color interpretation.}
\end{figure}

\begin{figure}[p]
\centering
\begin{minipage}[t]{0.76\linewidth}
\centering
\textbf{Training physics/constraint loss}\par\smallskip
\suppinclude[width=\linewidth,height=0.31\textheight]{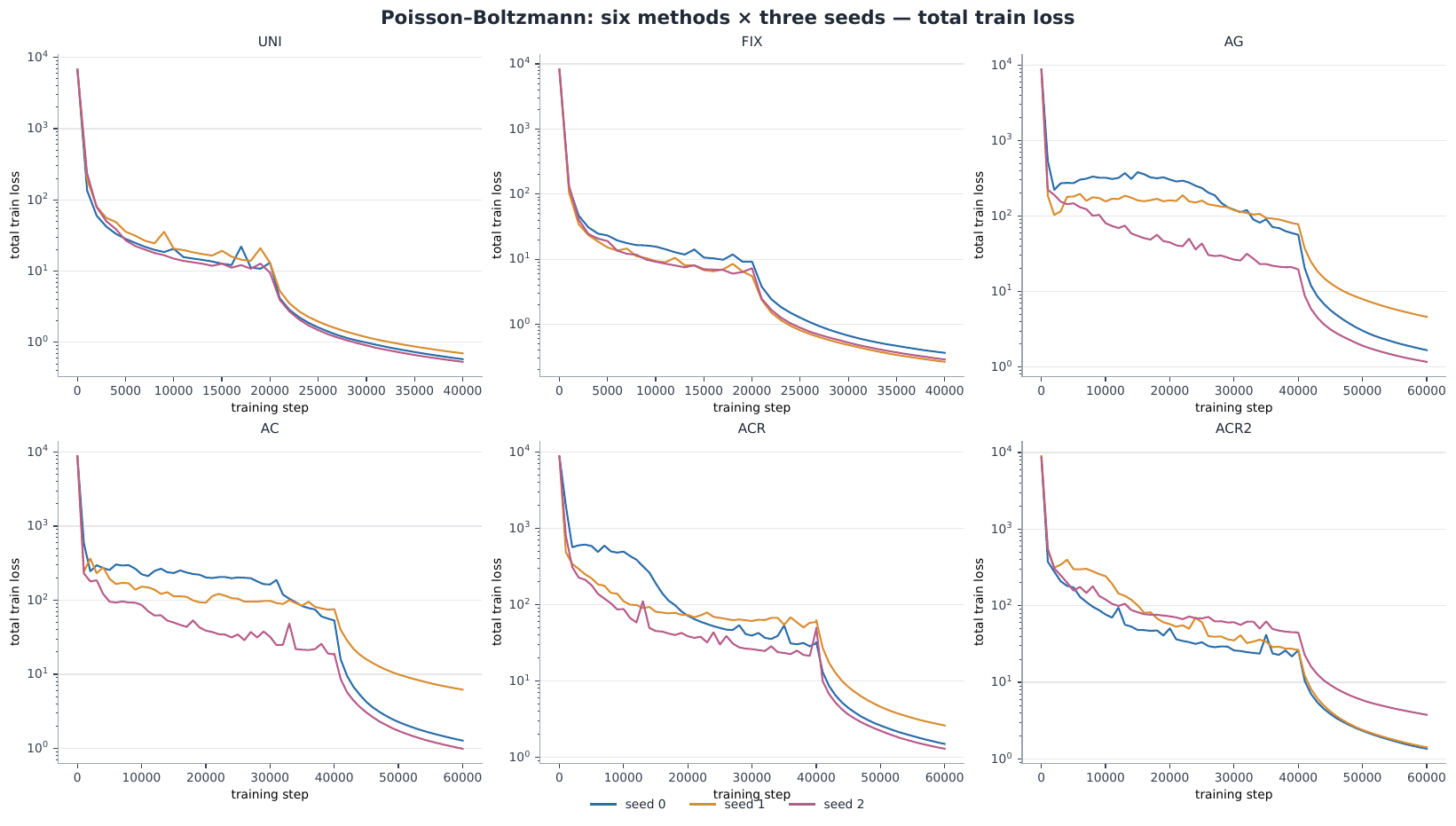}
\end{minipage}
\par\medskip
\begin{minipage}[t]{0.76\linewidth}
\centering
\textbf{Validation physics/constraint loss}\par\smallskip
\suppinclude[width=\linewidth,height=0.31\textheight]{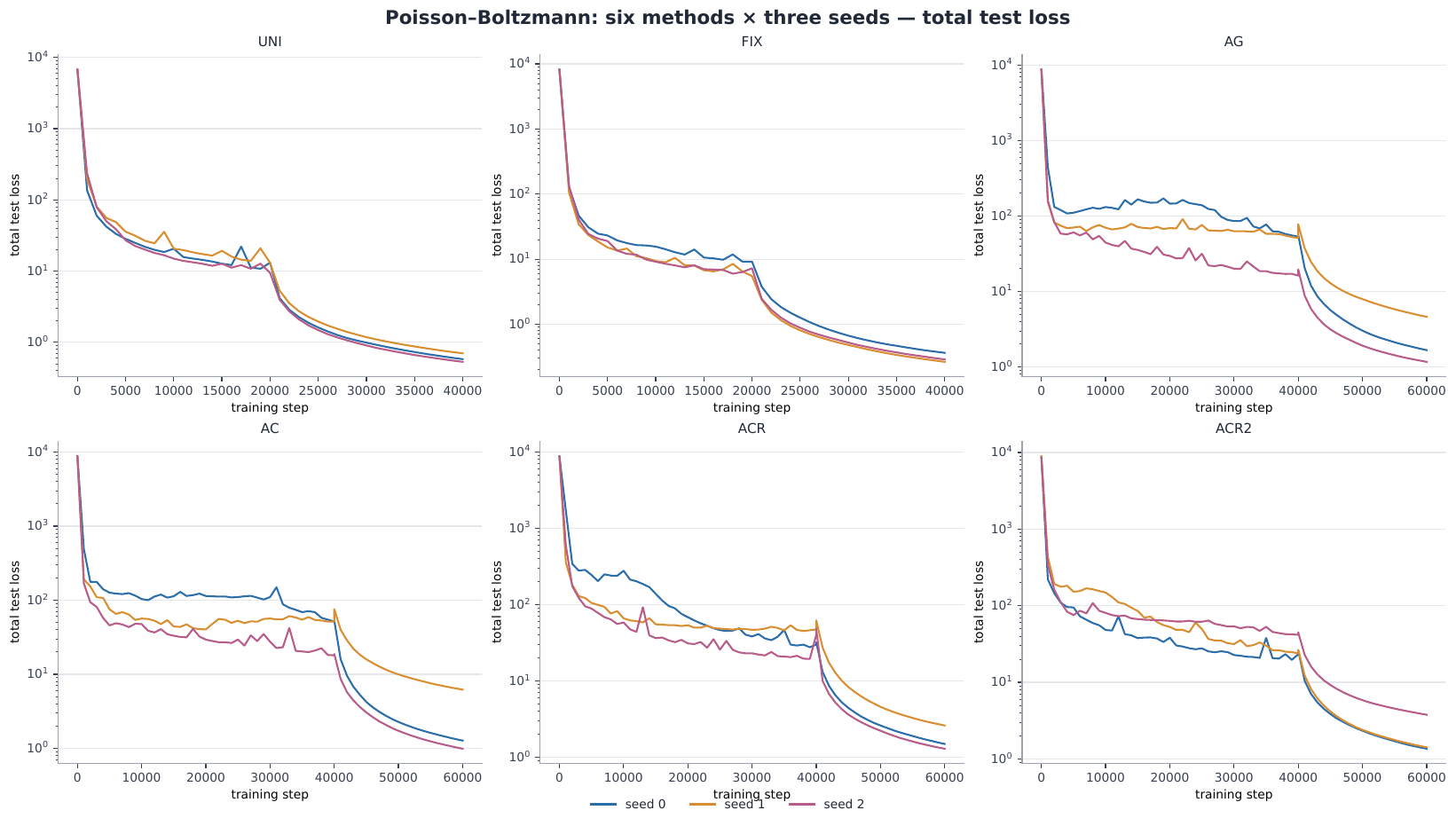}
\end{minipage}
\caption{Training and validation physics/constraint losses of the six methods on linearized Poisson--Boltzmann over three seeds. The final metric uses all $6^4=1296$ test-parameter tuples; the loss histories do not replace the full-grid reference-solution error.}
\label{fig:supp-expanded-poisson-loss}
\end{figure}

Figures~\ref{fig:supp-expanded-poisson-l2-baselines}, \ref{fig:supp-expanded-poisson-l2-active}, \ref{fig:supp-expanded-poisson-l2-replay}, and \ref{fig:supp-expanded-poisson-loss} show broadly distributed high-error regions for UNI and FIX, whereas the dynamic methods reduce error over most of the domain. ACR has the most consistently low-error regions, while ACR2-arch retains several local peaks. The patterns change markedly across $(k,A)$ conditions, indicating interactions among the four parameters that cannot be summarized by one marginal curve. Final comparison uses MSE, macro-averaged $E_{L_2}$, and worst $E_{L_2}$ over all 1,296 test tuples.

\FloatBarrier

\bibliographystyle{elsarticle-num}

\bibliography{references}